\def\cvprPaperID{11052}
\def\confName{ICCV}
\def\confYear{2023}

\def\paperTitle{VideoFACT: Detecting Video Forgeries Using Attention, Scene Context, and Forensic Traces}

\def\authorBlock{
    Tai D. Nguyen \qquad
    Shengbang Fang \qquad
    Matthew C. Stamm \qquad\\
    Drexel University \\
    {\tt\small \{tdn47, sf683, mcs382\}@drexel.edu}
}

\newif\ifreview 
\newif\ifarxiv \newcommand{\arxiv}{\arxivtrue}
\newif\ifcamera 
\newif\ifrebuttal 

\arxiv 

\pdfoutput=1
\documentclass[10pt,twocolumn,letterpaper]{article}
\ifreview \usepackage[review]{cvpr} \fi
\ifarxiv \usepackage[pagenumbers]{cvpr} \fi
\ifrebuttal \usepackage[rebuttal]{cvpr} \fi
\ifcamera \usepackage{cvpr} \fi

\usepackage{graphicx}
\usepackage{amsmath}
\usepackage{amssymb}
\usepackage{booktabs}
\usepackage{comment}

\usepackage{times}
\usepackage{microtype}
\usepackage{epsfig}
\usepackage[table,xcdraw]{xcolor}
\usepackage{caption}
\usepackage{cases}
\usepackage{float}
\usepackage{placeins}
\usepackage{color, colortbl}
\usepackage{stfloats}
\usepackage{enumitem}
\usepackage{tabularx}
\usepackage{xstring}
\usepackage{array}
\usepackage{graphicx}
\usepackage{multirow}
\usepackage{ragged2e}
\usepackage{xspace}
\usepackage{scalerel}
\usepackage{bbm}
\usepackage{url}
\usepackage{xcolor}
\usepackage{adjustbox}
\usepackage{balance}
\usepackage[hang,flushmargin]{footmisc}
\usepackage[nodisplayskipstretch]{setspace}
\usepackage[font=scriptsize]{caption}

\ifcamera \usepackage[accsupp]{axessibility} \fi

% ...

\newcommand{\smaller}{\fontsize{7pt}{8pt}\selectfont}

\newcommand{\subheader}[1]{\textbf{#1.}}
\newcommand{\skipresult}{N/A}
\newcommand{\boldred}[1]{\textcolor{red}{\textbf{#1}}}
\newcommand{\boldblue}[1]{\textcolor{blue}{\textbf{#1}}}

\DeclareMathOperator*{\bigplus}{\scalerel*{+}{\sum}}

\ifarxiv  \fi

\newcommand{\R}[1]{{%
    \textbf{%
        \ifstrequal{#1}{1}{\textcolor{red}{R#1}}{%
        \ifstrequal{#1}{2}{\textcolor{blue}{R#1}}{%
        \ifstrequal{#1}{3}{\textcolor{magenta}{R#1}}{%
        \ifstrequal{#1}{4}{\textcolor{teal}{R#1}}{%
                           \textcolor{cyan}{R#1}%
        }}}}%
    }%
}}

\setdisplayskipstretch{0.2}
\usepackage{xr-hyper}

\makeatletter
\newcommand*{\addFileDependency}[1]{
  \typeout{(#1)}
  \@addtofilelist{#1}
  \IfFileExists{#1}{}{\typeout{No file #1.}}
}

\makeatother

\usepackage[pagebackref,breaklinks,colorlinks]{hyperref}
\usepackage[capitalize]{cleveref}
\crefname{section}{Sec.}{Secs.}
\crefname{table}{Table}{Tables}
\crefname{figure}{Fig.}{Figs.}

\begin{document}
\title{\paperTitle}
\author{\authorBlock}
\maketitle
\begin{abstract}
\label{sec:abstract}

Fake videos represent an important misinformation threat. While existing forensic networks have demonstrated strong performance on image forgeries, recent results reported on the Adobe VideoSham dataset show that these networks fail to identify fake content in videos. 
In this paper, we show that this is due to video coding,  which introduces local variation into forensic traces. 
In response, 
we propose VideoFACT - a new network that is able to detect and localize a wide variety of video forgeries and manipulations. To overcome challenges that existing networks face when analyzing videos, our network utilizes both forensic embeddings to capture traces left by manipulation, context embeddings 
to control for variation in forensic traces introduced by video coding,
and a deep self-attention mechanism to estimate the quality and relative importance of local forensic embeddings.  
We create several new video forgery datasets and use these, along with publicly available data, to experimentally evaluate our network's performance. These results show that our proposed network is able to identify a diverse set of video forgeries, including those not encountered during training. 
Furthermore, we show that our network can be fine-tuned to achieve even stronger performance on challenging AI-based manipulations. 

\end{abstract}

%

\section{Introduction}
\label{sec:introduction}


Detecting fake and manipulated content in videos is critical in the fight against misinformation, online fraud, and many other threats.
Traditional  editing software enables users to convincingly add, remove, and alter  virtually any object in a video.  Furthermore, recent advances in AI-based video editing have caused dramatic advancements in how videos can be falsified.  For example, AI-based video inpainting makes it possible to seamlessly remove an object from a video and replace it with a visually convincing background.

To combat forgeries, researchers have developed a wide variety of general purpose forgery detection and localization techniques.  These networks operate by  directly learning to detect several known forgery types~\cite{Bayar_2018_TIFS, Zhou_2018_CVPR, Yang_2020_ICME, Zhang_2020_TIFS, Marra_2020_IEEEA, Yu_2019_IEEEA, Matern_2020_ICASSP, Bunk_2017_CVPRW, Bappy_2017_ICCV, Hu_2020_ECCV}, or by searching for localized anomalies in 
forensic traces~\cite{Noiseprint, EXIFnet, FSG, MVSS-Net, ManTra-Net}.
This research has focused nearly exclusively on images.


%
Most existing video-specific forensic techniques are aimed at detecting manipulations such as
frame deletion or addition~\cite{Gironi_2014_ICASSP, Gonzalez_2022_TITS}, speed manipulation~\cite{Hosler_2020_CVPRW}, etc. 
Other research makes specific assumptions about the video's content and forgery type, e.g. detecting deepfake videos of a human speaker's face~\cite{Wang_2019_ICCV, Guera_2018_AVSS, Zhao_2021_CVPR, Luo_2021_CVPR, Yang_2019_ICASSP, Jiang_2020_CVPR, Huang_2022_TIFS, Hsu_2020_JAS}, authenticating video with a still background~\cite{Shelke_2021_Vid_Survey}, etc.
%
Currently, there are no deep learning approaches designed to detect general content forgeries in modern video~\cite{Shelke_2021_Vid_Survey}.
This research is further limited by the lack public datasets of general video forgeries.  To the best of our knowledge, the first such dataset is Adobe's VideoSham dataset, which was published in 2023~\cite{VideoSham} .





\begin{figure}[!t]
    \centering
    \setlength{\fboxsep}{0pt}
    \begin{minipage}[c]{1\linewidth}
        \centering
        \makebox[0.190\textwidth]{}
        \makebox[0.252\textwidth]{Authentic}
        \makebox[0.252\textwidth]{Forged}
        \makebox[0.252\textwidth]{Predicted Mask}
        \smallskip
    \end{minipage}

	\vspace*{-0.2\baselineskip}

    \begin{minipage}[ct]{1\linewidth}
        \centering
		\makebox[0.190\textwidth][r]{\raisebox{13pt}{Deepfake\hspace{6pt}}}
        \fbox{\includegraphics[width=0.252\textwidth]{{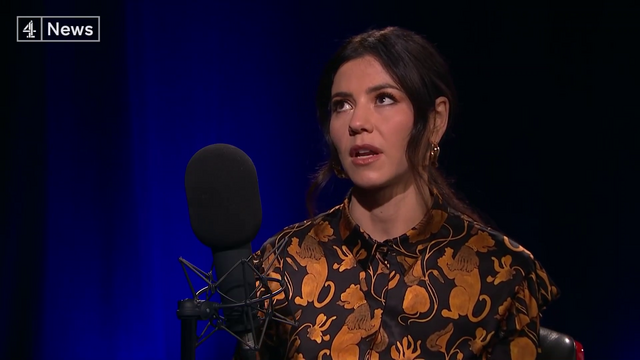}}}
        \fbox{\includegraphics[width=0.252\textwidth]{{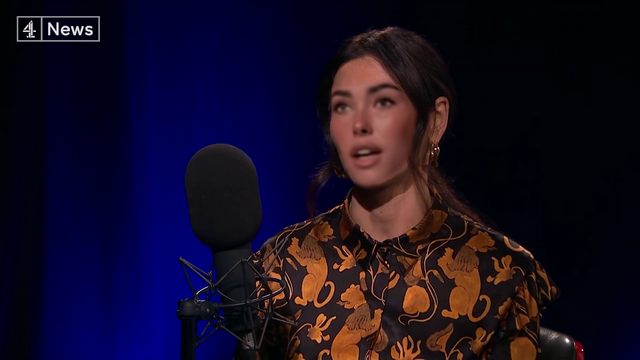}}}
        \fbox{\includegraphics[width=0.252\textwidth]{{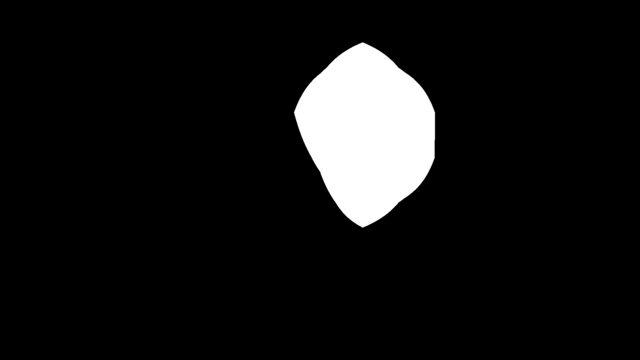}}}
        \smallskip
    \end{minipage}

	\vspace*{-0.2\baselineskip}
    
    \begin{minipage}[c]{1\linewidth}
        \centering
        \makebox[0.190\textwidth][r]{\raisebox{13pt}{Splicing\hspace{6pt}}}
        \fbox{\includegraphics[width=0.252\textwidth]{{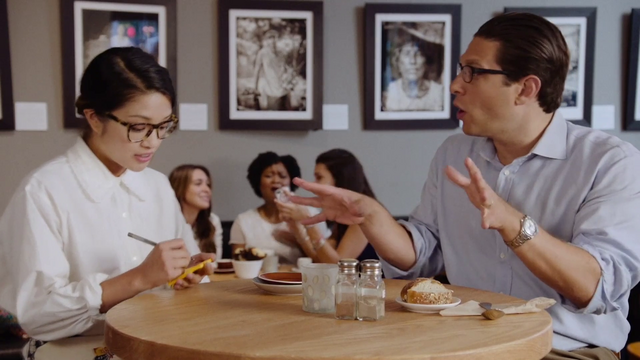}}}
        \fbox{\includegraphics[width=0.252\textwidth]{{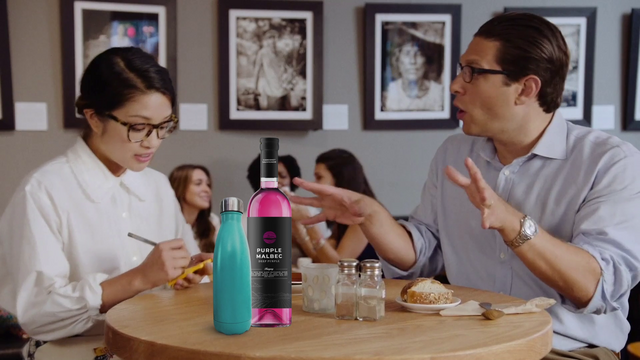}}}
        \fbox{\includegraphics[width=0.252\textwidth]{{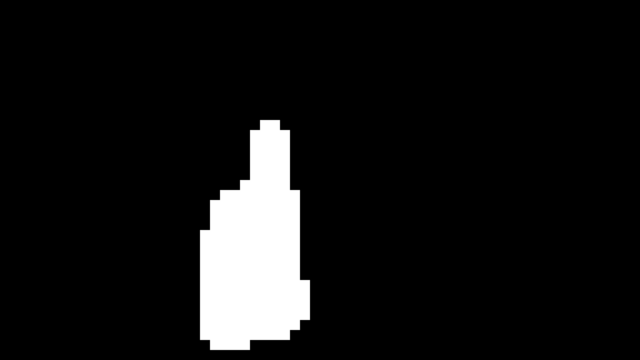}}}
        \smallskip
    \end{minipage}

	\vspace*{-0.2\baselineskip}

    \begin{minipage}[c]{1\linewidth}
        \centering
        \makebox[0.190\textwidth][r]{\raisebox{13pt}{Inpainting\hspace{6pt}}}
        \fbox{\includegraphics[width=0.252\textwidth]{{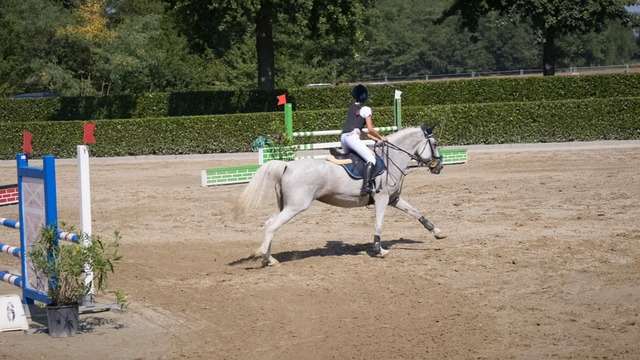}}}
        \fbox{\includegraphics[width=0.252\textwidth]{{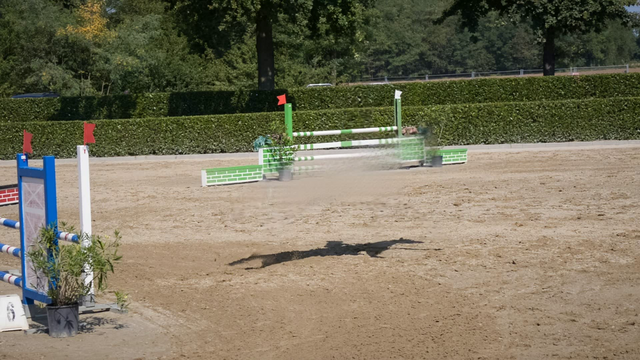}}}
        \fbox{\includegraphics[width=0.252\textwidth]{{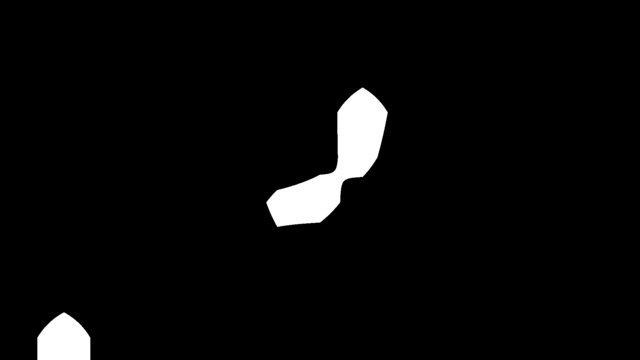}}}
        \smallskip
    \end{minipage}
    \vspace*{-0.8\baselineskip}

	\caption{\label{fig:front_page_graphics} Sample video forgery localization results obtained using our proposed VideoFACT network  on videos modified by deepfaking (Top), splicing in an object (Middle), and removing an object with inpainting (Bottom).}
	
	
    
    \vspace*{-\baselineskip}

\end{figure}

Suprisingly, VideoSham's benchmarking results have revealed that existing general forgery detection and localization networks
all fail when applied to video~\cite{VideoSham}.
This finding, which is further reinforced by results presented in this paper, can be attributed to the effects of video compression.
Modern video codecs, such as H.264, utilize different coding parameters and compression strengths for different macroblocks within a single frame~\cite{H264}.
This introduces local variations into the forensic traces in a frame.  Forensic networks misinterpret these variations as anomalous traces, causing them to false alarm.  Furthermore, stronger coding in some frame regions reduces the quality of local traces, which can cause forensic networks to make incorrect decisions.

In this paper, we propose a new general video forensic network that is able to detect and localize a wide variety of fake content in  video.
%
We name our network VideoFACT: Video Forensics using Attention, Context, and Traces.  To overcome the negative effects of video coding,  our network contains several critical and novel components.  We learn generic forensic feature embeddings for video that can capture traces left by a variety of manipulations. We observe that local video coding parameters are dependent upon contextual information, including  scene content and several other factors.
We exploit this by introducing the concept of context embeddings to forensics.  Our network uses these embeddings to control for local variation in forensic traces introduced by video coding.  We also observe that to correctly identify anomalous forensic traces left by forgery, it is better to analyze local forensic embeddings with respect to one another rather than independently.  Our network does this by utilizing a deep self-attention mechanism to estimate the quality and relative importance of local forensic embeddings.


The main contributions of our paper are as follows:

(1) We propose a new forensic network that is able to perform general purpose forgery detection and localization on video.  We overcome the negative effects of video coding by introducing novel network components, including: context embeddings to control for variation in forensic traces, 
and a deep self-attention mechanism to estimate the quality and relative importance of forensic embeddings.
%
We show that our network can detect a wide variety of manipulations, including those that it was not explicitly trained on.  

(2) We develop multiple new video forgery datasets,
composed of both standard video manipulations such as splicing, and advanced manipulations such as AI-based inpainting.
Currently, there are no  general video forgery datasets made for network training, and only  VideoSham for benchmarking.  Our datasets can be used to both train and benchmark video forensic algorithms, thus helping to enable further research. 



(3) We provide an extensive set of experiments to evaluate both our proposed network and state-of-the-art forensic networks.
We show that VideoFACT achieves the best reported video forgery detection and localization performance, while existing approaches fail due to video coding effects.
Furthermore, we show that our network can be fine-tuned to achieve even stronger performance on AI-based manipulations.
We perform an ablation study showing the importance of each component of VideoFACT.

\section{Background and Related Work}
\label{sec:related_work}

Researchers have developed multiple forensic algorithms to fight against fake content in images and videos.
These algorithms includes traditional signal-processing-based methods and recent deep-learning-based approaches.

\subheader{Signal-Processing-Based Methods}
Early image forensic algorithms used signal processing methods to extract forensic features~\cite{Farid_2005_TSP, Stamm_2008_ICIP, Fridrich_2012_TIFS, Fridrich_2008_TIFS, Kirchner_2008_fast, Popescu_2005_IWIH, Kirchner_2010_TSOP}. These algorithms utilize hand-designed features to detect specific types of image manipulations. Inspired by image forensics, researchers also developed several forensic algorithms for videos~\cite{Shelke_2021_Vid_Survey}. These algorithms either focus on specific types of forgeries, have restrictions on video contents, or only applied to early video codec with consistent compression.

\subheader{Deep-Learning-Based Methods}
Following the rise of deep learning, researchers developed a wide variety of new forensic networks capable of simultaneously identifying a broad range of editing operations~\cite{Bayar_2018_TIFS, Zhou_2018_CVPR, Yang_2020_ICME, Zhang_2020_TIFS, Marra_2020_IEEEA, Yu_2019_IEEEA, Matern_2020_ICASSP, Bunk_2017_CVPRW, Bappy_2017_ICCV, Hu_2020_ECCV}. These algorithms achieve strong performance detecting manipulations that they are trained on, but typically encounter difficulties on detecting new manipulations outside of the training set.
For videos, very few techniques are designed so far that can detect the general forgeries using deep learning~\cite{Shelke_2021_Vid_Survey}. Existing deep-learning based algorithms can only detect or localize specific types of falsified contents~\cite{Shanableh_2013_Elsevier, Gironi_2014_ICASSP, Long_2017_CVPRW, Zampoglou_2019_ICMMM}, like deepfakes
~\cite{Wang_2019_ICCV, Guera_2018_AVSS, Zhao_2021_CVPR, Luo_2021_CVPR, Yang_2019_ICASSP, Jiang_2020_CVPR, Huang_2022_TIFS}.

\subheader{Anomaly Detection}
Currently, the most relevant deep learning methods for general video forgery detection are algorithms using anomaly detection on images. Although these algorithms are not designed to performs on videos. These algorithms are designed to implicitly or explicitly detect the out-of-group forensic features within an image to perform general forgery detection and localization.
Specifically, Forensic Similarity Graph~\cite{FSG}, EXIF-Net~\cite{EXIFnet}, and Noiseprint~\cite{Noiseprint} do this by divided the full-size image into small patches, then compute patch-wise difference of forensic features and cluster patches into real and fake groups.
ManTraNet~\cite{ManTra-Net} computes the anomaly score by measuring the consistencies of the features within
a kernel. MVSS-Net~\cite{MVSS-Net} scores the consistencies of the features within an analysis window but it uses a pyramidal resolution approach.

\section{Effect of Video Coding on Forensic Traces}
\label{sec:effect_of_video_coding}

Though state-of-the-art forensic networks are able to correctly identify fake content in images, these networks all fail when used to analyze video. This surprising behavior is due to effects that video coding has on forensic traces.

One important effect is that video coding introduces localized variations into the forensic traces within a frame.  This is because modern video codecs, such as H.264, do not use the same coding parameters or compression strength for every macroblock in a frame~\cite{H264}.  Instead, these vary depending on several factors, including the content of a macroblock, the bit budget, the level of prediction error, etc.  Because compression is well-known to significantly alter forensic traces~\cite{Bayar_2017_ICASSP, Kirchner_2010_TSOP, Popescu_2005_ITSP}, this causes traces in some authentic regions to differ significantly from those in others.  These traces will then appear anomalous to forensic algorithms, which will cause them to false alarm.

Another important effect is that video coding reduces the quality of forensic traces in some regions.
This is because some macroblocks are subjected to stronger compression, which is well-known to reduce the quality of forensic traces~\cite{Jiang_2018_TIFS, Mandelli_2020_WIFS, Stamm_2011_TIFS, Stamm_2010_ICIP}.  
Existing forensic networks weight all local features equally and do not account for local variations in quality.  As a result, low-quality local forensic traces can cause forensic algorithms to make incorrect decision.

\begin{figure}[!t]
	\centering
	\includegraphics[width=\linewidth]{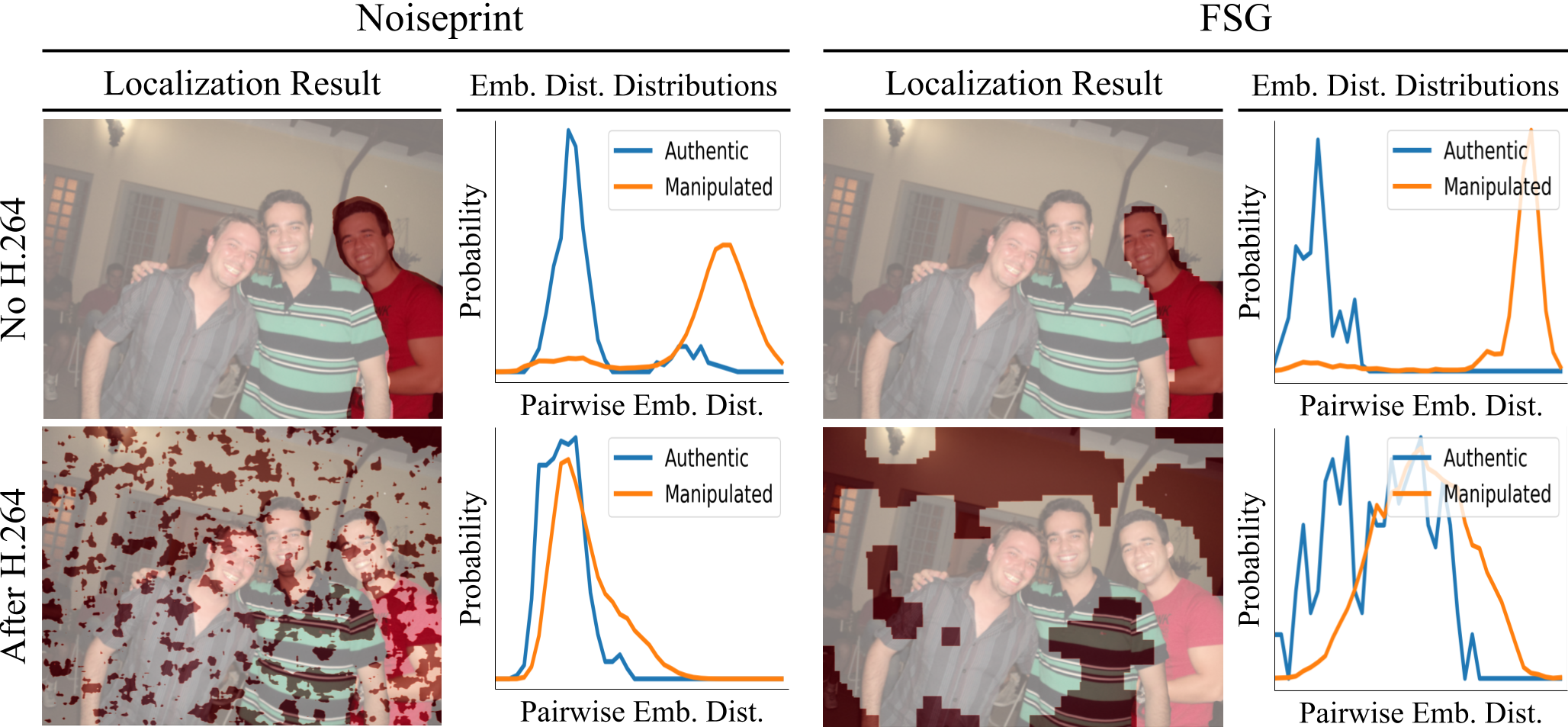}
	
	\vspace*{-0.5\baselineskip}
	
	
	\caption{Example from the Carvalho Dataset~\cite{Carvalho_DSO_1} illustrating the influence of video compression on forensic embeddings. 
	The distribution of pairwise forensic embedding distances across real and manipulated content as well as localization results are shown both before and after H.264 compression.  
%
	We can see that video compression makes manipulated regions become indistinguishable for both Noiseprint and FSG.}	
	
	\label{fig:effect_of_video_coding}
	
	\vspace*{-0.9\baselineskip}
	
\end{figure}

We can see the effects of video compression in Fig.~\ref{fig:effect_of_video_coding}.
This figure shows the distribution of pairwise differences between forensic embeddings in authentic and falsified regions obtained using Noiseprint~\cite{Noiseprint} and FSG~\cite{FSG} on the widely used Carvalho Dataset~\cite{Carvalho_DSO_1}.  The top row shows these distributions before H.264 compression, while the bottom row shows them after compression.
We can see
that video coding alters these distributions such that existing forensic networks can no longer
detect localized content forgeries.

\section{Proposed Approach}
\label{sec:proposed_approach}

\begin{figure*}
    \centering
    \includegraphics[width=0.96\linewidth,trim={0 5mm 0 5mm},clip]{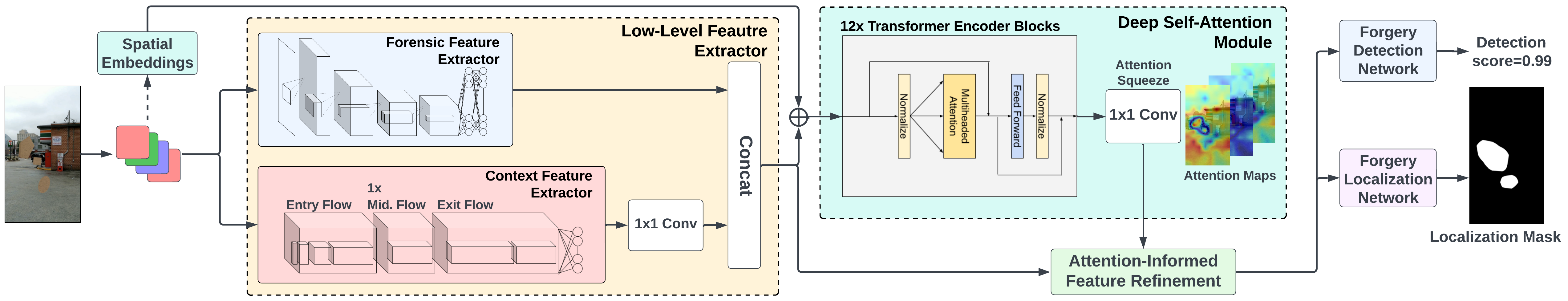}
    \caption{\label{fig:system_overview_diagram}
    Overview of our proposed VideoFACT network for video forgery detection and localization.  Our network extracts both forensic feature embeddings (FFE) and context feature embeddings (CFE) from local analysis blocks.  These embeddings are concatenated, then weighted by attention maps produced by a Deep Self-Attention Module (DSAM). The attention-refined joint feature embeddings are passed to two different subnetworks that produce frame-level detection decisions and pixel-level forgery masks.}
    \vspace*{-1.0\baselineskip}
\end{figure*}

An overview  VideoFACT's architecture is shown in Fig.~\ref{fig:system_overview_diagram}. 
VideoFACT is intentionally designed to overcome 
challenges posed by video compression, and to accurately detect and localize a variety of fake content in video.  To accomplish this, our network includes several important and novel aspects.

We utilize forensic feature embeddings specifically designed to capture traces in video.  These forensic feature embeddings are generic, 
i.e. they can be used to detect manipulations that were not explicitly seen during training.

We introduce the novel use of context embeddings to control for variation in forensic traces caused by video coding. 
These context embeddings capture relevant local information, such as scene content, texture, local compression strength, etc.  Later portions of our network are able to exploit 
this information by learning the distribution of forensic embeddings conditioned on local context. 
This allows our network to accurately distinguish between natural variation in authentic content and anomalies caused by forgery.


Additionally, we use a novel deep self-attention mechanism  to estimate the quality and relative importance of local forensic embeddings.  
This mechanism de-emphasizes embeddings from regions with low-quality traces, such as those strongly effected by compression.  Similarly, it emphasizes embeddings from regions that are important for forensic decision making, such as those with high-quality traces or potentially anomalous traces.
%
%
While some existing networks use attention, this is done to weight the relative importance of different forensic feature subsets within an embedding.  
This is significantly from self-attention, which captures how relevant forensic embeddings are with respect to one another.

\subsection{Low-Level Feature Extraction}
\label{subsec:low_level_feature_extraction}

VideoFACT consists of two low-level feature extractors working in tandem: the Forensic Feature Extractor (FFE) and the Context Feature Extractor (CFE).
Our method first divides a frame into non-overlapping analysis blocks of size $128 \times 128$ pixels, then passes each analysis block $b_k$ through both extractors to produce a forensic feature embedding $f_k$ and a context feature embedding $c_k$.

\subheader{Forensic Feature Embeddings}
We use Bayar and Stamm's forensic network $g(\cdot)$ with learned high-pass filters~\cite{MISLnet} to produce dedicated forensic feature embeddings.
Importantly, $g$ is first pre-trained with a cross-entropy loss to discriminate between a video's source camera using a separate camera model identification dataset~\cite{Video-ACID}.
After pre-training, the final classification layer is discarded.
This pre-training approach has been shown to be important for learning 
transferrable forensic embeddings~\cite{Noiseprint, EXIFnet, Splicebuster, Mayer_2018_IHMMS, FSM, FSG}.
We note that this branch is fixed during Training Stage 1, but is allowed to evolve during subsequent fine-tuning stages.


\subheader{Context Feature Embeddings}
The context feature extractor $h(\cdot)$ produces embeddings $c$ that provide information about a scene that contextualizes the forensic embeddings.
To extract context information, we implement $h$ using an Xception~\cite{Xception} network backbone that is modified to use only a single middle flow module, followed by a $1\times 1$ layer to reduce the feature embedding dimension.
Importantly,  $h$ is not trained until after $g$'s pre-training is complete, so that the resulting context embeddings can provide conditional information about the distribution of the  $f_k$'s. Denote $\theta$ as other layers in the network, this is equivalent to:
\begin{equation}
\min_{h, \theta} \mathcal{L}_T (h, g, \theta; b_1,.\hfil.\hfil.,b_k,.\hfil.\hfil.,b_N)
\end{equation}
where $\mathcal{L}_T$ is VideoFACT's total loss defined in~\eqref{LT}.

\subheader{Joint Feature Extraction}
After obtaining both embeddings, we produce the joint feature embeddings $x$ by concatenating $f$ and $c$, i.e.,
\vspace{-0.5em}
\begin{align}
        x_k = \text{concat}(f_k, c_k)
\end{align}
This process is repeated for every analysis block in the video frame to produce $N$ arrays of joint feature embedding with a dimension of 768.
By jointly analyzing forensic and context feature embeddings, later portions of our network can adjust how they interpret forensic features in a particular block to make better forensic decisions.
\subsection{Deep Self-Attention Module}
\label{subsec:deep_self_attention_module}

\vspace*{-0.2\baselineskip}

The sequence of joint embeddings is passed to the Deep Self-Attention Module (DSAM), which is designed to produce a series of $L$ different spatial attention maps that estimate the quality of contextualized forensic traces at a particular location. A vector of 1-D learnable positional embeddings is added to the joint embeddings before being passed through twelve Transformer encoder blocks stacked on top of one another. The output of these blocks is passed to an ``Attention Squeeze'' layer that consists of $L$, $1\times 1$ convolutional kernels. This ``squeezes'' down the high dimensional output to a series of $L$ different $M\times N$ spatial attention maps $m_l$. Each entry of the $L$
maps $m_l$ is the network's attention score for the corresponding joint feature embedding at that position.

\begin{figure}
    \centering
    \vspace*{-0.1\baselineskip}
    \includegraphics[width=0.9\linewidth]{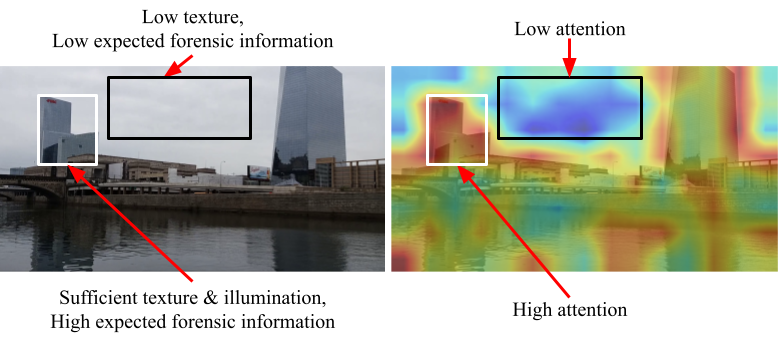}

    \vspace*{-0.7\baselineskip}

    \caption{Example showing the effect of our attention module.  The attention map produced for this frame gives large weights (shown in red) to regions with high quality forensic information - i.e. it has sufficient texture, illumination, no blurring, etc.  Regions with low-quality forensic information are given small weight (shown in blue).}

    \label{fig:spatial_attn_example}

    \vspace*{-0.8\baselineskip}

\end{figure}

Fig.~\ref{fig:spatial_attn_example} shows an example of a spatial attention map produced by our DSAM, as well as the corresponding the unaltered video frame.
We can see that VideoFACT attends to regions that contain high-quality forensic traces, i.e. regions with sufficient texture, illumination, etc.  By contrast, regions known to contain low-quality forensic information such as the sky are de-emphasized. Often, these regions will cause false alarms because their low-quality forensic traces will appear anomalous with respect to the regions with high-quality forensic traces.
By reducing the relative importance of the traces in these regions, our network can prevent false alarms.

\subsection{Attention-Informed Feature Refinement}
\label{subsec:attn_informed_feat_refine}

\vspace*{-0.2\baselineskip}

This module enables the network to attend to feature embeddings that contain higher-quality forensic information.
This is achieved by using $L$ attention maps, produced by DSAM,
to weight the low-level joint feature embeddings.
The joint feature embedding $x_k$ at the $k^{th}$ spatial location is weighted by the corresponding $k^{th}$ entry in the $l^{th}$ attention map.  This process is repeated for all $L$ spatial attention maps. Then the resulting features are element-wise summed to produce the attention-refined features $y$, such that:
\begin{equation}
        y_k = \sum\nolimits_{\ell = 1}^{L} x_k m_{k,l}
\end{equation}

\vspace*{-0.5\baselineskip}
\subsection{Detection and Localization}
\label{subsec:detection_and_localization}

VideoFACT produces two outputs: a frame-level detection score and a pixel-level localization mask. The localization mask is disregarded if the detection score 
indicates no forged content exists. Our network separately analyzes the attention-refined feature embeddings using two different subnetworks for localization and detection.

\subheader{Detection Network}
The detection network $p(\cdot)$ consists of two $1\times1$ Conv + ReLU layers for dimension reduction with 200, and 2 kernels, respectively, followed by a fully connected and a softmax layer to output two neurons, one corresponds to pristine and the other corresponds to being falsified. The detection loss for this subnetwork is the cross-entropy loss between the label and the predicted output:
\begin{align}
\mathcal{L}_{D} = - \sum\nolimits_{n=1}^{2} w_n log(p_n)
\end{align}
where $p_n$ is the output of $p(\cdot)$ and the $w_n$ is the one-hot vector indicate whether the video is manipulated.

\subheader{Localization Network}
The localization network $q(\cdot)$ is composed of four $1\times1$ Convolutional layers with 192, 96, 12, and 1 kernel, respectively. We use ReLU activation except in the last layer, which uses sigmoid activation to output the block-wise probability of being manipulated.

The localization loss is built on top of this subnetwork which we define as

\begin{align}
	\mathcal{L}_{L} =& - \sum_{k}\Bigl[\Bigl(\sum_{i,j \in \mathbbm{P}_k}\frac{M_{i,j}}{|\mathbbm{P}_k|}\Bigr) log(q_k) \nonumber \\
	&+ \Bigl(1 - \sum_{i,j \in \mathbbm{P}_k}\frac{M_{i,j}}{|\mathbbm{P}_k|}\Bigr)log(1 - q_k)\Bigr]
\end{align}

where $q_k$ is the prediction of $q(\cdot)$ for the corresponding block $b_k$. Because a block could contain a partial of manipulated region, $\mathbbm{P}_k$ is the set of pixel coordinates that belong to block $k$, and $M$ is the ground-truth binary mask.

\subheader{Localization Mask Generation}
During inference, we achieve the pixel-level predicted mask from the block-wise prediction. We first threshold the block-level prediction probabilities. Typically, there are two peaks in this histogram of block-level probabilities, one due to unaltered blocks and the other due to manipulated blocks. The threshold is chosen as the location of the first minima to the right of the first histogram peak (i.e. the one induced by unaltered blocks). After thresholding, we use the flood-fill morphological algorithm to remove holes from the localization mask. A final pixel-level mask is produced by scaling the block-level mask to the full video frame size using bilinear interpolation.

\subheader{Total Loss}
To train the entire network, we define the total loss as a linear combination of the detection and localization losses:
\begin{align}
        \mathcal{L}_T = \alpha  \mathcal{L}_D + (1 - \alpha)  \mathcal{L}_L
\label{LT}
\end{align}
where $\alpha \in (0,1)$ is the weight to balance the frame-level detection loss and the block-level localization loss.

\subsection{Multi-Stage Training Protocol}
\label{subsec:training_protocol}

Our proposed network training protocol consists of five stages.  
In the first three stages, the training datasets consist of videos falsified with manipulations that are successively more difficult to detect.  
This enables the network to progressively learn better features by refining those learned in the previous stage. In Stage 1, we use the Video Camera Model Splicing dataset (VCMS), which contains spliced content from other videos. In Stages 2 and 3,  we train the network with the Video Perceptually Visible Manipulation dataset (VPVM) and Video Perceptually Invisible Manipulation datasets (VPIM), respectively. As described in Section~\ref{sec:training_datasets}, these videos contains basic manipulations with different strengths. In Stage 4, we fine-tune the network using all three previous datasets simultaneously (VCMS, VPVM and VPIM). In Stage 5, we further fine-tune the model by incorporating three auxiliary datasets made from images in the Camera Model Identification Database~\cite{Bayar_2018_ICASSP, MISLnet} to 
diversify our content distribution.
These datasets are made using the same process used to create the three video datasets, resulting in one dataset with spliced content (ICMS), one with perceptually visible manipulations (IPVM), and one with perceptually invisible manipulations (IPIM).

\section{Video Forgery Datasets}
\label{sec:training_datasets}

Currently, there are no publicly available datasets of manipulated video large enough to train general content forgery detection and localization networks.  Similarly, there are almost no datasets suitable to evaluate such networks, with the notable exception of the Adobe VideoSham dataset, which was released in 2023~\cite{VideoSham}. To address this issue, we created a series of new video manipulation datasets for training and evaluating our network. These are divided into two subsets. Set A contains videos modified using standard manipulations, e.g. splicing and local editing, etc.  Set B contains ``In-the-Wild'' videos made using sophisticated editing operations such as inpainting, deepfakes, etc.  All datasets will be made publicly available upon publication of this paper.
\subsection{Set A: Standard Video Manipulations Datasets}

We made three  datasets by applying different sets of standard manipulations to videos from the Video-ACID~\cite{hosler2019video} dataset. All three datasets were made using a common procedure. First, we created binary ground-truth masks specifying the tamper regions for each video.  These tamper regions correspond to multiple randomly chosen shapes with random sizes, orientations, and placements within a frame. Fake videos were created by choosing a mask, then manipulating content within the tamper region.  
Original videos were retained to form the set of authentic videos.
All video frames 
of both sets
were re-encoded as H.264 videos using FFmpeg~\cite{FFmpeg} with 30 FPS and constant rate factor of 23.

Each dataset in Set A corresponds to a different manipulation type.  The \textbf{Video Camera Model Splicing (VCMS)} dataset  contains videos with content spliced in from other videos.  The \textbf{Video Perceptually Visible Manipulation (VPVM)} dataset contains content modified using common editing operations, e.g. contrast enhancement, smoothing, sharpening, blurring, etc. applied with strengths that can be visually detected. The \textbf{Video Perceptually Invisible Manipulation (VPIM)} dataset was made in a similar fashion to VPVM, but with much smaller manipulation strengths to create challenging forgeries. For each dataset, we made 3200 videos (96000 frames) for training, 520 videos (15600 frames) for validation, 280 videos (8400 frames) for testing.

\begin{table}[!t]
    \centering
    \setlength\extrarowheight{-2pt}
    \resizebox{0.9\linewidth}{!}{%
    \begin{tabular}{
        >{\centering\hspace{0pt}}m{0.100\linewidth}|
        >{\centering\hspace{0pt}}m{0.280\linewidth}
        >{\centering\hspace{0pt}}m{0.110\linewidth}
        >{\centering\hspace{0pt}}m{0.150\linewidth}
        >{\centering\hspace{0pt}}m{0.130\linewidth}
        >{\centering\hspace{0pt}}m{0.110\linewidth}
        >{\centering\arraybackslash\hspace{0pt}}m{0.110\linewidth}
    } 
        \hline
        \textbf{Stage} & \textbf{Dataset} & \textbf{Optimizer} & \textbf{Epochs} & \textbf{Initial Lr} & \textbf{Decay rate} & \textbf{Decay step} \\ 
        \hline
        1 & A & SGD & 6 & $1.0\mathrm{e}{-4}$ & 0.75 & 2 \\
        2 & B & SGD & 6 & $8.5\mathrm{e}{-5}$ & 0.85 & 2 \\
        3 & C & SGD & 23 & $8.5\mathrm{e}{-5}$ & 0.85 & 2 \\
        4 & A, B, C & SGD & 10 & $8.5\mathrm{e}{-5}$ & 0.85 & 2 \\
        5 & A, B, C, D, E, F & SGD & 9 & $5.0\mathrm{e}{-5}$ & 0.85 & 2 \\
        \hline
    \end{tabular}
    }
    \vspace*{-0.5\baselineskip}

    \caption{\label{table:training_stages_parameters} Training parameters for different training stages of our model. We denote: A=VCMS, B=VPVM, C=VPIM, D=ICMS, E=IPVM, F=IPIM.}

    \vspace*{-0.7\baselineskip}

\end{table}

\subsection{Set B: In-the-Wild Manipulated Datasets}

We use both publicly available datasets and datasets created by us to evaluate our network. These datasets contain advanced, challenging video forgeries with scene content that significantly differs from our training datasets.

\subheader{Public Datasets}
We evaluate on three publicly available datasets: VideoSham~\cite{VideoSham}, DeepfakeDetectionDataset (DFD)~\cite{DFD}, and FaceForensics++ (FF++)~\cite{FF++}. VideoSham contains high-quality videos manipulated by professional editors using multiple techniques. Because this work focuses exclusively on identifying fake content, we excluded videos with audio track or temporal manipulations. Both DFD and FF++ are popular deepfake benchmarking datasets that contain original videos and videos which were deepfaked using different algorithms (Face2Face~\cite{Face2Face}, FaceSwap~\cite{FaceSwap}, etc.).

\subheader{Datasets Created By Us}
We also created three additional datasets to evaluate our proposed approach more comprehensively, including: E2FGVI Inpainted Videos, FuseFormer Inpainted Videos, and DeepFaceLab Deepfake Videos. The two inpainting datasets were made by using SOTA AI-aided video inpainting algorithms, E2FGVI~\cite{E2FGVI} and FuseFormer~\cite{FuseFormer}, to remove objects specified by segmentation masks from videos in the Densely Annotated Video Segmentation (DAVIS) dataset~\cite{pont2017DAVIS}. Each datasets have authentic and manipulated subsets with each contain of 6208 frames from 90 videos.
Additionally, The DeepFaceLab Deepfake Videos dataset was made by applying DeepFaceLab~\cite{DeepFaceLab} - a popular, high quality deepfake algorithm - on a set of publicly available videos of celebrities downloaded from YouTube. This dataset consists of authentic and manipulated subsets with each having 300 frames from 10 videos.

\begin{table*}[!t]

	\vspace*{-0.5\baselineskip}

    \centering
    \setlength\extrarowheight{-1.0pt}
    \resizebox{0.600\linewidth}{!}{%
    \begin{tabular}{
        >{\hspace{0pt}}m{0.180\linewidth}
        >{\hspace{0pt}}m{0.055\linewidth}
        >{\hspace{0pt}}m{0.055\linewidth}
        >{\hspace{0pt}}m{0.055\linewidth}
        >{\hspace{0pt}}m{0.055\linewidth}
        >{\hspace{0pt}}m{0.000\linewidth}
        >{\hspace{0pt}}m{0.055\linewidth}
        >{\hspace{0pt}}m{0.055\linewidth}
        >{\hspace{0pt}}m{0.055\linewidth}
        >{\hspace{0pt}}m{0.055\linewidth}
        >{\hspace{0pt}}m{0.000\linewidth}
        >{\hspace{0pt}}m{0.055\linewidth}
        >{\hspace{0pt}}m{0.055\linewidth}
        >{\hspace{0pt}}m{0.055\linewidth}
        >{\hspace{0pt}}m{0.055\linewidth}
    }
        \hline

        \multicolumn{1}{>{\hspace{0pt}}m{0.180\linewidth}}{\multirow{2}{0.700\linewidth}{\hspace{0pt}\Centering{}\textbf{Method}}} &
        \multicolumn{4}{>{\Centering\hspace{-30pt}}m{0.250\linewidth}}{\textbf{VCMS}} &  &
        \multicolumn{4}{>{\Centering\hspace{-30pt}}m{0.250\linewidth}}{\textbf{VPVM}} &  &
        \multicolumn{4}{>{\Centering\hspace{-30pt}}m{0.250\linewidth}}{\textbf{VPIM}} \\

        \cline{2-5}\cline{7-10}\cline{12-15}

        \multicolumn{1}{>{\Centering\hspace{0pt}}m{0.040\linewidth}}{} &
        \textit{Det. mAP} & \textit{Det. ACC} & \textit{Loc. MCC} & \textit{Loc. F1} &  &
        \textit{Det. mAP} & \textit{Det. ACC} & \textit{Loc. MCC} & \textit{Loc. F1} &  &
        \textit{Det. mAP} & \textit{Det. ACC} & \textit{Loc. MCC} & \textit{Loc. F1} \\

        \hline

        FSG \cite{FSG}
        & 0.445 & 0.497 & 0.001 & 0.064 &
        & 0.431 & 0.480 & 0.004 & 0.067 &
        & 0.485 & 0.494 & 0.011 & 0.065 \\
        EXIFnet \cite{EXIFnet}
        & 0.610 & 0.502 & 0.208 & 0.230 &
        & 0.568 & 0.501 & 0.213 & 0.236 &
        & 0.509 & 0.500 & 0.026 & 0.124 \\
        Noiseprint \cite{Noiseprint}
        & 0.521 & 0.500 & 0.041 & 0.030 &
        & 0.495 & 0.500 & 0.012 & 0.013 &
        & 0.511 & 0.500 & 0.010 & 0.010 \\
        ManTra-Net \cite{ManTra-Net}
        & 0.451 & 0.500 & 0.079 & 0.114 &
        & 0.526 & 0.500 & 0.110 & 0.145 &
        & 0.513 & 0.500 & 0.025 & 0.064 \\
        MVSS-Net \cite{MVSS-Net}
        & 0.883 & 0.602 & \boldblue{0.545} & \boldblue{0.557} &
        & 0.644 & 0.529 & 0.267 & 0.279 &
        & 0.482 & 0.492 & 0.018 & 0.042 \\

        \hline

        E.ViT  \cite{EffViT}
        & 0.491 & 0.500 & \skipresult & \skipresult &
        & 0.507 & 0.507 & \skipresult & \skipresult &
        & 0.503 & 0.504 & \skipresult & \skipresult \\

        CCE.ViT  \cite{EffViT}
        & 0.472 & 0.461 & \skipresult & \skipresult &
        & 0.503 & 0.500 & \skipresult & \skipresult &
        & 0.509 & 0.507 & \skipresult & \skipresult \\

        CNN Ensemble \cite{CNNEnsemble}
        & 0.506 & 0.521 & \skipresult & \skipresult &
        & 0.495 & 0.493 & \skipresult & \skipresult &
        & 0.486 & 0.487 & \skipresult & \skipresult \\

        \hline

%
%

        \textbf{VideoFACT}
        & \boldblue{0.995} & \boldblue{0.987} & 0.530          & 0.526          &
        & \boldblue{0.980} & \boldblue{0.950} & \boldblue{0.676} & \boldblue{0.697} &
        & \boldblue{0.869} & \boldblue{0.797} & \boldblue{0.515} & \boldblue{0.547} \\

        \hline
    \end{tabular}
    }

	\vspace*{-0.5\baselineskip}

    \caption{\label{table:results_on_our_datasets} Frame-level detection and pixel-level localization performance on Set A Standard Video Manipulations datasets - VCMS, VPVM, VPIM.}

\end{table*}

\begin{table*}[t!]
    \centering
    \setlength\extrarowheight{-1.0pt}
    \resizebox{1.0\linewidth}{!}{%
    \begin{tabular}{
        >{\hspace{0pt}}m{0.180\linewidth}
        >{\hspace{0pt}}m{0.055\linewidth}
        >{\hspace{0pt}}m{0.055\linewidth}
        >{\hspace{0pt}}m{0.055\linewidth}
        >{\hspace{0pt}}m{0.055\linewidth}
        >{\hspace{0pt}}m{0.000\linewidth}
        >{\hspace{0pt}}m{0.055\linewidth}
        >{\hspace{0pt}}m{0.055\linewidth}
        >{\hspace{0pt}}m{0.055\linewidth}
        >{\hspace{0pt}}m{0.055\linewidth}
        >{\hspace{0pt}}m{0.000\linewidth}
        >{\hspace{0pt}}m{0.055\linewidth}
        >{\hspace{0pt}}m{0.055\linewidth}
        >{\hspace{0pt}}m{0.055\linewidth}
        >{\hspace{0pt}}m{0.055\linewidth}
        >{\hspace{0pt}}m{0.000\linewidth}
        >{\hspace{0pt}}m{0.055\linewidth}
        >{\hspace{0pt}}m{0.055\linewidth}
        >{\hspace{0pt}}m{0.055\linewidth}
        >{\hspace{0pt}}m{0.055\linewidth}
        >{\hspace{0pt}}m{0.000\linewidth}
        >{\hspace{0pt}}m{0.055\linewidth}
        >{\hspace{0pt}}m{0.055\linewidth}
        >{\hspace{0pt}}m{0.055\linewidth}
        >{\hspace{0pt}}m{0.055\linewidth}
        >{\hspace{0pt}}m{0.000\linewidth}
        >{\hspace{0pt}}m{0.055\linewidth}
        >{\hspace{0pt}}m{0.055\linewidth}
        >{\hspace{0pt}}m{0.055\linewidth}
        >{\hspace{0pt}}m{0.055\linewidth}
    }
        \hline

        \multicolumn{1}{>{\hspace{0pt}}m{0.180\linewidth}}{\multirow{2}{0.700\linewidth}{\hspace{0pt}\Centering{}\textbf{Method}}} &
        \multicolumn{4}{>{\Centering\hspace{-20pt}}m{0.250\linewidth}}{\textbf{E2FGVI Inpainted Videos}} &  &
        \multicolumn{4}{>{\Centering\hspace{-20pt}}m{0.250\linewidth}}{\textbf{FuseFormer Inpainted Videos}} &  &
        \multicolumn{4}{>{\Centering\hspace{-20pt}}m{0.250\linewidth}}{\textbf{VideoSham}~\cite{VideoSham}} &  &
        \multicolumn{4}{>{\Centering\hspace{-10pt}}m{0.250\linewidth}}{\textbf{DeepFaceLab Deepfake Videos}} &  &
        \multicolumn{4}{>{\Centering\hspace{-20pt}}m{0.250\linewidth}}{\textbf{DFD}~\cite{DFD}} &  &
        \multicolumn{4}{>{\Centering\hspace{-20pt}}m{0.250\linewidth}}{\textbf{FF++}~\cite{FF++}} \\

        \cline{2-5}\cline{7-10}\cline{12-15}\cline{17-20}\cline{22-25}\cline{27-30}

        \multicolumn{1}{>{\Centering\hspace{0pt}}m{0.040\linewidth}}{} &
        \textit{Det. mAP} & \textit{Det. ACC} & \textit{Loc. MCC} & \textit{Loc. F1} &  &
        \textit{Det. mAP} & \textit{Det. ACC} & \textit{Loc. MCC} & \textit{Loc. F1} &  &
        \textit{Det. mAP} & \textit{Det. ACC} & \textit{Loc. MCC} & \textit{Loc. F1} &  &
        \textit{Det. mAP} & \textit{Det. ACC} & \textit{Loc. MCC} & \textit{Loc. F1} &  &
        \textit{Det. mAP} & \textit{Det. ACC} & \textit{Loc. MCC} & \textit{Loc. F1} &  &
        \textit{Det. mAP} & \textit{Det. ACC} & \textit{Loc. MCC} & \textit{Loc. F1} \\

        \hline

        FSG \cite{FSG}
        & 0.386 & 0.452 & 0.208 & 0.302 &
        & 0.351 & 0.484 & \boldblue{0.241} & \boldblue{0.290} &
        & 0.596 & 0.538 & 0.162 & 0.246 &
        & 0.450 & 0.515 & 0.204 & 0.137 &
        & 0.449 & 0.325 & 0.097 & 0.043 &
        & 0.509 & \boldblue{0.519} & 0.144 & 0.113 \\

        EXIFnet \cite{EXIFnet}
        & 0.635 & 0.501 & 0.160 & 0.244 &
        & 0.506 & 0.507 & 0.146 & 0.225 &
        & 0.584 & 0.555 & 0.148 & 0.246 &
        & 0.447 & 0.492 & 0.180 & 0.133 &
        & 0.489 & 0.258 & 0.095 & 0.051 &
        & 0.487 & \boldblue{0.519} & 0.141 & 0.073 \\

        Noiseprint \cite{Noiseprint}
        & 0.601 & 0.500 & 0.091 & 0.232 &
        & 0.471 & 0.500 & 0.001 & 0.200 &
        & 0.422 & 0.447 & 0.034 & 0.206 &
        & 0.591 & 0.500 & 0.010 & 0.062 &
        & 0.489 & 0.252 & 0.000 & 0.021 &
        & 0.486 & 0.518 & 0.000 & 0.066 \\

        ManTra-Net \cite{ManTra-Net}
        & 0.499 & 0.500 & 0.009 & 0.055 &
        & 0.613 & 0.500 & 0.031 & 0.204 &
        & 0.551 & 0.553 & 0.009 & 0.058 &
        & 0.450 & 0.500 & 0.004 & 0.042 &
        & 0.476 & 0.253 & 0.017 & 0.025 &
        & 0.504 & 0.514 & 0.070 & 0.091 \\

        MVSS-Net \cite{MVSS-Net}
        & 0.341 & 0.435 & 0.058 & 0.227 &
        & 0.230 & 0.359 & 0.029 & 0.206 &
        & 0.595 & 0.449 & 0.142 & 0.096 &
        & 0.464 & 0.498 & 0.199 & 0.189 &
        & \boldblue{0.513} & \boldblue{0.532} & \boldblue{0.152} & \boldblue{0.108} &
        & 0.499 & 0.487 & 0.133 & 0.164 \\

        \hline

        \textbf{VideoFACT}
        & \boldblue{0.782} & \boldblue{0.687} & \boldblue{0.225} & \boldblue{0.309} &
        & \boldblue{0.652} & \boldblue{0.527} & 0.118 & 0.237 &
        & \boldblue{0.691} & \boldblue{0.656} & \boldblue{0.193} & \boldblue{0.312} &
        & \boldblue{0.666} & \boldblue{0.648} & \boldblue{0.415} & \boldblue{0.410} &
        & 0.468 & 0.444 & 0.081 & 0.077 &
        & \boldblue{0.529} & \boldblue{0.519} & \boldblue{0.160} & \boldblue{0.167} \\

        \hline

        \textbf{VideoFACT-FT}
        & \boldred{0.908} & \boldred{0.820} & \boldred{0.411} & \boldred{0.445} &
        & \boldred{0.948} & \boldred{0.846} & \boldred{0.361} & \boldred{0.411} &
        & \skipresult   & \skipresult   & \skipresult   & \skipresult   &
        & \boldred{0.988} & \boldred{0.922} & \boldred{0.745} & \boldred{0.732} &
        & \boldred{0.937} & \boldred{0.804} & \boldred{0.536} & \boldred{0.490} &
        & \boldred{0.916} & \boldred{0.837} & \boldred{0.661} & \boldred{0.645} \\

        \hline

        E.ViT  \cite{EffViT}
        & 0.557 & 0.528 & \skipresult & \skipresult &
        & 0.535 & 0.509 & \skipresult & \skipresult &
        & 0.497 & 0.499 & \skipresult & \skipresult &
        & 0.896 & 0.805 & \skipresult & \skipresult &
        & 0.811 & 0.737 & \skipresult & \skipresult &  
        & 0.764 & 0.676 & \skipresult & \skipresult \\ 

        CCE.ViT  \cite{EffViT}
        & 0.564 & 0.550 & \skipresult & \skipresult &
        & 0.653 & 0.586 & \skipresult & \skipresult &
        & 0.489 & 0.493 & \skipresult & \skipresult &
        & 0.962 & 0.837 & \skipresult & \skipresult &
        & 0.816 & 0.761 & \skipresult & \skipresult &  
        & 0.796 & 0.719 & \skipresult & \skipresult \\ 

        CNN Ensemble \cite{CNNEnsemble}
        & 0.595 & 0.556 & \skipresult & \skipresult &
        & 0.579 & 0.543 & \skipresult & \skipresult &
        & 0.551 & 0.552 & \skipresult & \skipresult &
        & 0.936 & 0.857 & \skipresult & \skipresult &
        & 0.829 & 0.745 & \skipresult & \skipresult &  
        & 0.713 & 0.672 & \skipresult & \skipresult \\ 

        \hline

%
%

%

        \hline
    \end{tabular}
    }

	\vspace*{-0.5\baselineskip}

    \caption{\label{table:results_on_itw_datasets} Frame-level detection and pixel-level localization performance on Set B's ``In-the-Wild'' datasets - E2FGVI Inpainted Videos, FuseFormer Inpainted Videos, VideoSham~\cite{VideoSham}, DeepFaceLab Deepfake Videos, DFD~\cite{DFD}, and FF++~\cite{FF++}.}

    \vspace*{-0.5\baselineskip}

\end{table*}

\section{Experiments}
\label{sec:experiments}

\subheader{Training Implementation} We implemented our network using PyTorch and trained it using an NVIDIA RTX 3090. The network input size is $1080 \times 1920$ pixels (a single 1080p frame). We first pre-trained the FFE on the Video-ACID dataset using the Stochastic Gradient Descent (SGD) optimizer with an initial learning rate of $1.0\mathrm{e}{-3}$, momentum of $0.95$, and an exponential decay-rate of $0.5$ for every $2$ epochs. After pretraining the FFE, we trained the entire network following the five stages described in Section \ref{subsec:training_protocol} with different training parameters shown in Table \ref{table:training_stages_parameters}. Throughout the stages, we set $\alpha = 0.4$.

\subheader{Evaluation Datasets} 
We evaluated the frame-level performance of our proposed network and competing networks on the nine datasets described in Section~\ref{sec:training_datasets}.

\subheader{Evaluation Metrics} 
For frame-level manipulation detection, we report the mean average precision (mAP) for each dataset. Also, we provide the average accuracy (ACC) per datasets using a unified threshold of 0.5 to reflect real-world's performance. For forgery localization (i.e. pixel-level manipulation detection) we use F1 and MCC scores to evaluate the correlation between the ground-truth and predicted masks, which are binarized with a threshold of~0.5.












\begin{figure*}[!t]

	\vspace*{-0.5\baselineskip}

    \centering
    \setlength{\fboxsep}{0pt}
    \begin{minipage}[t]{1\textwidth}
        \makebox[0.110\textwidth][s]{    }
        \makebox[0.120\textwidth]{\smaller VCMS}
        \makebox[0.120\textwidth]{\smaller VPVM}
        \makebox[0.120\textwidth]{\smaller VPIM}
        \makebox[0.120\textwidth]{\smaller Deepfake Video}
        \makebox[0.120\textwidth]{\smaller Inpainted Video}
        \makebox[0.120\textwidth]{\smaller VideoSham}
        \smallskip
    \end{minipage}

	\vspace*{-0.1\baselineskip}

    \begin{minipage}[t]{1\textwidth}
        \makebox[0.110\textwidth][r]{\raisebox{15pt}{\smaller Frame\hspace{6pt}}}
        \fbox{\includegraphics[width=0.120\textwidth]{{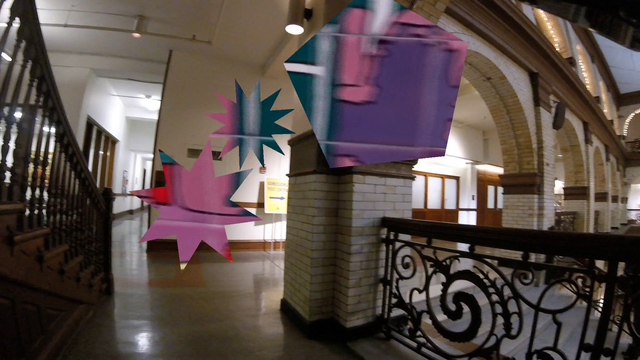}}}
        \fbox{\includegraphics[width=0.120\textwidth]{{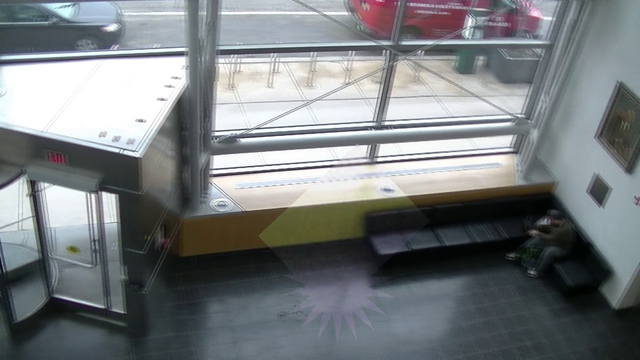}}}
        \fbox{\includegraphics[width=0.120\textwidth]{{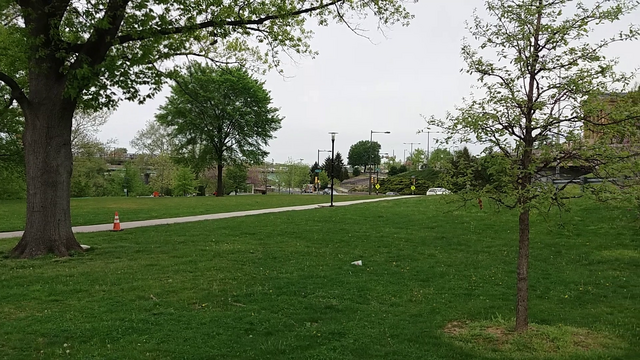}}}
        \fbox{\includegraphics[width=0.120\textwidth]{{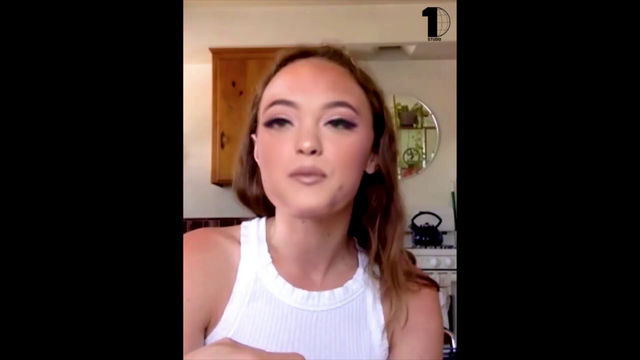}}}
        \fbox{\includegraphics[width=0.120\textwidth]{{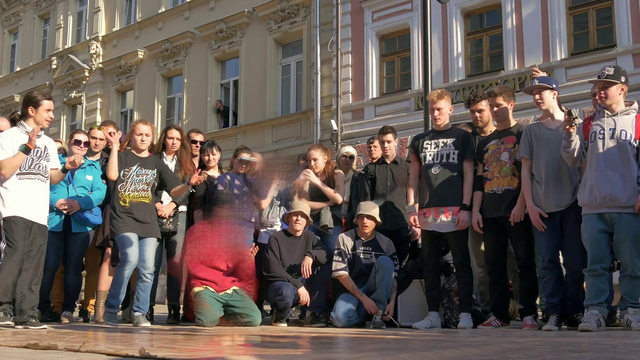}}}
        \fbox{\includegraphics[width=0.120\textwidth]{{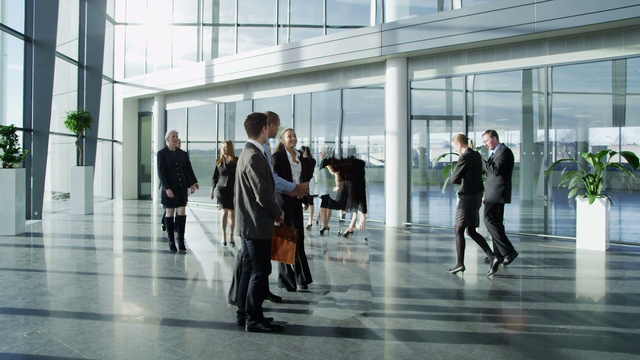}}}
        \smallskip
    \end{minipage}

	\vspace*{-0.1\baselineskip}

    \begin{minipage}[t]{1\textwidth}
        \makebox[0.110\textwidth][r]{\raisebox{15pt}{\smaller Ground-truth Mask\hspace{6pt}}}
        \fbox{\includegraphics[width=0.120\textwidth]{{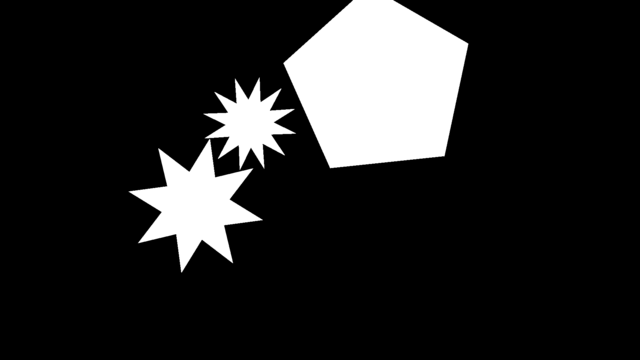}}}
        \fbox{\includegraphics[width=0.120\textwidth]{{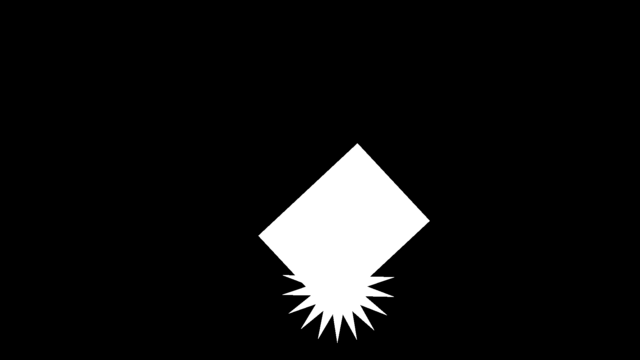}}}
        \fbox{\includegraphics[width=0.120\textwidth]{{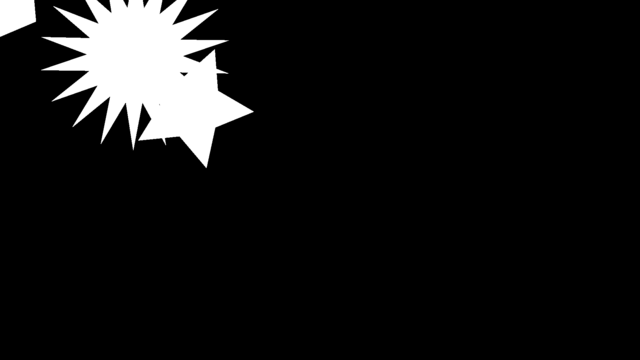}}}
        \fbox{\includegraphics[width=0.120\textwidth]{{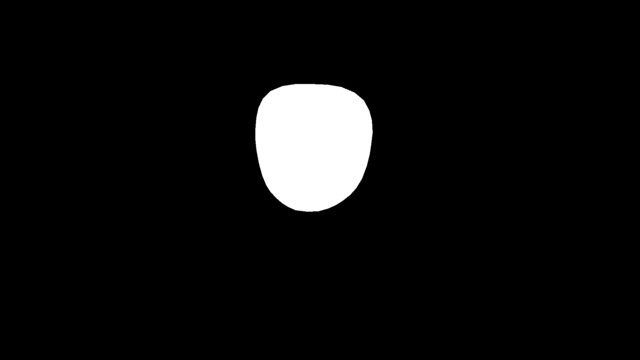}}}
        \fbox{\includegraphics[width=0.120\textwidth]{{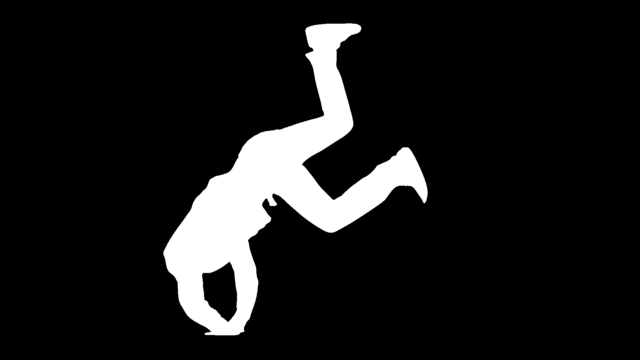}}}
        \fbox{\includegraphics[width=0.120\textwidth]{{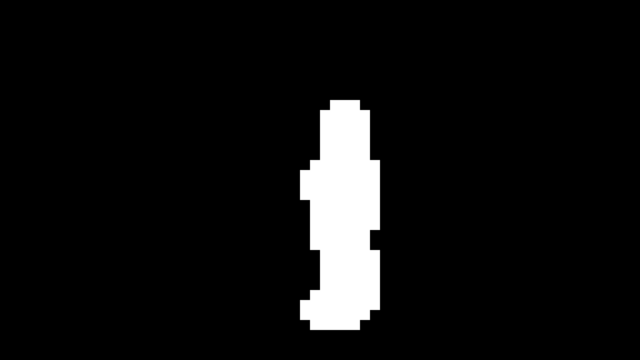}}}
        \smallskip
    \end{minipage}

	\vspace*{-0.1\baselineskip}

    \begin{minipage}[t]{1\textwidth}
        \makebox[0.110\textwidth][r]{\raisebox{15pt}{\smaller VideoFACT (Ours)\hspace{6pt}}}
        \fbox{\includegraphics[width=0.120\textwidth]{{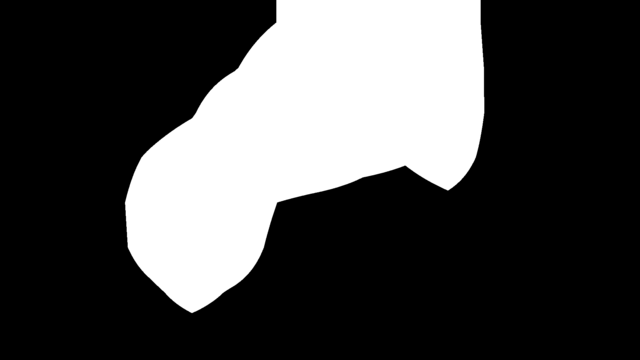}}}
        \fbox{\includegraphics[width=0.120\textwidth]{{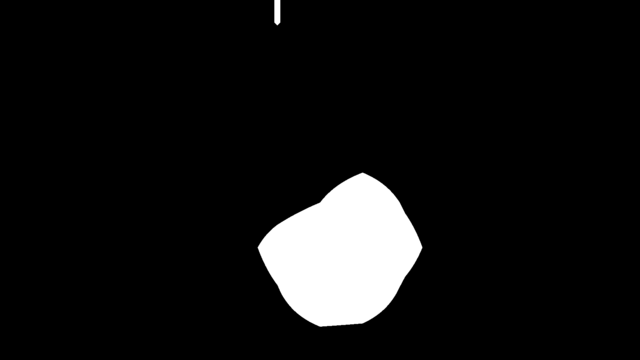}}}
        \fbox{\includegraphics[width=0.120\textwidth]{{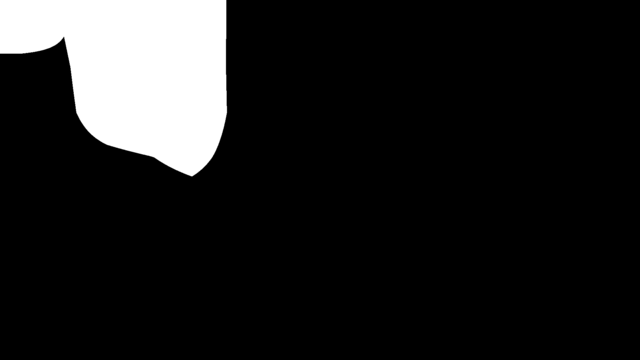}}}
        \fbox{\includegraphics[width=0.120\textwidth]{{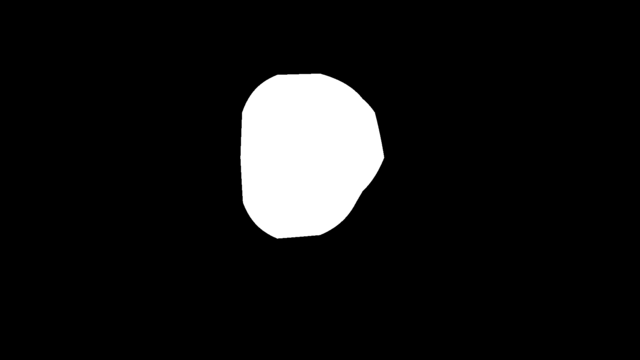}}}
        \fbox{\includegraphics[width=0.120\textwidth]{{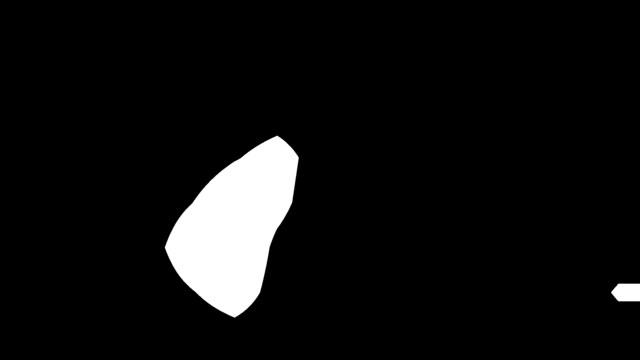}}}
        \fbox{\includegraphics[width=0.120\textwidth]{{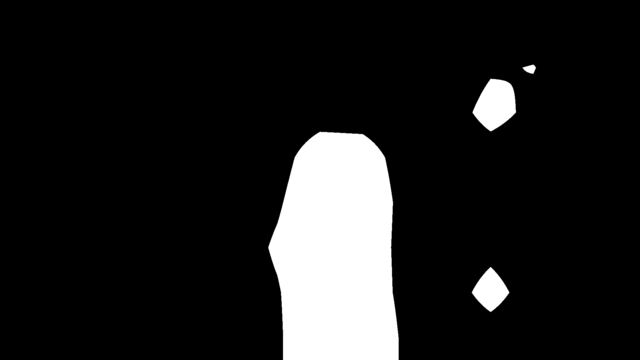}}}
        \smallskip
    \end{minipage}

	\vspace*{-0.1\baselineskip}

    \begin{minipage}[t]{1\textwidth}
        \makebox[0.110\textwidth][r]{\raisebox{15pt}{\smaller FSG\hspace{6pt}}}
        \fbox{\includegraphics[width=0.120\textwidth]{{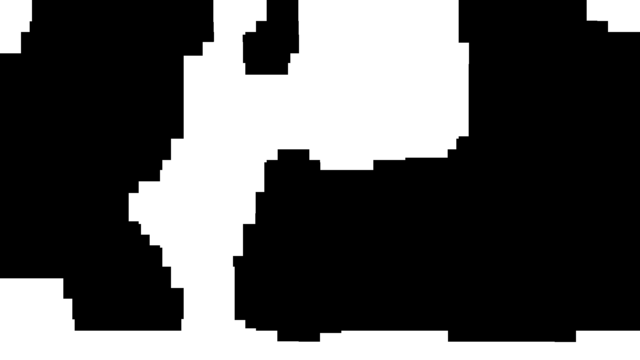}}}
        \fbox{\includegraphics[width=0.120\textwidth]{{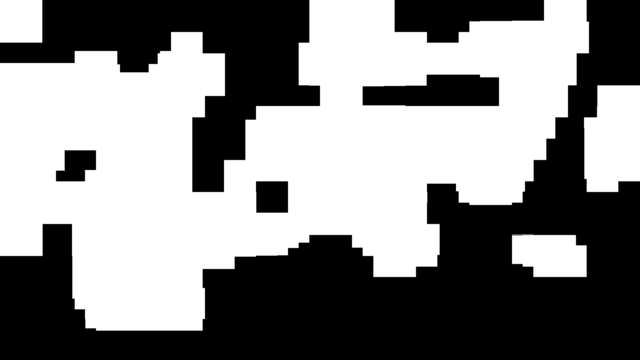}}}
        \fbox{\includegraphics[width=0.120\textwidth]{{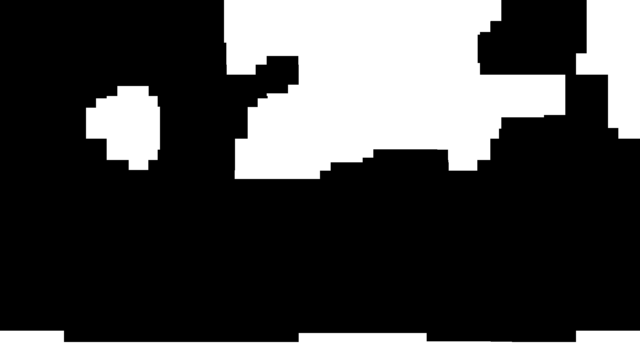}}}
        \fbox{\includegraphics[width=0.120\textwidth]{{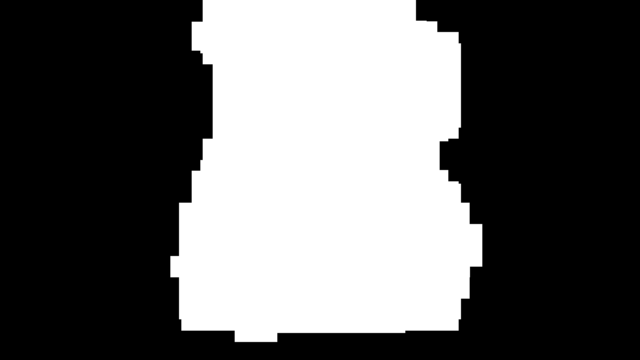}}}
        \fbox{\includegraphics[width=0.120\textwidth]{{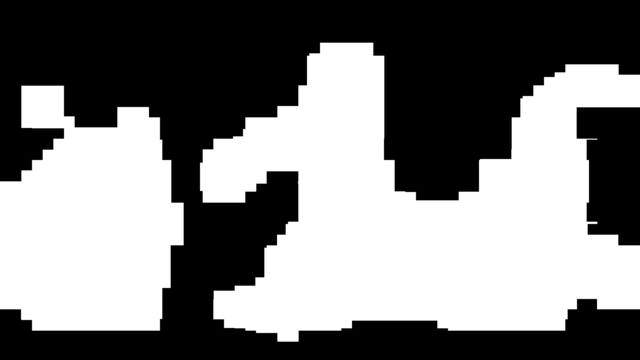}}}
        \fbox{\includegraphics[width=0.120\textwidth]{{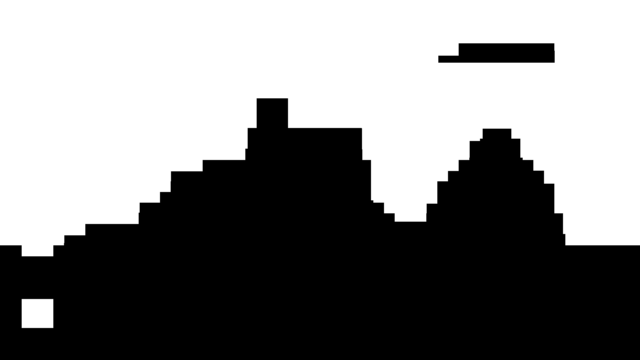}}}
        \smallskip
    \end{minipage}

	\vspace*{-0.1\baselineskip}

    \begin{minipage}[t]{1\textwidth}
        \makebox[0.110\textwidth][r]{\raisebox{15pt}{\smaller EXIFnet\hspace{6pt}}}
        \fbox{\includegraphics[width=0.120\textwidth]{{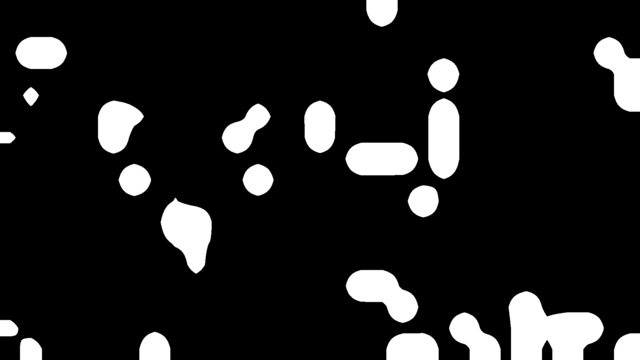}}}
        \fbox{\includegraphics[width=0.120\textwidth]{{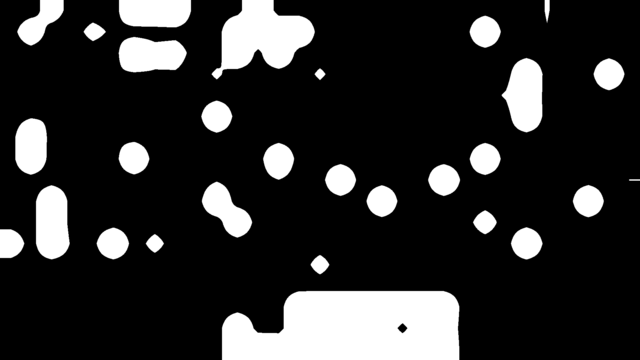}}}
        \fbox{\includegraphics[width=0.120\textwidth]{{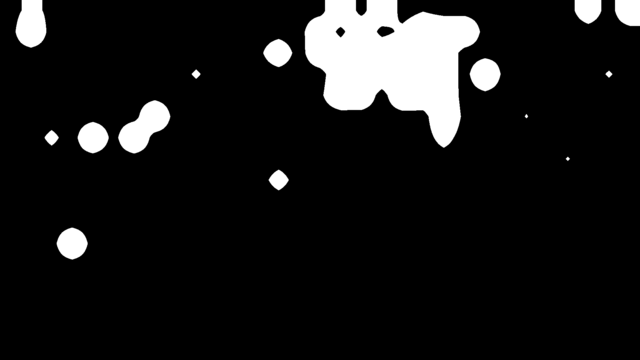}}}
        \fbox{\includegraphics[width=0.120\textwidth]{{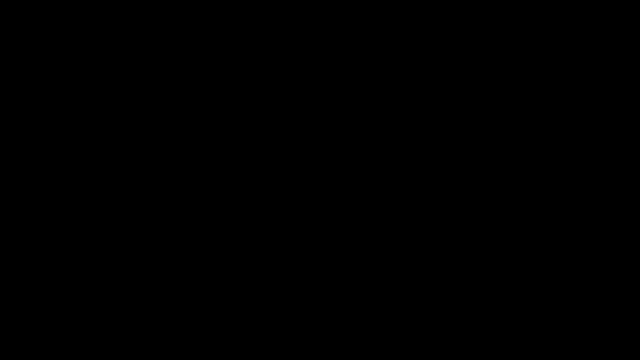}}}
        \fbox{\includegraphics[width=0.120\textwidth]{{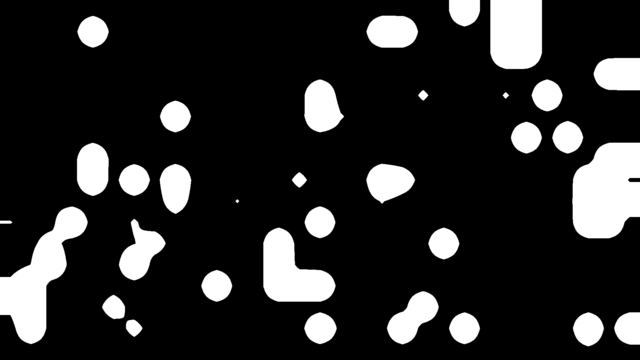}}}
        \fbox{\includegraphics[width=0.120\textwidth]{{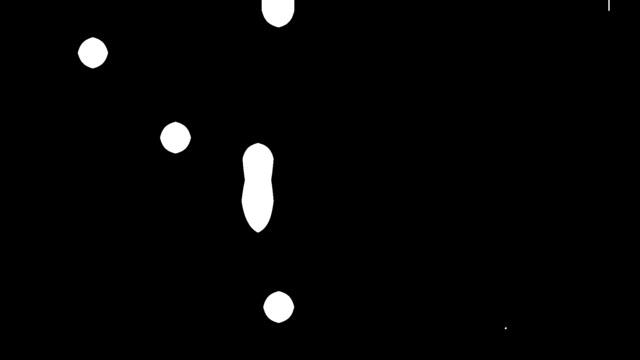}}}
        \smallskip
    \end{minipage}

	\vspace*{-0.1\baselineskip}

    \begin{minipage}[t]{1\textwidth}
        \makebox[0.110\textwidth][r]{\raisebox{15pt}{\smaller Noiseprint\hspace{6pt}}}
        \fbox{\includegraphics[width=0.120\textwidth]{{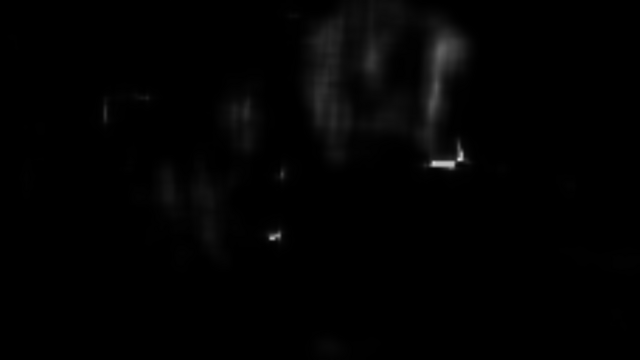}}}
        \fbox{\includegraphics[width=0.120\textwidth]{{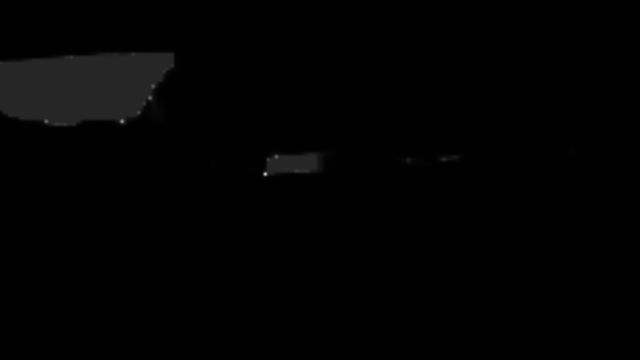}}}
        \fbox{\includegraphics[width=0.120\textwidth]{{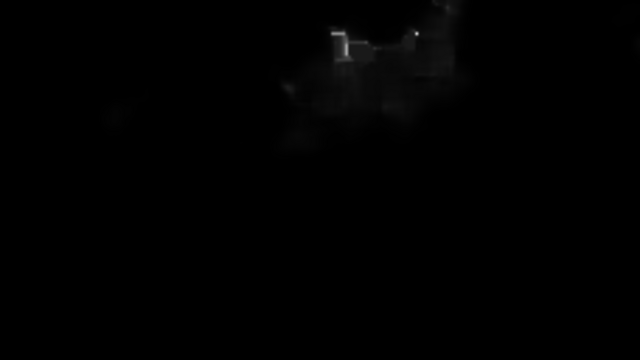}}}
        \fbox{\includegraphics[width=0.120\textwidth]{{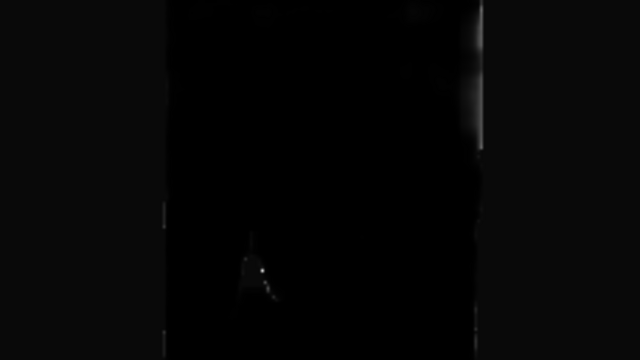}}}
        \fbox{\includegraphics[width=0.120\textwidth]{{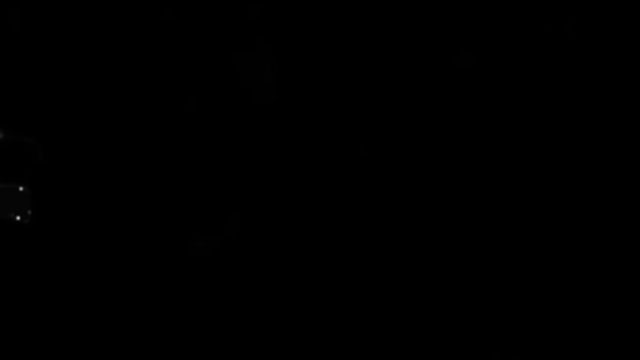}}}
        \fbox{\includegraphics[width=0.120\textwidth]{{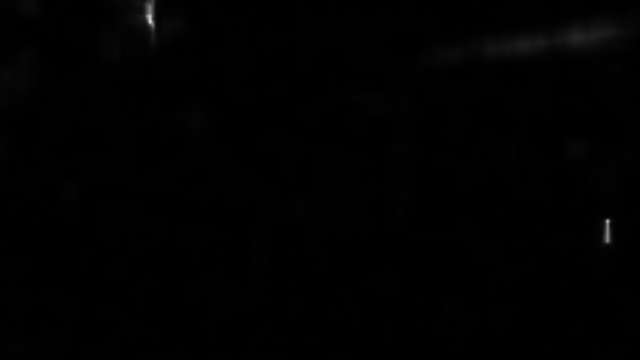}}}
        \smallskip
    \end{minipage}

	\vspace*{-0.1\baselineskip}

    \begin{minipage}[t]{1\textwidth}
        \makebox[0.110\textwidth][r]{\raisebox{15pt}{\smaller ManTra-Net\hspace{6pt}}}
        \fbox{\includegraphics[width=0.120\textwidth]{{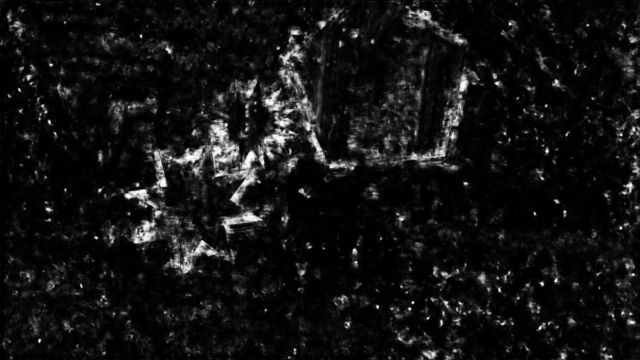}}}
        \fbox{\includegraphics[width=0.120\textwidth]{{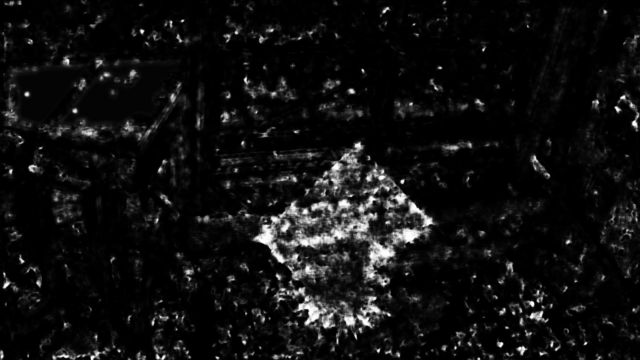}}}
        \fbox{\includegraphics[width=0.120\textwidth]{{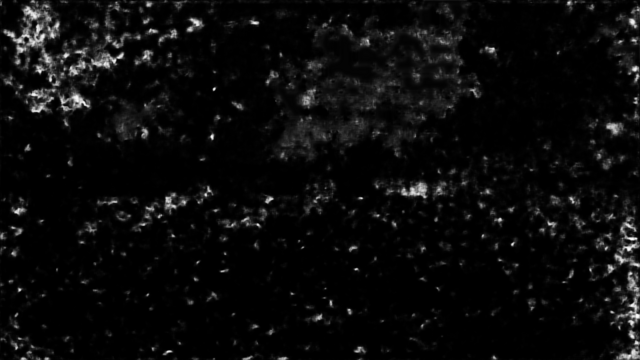}}}
        \fbox{\includegraphics[width=0.120\textwidth]{{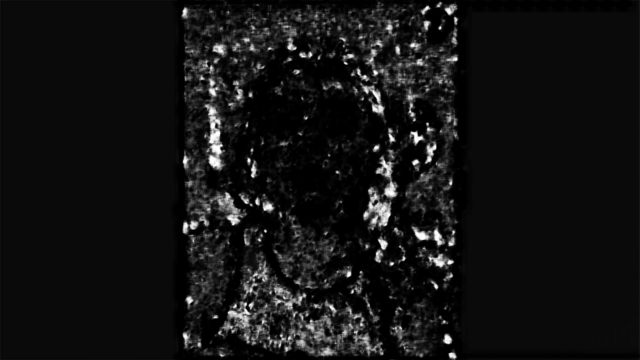}}}
        \fbox{\includegraphics[width=0.120\textwidth]{{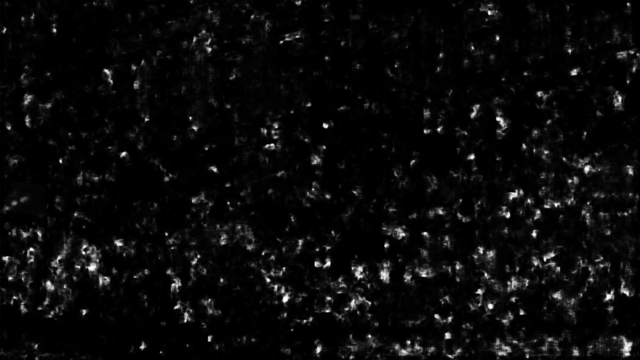}}}
        \fbox{\includegraphics[width=0.120\textwidth]{{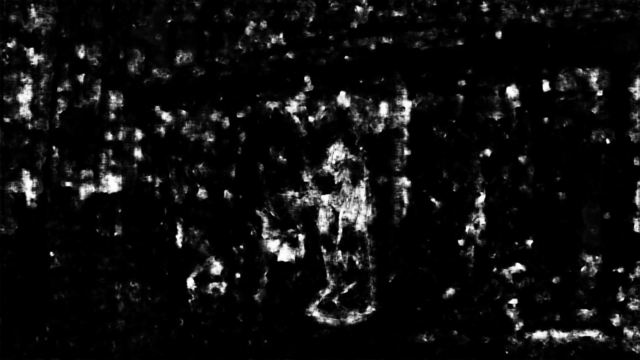}}}
        \smallskip
    \end{minipage}

	\vspace*{-0.1\baselineskip}

    \begin{minipage}[t]{1\textwidth}
        \makebox[0.110\textwidth][r]{\raisebox{15pt}{\smaller MVSS-Net\hspace{6pt}}}
        \fbox{\includegraphics[width=0.120\textwidth]{{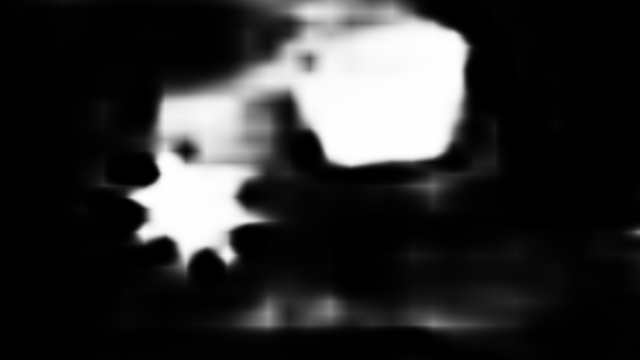}}}
        \fbox{\includegraphics[width=0.120\textwidth]{{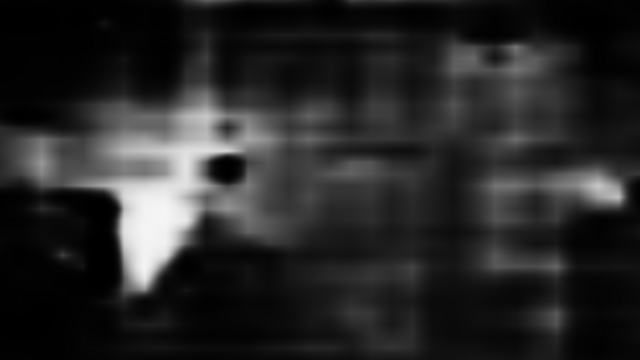}}}
        \fbox{\includegraphics[width=0.120\textwidth]{{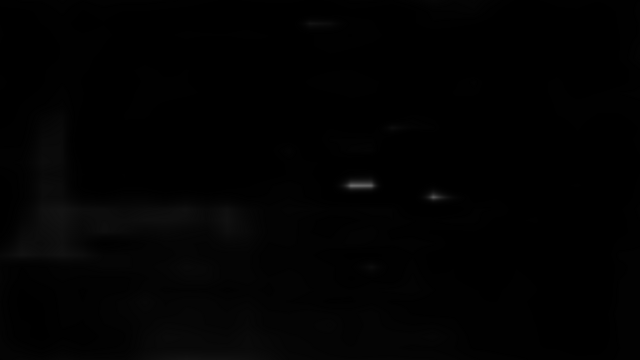}}}
        \fbox{\includegraphics[width=0.120\textwidth]{{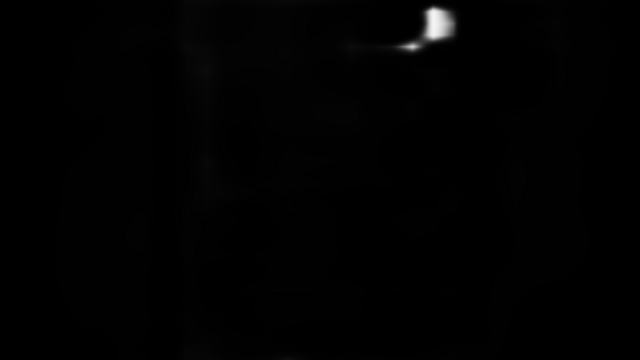}}}
        \fbox{\includegraphics[width=0.120\textwidth]{{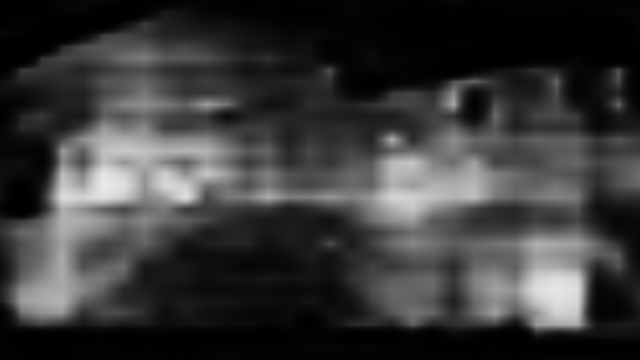}}}
        \fbox{\includegraphics[width=0.120\textwidth]{{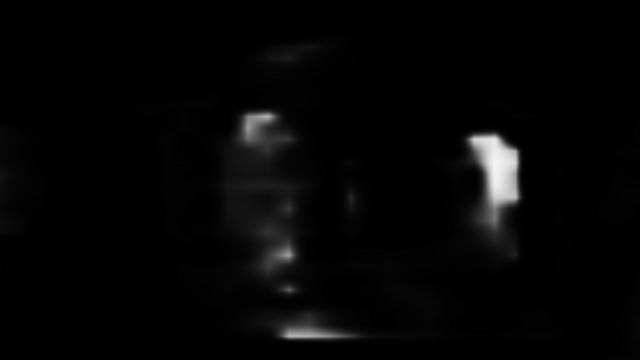}}}
        \smallskip
    \end{minipage}

	\vspace*{-0.5\baselineskip}

    \caption{\label{fig:localization_results} This figure shows localization results from our proposed network as well as FSG \cite{FSG}, EXIFnet \cite{EXIFnet}, Noiseprint \cite{Noiseprint}, ManTra-Net \cite{ManTra-Net}, and MVSS-Net \cite{MVSS-Net} on six different datasets, VCMS, VPVM, VPIM, Deepfake Video, Inpainted Video and VideoSham \cite{VideoSham}. Our proposed network correctly identifies the manipulated area in videos
    falsified using a wide variety of forgery operations. We note that we do not provide localization results for deepfake detectors because these algorithms only perform detection.
    }

    \vspace*{-0.5\baselineskip}

\end{figure*}

\subsection{Detection and Localization Performance}
\label{subsec:comparison_with_prior_works}

We compared the performance of VideoFACT to several 
state-of-the-art (SOTA)
image forensic networks including Forensic Similarity Graphs (FSG) \cite{FSG}, EXIFnet \cite{EXIFnet}, Noiseprint \cite{Noiseprint}, ManTra-Net \cite{ManTra-Net}, and MVSS-Net \cite{MVSS-Net}, representing a broad spectrum of successful techniques for performing general content forgery detection and localization in images.
Frame-level detection scores are calculated for Noiseprint and ManTra-Net by computing the average normalized per-pixel detection probability.
Additionally, we benchmarked against three SOTA deepfake detectors: Efficient ViT (E.ViT)~\cite{EffViT}, Convolutional Cross Efficient ViT (CCE.ViT)~\cite{EffViT}, and CNN Ensemble~\cite{CNNEnsemble}.
Since these only perform detection, no localization results are presented.

\subheader{Set A: Standard Video Manipulations}
Table \ref{table:results_on_our_datasets} shows the performance of both our proposed network and competing networks on the three Standard Video Manipulations datasets in Set A. These results show that VideoFACT achieves the best performance by a large margin on these datasets. The only exception is for the VCMS dataset, where MVSS-Net's localization performance is slightly better than ours, though still comparable. Except for this case, existing networks largely do no better than a random guess (i.e. $\mbox{mAP}=0.5$ and $\mbox{MCC}=0$).  This phenomenon can be clearly seen in both Table~\ref{table:results_on_our_datasets} and the qualitative results presented in Fig~\ref{fig:localization_results}.  These results reinforce similar findings reported in the VideoSham paper~\cite{VideoSham}, i.e. existing forensic networks
fail when analyzing video forgeries, 
and deepfake detectors cannot transfer to forgeries other than deepfakes.

\subheader{Set B: ``In-the-Wild'' Datasets}
Table \ref{table:results_on_itw_datasets} shows the performance of  both VideoFACT and competing networks on the six ``In-the-Wild'' datasets, which contains complex and challenging forgeries.
From these results, we see that VideoFACT outperforms existing general  forgery detectors and localizers. Additionally, VideoFACT can transfer to advanced manipulations
by finetuning on a small amount of data.

Specifically, on the two Inpainted Videos datasets, VideoFACT outperforms  existing  forensic networks and deepfake detectors by a large margin. For example, we achieved $\mbox{mAP}=0.782$ on E2FGVI  and $0.652$ on FuseFormer Inpainted Videos.
Additionally, we obtain strong performance and outperform competing networks on VideoSham, which contains four difficult forgery types.
Here, we achieved $\mbox{mAP}=0.691$ for detection and $\mbox{F1}=0.312$  for localization.
On the three deepfake datasets, we did not outperform deepfake detectors.
This is expected because
they
leverage significant problem-specific information, while VideoFACT does not.
Furthermore, the traces left by deepfakes and the content of these datasets differs significantly from our training data.
However, through finetuning, we will show that VideoFACT can
achieve strong performance on deepfakes.




Next, our experiments show that VideoFACT can transfer to advanced manipulations by finetuning using only a small portion of data.  Here, VideoFACT-FT denotes a version of VideoFACT  finetuned using only 10\% of each relevant dataset.
By finetuning on the Inpainted Videos datasets, we achieved very strong performance: $\mbox{mAP}=0.908$ on E2FGVI and $\mbox{mAP}=0.948$ on FuseFormer.
%
%
Finetuning for deepfake detection using DFD and FF++ data, we were able to achieve strong performance comparable with SOTA deepfake detectors.  Notably, we achieved $\mbox{mAP}=0.937$ on DFD, $0.916$ on FF++, and $0.988$ on our DeepFaceLab dataset.
Here, we report frame-level deepfake detection results for fair comparison between algorithms.
We note that no finetuning experiments were performed on VideoSham because it has only an evaluation set and no training set.

\begin{figure}[!htb]

    \centering
    \setlength{\fboxsep}{0pt}
    \begin{minipage}[c]{1\linewidth}
        \centering
        \makebox[0.230\textwidth]{\smaller Frame}
        \makebox[0.230\textwidth]{\smaller Ground-truth Mask}
        \makebox[0.230\textwidth]{\smaller VideoFACT (Ours)}
        \smallskip
    \end{minipage}

	\vspace*{-0.3\baselineskip}

    \begin{minipage}[c]{1\linewidth}
        \centering
        \fbox{\includegraphics[width=0.230\textwidth]{{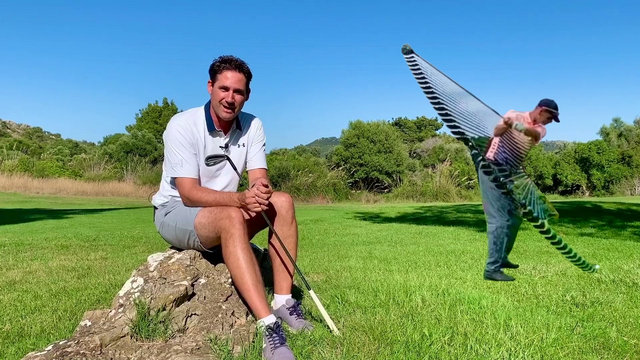}}}
        \fbox{\includegraphics[width=0.230\textwidth]{{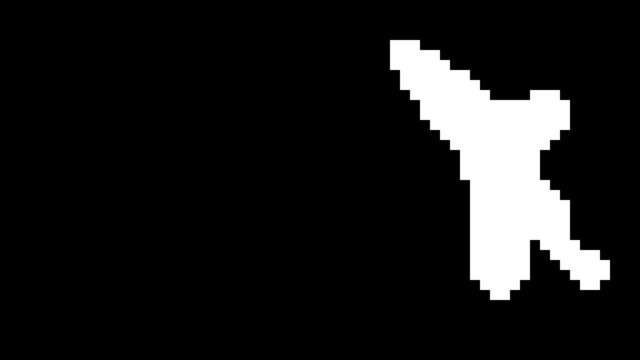}}}
        \fbox{\includegraphics[width=0.230\textwidth]{{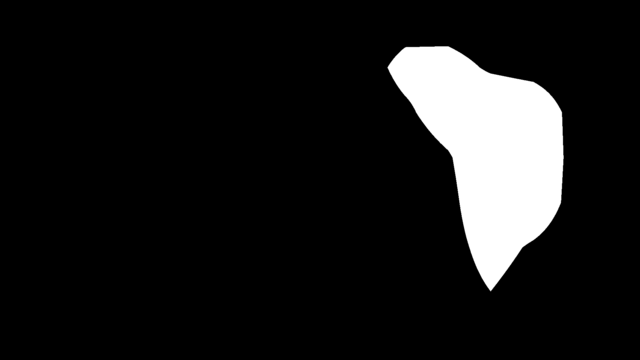}}}
        \smallskip
    \end{minipage}

    \begin{minipage}[c]{\linewidth}
        \centering
        \makebox[0.230\textwidth]{\smaller FSG}
        \makebox[0.230\textwidth]{\smaller EXIFnet}
        \makebox[0.230\textwidth]{\smaller Noiseprint}
        \smallskip
    \end{minipage}

	\vspace*{-0.3\baselineskip}

    \begin{minipage}[c]{1\linewidth}
        \centering
        \fbox{\includegraphics[width=0.230\textwidth]{{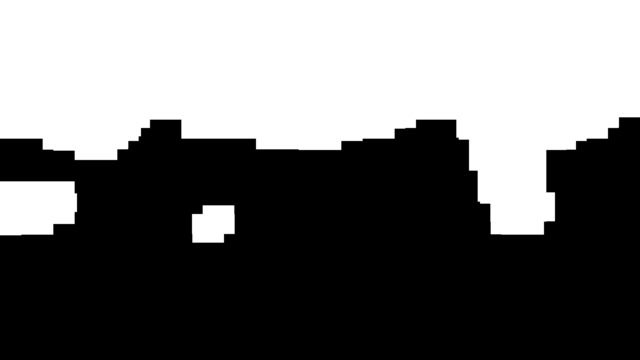}}}
        \fbox{\includegraphics[width=0.230\textwidth]{{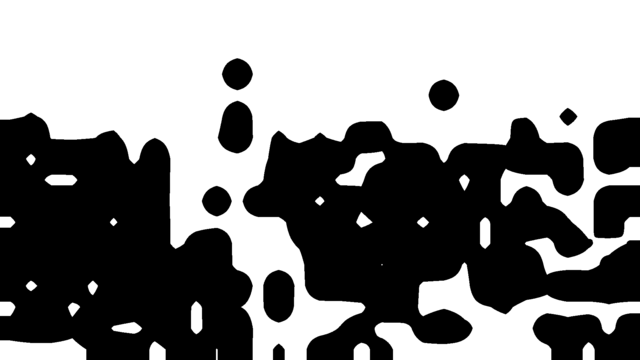}}}
        \fbox{\includegraphics[width=0.230\textwidth]{{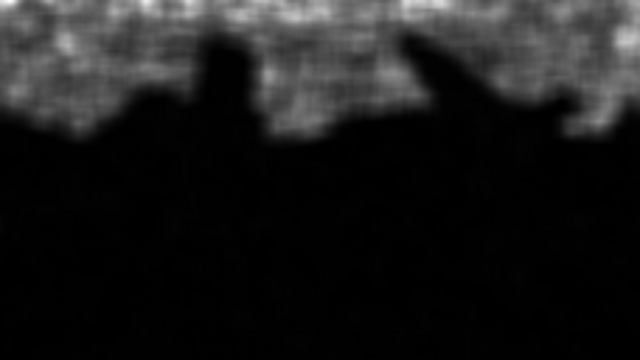}}}
        \smallskip
    \end{minipage}

    \begin{minipage}[c]{\linewidth}
        \centering
        \makebox[0.230\textwidth]{\smaller ManTra-Net}
        \makebox[0.230\textwidth]{\smaller MVSS-Net}
        \smallskip
    \end{minipage}

	\vspace*{-0.2\baselineskip}

    \begin{minipage}[c]{1\linewidth}
        \centering
        \fbox{\includegraphics[width=0.230\textwidth]{{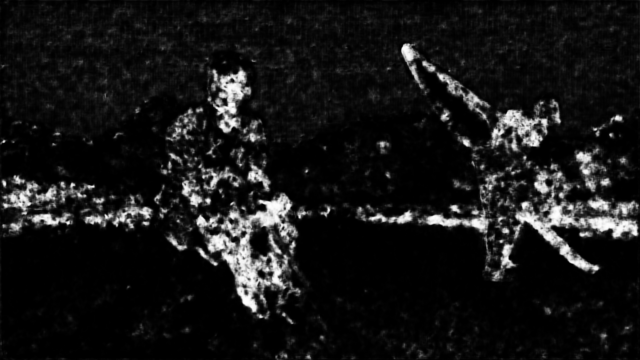}}}
        \fbox{\includegraphics[width=0.230\textwidth]{{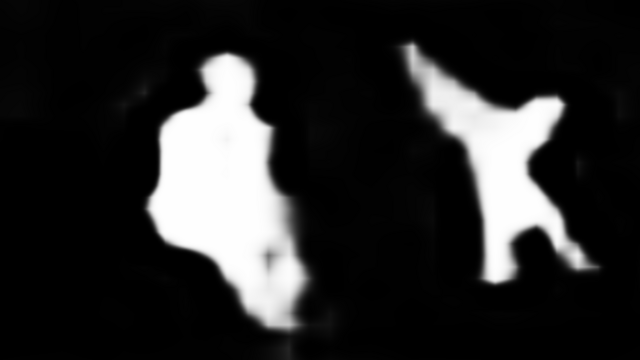}}}
        \smallskip
    \end{minipage}

    \vspace*{-0.5\baselineskip}

    \caption{\label{fig:falsification_example}
Example showing scene conditions that typically cause competing approaches to false alarm.  Systems like FSG, EXIFnet, and Noiseprint will mistake traces in smooth regions such as the sky for anomalous traces due to editing.  Networks like ManTra-Net and MVSS-Net mistake naturally occurring differences in noise statistics between foreground and background  objects as caused by editing.  Our network is able to use scene context and attention to control for these effects.
%
%
    }

    \vspace*{-0.5\baselineskip}

\end{figure}
\subsection{Discussion}
\label{subsec:discussion}

From the results presented in Tables~\ref{table:results_on_our_datasets} and \ref{table:results_on_itw_datasets}, we can see that existing approaches, which reported strong performance on image forgeries, largely fail on video.  
There are multiple reasons why this may occur.

The first reason, as mentioned in Section~\ref{sec:effect_of_video_coding}, is the effects of video compression.  Video coding parameters vary for each macroblock, which induces localized inconsistencies in forensic traces.


Another important reason is the false alarms effects in authentic regions.
An example of this can be seen in Fig.~\ref{fig:falsification_example}. Here, existing networks produce false alarms in different ways 
depending on each network's design.
One set of existing approaches, such as EXIFnet, FSG, and Noiseprint, identify fake content by searching for anomalous forensic traces. Hence, they produce false alarms in scene regions that contain low-quality forensic information such as the sky, which is consistent with previous research~\cite{FSM}. 
%
Another set of approaches, such as ManTra-Net and MVSS-Net,  seek to learn good forgery detection features by analyzing noise residuals and edge information.  These techniques can false alarm when highly salient foreground objects
, like the man on the rock in Fig.~\ref{fig:falsification_example},
naturally exhibit different statistics, i.e. sharper edges, from background objects. 

\subheader{How Our Network Overcomes These Effects}
Our network’s use of context embeddings and self-attention enables it to be much more resilient to the effects described above. 
Our Deep Self-Attention Module 
allows our network to learn which scene regions contain high quality forensic information. Therefore, VideoFACT relies more heavily on these regions when making decisions and avoids many of the false alarm issues faced by other networks. Additionally, local context information enables our network to control for local variations in forensic traces induced by video compression. 


\subheader{Failure Cases and Limitations}
We identified several common failure cases experienced by our network. 
Our network often misses falsified regions that are very small, particularly 
smaller than our $128\times 128$ pixels analysis window.  Our network produces poor decisions when both  the manipulated region and background have similar poor lighting conditions.  It also has diffficulty detecting regions altered by color swaps. 
Additionally, our network's current implementation is limited to 
analyzing 
video resolution of 1080p.



\section{Ablation Study}
\label{sec:ablation_study}
%
\begin{table}[!t]

    \centering
    \setlength\extrarowheight{-0.3pt}
    \resizebox{\linewidth}{!}{%
    \begin{tabular}{
        >{\hspace{0pt}}m{0.410\linewidth}
        >{\centering\hspace{0pt}}m{0.070\linewidth}
        >{\centering\hspace{0pt}}m{0.070\linewidth}
        >{\centering\hspace{0pt}}m{0.070\linewidth}
        >{\centering\hspace{0pt}}m{0.070\linewidth}
        >{\centering\hspace{0pt}}m{0.120\linewidth}
        >{\hspace{0pt}}m{0.001\linewidth}
        >{\hspace{0pt}}m{0.100\linewidth}
        >{\hspace{0pt}}m{0.100\linewidth}
        >{\hspace{0pt}}m{0.100\linewidth}
        >{\hspace{0pt}}m{0.100\linewidth}
    } 
        \hline

        \multirow{2}{0.300\linewidth}{\hspace{0pt}\textbf{Setup}} & 
        \multicolumn{5}{>{\centering\hspace{-50pt}}m{0.440\linewidth}}{\textbf{Component}} &  & 
        \multicolumn{4}{>{\centering\arraybackslash\hspace{-65pt}}m{0.257\linewidth}}{\textbf{VideoSham}} \\ 
        
        \cline{2-6}\cline{8-11} &

        \multicolumn{1}{>{\Centering\hspace{0pt}}m{0.070\linewidth}}{\textit{FFE}} & 
        \multicolumn{1}{>{\Centering\hspace{0pt}}m{0.070\linewidth}}{\textit{CFE}} & 
        \multicolumn{1}{>{\Centering\hspace{0pt}}m{0.120\linewidth}}{\textit{Transformer}} & 
        \multicolumn{1}{>{\Centering\hspace{0pt}}m{0.100\linewidth}}{\textit{Attn. maps}} & 
        \multicolumn{1}{>{\Centering\hspace{0pt}}m{0.120\linewidth}}{\textit{Data comb.}} &  
        
        & 

        \multicolumn{1}{>{\centering\hspace{0pt}}m{0.100\linewidth}}{\textit{Det. ACC}} & 
        \multicolumn{1}{>{\centering\hspace{0pt}}m{0.100\linewidth}}{\textit{Det. mAP}} & 
        \multicolumn{1}{>{\centering\hspace{0pt}}m{0.100\linewidth}}{\textit{Loc. F1}} & 
        \multicolumn{1}{>{\centering\arraybackslash\hspace{0pt}}m{0.100\linewidth}}{Loc. MCC} \\ 

        \hline

        \textbf{Proposed} 
        & $\bigplus$ & $\bigplus$ & $\bigplus$ & 3 & Add &  
        & \textbf{0.656} & \textbf{0.691} & \textbf{0.258} & \textbf{0.168} \\ 

        \hline

        No FFE 
        & — & $\bigplus$ & $\bigplus$ & 3 & Add &  
        & 0.610 & 0.646 & 0.209 & 0.118 \\

        No CFE 
        & $\bigplus$ & — & $\bigplus$ & 3 & Add &  
        & 0.586 & 0.635 & 0.163 & 0.043 \\

        No DSAM 
        & $\bigplus$ & $\bigplus$ & — & — & — &  
        & 0.601 & 0.626 & 0.144 & 0.000 \\

        No Transformer 
        & $\bigplus$ & $\bigplus$ & — & 3 & Add &  
        & 0.533 & 0.538 & 0.140 & 0.048 \\

        No Attention Squeeze 
        & $\bigplus$ & $\bigplus$ & $\bigplus$ & — & — &  
        & 0.622 & 0.656 & 0.254 & 0.120 \\

        1 Attention Map 
        & $\bigplus$ & $\bigplus$ & $\bigplus$ & 1 & Add &  
        & 0.610 & 0.655 & 0.175 & 0.121 \\

        10 Attention Maps 
        & $\bigplus$ & $\bigplus$ & $\bigplus$ & 10 & Add &  
        & 0.622 & 0.676 & 0.212 & 0.127 \\

        Diff. Feat. Refine
        & $\bigplus$ & $\bigplus$ & $\bigplus$ & 3 & Concat &  
        & 0.614 & 0.684 & 0.162 & 0.091 \\
        
        \hline
        \end{tabular}
    }
    
    \vspace*{-0.5\baselineskip}

    \caption{\label{table:ablation_evaluations} Ablation study of the components in our proposed network and their performance evaluations.}
	
	\vspace*{-0.9\baselineskip}
        
\end{table}

We conducted multiple ablation experiments to validate the importance of various components in VideoFACT's architecture. We trained each of the network variants using the same settings as the proposed method and assessed their performance on VideoSham. A summary of our experiments is shown in Table \ref{table:ablation_evaluations}.

The results show that each proposed components improve the performance of the model. By removing either FFE or CFE, every metric has a significant reduction. When removing the entire DSAM, the joint embeddings were fed directly into the detector and the localizer networks. This variant still resulted in a substantial performance drop across the board. The detection metrics suggest that this network only does slightly better than random guess. 

We also measure the importance of the Transformer and Attention Squeeze in the DSAM. We first replace the Transformer encoders with six fully connected layers with ReLU as the activation function. In this scenario, the model performs at a level close to random guess. Therefore, this cements the necessity of the Transformer for our network. Besides, we also try removing Attention Squeeze and connect the output of the Transformer encoders directly to the detector and localizer. We see that this variant under-performs in both detection and localization. Finally, we try using 1, 3 (proposed) and 10 attention maps. Results show that using 3 attention maps yields the best performance.

For Feature Refinement approaches, we also try concatenating all three sets of weighted spatially contextualized forensic embeddings, instead of proposed approach. Results show that both detection and localization performance drop significantly. Therefore, the proposed method is optimal.

\section{Conclusion}
\label{sec:conclusion}

In this paper, we propose a new network, VideoFACT, to detect and localize a broad range of video forgeries and manipulations.   Our network does this by utilizing both forensic embeddings to capture traces left by manipulation, context embeddings to 
control for variation in forensic traces caused by video coding, 
and a deep self-attention mechanism to estimate the local quality and relevance of forensic embeddings. 
We create several new video forgery datasets, which we used along with the Adobe VideoSham dataset to experimentally evaluate our network's performance. Our results show that our proposed network is able to identify a diverse set of video forgeries, including those not encountered during training.  Furthermore, our results show that existing image forensic networks largely fail to identify fake content in video.

{
    \small
    \bibliographystyle{ieee_fullname}
    \bibliography{bib/references,bib/MCS_CVPR23_Refs}
}

\ifarxiv \clearpage \appendix
\label{sec:appendix}

\section{Code}
\label{appendix:code}

In order to provide transparency and reproducibility towards the results presented in this paper, we have included VideoFACT's network code and example code for training/evaluating our model. These files are stored in a zip file called \texttt{code.zip}. The whole code for this project and trained weights for the network will be publicly available under our research group's gitlab page upon the publication of this work [\textit{this link is temporary hidden to preserve our anonimity under the double blind review process}].
\section{Dataset Creation}
\label{appendix:dataset_creation}

\setcounter{table}{0}
\renewcommand{\thetable}{B\arabic{table}}%
\setcounter{figure}{0}
\renewcommand{\thefigure}{B\arabic{figure}}%

\subsection{Set A: Standard Video Manipulations Datasets}

In this section, we presents more details regarding the creation of the Standard Video Manipulations datasets used in this paper, specifically, those used for training and validation of our network - VCMS, VPVM, VPIM. 

In order to generate samples for each of these datasets, we first needed to make binary masks which specify the to-be-manipulated regions in video frames/images. 
To facilitate this process, we created a library of 10 basic shapes: rectangle, circle, ellipse, triangle, pentagon, heptagon, 5-pointed star, 8-pointed star, 12-pointed star, and 18-pointed star. Since we wanted to avoid our network overfitting to any particular shapes, we further constructed complex compound shapes by overlapping up-to three basic shapes in a random manner. After the compound shapes for each mask were finalized, it was randomly resized, randomly rotated, and translated to a random location inside the mask's boundary ($1080 \times 1920$). We note that masks with the manipulated area greater than 75\% of the total area were rejected and regenerated using the same procedure.

With the binary masks ready, we then applied either splicing operations (copying content from a source video/image and pasting it into a destination video/image), or in-place editing operation (manipulating each frames of a video or image). With regards to in-place editing operations, we manipulated the regions specified by the mask with at least one of the following operations: changing brightness, contrast, saturation, and hue, adding random Gaussian blur, random motion blur, random box blur, and random Gaussian noise. We used the open-source differentiable computer vision library Kornia to perform these edits. The parameters used for the perceptuall visible datasets (VPVM) and the perceptually invisible datasets (VPIM) are listed in the Table \ref{table:kornia_manipulation_parameters}.

\begin{table}[!t]
    \centering
    \setlength\extrarowheight{-1.5pt}
    \resizebox{1.0\linewidth}{!}{%
    \begin{tabular}{
        >{\hspace{0pt}}m{0.150\linewidth}
        |>{\hspace{0pt}}m{0.300\linewidth}
        >{\hspace{0pt}}m{0.500\linewidth}
        >{\hspace{0pt}}m{0.180\linewidth}
    }
        \hline

        \multicolumn{1}{>{\hspace{0pt}}m{0.150\linewidth}}{\textbf{Visible?}} & 
        \multicolumn{1}{>{\hspace{0pt}}m{0.300\linewidth}}{\textbf{Manip. Type}} & 
        \multicolumn{1}{>{\hspace{0pt}}m{0.500\linewidth}}{\textbf{Manip. Parameters}} & 
        \multicolumn{1}{>{\centering\hspace{-30pt}}m{0.180\linewidth}}{\textbf{Manip. Prob.}} \\ 

        \hline

        \multirow{8}{0.150\linewidth}{\vspace{-50pt}Yes} 
        & Brightness        & \texttt{range=[0.8, 1.6]}                     & 1.0 \\ 
        \cline{2-4}
        & Contrast          & \texttt{range=[0.7, 1.3]}                     & 1.0 \\
        \cline{2-4}
        & Saturation        & \texttt{range=[0.8, 1.1]}                     & 1.0 \\
        \cline{2-4}
        & Hue               & \texttt{range=[-0.2, 0.2]}                    & 1.0 \\
        \cline{2-4}
        & Gaussian Blur     & \texttt{kernel\_size=(5,5), sigma=(2,2)}      & 0.7 \\
        \cline{2-4}
        & Motion Blur       & \texttt{kernel\_size=(5,5), angle=[-25, 25], direction=[-1, 1], resample='BICUBIC'} & 0.7 \\
        \cline{2-4}
        & Box Blur          & \texttt{kernel\_size=(5,5)}                   & 0.7 \\
        \cline{2-4}
        & Gaussian Noise    & \texttt{std=0.05}                             & 1.0 \\

        \specialrule{.3em}{.2em}{.2em}

        \multirow{8}{0.150\linewidth}{\vspace{-40pt}No} 
        & Brightness        & \texttt{range=[0.95, 1.05]}                   & 0.9 \\ 
        \cline{2-4}
        & Contrast          & \texttt{range=[0.95, 1.05]}                   & 0.9 \\
        \cline{2-4}
        & Saturation        & \texttt{range=[0.95, 1.05]}                   & 0.9 \\
        \cline{2-4}
        & Gaussian Blur     & \texttt{kernel\_size=(3,3), sigma=(1.2,1.2)}  & 0.7 \\
        \cline{2-4}
        & Motion Blur       & \texttt{kernel\_size=(3,3), angle=[-20, 20], direction=[-1, 1], resample='BICUBIC'} & 0.7 \\
        \cline{2-4}
        & Box Blur          & \texttt{kernel\_size=(3,3)}                   & 0.7 \\
        \cline{2-4}
        & Gaussian Noise    & \texttt{std=0.006}                            & 0.9 \\

        \hline
    \end{tabular}
    }
    \caption{\label{table:kornia_manipulation_parameters} This table lists the different parameters used to manipulate a victim video frame or victim image so that the manipulated area is either perceptually visible or invisible. These parameters are passed into different augmentation modules available in the open-source differentiable computer vision library Kornia.}
\end{table}

Finally, we re-encoded the authentic video frames of each authentic video into the set of authentic videos and the manipulated videos frames of each manipulated video into the set of manipulated videos. All videos were re-encoded as H.264 videos using FFmpeg with the constant rate factor of 23 and the frame rate of 30 FPS. 

Note that, in addition to the three video datasets made using the process described above, we also created three Standard Image Manipulation datasets in the exact same procedure. This resulted in three auxilary image datasets: ICMS, IPVM and IPIM. 

A summary of all datasets are listed in Table \ref{table:training_datasets_information}, \ref{table:validating_datasets_information}, \ref{table:testing_datasets_information}

\subsection{Set B: In-the-Wild Manipulated Datasets}

In addition to evaluating on our Standard Video Manipulation datasets, we tested our proposed network and other network on three In-the-Wild Manipulated datasets: VideoSham, Inpainted Video and Deepfake Video. 

The \textbf{VideoSham} dataset used in this paper was a subset of the one published by Adobe Research \cite{VideoSham}. We excluded videos with audio track or temporal manipulations and videos with resolution less than 1080p. Hence, the remaining manipulated videos were attacked by 1) adding objects/subjects, 2) removing objects/subjects, 3) background/object's color change and 4) adding/removing text. 
Since the masks indicating the manipulation regions were missing from the original dataset, we created them by 
first dividing both the manipulated frame and the original frame into small, non-overlapping blocks ($16 \times 16$), 
then computing the average luminance-value absolute difference across 3 channels (R, G, B) between each original and manipulated block. 
After we produced a scalar value for every block, we normalized these values so that they were between 0 and 1 and thresholded them so that they became binarized. We then projected the binary value of each blocks to all the pixel locations belong to that block in order to create a binarized pixel-level masks indicating the manipulated region.

The \textbf{Inpainted Video} dataset was generated by replacing objects in a scene with their background. In order to leverage existing state-of-the-art video inpainting algorithms like E2FGVI-HQ, we needed videos in which some to-be-removed objects were segmented across multiple frames. Therefore, we chose the Densely Annotated Video Segmentation (DAVIS) dataset \cite{pont2017DAVIS} for this purpose because it contains both the original videos and the ground-truth segmentation masks for foreground objects in those videos. Inputing a video and its masks into the network code provided publicly by the authors of E2FGVI-HQ \cite{E2FGVI}, and using the recommended settings, we generated videos in which the segmented objects were inpainted over. Since the resolution of the output videos were 540p, we resized them up to 1080p so that they could be analyzed by our proposed network. We applied this process over all 90 videos in the DAVIS dataset to get 90 inpainted videos, which we used for evaluation.

The \textbf{Deepfake Video} dataset was made by creating deepfaked videos of celebrity interview videos downloaded from YouTube. These downloaded videos were first trimmed down to maximum of 30 seconds, where the only primary subject on the scene was a human body with its face clearly visible. We then chose one source-destination video pair from our set of downloaded videos to perform face swapping. Note that faces of the similar skin tone and gender were more likely to result in better quality deepfakes. We then extracted all faces from the frames of the source and destination video. These faces were then aligned and learned by a deep neural network architecture from DeepFaceLab \cite{DeepFaceLab}. After we trained the deepfake network, we used it to perform face-swapping for each uncompressed frame. Finally, these frames were compressed into an H.264 video using FFmpeg with the constant rate factor of 23 and the frame rate of 30 FPS. Using this procedure, we generated 10 deepfaked videos, in which we took out continuous 30 frames chunk from each video to make the Deepfake Video dataset.

\begin{table}[!htb]
    \centering
    \setlength\extrarowheight{-1.5pt}
    \resizebox{1.0\linewidth}{!}{%
    \begin{tabular}{
        >{\hspace{0pt}}m{0.200\linewidth}
        >{\hspace{0pt}}m{0.200\linewidth}
        >{\hspace{0pt}}m{0.200\linewidth}
        >{\hspace{0pt}}m{0.001\linewidth}
        >{\hspace{0pt}}m{0.200\linewidth}
        >{\hspace{0pt}}m{0.200\linewidth}
    }
        \hline

        \multirow{2}{0.146\linewidth}{\hspace{0pt}\textbf{Dataset}} & 
        \multicolumn{2}{>{\centering\hspace{-30pt}}m{0.320\linewidth}}{\textbf{Original}} & 
        \multicolumn{1}{>{\centering\hspace{0pt}}m{0.001\linewidth}}{} & 
        \multicolumn{2}{>{\centering\arraybackslash\hspace{-30pt}}m{0.320\linewidth}}{\textbf{Manipulated}} \\

        \cline{2-3}\cline{5-6} & 
        \multicolumn{1}{>{\centering\hspace{0pt}}m{0.200\linewidth}}{\# of Frames} &
        \multicolumn{1}{>{\centering\hspace{0pt}}m{0.200\linewidth}}{\# of Videos} &
        \multicolumn{1}{>{\centering\hspace{0pt}}m{0.001\linewidth}}{} &
        \multicolumn{1}{>{\centering\hspace{0pt}}m{0.200\linewidth}}{\# of Frames} &
        \multicolumn{1}{>{\centering\arraybackslash\hspace{0pt}}m{0.200\linewidth}}{\# of Videos} \\ 

        \hline

        VCMS 
        & 48000 & 1600 &  
        & 48000 & 1600 \\
        VPVM 
        & 48000 & 1600 &  
        & 48000 & 1600 \\
        VPIM 
        & 48000 & 1600 &  
        & 48000 & 1600 \\ 

        \hline

        ICMS 
        & 48000 & N/A & 
        & 48000 & N/A \\
        IPVM 
        & 48000 & N/A &  
        & 48000 & N/A \\
        IPIM 
        & 48000 & N/A & 
        & 48000 & N/A \\ 

        \hline
    \end{tabular}
    }
    \caption{\label{table:training_datasets_information} Summary of our training datasets.}
\end{table}

\begin{table}[!htb]
    \centering
    \setlength\extrarowheight{-1.5pt}
    \resizebox{1.0\linewidth}{!}{%
    \begin{tabular}{
        >{\hspace{0pt}}m{0.200\linewidth}
        >{\hspace{0pt}}m{0.200\linewidth}
        >{\hspace{0pt}}m{0.200\linewidth}
        >{\hspace{0pt}}m{0.001\linewidth}
        >{\hspace{0pt}}m{0.200\linewidth}
        >{\hspace{0pt}}m{0.200\linewidth}
    }
        \hline

        \multirow{2}{0.146\linewidth}{\hspace{0pt}\textbf{Dataset}} & 
        \multicolumn{2}{>{\centering\hspace{-30pt}}m{0.320\linewidth}}{\textbf{Original}} & 
        \multicolumn{1}{>{\centering\hspace{0pt}}m{0.001\linewidth}}{} & 
        \multicolumn{2}{>{\centering\arraybackslash\hspace{-30pt}}m{0.320\linewidth}}{\textbf{Manipulated}} \\

        \cline{2-3}\cline{5-6} & 
        \multicolumn{1}{>{\centering\hspace{0pt}}m{0.200\linewidth}}{\# of Frames} &
        \multicolumn{1}{>{\centering\hspace{0pt}}m{0.200\linewidth}}{\# of Videos} &
        \multicolumn{1}{>{\centering\hspace{0pt}}m{0.001\linewidth}}{} &
        \multicolumn{1}{>{\centering\hspace{0pt}}m{0.200\linewidth}}{\# of Frames} &
        \multicolumn{1}{>{\centering\arraybackslash\hspace{0pt}}m{0.200\linewidth}}{\# of Videos} \\ 

        \hline

        VCMS 
        & 7800 & 260 &  
        & 7800 & 260 \\
        VPVM 
        & 7800 & 260 &  
        & 7800 & 260 \\
        VPIM 
        & 7800 & 260 &  
        & 7800 & 260 \\ 

        \hline

        ICMS 
        & 7800 & N/A & 
        & 7800 & N/A \\
        IPVM 
        & 7800 & N/A &  
        & 7800 & N/A \\
        IPIM 
        & 7800 & N/A & 
        & 7800 & N/A \\ 

        \hline
    \end{tabular}
    }
    \caption{\label{table:validating_datasets_information} Summary of our validation datasets.}
\end{table}

\begin{table}[!htb]
    \centering
    \setlength\extrarowheight{-1.5pt}
    \resizebox{1.0\linewidth}{!}{%
    \begin{tabular}{
        >{\hspace{0pt}}m{0.100\linewidth}
        >{\hspace{0pt}}m{0.270\linewidth}
        >{\hspace{0pt}}m{0.200\linewidth}
        >{\hspace{0pt}}m{0.200\linewidth}
        >{\hspace{0pt}}m{0.001\linewidth}
        >{\hspace{0pt}}m{0.200\linewidth}
        >{\hspace{0pt}}m{0.200\linewidth}
    }
        \hline

        \multirow{2}{0.100\linewidth}{\hspace{0pt}\textbf{Group}} & 
        \multirow{2}{0.146\linewidth}{\hspace{0pt}\textbf{Dataset}} & 
        \multicolumn{2}{>{\centering\hspace{-30pt}}m{0.320\linewidth}}{\textbf{Original}} & 
        \multicolumn{1}{>{\centering\hspace{0pt}}m{0.001\linewidth}}{} & 
        \multicolumn{2}{>{\centering\arraybackslash\hspace{-30pt}}m{0.320\linewidth}}{\textbf{Manipulated}} \\

        \cline{3-4}\cline{6-7} &  & 
        \multicolumn{1}{>{\centering\hspace{0pt}}m{0.200\linewidth}}{\# of Frames} &
        \multicolumn{1}{>{\centering\hspace{0pt}}m{0.200\linewidth}}{\# of Videos} &
        \multicolumn{1}{>{\centering\hspace{0pt}}m{0.001\linewidth}}{} &
        \multicolumn{1}{>{\centering\hspace{0pt}}m{0.200\linewidth}}{\# of Frames} &
        \multicolumn{1}{>{\centering\arraybackslash\hspace{0pt}}m{0.200\linewidth}}{\# of Videos} \\ 

        \hline

        \multirow{3}{0.130\linewidth}{\vspace{0pt}A} 
        & VCMS & 4200 & 140 &  & 4200 & 140 \\
        & VPVM & 4200 & 140 &  & 4200 & 140 \\
        & VPIM & 4200 & 140 &  & 4200 & 140 \\ 

        \hline

        \multirow{3}{0.130\linewidth}{\vspace{0pt}B} 
        & VideoSham
        & 7897 & 32 & 
        & 12746 & 64 \\

        & Inpainted Video 
        & 9312 & 90 & 
        & 9312 & 90 \\ 

        & Deepfake Video 
        & 300 & 10 &  
        & 300 & 10 \\

        \hline
    \end{tabular}
    }
    \caption{\label{table:testing_datasets_information} Summary of the datasets used for evaluating the performance of our network and others.}
\end{table}

\section{Run Time Analysis}
\label{appendix:runtime_analysis}

\setcounter{table}{0}
\renewcommand{\thetable}{C\arabic{table}}%
\setcounter{figure}{0}
\renewcommand{\thefigure}{C\arabic{figure}}%

In this section, we provide a preliminary run time analysis of our network and other competing networks: FSG, EXIFnet, Noiseprint, ManTra-Net and MVSS-Net. These run time benchmarks are gathered using unoptimized code, taken directly from other authors' publicly available code repositories. Therefore, these results do not represent the best potential run time of these algorithms. They only show what one may experience the run time of each algorithm by directly using other authors' publicly available code. We performed this analysis on a machine with an NVIDIA RTX 3090 GPU, a 12th Gen Intel i9-12900KF CPU (24 cores at 5.200 Ghz), 64 GB of DDR4 (2800Mhz) RAM, running Ubuntu 22.04.1 with Kernel version 5.15.0-52-generic. We ran 1000 samples individually, each of size $1080 \times 1920 \times 3$, through each network, recorded the total run time, then reported the average run time as frames per second (FPS).

From Table \ref{table:runtime_analysis}, we see that our network achieves the highest FPS, or fastest run time when compared to competing networks. Additionally, with a FPS of over 30, our network is capable of real-time video processing, which can be very useful in practical scenarios.

\begin{table}
    \centering
    \setlength\extrarowheight{-1.0pt}
    \resizebox{\linewidth}{!}{%
    \begin{tabular}{
        >{\hspace{0pt}}m{0.452\linewidth}
        |>{\centering\arraybackslash\hspace{0pt}}m{0.700\linewidth}
    } 
        \hline

        \textbf{Network} & 
        \multicolumn{1}{>{\centering\arraybackslash\hspace{0pt}}m{0.700\linewidth}}{\textbf{Average Analysis Frame Rate}\par{}\textbf{(FPS)}} \\ 

        \hline

        \textbf{Proposed} & \textbf{30.80} \\ 

        \hline

        FSG & 1.82 \\
        EXIFnet & 0.04 \\
        Noiseprint & 0.57 \\
        ManTra-Net & 0.40 \\
        MVSS-Net & 26.06 \\
        \hline
        \end{tabular}
    }

    \caption{\label{table:runtime_analysis} This table shows the average run time (number of samples per second) of our network and competing networks.}
\end{table}
\section{Additional Examples From Each Dataset and Their Localization Results}
\label{appendix:additional_loc_examples}

\setcounter{table}{0}
\renewcommand{\thetable}{D\arabic{table}}%
\setcounter{figure}{0}
\renewcommand{\thefigure}{D\arabic{figure}}%

In this section, we show additional representative examples from each dataset along with their localization results from our proposed network and other competing networks: FSG, EXIFnet, Noiseprint, ManTra-Net and MVSS-Net. 

A brief discussion and interpretation of the results for each dataset is provided in each figure's caption. Results for the VCMS dataset are presented in Figure \ref{fig:vcms_localization_results}, results for the VPVM dataset are presented in Figure \ref{fig:vpvm_localization_results}, results for the VPIM dataset are presented in Figure \ref{fig:vpim_localization_results}, results for the Deepfake Video dataset are presented in Figure \ref{fig:deepfake_localization_results}, results for the Inpainted Video dataset are presented in Figure \ref{fig:inpaint_localization_results} and results for the VideoSham dataset are presented in Figure \ref{fig:videosham_localization_results}.
\balance

\begin{figure*}[!b]
	\vspace*{-0.1\baselineskip}

    \centering
    \setlength{\fboxsep}{0pt}

    \begin{minipage}[t]{1\textwidth}
        \makebox[0.083\textwidth][r]{\raisebox{15pt}{\smaller Frame\hspace{6pt}}}
        \fbox{\includegraphics[width=0.100\textwidth]{{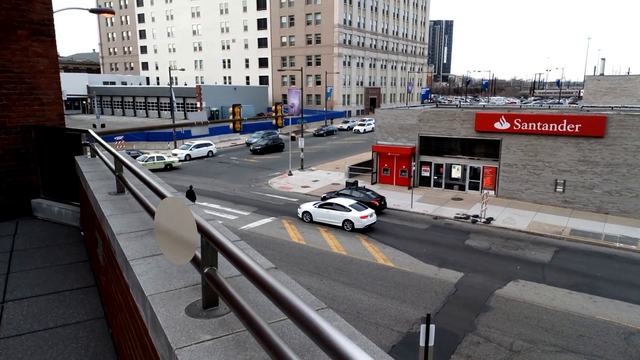}}}
        \fbox{\includegraphics[width=0.100\textwidth]{{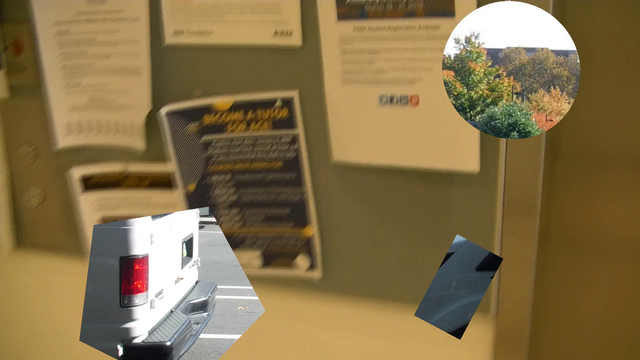}}}
        \fbox{\includegraphics[width=0.100\textwidth]{{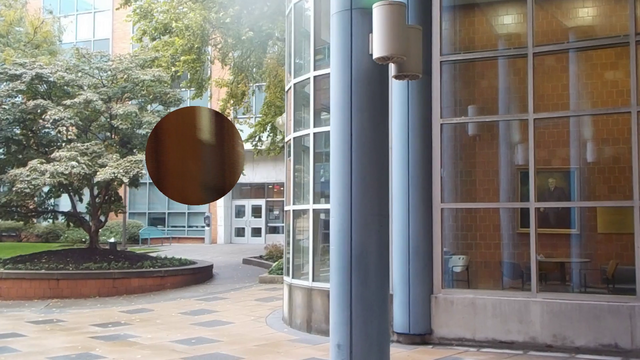}}}
        \fbox{\includegraphics[width=0.100\textwidth]{{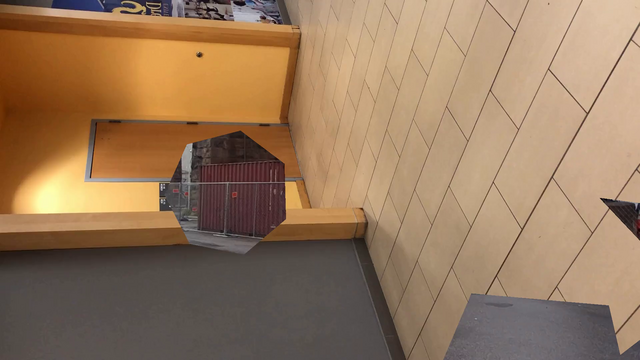}}}
        \fbox{\includegraphics[width=0.100\textwidth]{{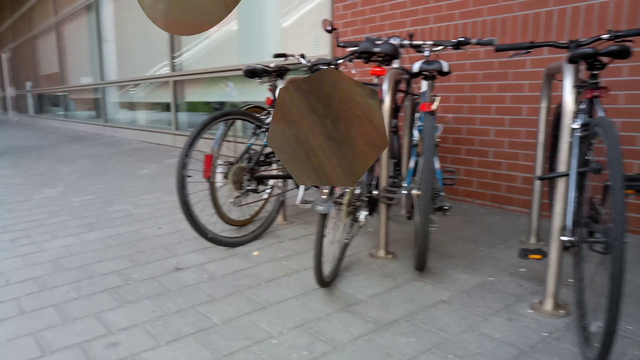}}}
        \fbox{\includegraphics[width=0.100\textwidth]{{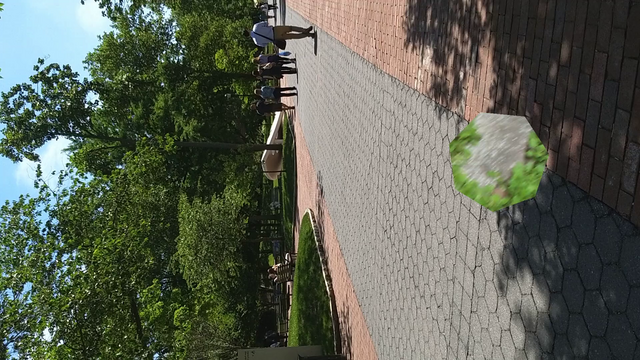}}}
        \fbox{\includegraphics[width=0.100\textwidth]{{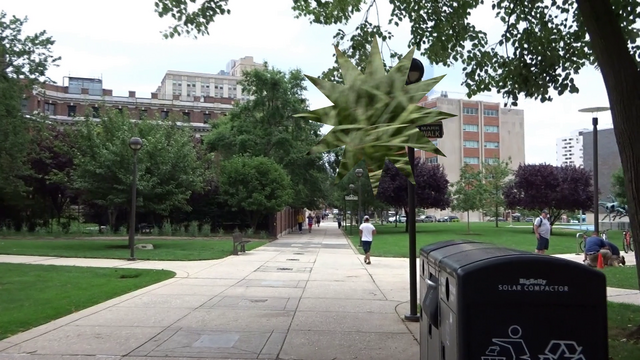}}}
        \fbox{\includegraphics[width=0.100\textwidth]{{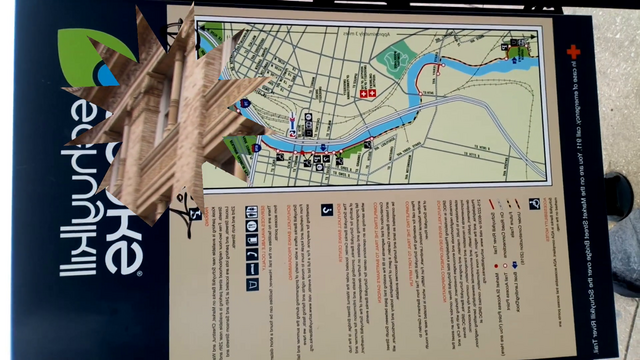}}}
        \smallskip
    \end{minipage}

	\vspace*{-0.1\baselineskip}

    \begin{minipage}[t]{1\textwidth}
        \makebox[0.083\textwidth][r]{\raisebox{15pt}{\smaller Mask\hspace{6pt}}}
        \fbox{\includegraphics[width=0.100\textwidth]{{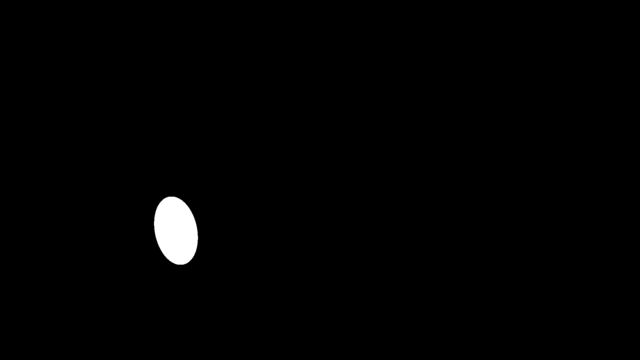}}}
        \fbox{\includegraphics[width=0.100\textwidth]{{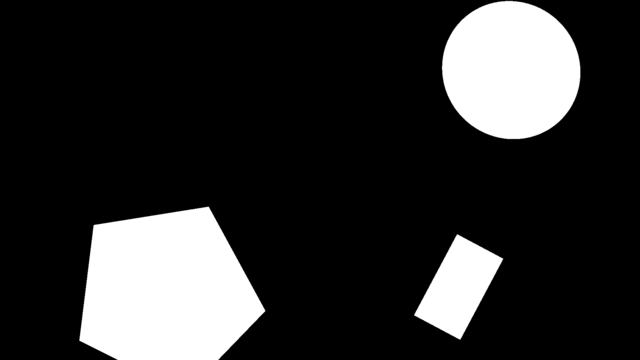}}}
        \fbox{\includegraphics[width=0.100\textwidth]{{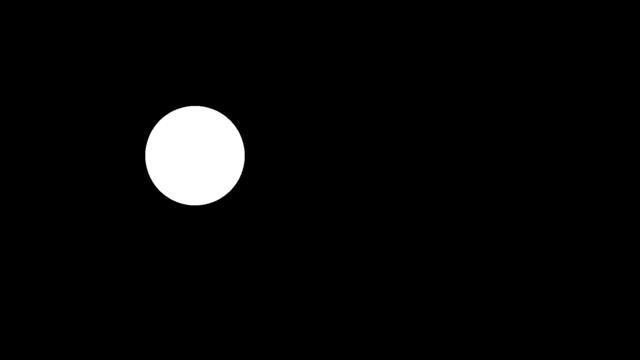}}}
        \fbox{\includegraphics[width=0.100\textwidth]{{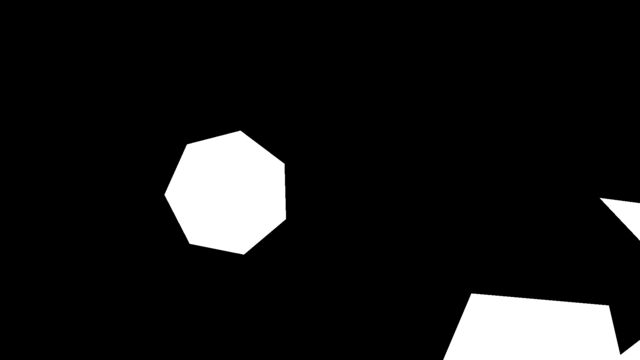}}}
        \fbox{\includegraphics[width=0.100\textwidth]{{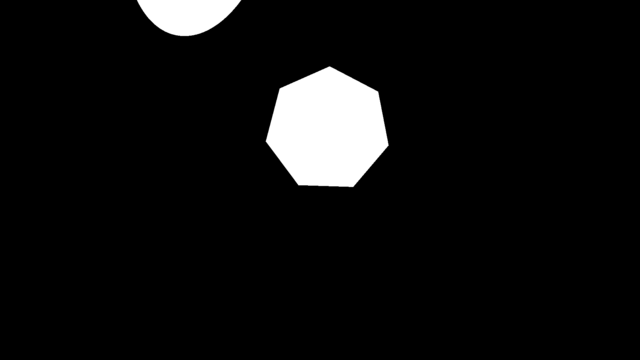}}}
        \fbox{\includegraphics[width=0.100\textwidth]{{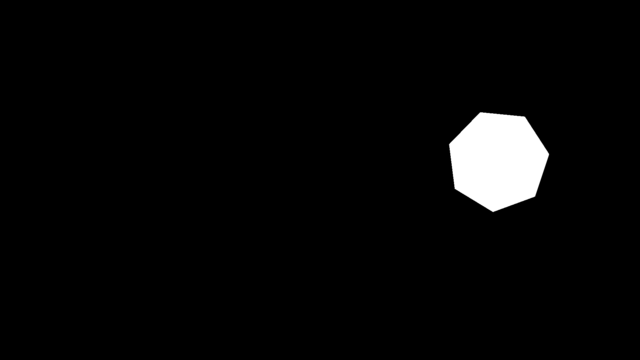}}}
        \fbox{\includegraphics[width=0.100\textwidth]{{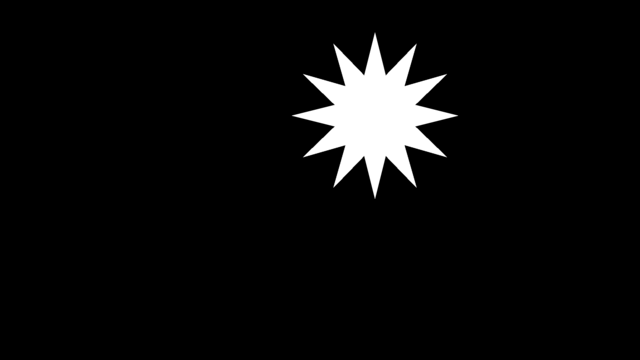}}}
        \fbox{\includegraphics[width=0.100\textwidth]{{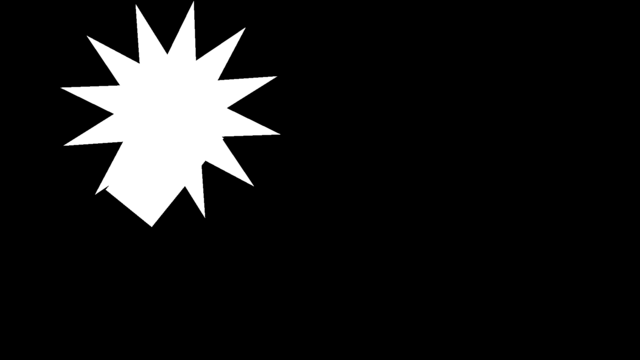}}}
        \smallskip
    \end{minipage}

	\vspace*{-0.1\baselineskip}

    \begin{minipage}[t]{1\textwidth}
        \makebox[0.083\textwidth][r]{\raisebox{15pt}{\smaller Proposed\hspace{6pt}}}
        \fbox{\includegraphics[width=0.100\textwidth]{{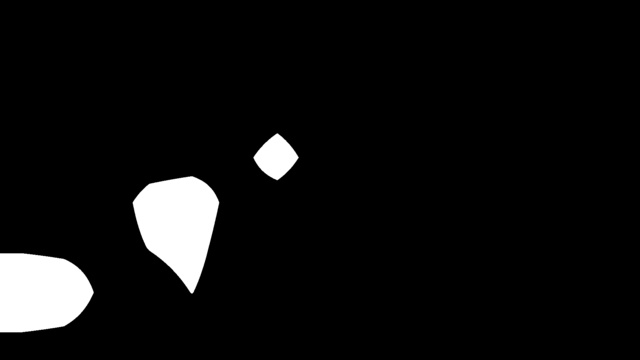}}}
        \fbox{\includegraphics[width=0.100\textwidth]{{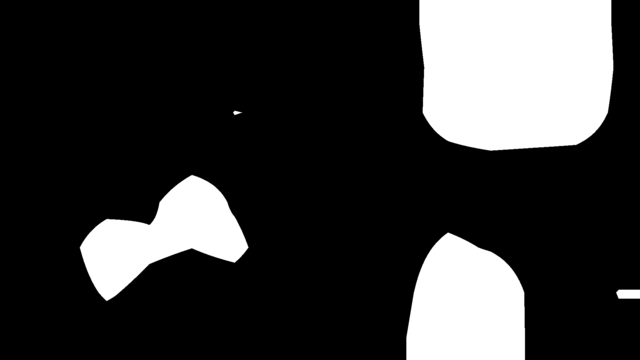}}}
        \fbox{\includegraphics[width=0.100\textwidth]{{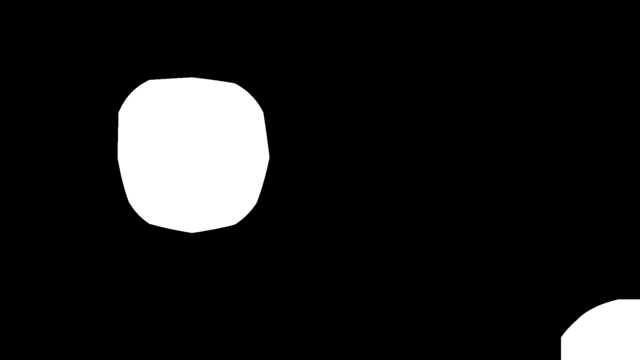}}}
        \fbox{\includegraphics[width=0.100\textwidth]{{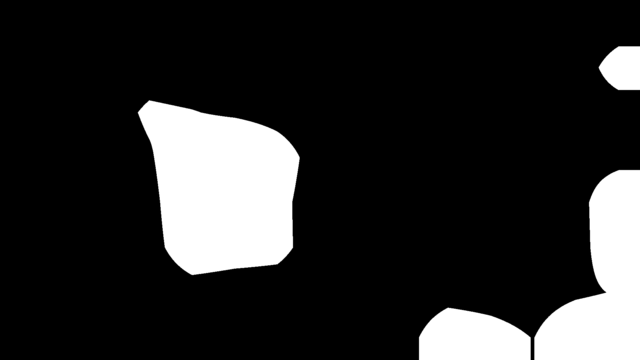}}}
        \fbox{\includegraphics[width=0.100\textwidth]{{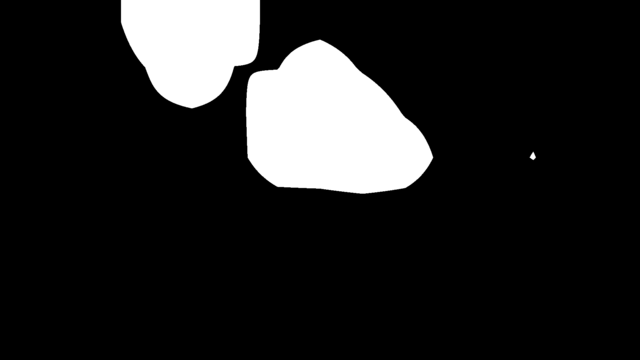}}}
        \fbox{\includegraphics[width=0.100\textwidth]{{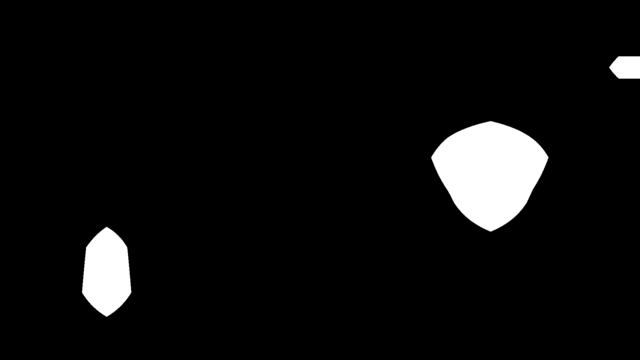}}}
        \fbox{\includegraphics[width=0.100\textwidth]{{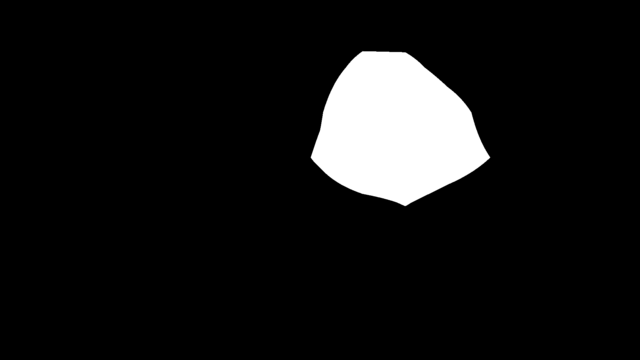}}}
        \fbox{\includegraphics[width=0.100\textwidth]{{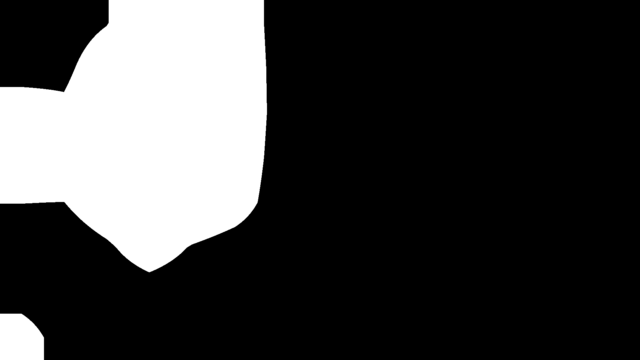}}}
        \smallskip
    \end{minipage}

	\vspace*{-0.1\baselineskip}

    \begin{minipage}[t]{1\textwidth}
        \makebox[0.083\textwidth][r]{\raisebox{15pt}{\smaller FSG\hspace{6pt}}}
        \fbox{\includegraphics[width=0.100\textwidth]{{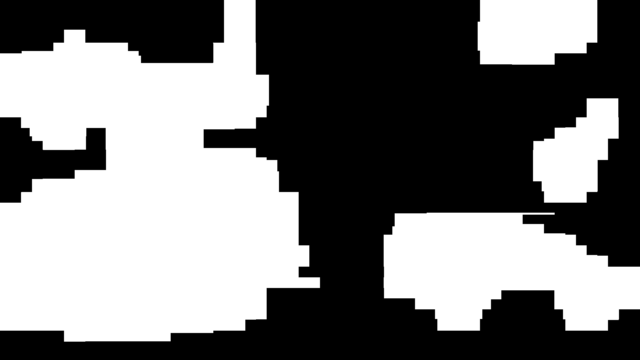}}}
        \fbox{\includegraphics[width=0.100\textwidth]{{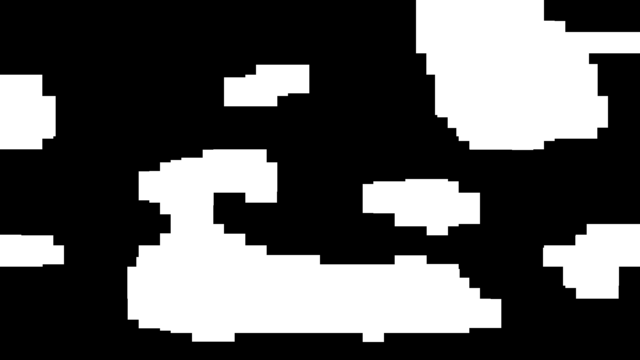}}}
        \fbox{\includegraphics[width=0.100\textwidth]{{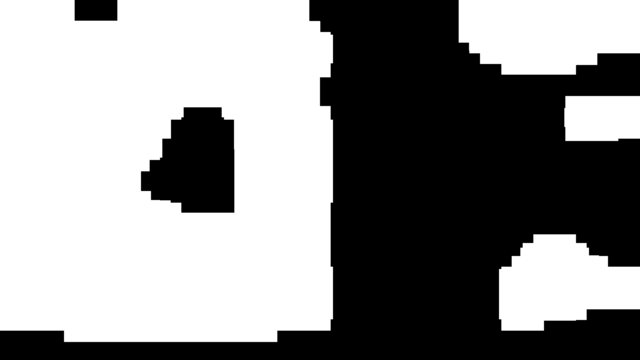}}}
        \fbox{\includegraphics[width=0.100\textwidth]{{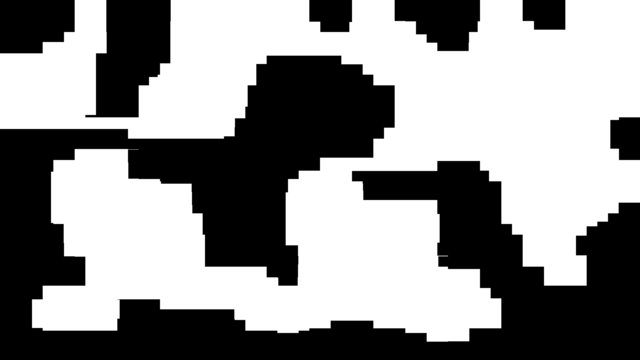}}}
        \fbox{\includegraphics[width=0.100\textwidth]{{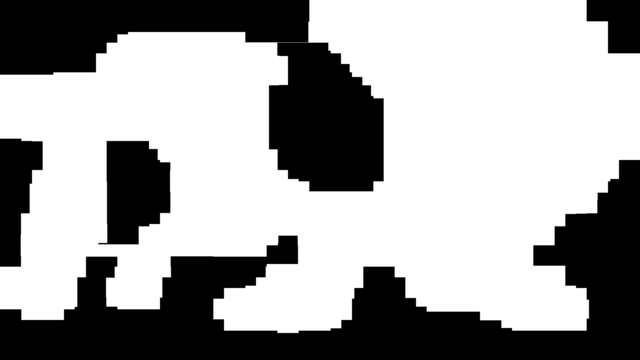}}}
        \fbox{\includegraphics[width=0.100\textwidth]{{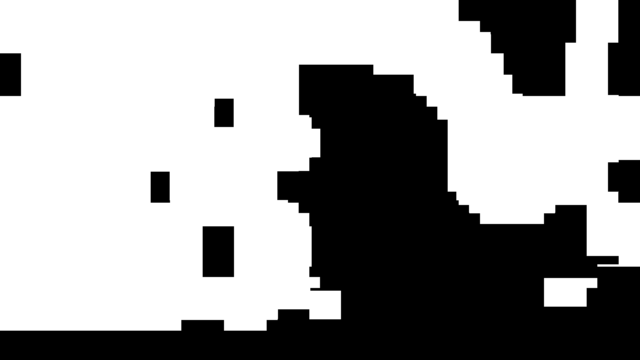}}}
        \fbox{\includegraphics[width=0.100\textwidth]{{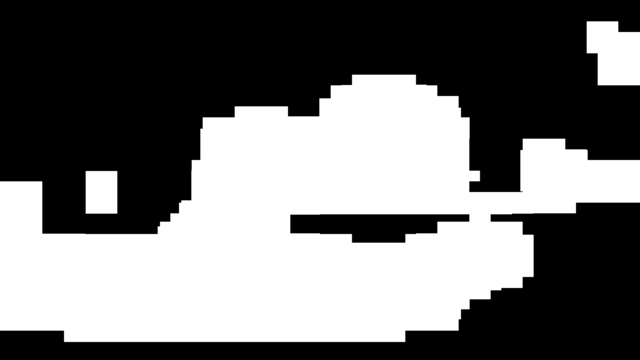}}}
        \fbox{\includegraphics[width=0.100\textwidth]{{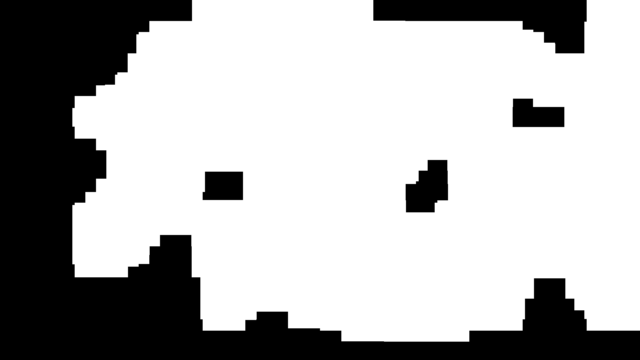}}}
        \smallskip
    \end{minipage}

	\vspace*{-0.1\baselineskip}

    \begin{minipage}[t]{1\textwidth}
        \makebox[0.083\textwidth][r]{\raisebox{15pt}{\smaller EXIFnet\hspace{6pt}}}
        \fbox{\includegraphics[width=0.100\textwidth]{{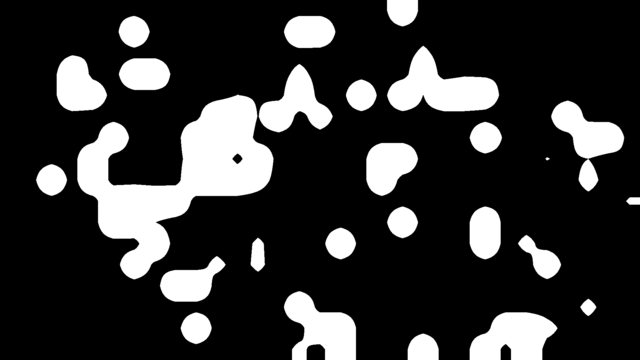}}}
        \fbox{\includegraphics[width=0.100\textwidth]{{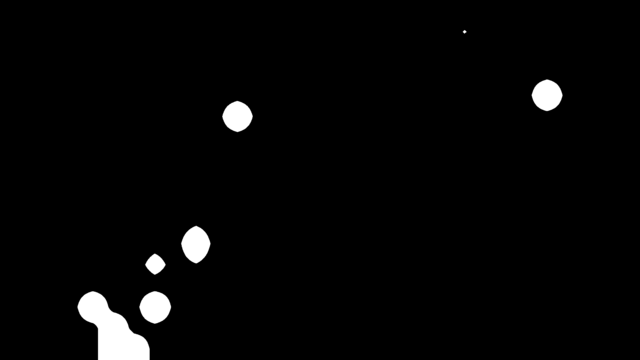}}}
        \fbox{\includegraphics[width=0.100\textwidth]{{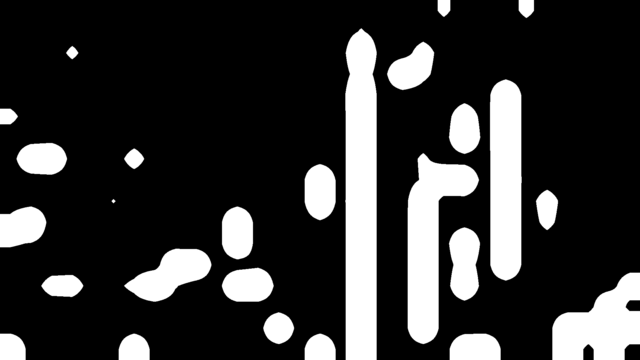}}}
        \fbox{\includegraphics[width=0.100\textwidth]{{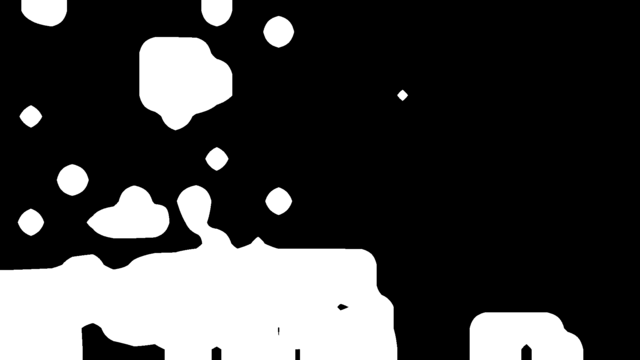}}}
        \fbox{\includegraphics[width=0.100\textwidth]{{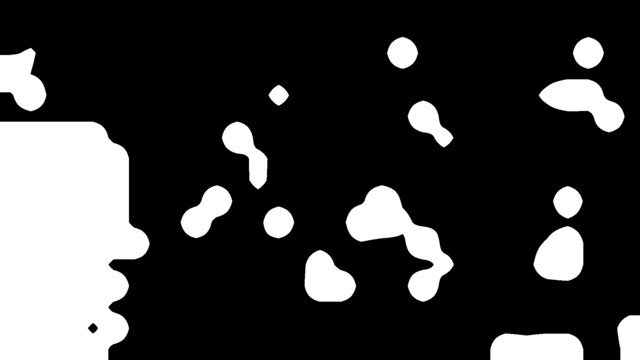}}}
        \fbox{\includegraphics[width=0.100\textwidth]{{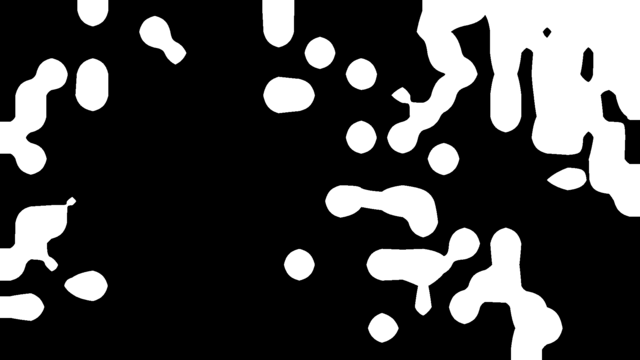}}}
        \fbox{\includegraphics[width=0.100\textwidth]{{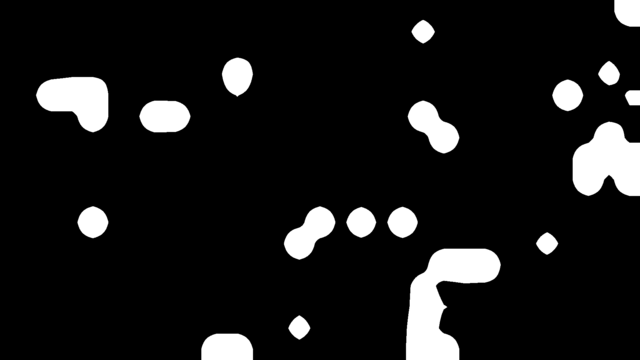}}}
        \fbox{\includegraphics[width=0.100\textwidth]{{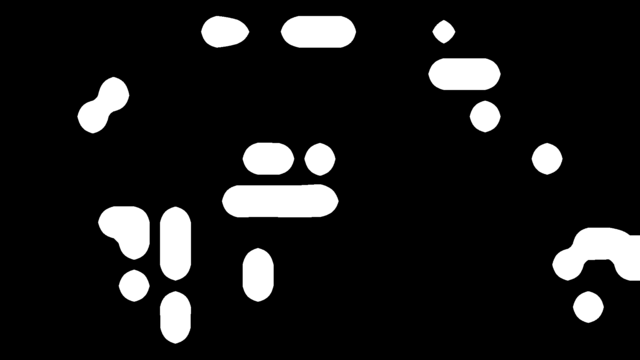}}}
        \smallskip
    \end{minipage}

	\vspace*{-0.1\baselineskip}

    \begin{minipage}[t]{1\textwidth}
        \makebox[0.083\textwidth][r]{\raisebox{15pt}{\smaller Noiseprint\hspace{6pt}}}
        \fbox{\includegraphics[width=0.100\textwidth]{{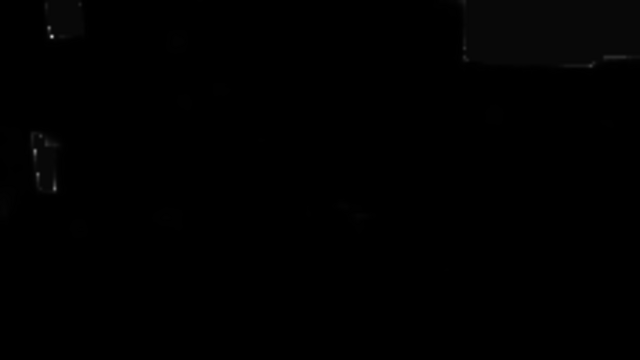}}}
        \fbox{\includegraphics[width=0.100\textwidth]{{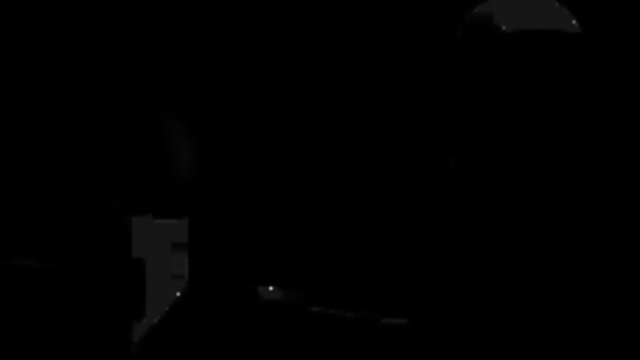}}}
        \fbox{\includegraphics[width=0.100\textwidth]{{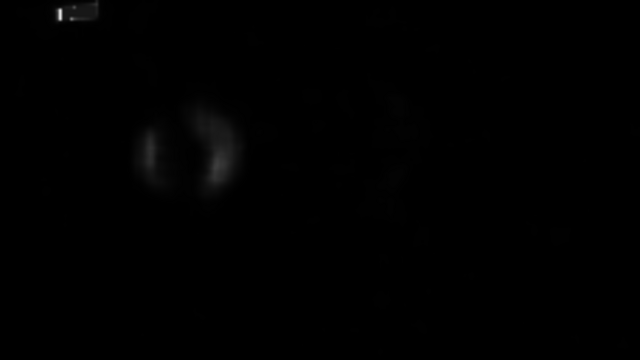}}}
        \fbox{\includegraphics[width=0.100\textwidth]{{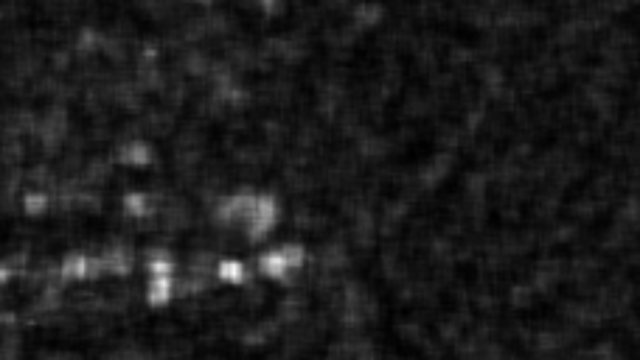}}}
        \fbox{\includegraphics[width=0.100\textwidth]{{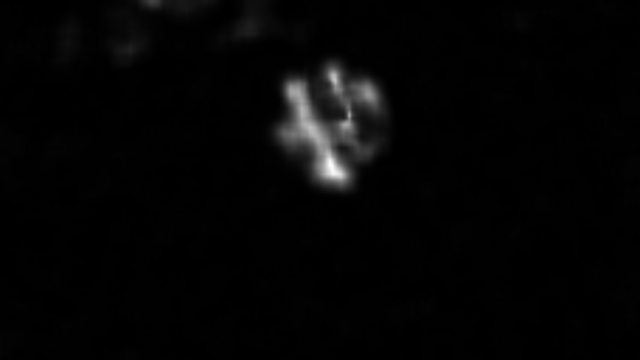}}}
        \fbox{\includegraphics[width=0.100\textwidth]{{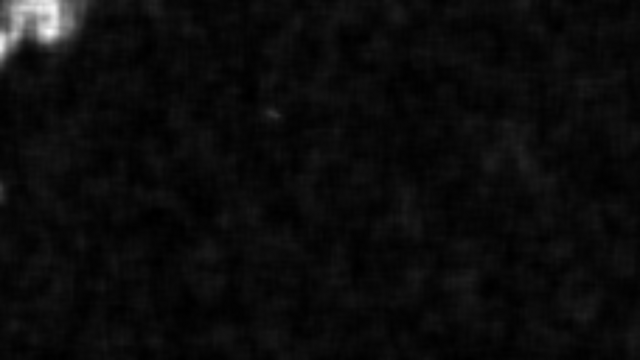}}}
        \fbox{\includegraphics[width=0.100\textwidth]{{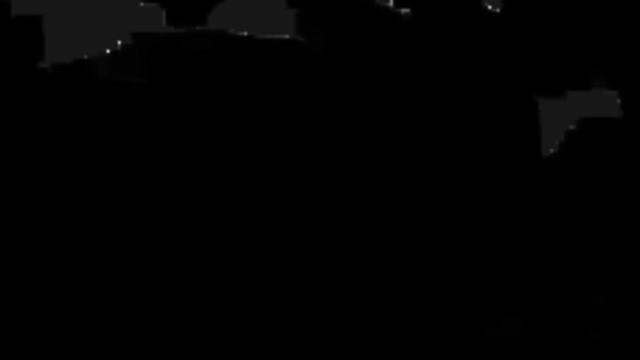}}}
        \fbox{\includegraphics[width=0.100\textwidth]{{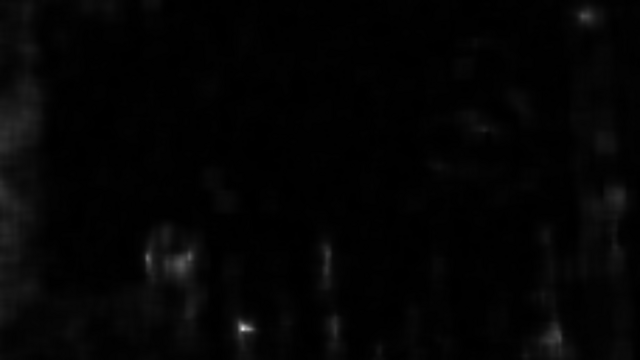}}}
        \smallskip
    \end{minipage}

	\vspace*{-0.1\baselineskip}

    \begin{minipage}[t]{1\textwidth}
        \makebox[0.083\textwidth][r]{\raisebox{15pt}{\smaller ManTra-Net\hspace{6pt}}}
        \fbox{\includegraphics[width=0.100\textwidth]{{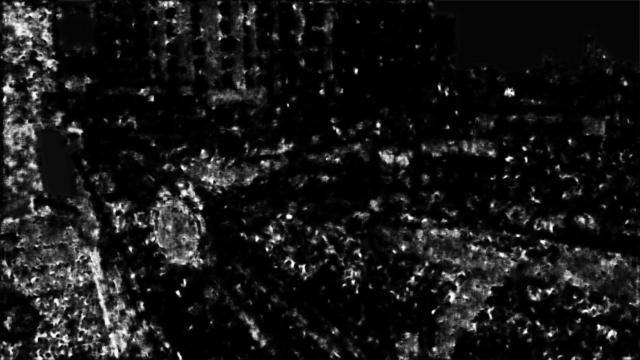}}}
        \fbox{\includegraphics[width=0.100\textwidth]{{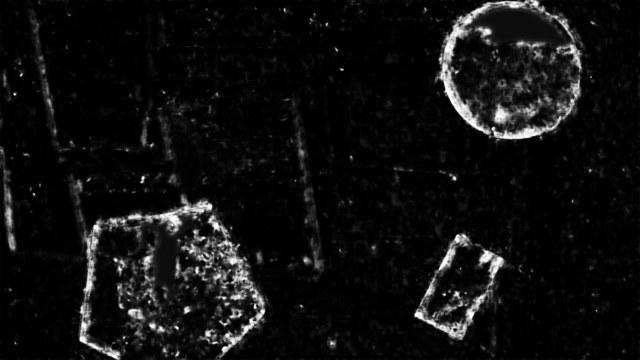}}}
        \fbox{\includegraphics[width=0.100\textwidth]{{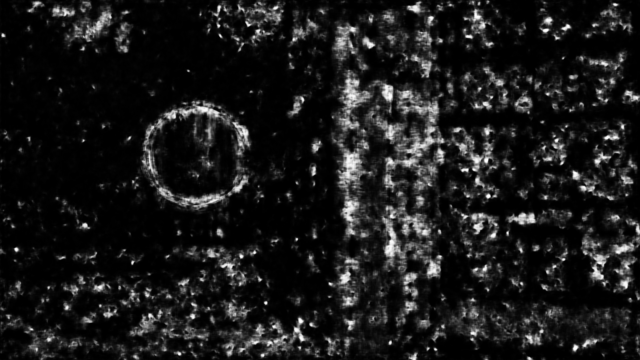}}}
        \fbox{\includegraphics[width=0.100\textwidth]{{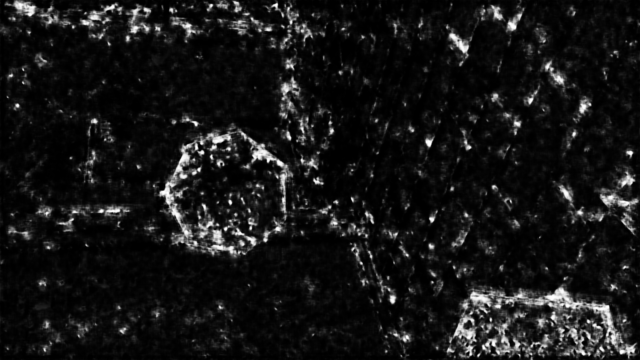}}}
        \fbox{\includegraphics[width=0.100\textwidth]{{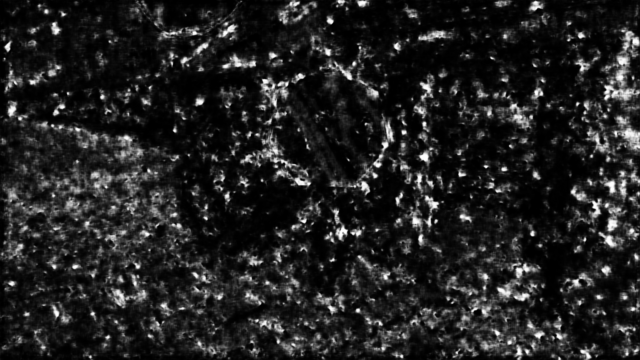}}}
        \fbox{\includegraphics[width=0.100\textwidth]{{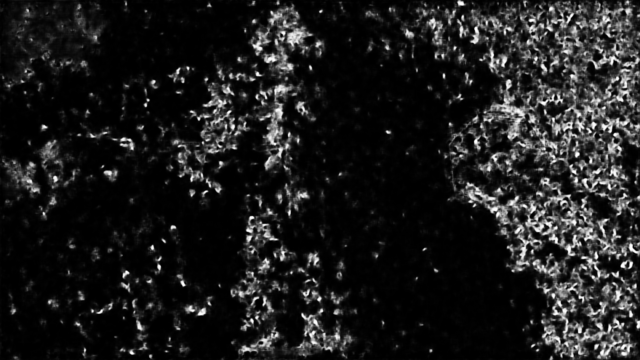}}}
        \fbox{\includegraphics[width=0.100\textwidth]{{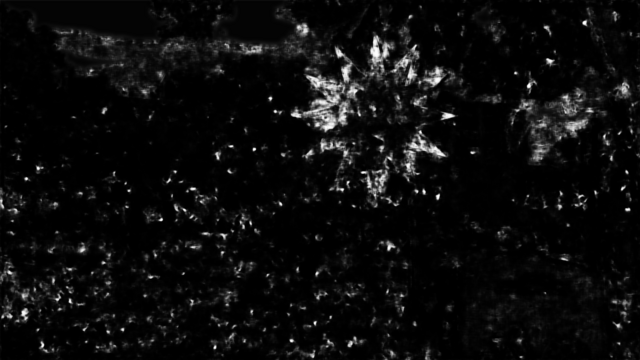}}}
        \fbox{\includegraphics[width=0.100\textwidth]{{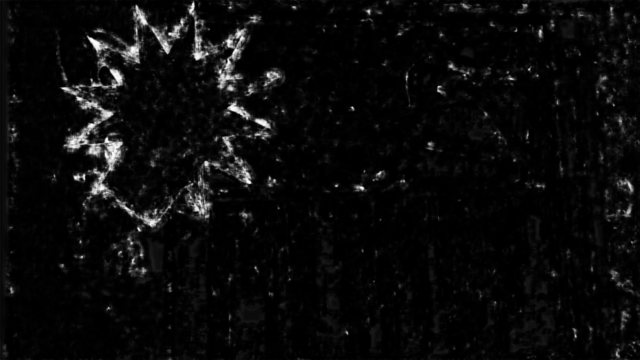}}}
        \smallskip
    \end{minipage}

	\vspace*{-0.1\baselineskip}

    \begin{minipage}[t]{1\textwidth}
        \makebox[0.083\textwidth][r]{\raisebox{15pt}{\smaller MVSS-Net\hspace{6pt}}}
        \fbox{\includegraphics[width=0.100\textwidth]{{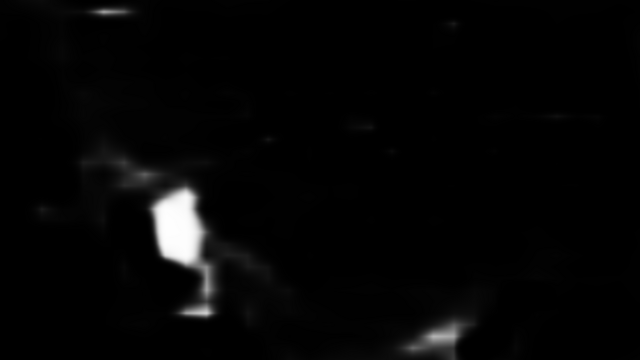}}}
        \fbox{\includegraphics[width=0.100\textwidth]{{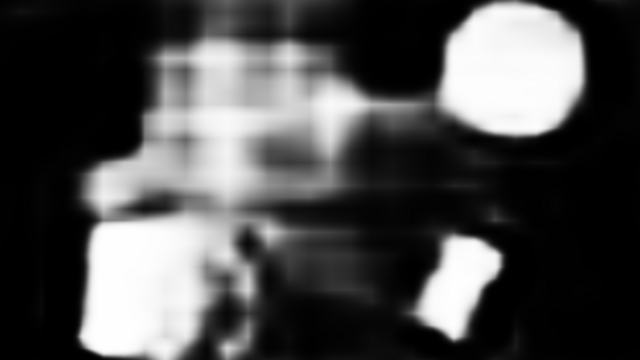}}}
        \fbox{\includegraphics[width=0.100\textwidth]{{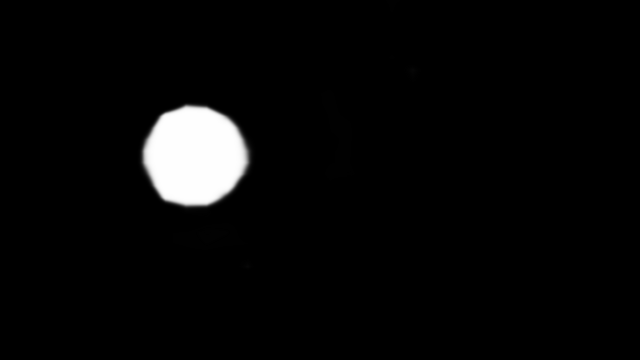}}}
        \fbox{\includegraphics[width=0.100\textwidth]{{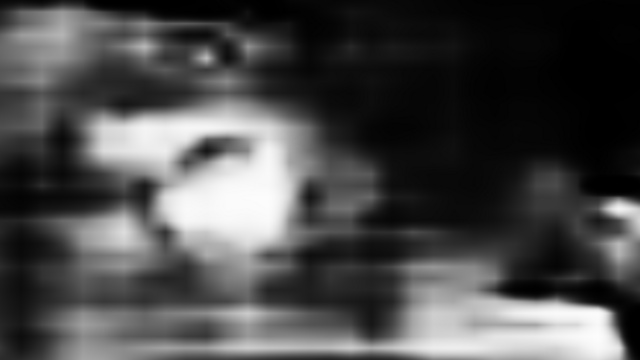}}}
        \fbox{\includegraphics[width=0.100\textwidth]{{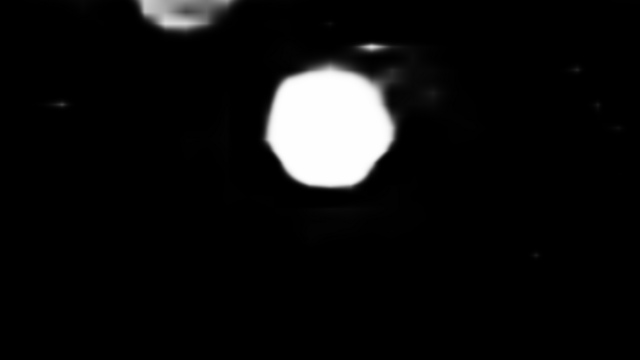}}}
        \fbox{\includegraphics[width=0.100\textwidth]{{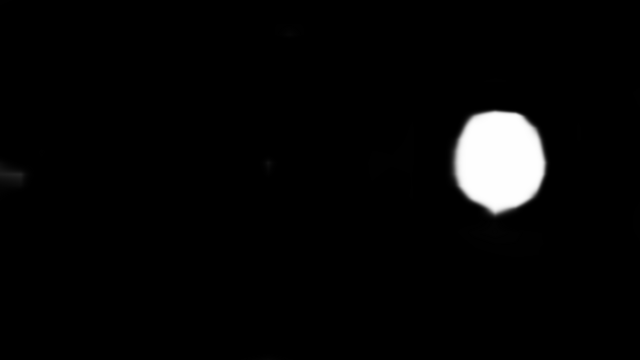}}}
        \fbox{\includegraphics[width=0.100\textwidth]{{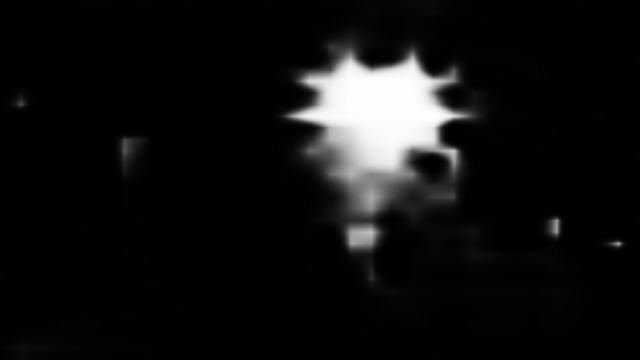}}}
        \fbox{\includegraphics[width=0.100\textwidth]{{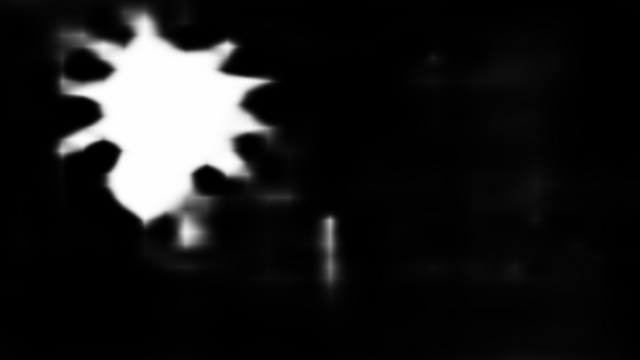}}}
        \smallskip
    \end{minipage}
    
	\vspace*{-0.5\baselineskip}

    \caption{\label{fig:vcms_localization_results} 
    This figure shows the localization results of different networks on the VCMS dataset. Our proposed network's localization results are good, with some minor false alarms on column 1, 3, 6 and 9. We note that, our predicted masks are blobs, which means a large source of our pixel-level localization error comes from the fact that we cannot predict shapes with sharp, concave edges. By contrast, networks, which leverage edge information, like ManTra-Net and MVSS-Net were able to produce masks with sharp edges. Hence, MVSS-Net were comparable to our network in terms of localization performance. On the other hand, FSG, EXIFnet and Noiseprint did not seem like they could make reasonable predictions about the manipulation region.
    }

\end{figure*}


\begin{figure*}[!t]

    \centering
    \setlength{\fboxsep}{0pt}
    \begin{minipage}[t]{1\textwidth}
        \makebox[0.083\textwidth][r]{\raisebox{15pt}{\smaller Frame\hspace{6pt}}}
        \fbox{\includegraphics[width=0.100\textwidth]{{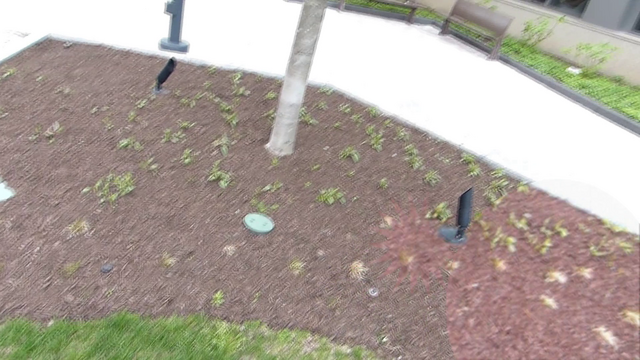}}}
        \fbox{\includegraphics[width=0.100\textwidth]{{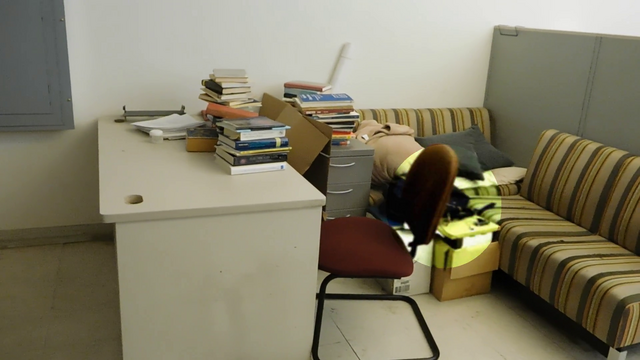}}}
        \fbox{\includegraphics[width=0.100\textwidth]{{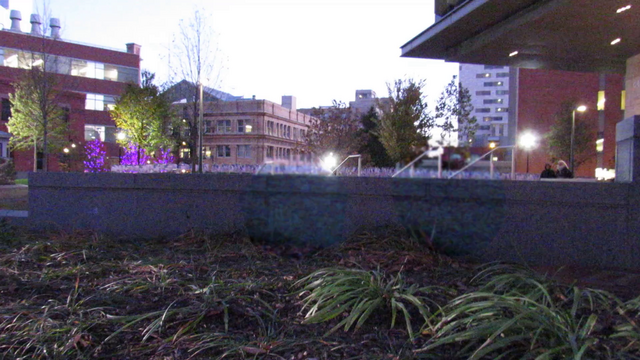}}}
        \fbox{\includegraphics[width=0.100\textwidth]{{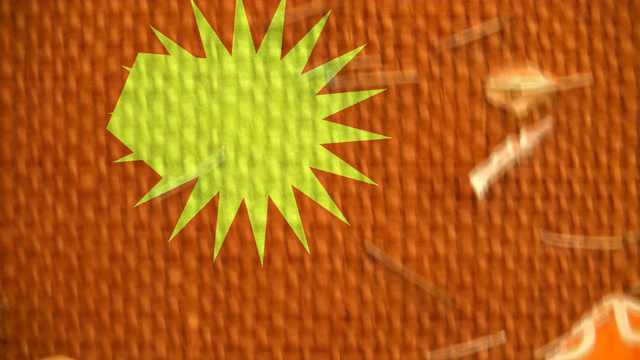}}}
        \fbox{\includegraphics[width=0.100\textwidth]{{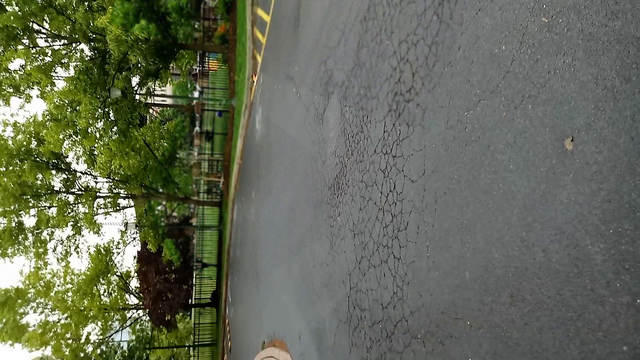}}}
        \fbox{\includegraphics[width=0.100\textwidth]{{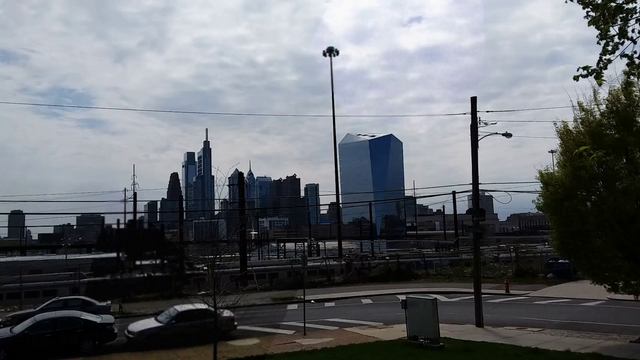}}}
        \fbox{\includegraphics[width=0.100\textwidth]{{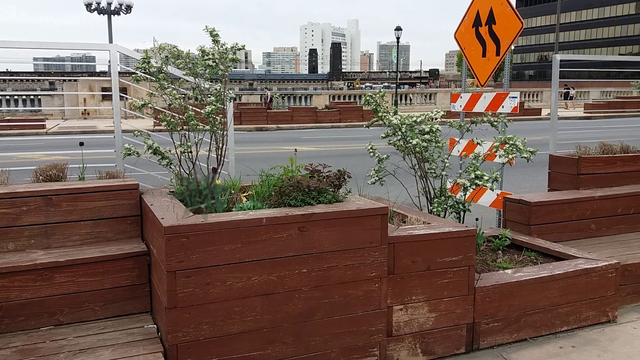}}}
        \fbox{\includegraphics[width=0.100\textwidth]{{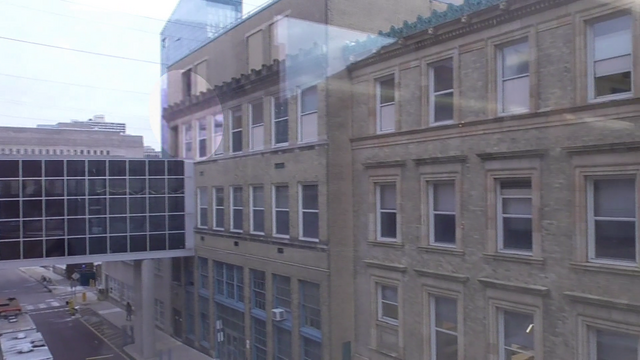}}}
        \smallskip
    \end{minipage}

	\vspace*{-0.1\baselineskip}

    \begin{minipage}[t]{1\textwidth}
        \makebox[0.083\textwidth][r]{\raisebox{15pt}{\smaller Mask\hspace{6pt}}}
        \fbox{\includegraphics[width=0.100\textwidth]{{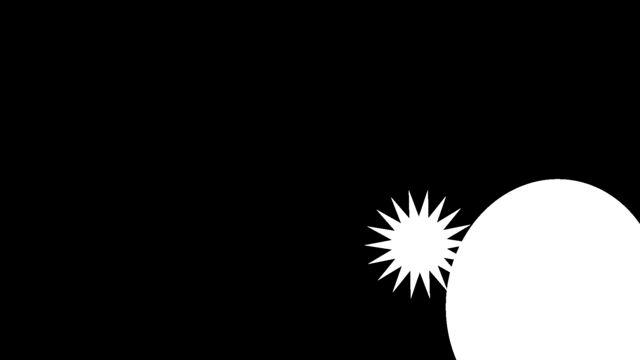}}}
        \fbox{\includegraphics[width=0.100\textwidth]{{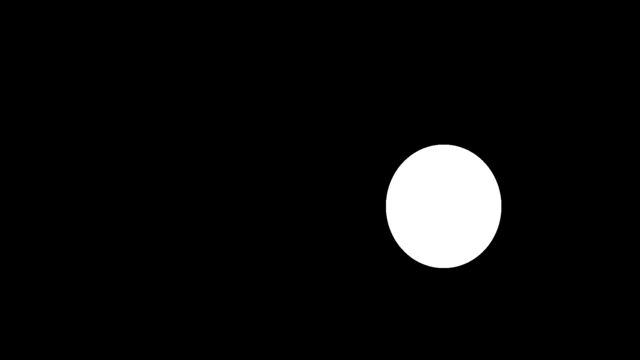}}}
        \fbox{\includegraphics[width=0.100\textwidth]{{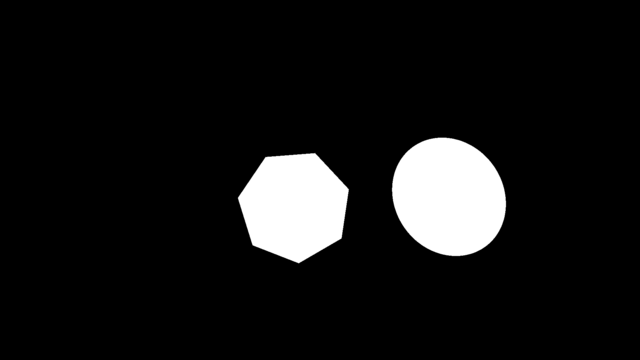}}}
        \fbox{\includegraphics[width=0.100\textwidth]{{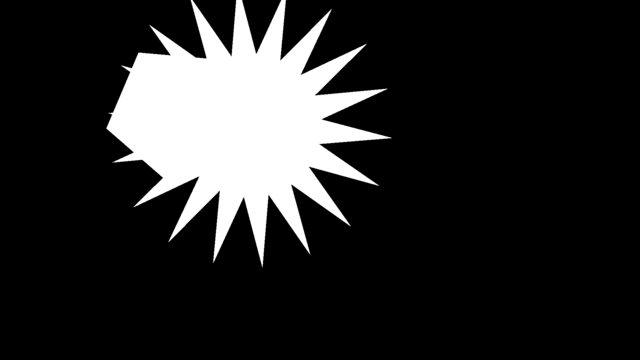}}}
        \fbox{\includegraphics[width=0.100\textwidth]{{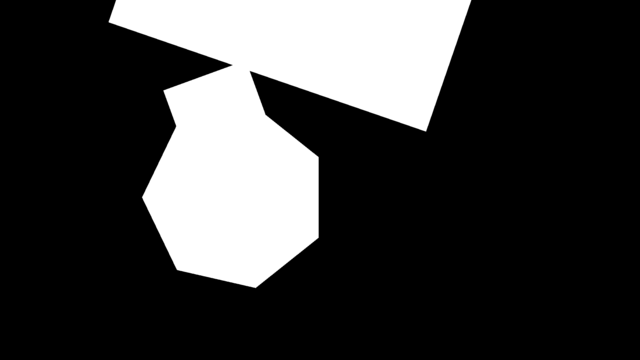}}}
        \fbox{\includegraphics[width=0.100\textwidth]{{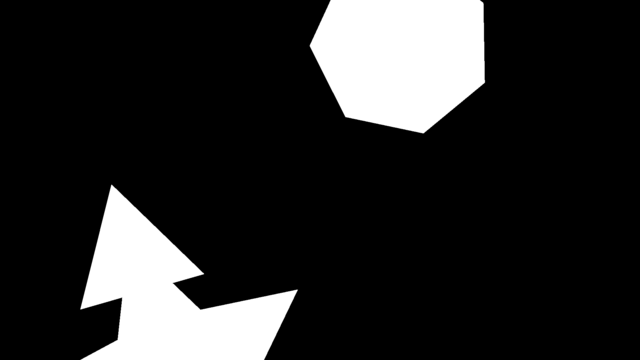}}}
        \fbox{\includegraphics[width=0.100\textwidth]{{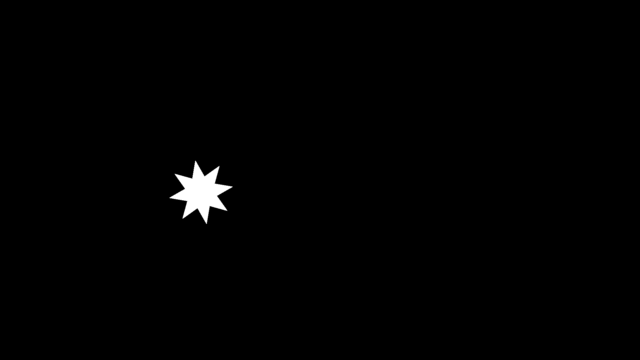}}}
        \fbox{\includegraphics[width=0.100\textwidth]{{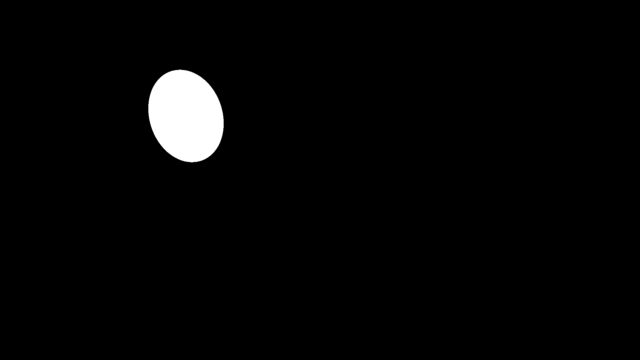}}}
        \smallskip
    \end{minipage}

	\vspace*{-0.1\baselineskip}

    \begin{minipage}[t]{1\textwidth}
        \makebox[0.083\textwidth][r]{\raisebox{15pt}{\smaller Proposed\hspace{6pt}}}
        \fbox{\includegraphics[width=0.100\textwidth]{{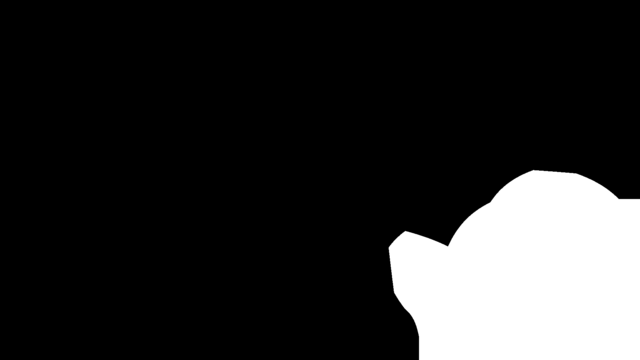}}}
        \fbox{\includegraphics[width=0.100\textwidth]{{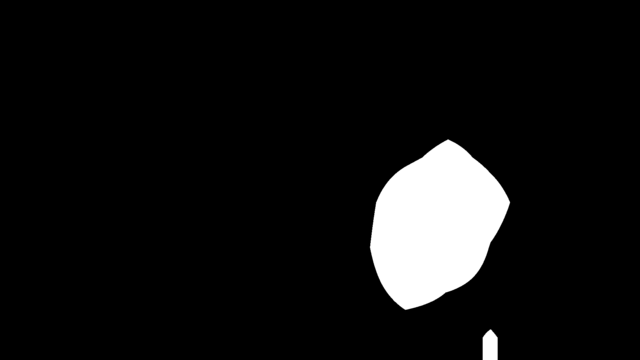}}}
        \fbox{\includegraphics[width=0.100\textwidth]{{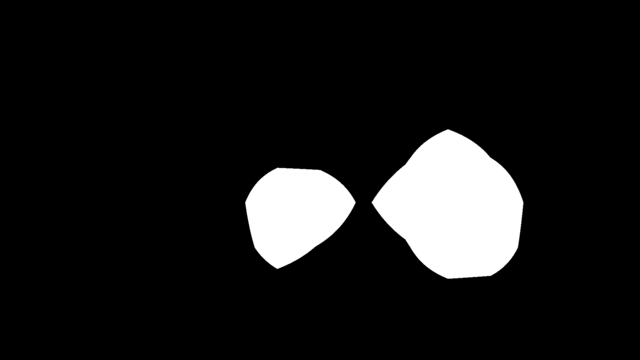}}}
        \fbox{\includegraphics[width=0.100\textwidth]{{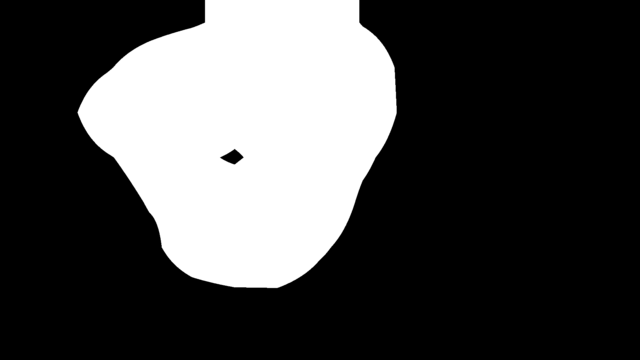}}}
        \fbox{\includegraphics[width=0.100\textwidth]{{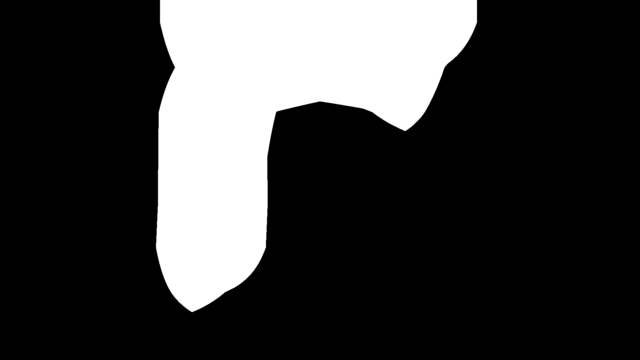}}}
        \fbox{\includegraphics[width=0.100\textwidth]{{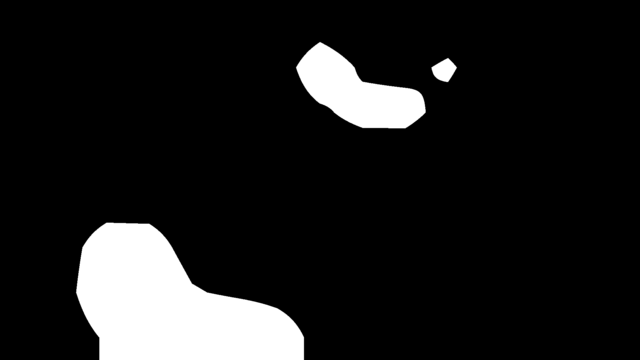}}}
        \fbox{\includegraphics[width=0.100\textwidth]{{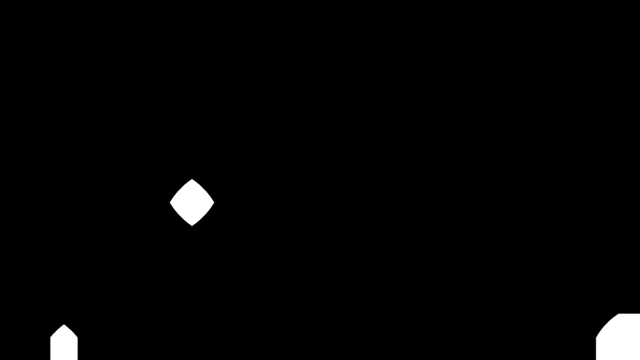}}}
        \fbox{\includegraphics[width=0.100\textwidth]{{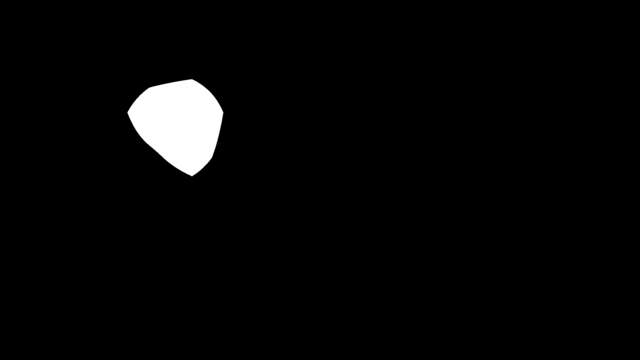}}}
        \smallskip
    \end{minipage}

	\vspace*{-0.1\baselineskip}

    \begin{minipage}[t]{1\textwidth}
        \makebox[0.083\textwidth][r]{\raisebox{15pt}{\smaller FSG\hspace{6pt}}}
        \fbox{\includegraphics[width=0.100\textwidth]{{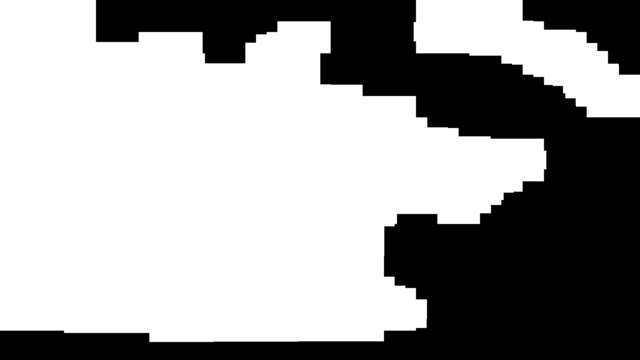}}}
        \fbox{\includegraphics[width=0.100\textwidth]{{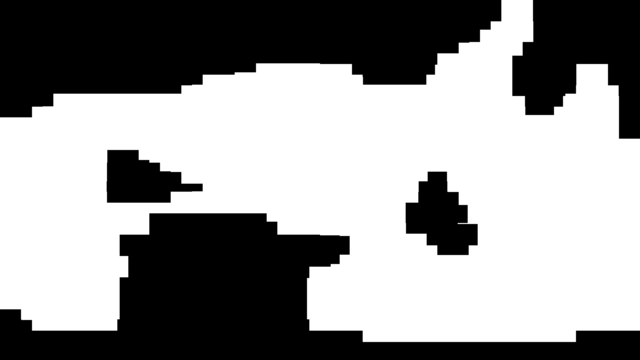}}}
        \fbox{\includegraphics[width=0.100\textwidth]{{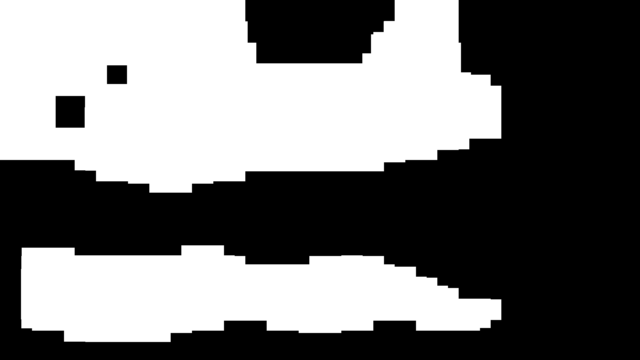}}}
        \fbox{\includegraphics[width=0.100\textwidth]{{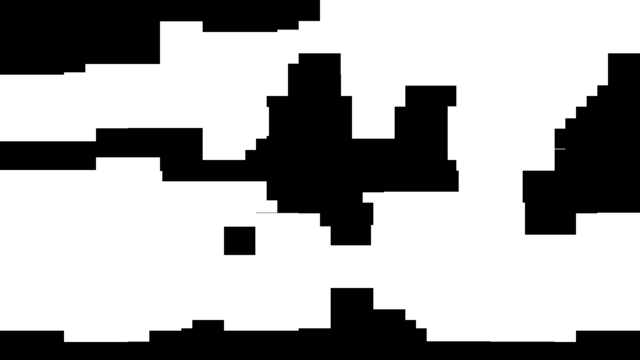}}}
        \fbox{\includegraphics[width=0.100\textwidth]{{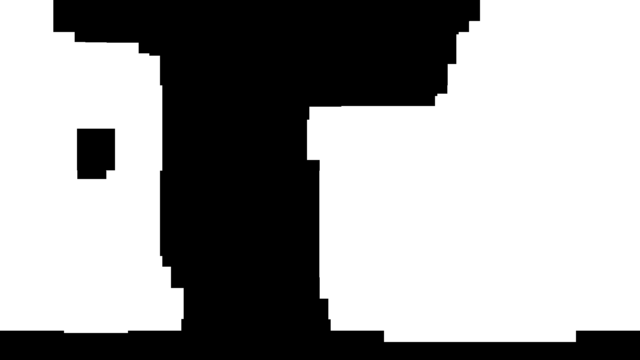}}}
        \fbox{\includegraphics[width=0.100\textwidth]{{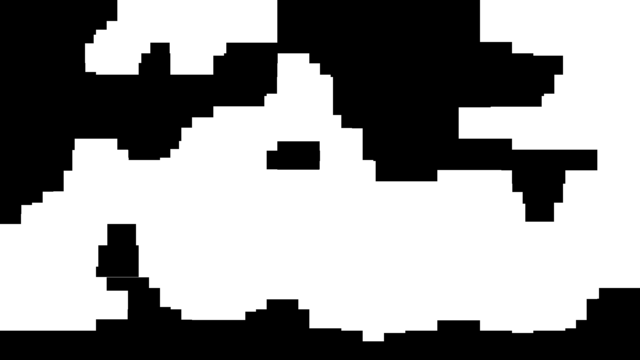}}}
        \fbox{\includegraphics[width=0.100\textwidth]{{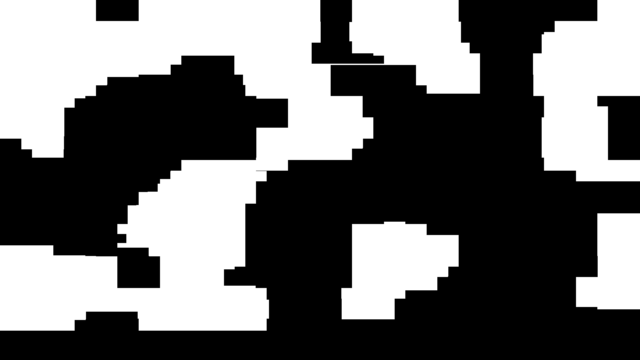}}}
        \fbox{\includegraphics[width=0.100\textwidth]{{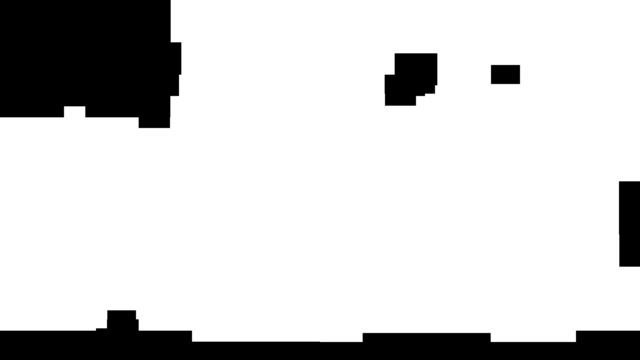}}}
        \smallskip
    \end{minipage}

	\vspace*{-0.1\baselineskip}

    \begin{minipage}[t]{1\textwidth}
        \makebox[0.083\textwidth][r]{\raisebox{15pt}{\smaller EXIFnet\hspace{6pt}}}
        \fbox{\includegraphics[width=0.100\textwidth]{{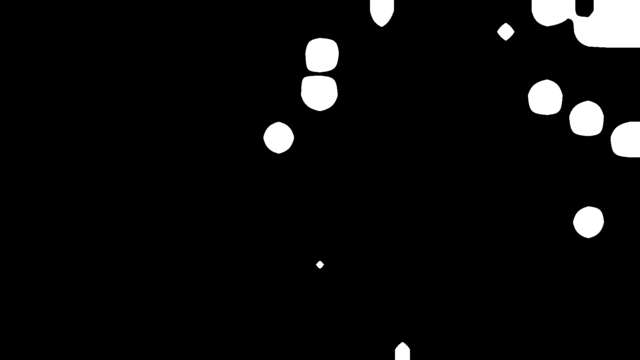}}}
        \fbox{\includegraphics[width=0.100\textwidth]{{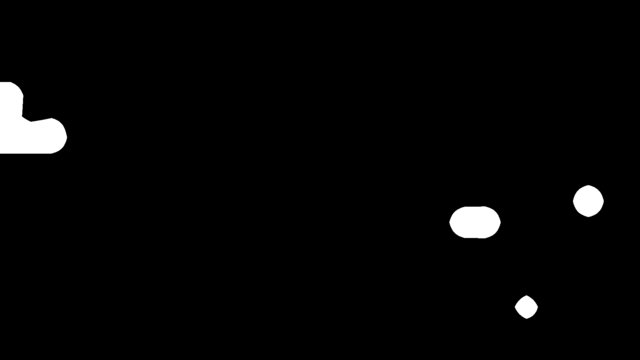}}}
        \fbox{\includegraphics[width=0.100\textwidth]{{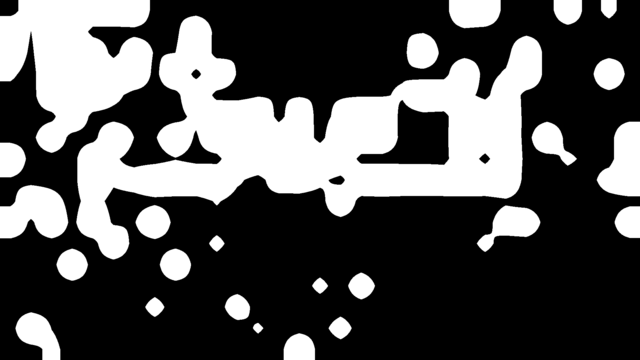}}}
        \fbox{\includegraphics[width=0.100\textwidth]{{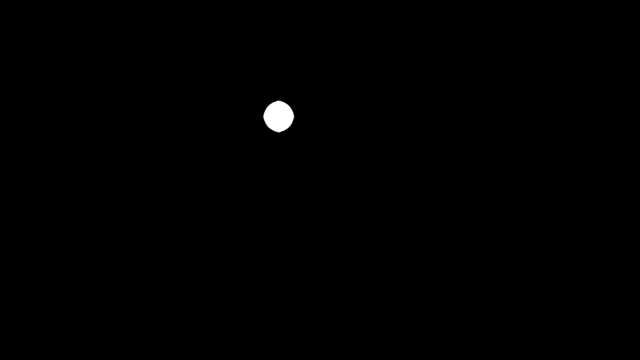}}}
        \fbox{\includegraphics[width=0.100\textwidth]{{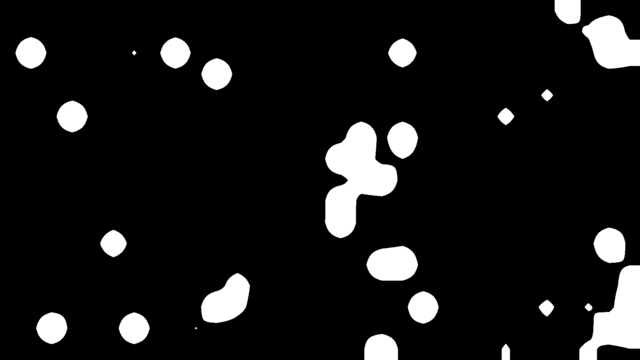}}}
        \fbox{\includegraphics[width=0.100\textwidth]{{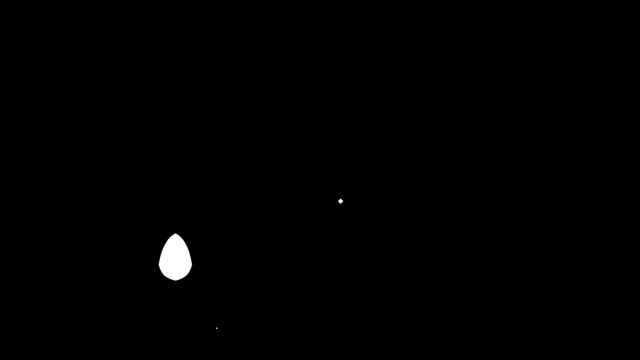}}}
        \fbox{\includegraphics[width=0.100\textwidth]{{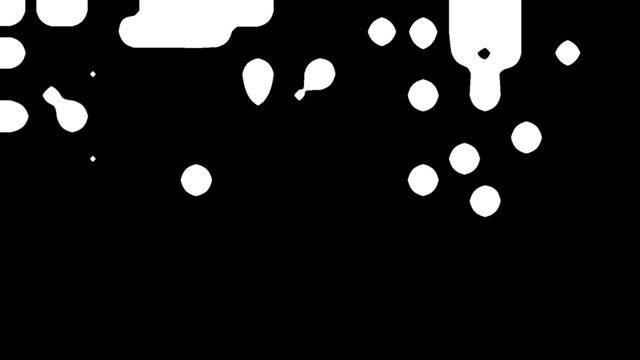}}}
        \fbox{\includegraphics[width=0.100\textwidth]{{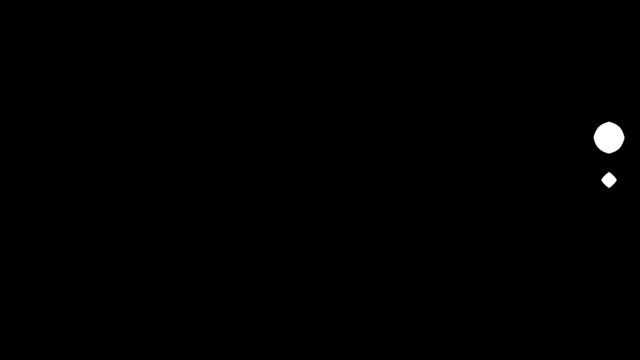}}}
        \smallskip
    \end{minipage}

	\vspace*{-0.1\baselineskip}

    \begin{minipage}[t]{1\textwidth}
        \makebox[0.083\textwidth][r]{\raisebox{15pt}{\smaller Noiseprint\hspace{6pt}}}
        \fbox{\includegraphics[width=0.100\textwidth]{{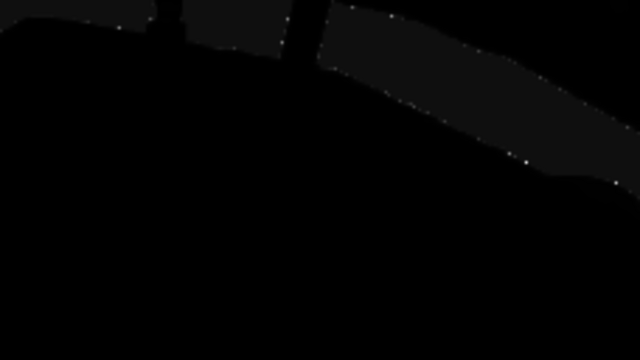}}}
        \fbox{\includegraphics[width=0.100\textwidth]{{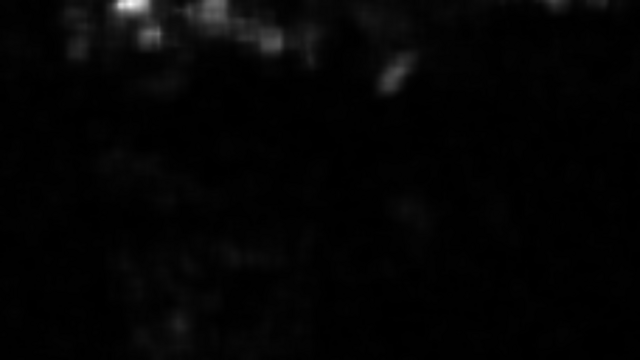}}}
        \fbox{\includegraphics[width=0.100\textwidth]{{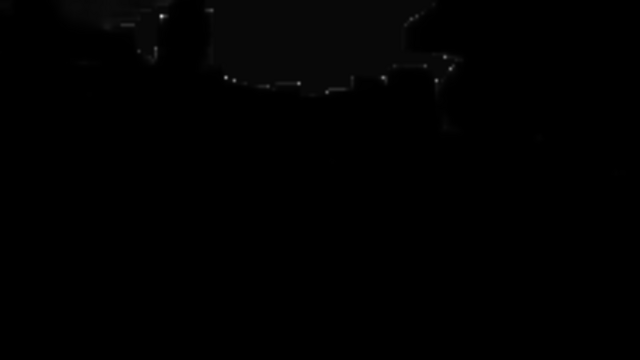}}}
        \fbox{\includegraphics[width=0.100\textwidth]{{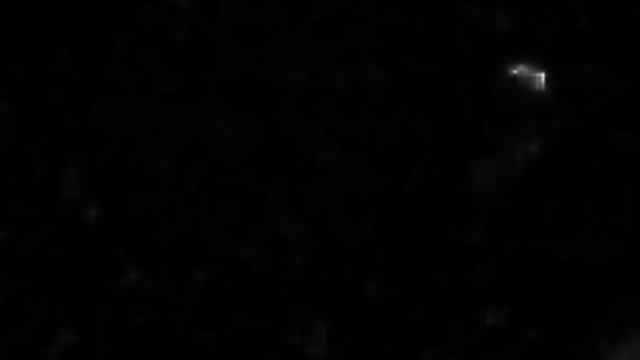}}}
        \fbox{\includegraphics[width=0.100\textwidth]{{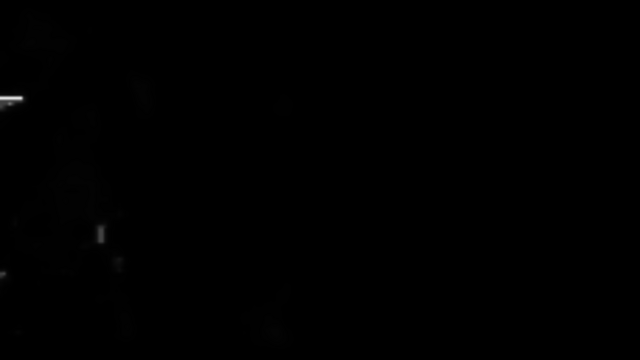}}}
        \fbox{\includegraphics[width=0.100\textwidth]{{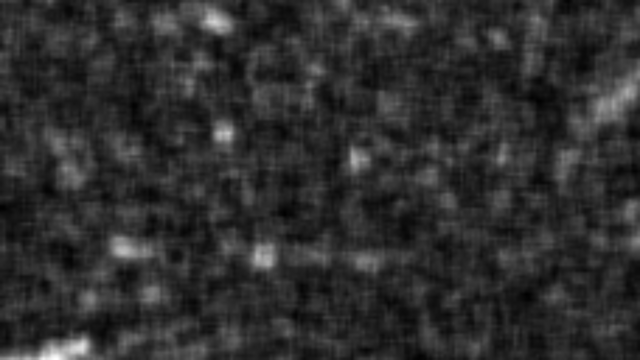}}}
        \fbox{\includegraphics[width=0.100\textwidth]{{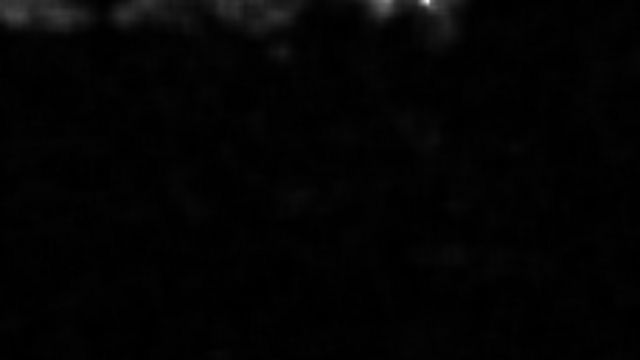}}}
        \fbox{\includegraphics[width=0.100\textwidth]{{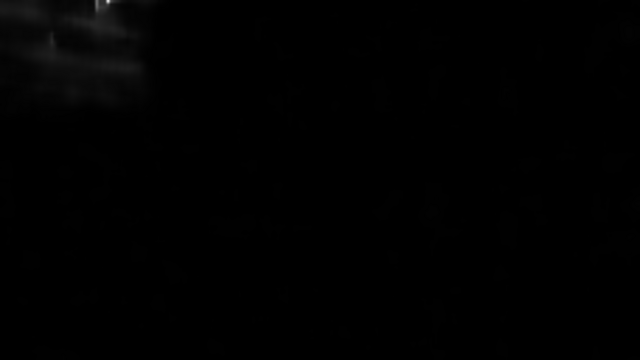}}}
        \smallskip
    \end{minipage}

	\vspace*{-0.1\baselineskip}

    \begin{minipage}[t]{1\textwidth}
        \makebox[0.083\textwidth][r]{\raisebox{15pt}{\smaller ManTra-Net\hspace{6pt}}}
        \fbox{\includegraphics[width=0.100\textwidth]{{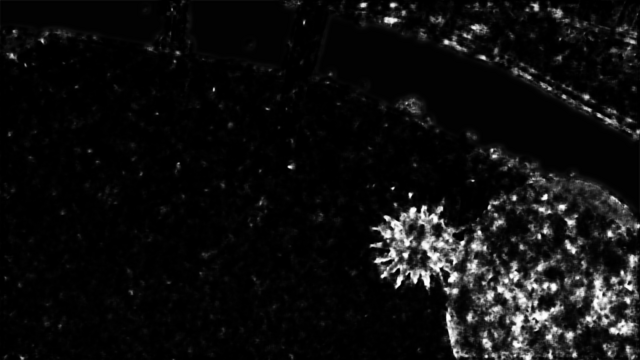}}}
        \fbox{\includegraphics[width=0.100\textwidth]{{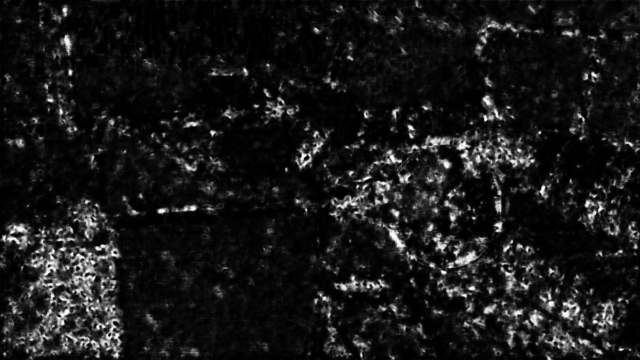}}}
        \fbox{\includegraphics[width=0.100\textwidth]{{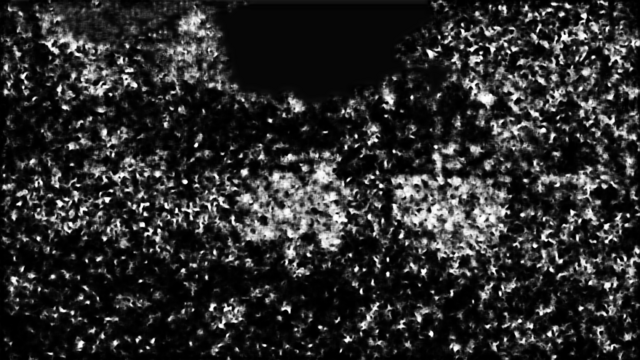}}}
        \fbox{\includegraphics[width=0.100\textwidth]{{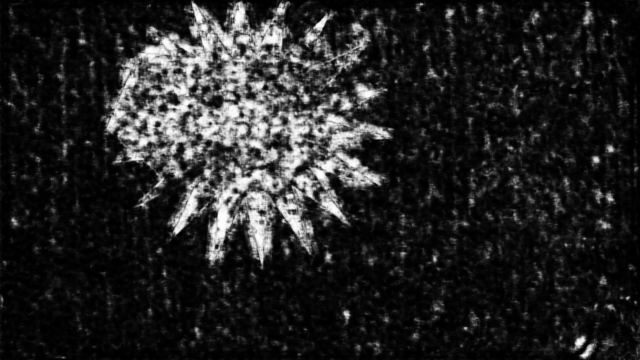}}}
        \fbox{\includegraphics[width=0.100\textwidth]{{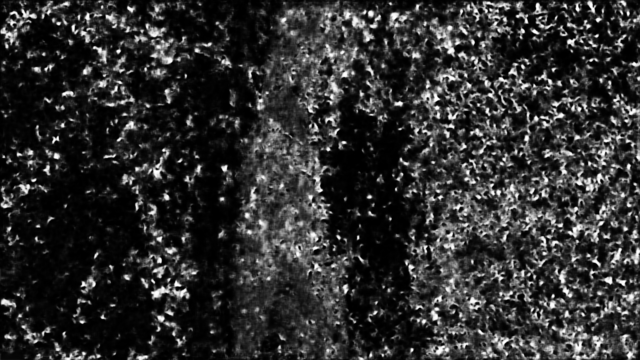}}}
        \fbox{\includegraphics[width=0.100\textwidth]{{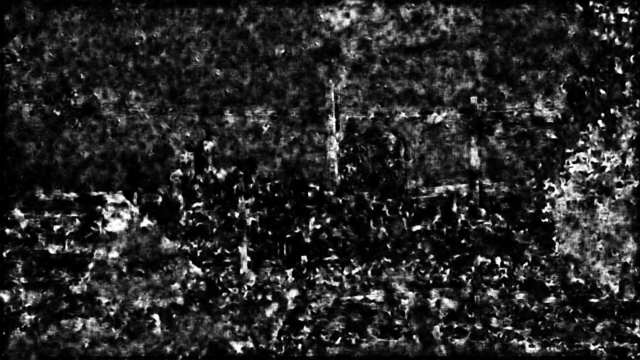}}}
        \fbox{\includegraphics[width=0.100\textwidth]{{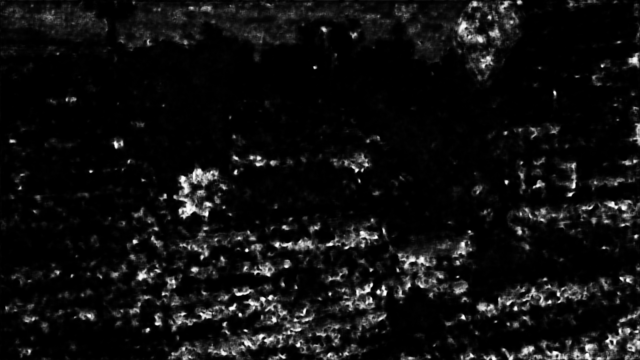}}}
        \fbox{\includegraphics[width=0.100\textwidth]{{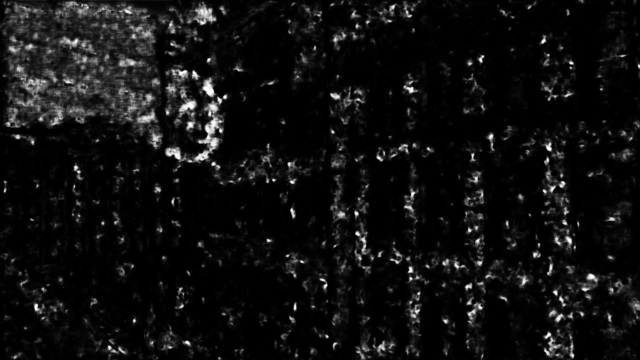}}}
        \smallskip
    \end{minipage}

	\vspace*{-0.1\baselineskip}

    \begin{minipage}[t]{1\textwidth}
        \makebox[0.083\textwidth][r]{\raisebox{15pt}{\smaller MVSS-Net\hspace{6pt}}}
        \fbox{\includegraphics[width=0.100\textwidth]{{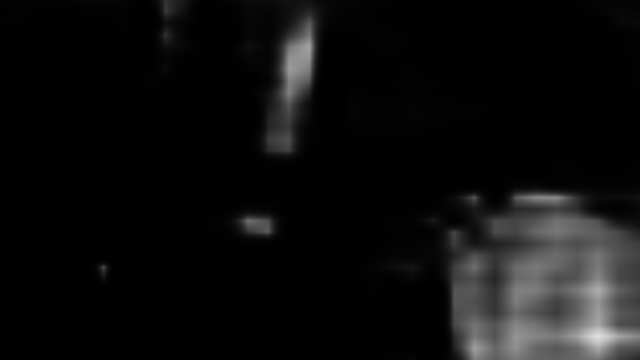}}}
        \fbox{\includegraphics[width=0.100\textwidth]{{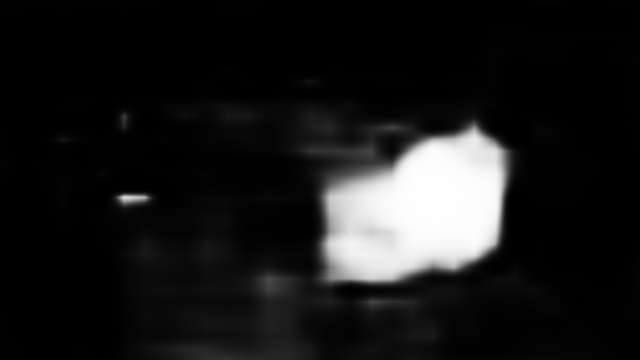}}}
        \fbox{\includegraphics[width=0.100\textwidth]{{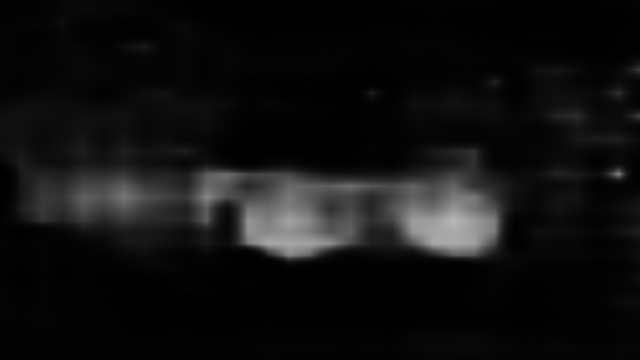}}}
        \fbox{\includegraphics[width=0.100\textwidth]{{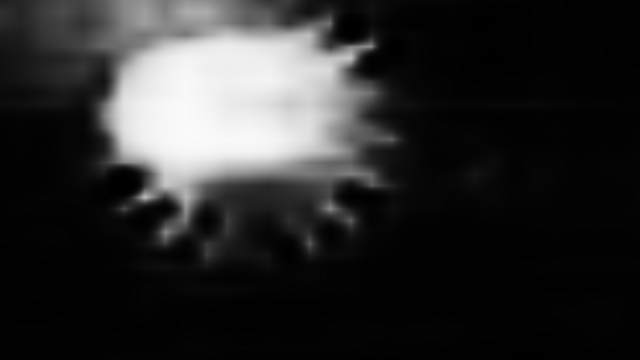}}}
        \fbox{\includegraphics[width=0.100\textwidth]{{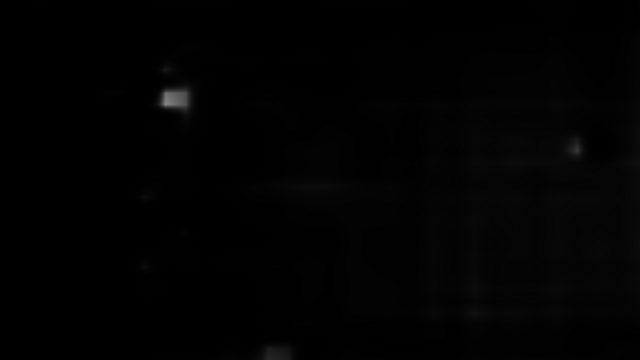}}}
        \fbox{\includegraphics[width=0.100\textwidth]{{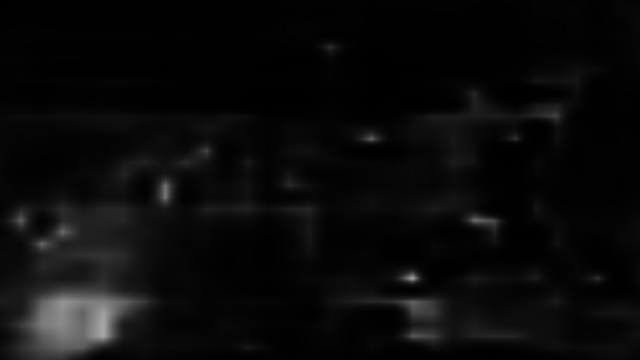}}}
        \fbox{\includegraphics[width=0.100\textwidth]{{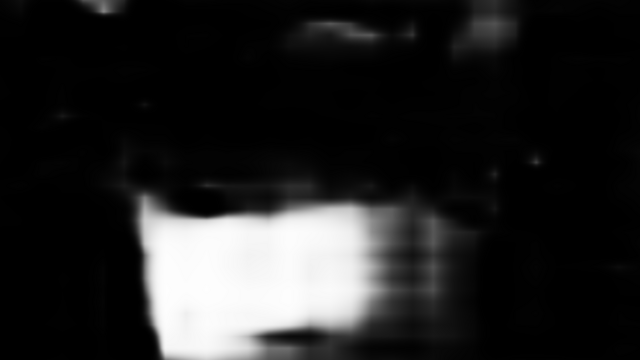}}}
        \fbox{\includegraphics[width=0.100\textwidth]{{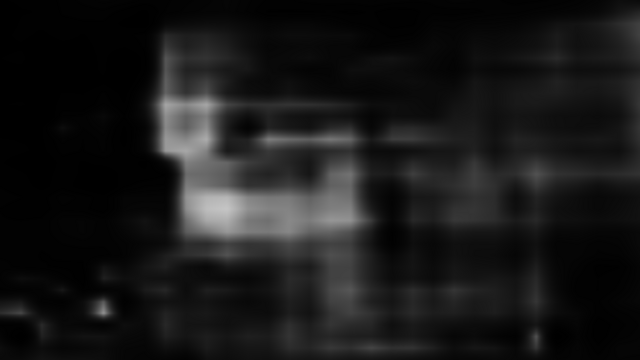}}}
        \smallskip
    \end{minipage}
    
	\vspace*{-0.5\baselineskip}

    \caption{\label{fig:vpvm_localization_results} 
    This figure shows the localization results of different networks on the VPVM dataset. Our proposed network's localization results are good, with some minor false alarms on column 2 and 7. Again, our predicted masks are blobs, which means a large source of our error comes from the fact that we cannot predict shapes with sharp, concave edges. Nonetheless, contrast to results on VCMS, strong competitors like MVSS-Net and Mantra-Net could not identify the manipulated region unless it was extremely visible (e.g column 1, 2, 3 and 4). Other examples contained manipulations which resulted in perceptually visible edges but it seemed like competing algorithms false alarmed on regions which had distinct textures versus the rest of the frame, such as they sky (column 3, 6, 8), the colored bricks (column 7), and the building (column 8).
    }
    
\end{figure*}


\begin{figure*}[!t]
    \centering
    \setlength{\fboxsep}{0pt}

    \begin{minipage}[t]{1\textwidth}
        \makebox[0.083\textwidth][r]{\raisebox{15pt}{\smaller Frame\hspace{6pt}}}
        \fbox{\includegraphics[width=0.100\textwidth]{{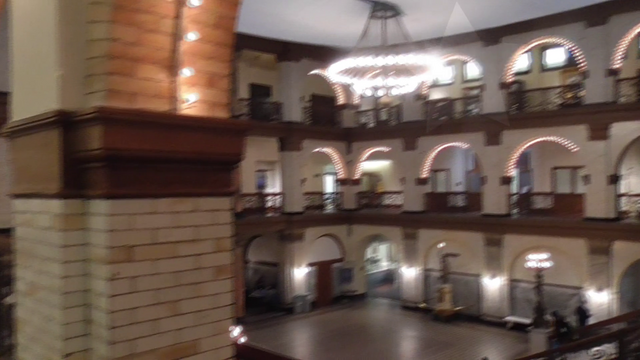}}}
        \fbox{\includegraphics[width=0.100\textwidth]{{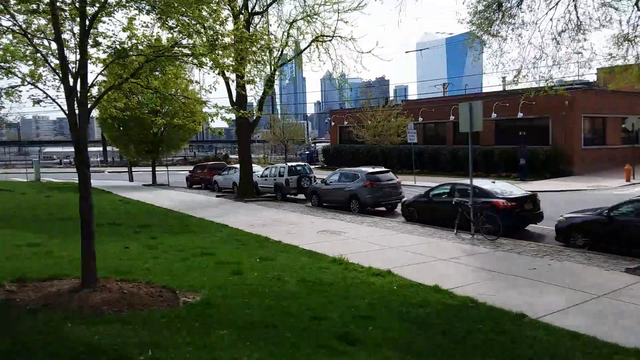}}}
        \fbox{\includegraphics[width=0.100\textwidth]{{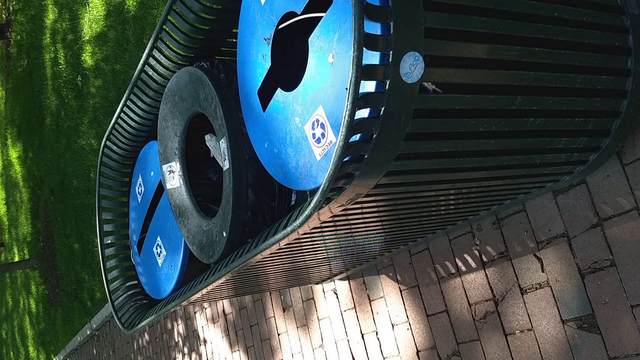}}}
        \fbox{\includegraphics[width=0.100\textwidth]{{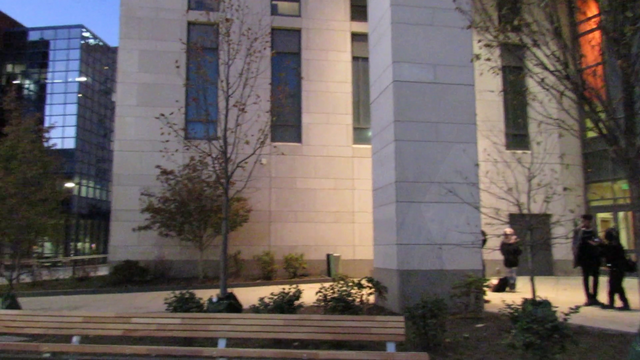}}}
        \fbox{\includegraphics[width=0.100\textwidth]{{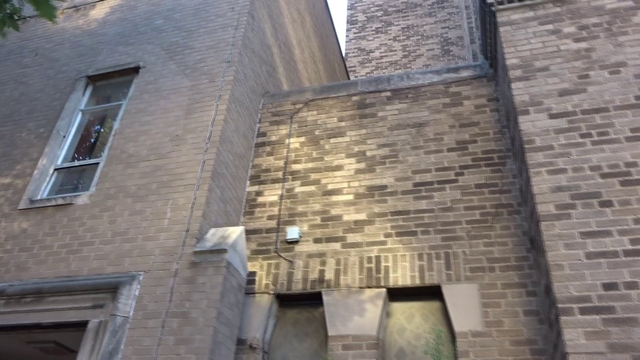}}}
        \fbox{\includegraphics[width=0.100\textwidth]{{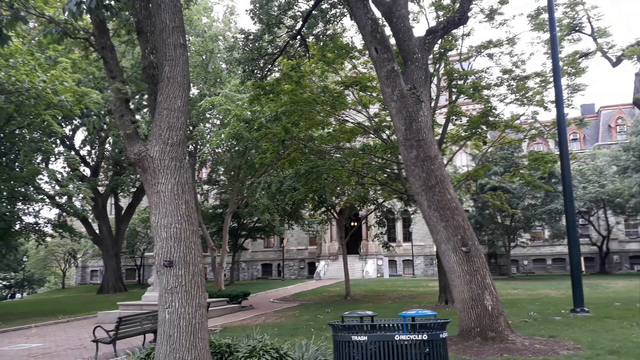}}}
        \fbox{\includegraphics[width=0.100\textwidth]{{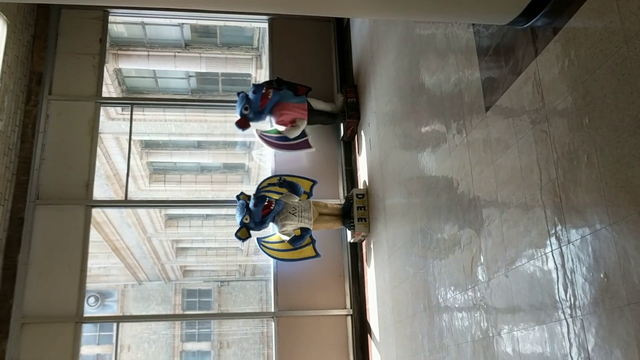}}}
        \fbox{\includegraphics[width=0.100\textwidth]{{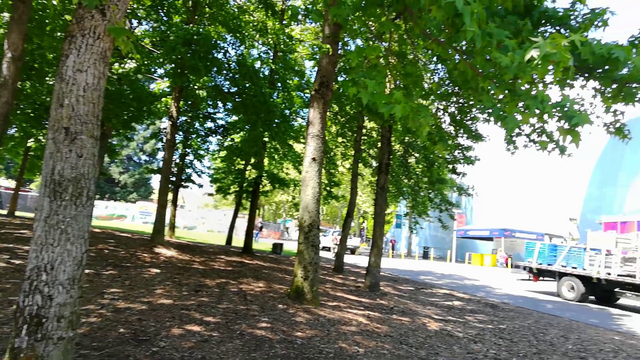}}}
        \smallskip
    \end{minipage}

	\vspace*{-0.1\baselineskip}

    \begin{minipage}[t]{1\textwidth}
        \makebox[0.083\textwidth][r]{\raisebox{15pt}{\smaller Mask\hspace{6pt}}}
        \fbox{\includegraphics[width=0.100\textwidth]{{}}}
        \fbox{\includegraphics[width=0.100\textwidth]{{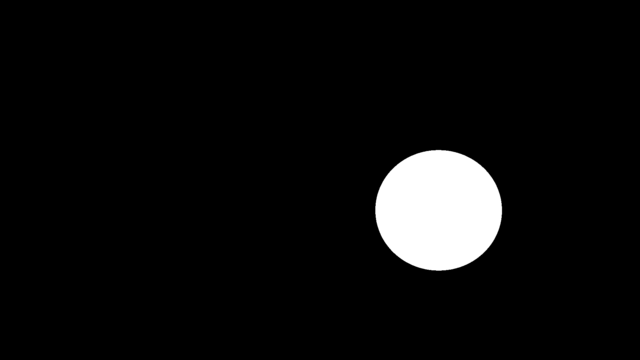}}}
        \fbox{\includegraphics[width=0.100\textwidth]{{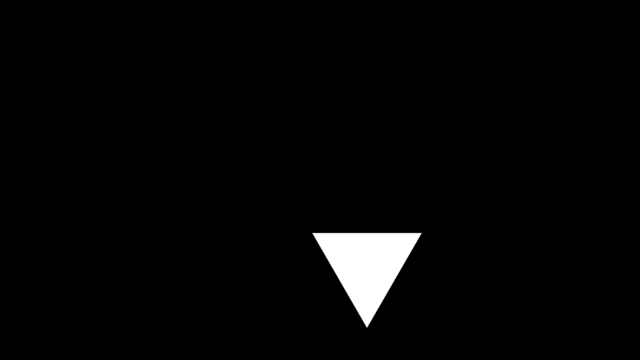}}}
        \fbox{\includegraphics[width=0.100\textwidth]{{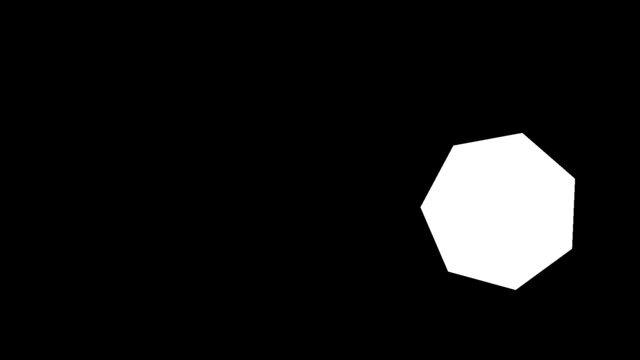}}}
        \fbox{\includegraphics[width=0.100\textwidth]{{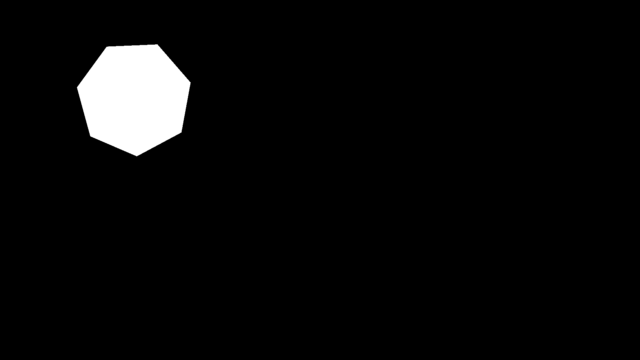}}}
        \fbox{\includegraphics[width=0.100\textwidth]{{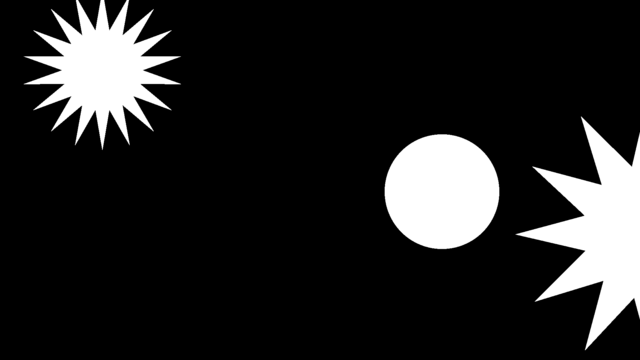}}}
        \fbox{\includegraphics[width=0.100\textwidth]{{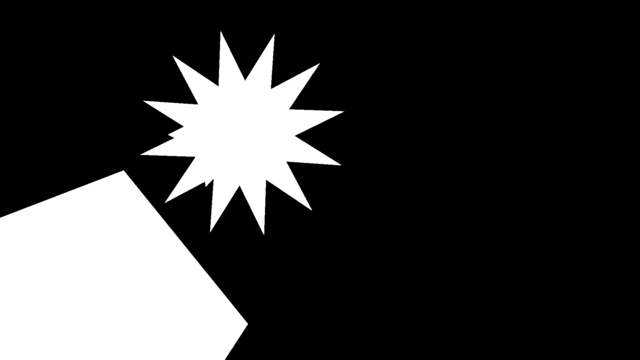}}}
        \fbox{\includegraphics[width=0.100\textwidth]{{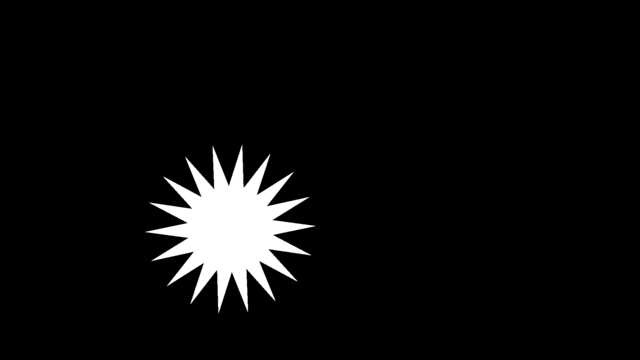}}}
        \smallskip
    \end{minipage}

	\vspace*{-0.1\baselineskip}

    \begin{minipage}[t]{1\textwidth}
        \makebox[0.083\textwidth][r]{\raisebox{15pt}{\smaller Proposed\hspace{6pt}}}
        \fbox{\includegraphics[width=0.100\textwidth]{{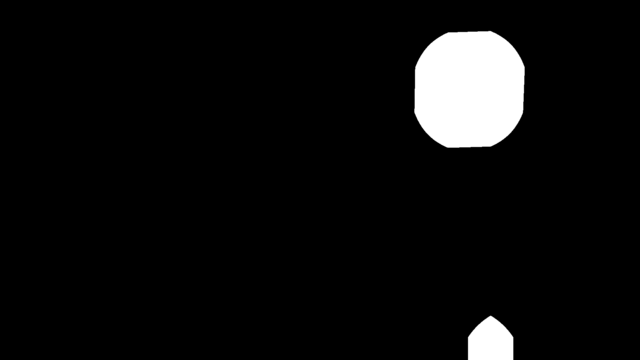}}}
        \fbox{\includegraphics[width=0.100\textwidth]{{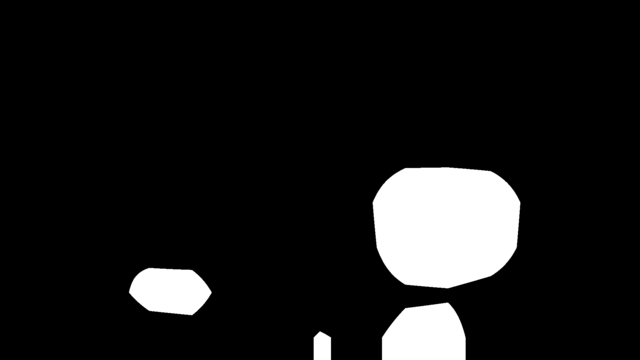}}}
        \fbox{\includegraphics[width=0.100\textwidth]{{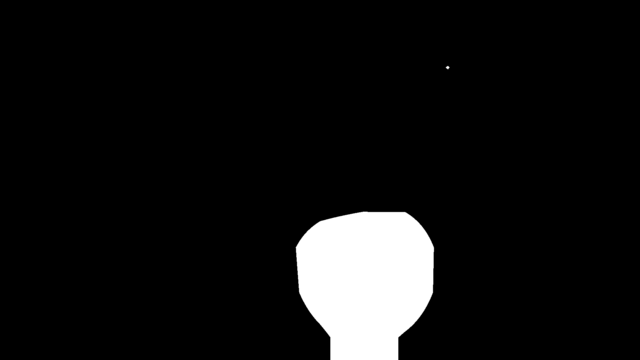}}}
        \fbox{\includegraphics[width=0.100\textwidth]{{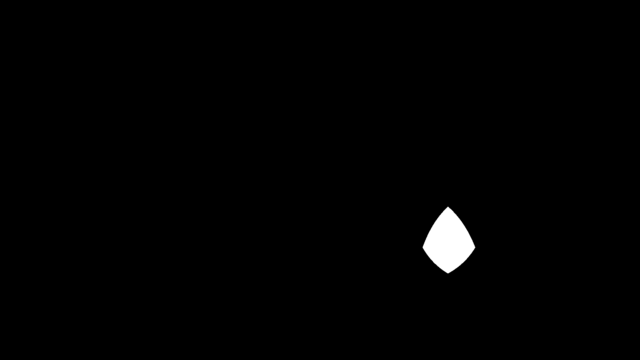}}}
        \fbox{\includegraphics[width=0.100\textwidth]{{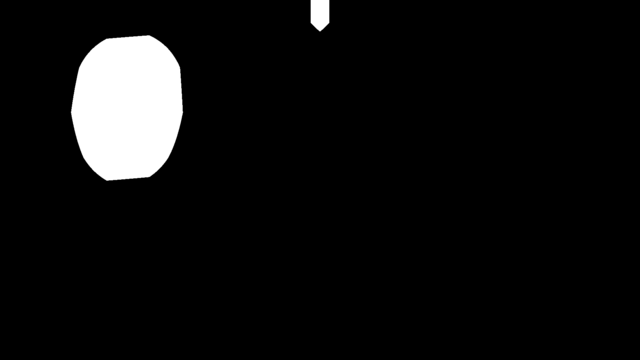}}}
        \fbox{\includegraphics[width=0.100\textwidth]{{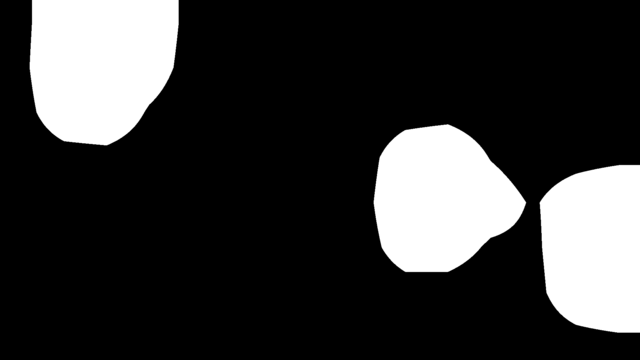}}}
        \fbox{\includegraphics[width=0.100\textwidth]{{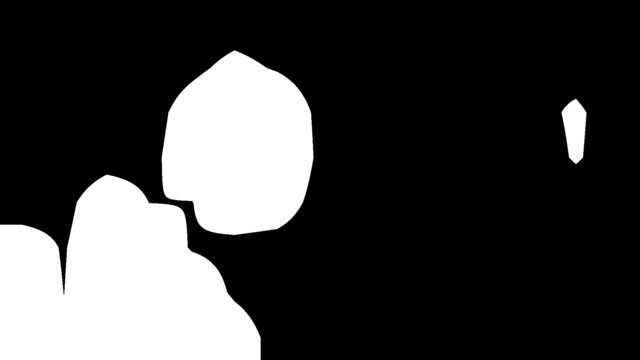}}}
        \fbox{\includegraphics[width=0.100\textwidth]{{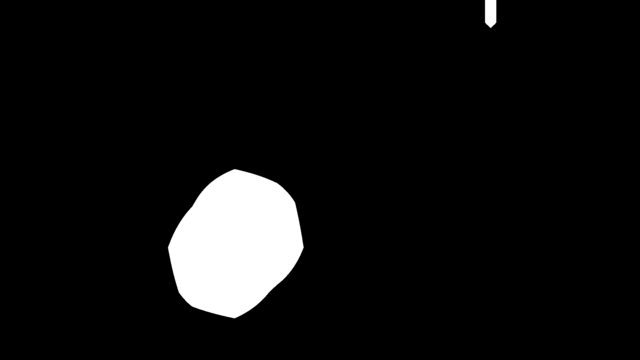}}}
        \smallskip
    \end{minipage}

	\vspace*{-0.1\baselineskip}

    \begin{minipage}[t]{1\textwidth}
        \makebox[0.083\textwidth][r]{\raisebox{15pt}{\smaller FSG\hspace{6pt}}}
        \fbox{\includegraphics[width=0.100\textwidth]{{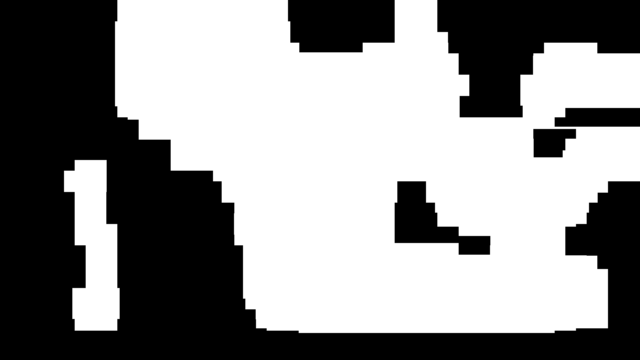}}}
        \fbox{\includegraphics[width=0.100\textwidth]{{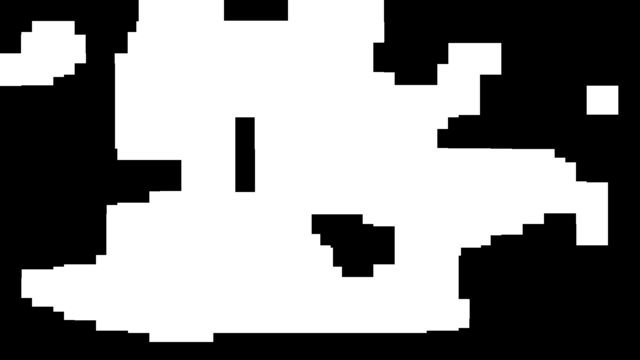}}}
        \fbox{\includegraphics[width=0.100\textwidth]{{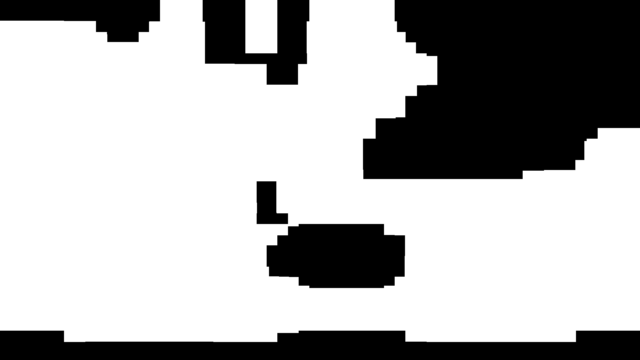}}}
        \fbox{\includegraphics[width=0.100\textwidth]{{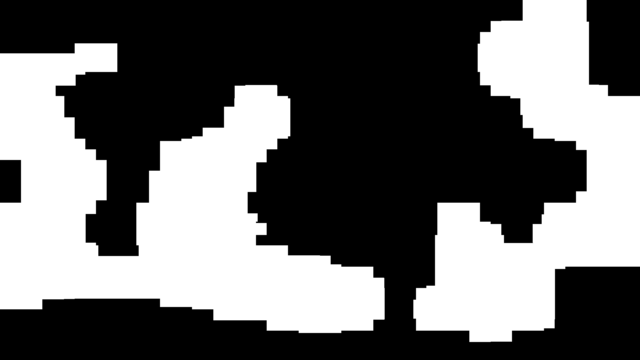}}}
        \fbox{\includegraphics[width=0.100\textwidth]{{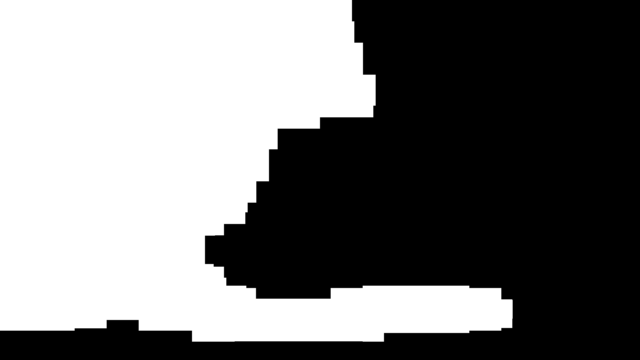}}}
        \fbox{\includegraphics[width=0.100\textwidth]{{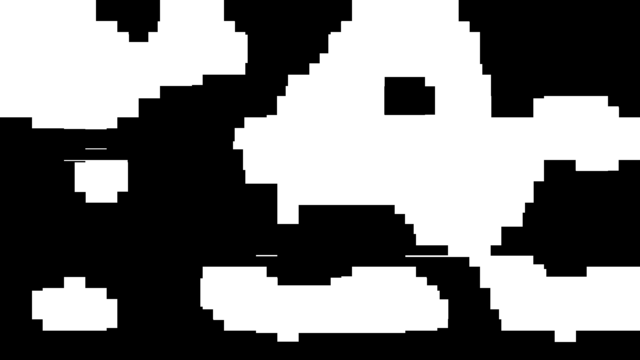}}}
        \fbox{\includegraphics[width=0.100\textwidth]{{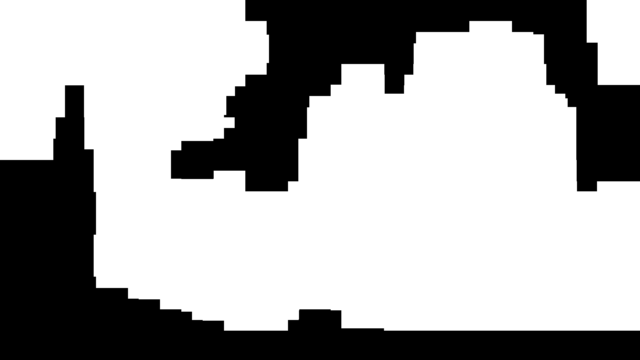}}}
        \fbox{\includegraphics[width=0.100\textwidth]{{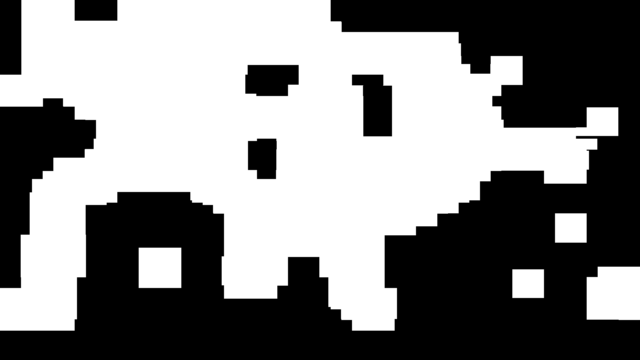}}}
        \smallskip
    \end{minipage}

	\vspace*{-0.1\baselineskip}

    \begin{minipage}[t]{1\textwidth}
        \makebox[0.083\textwidth][r]{\raisebox{15pt}{\smaller EXIFnet\hspace{6pt}}}
        \fbox{\includegraphics[width=0.100\textwidth]{{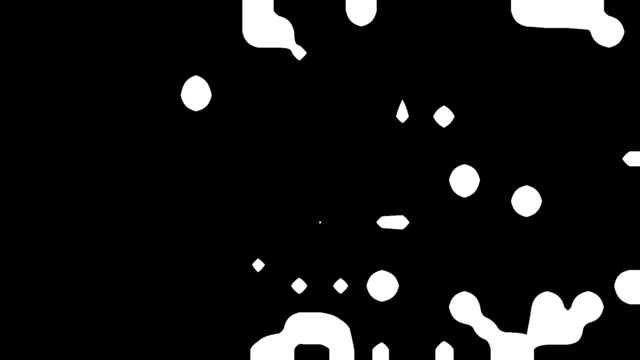}}}
        \fbox{\includegraphics[width=0.100\textwidth]{{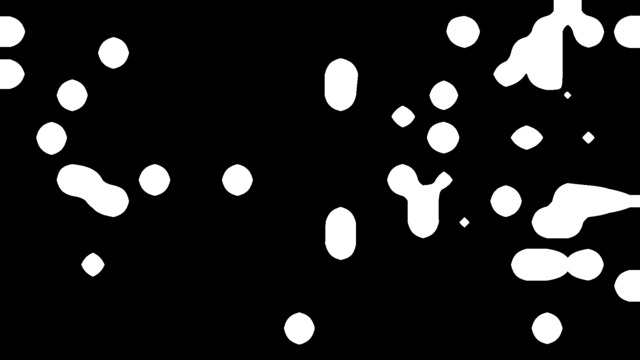}}}
        \fbox{\includegraphics[width=0.100\textwidth]{{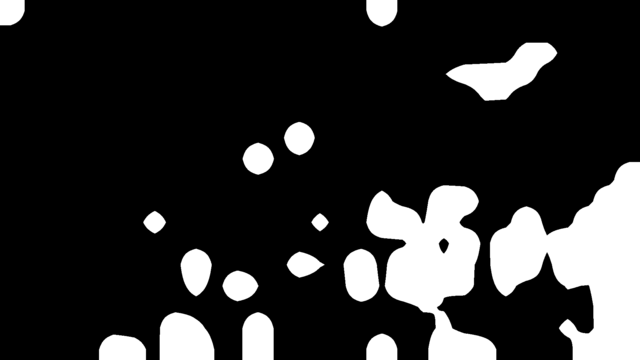}}}
        \fbox{\includegraphics[width=0.100\textwidth]{{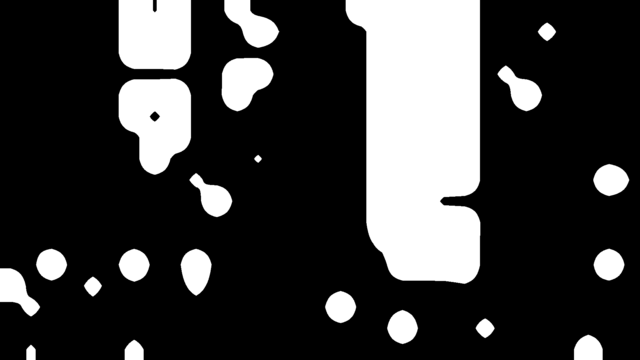}}}
        \fbox{\includegraphics[width=0.100\textwidth]{{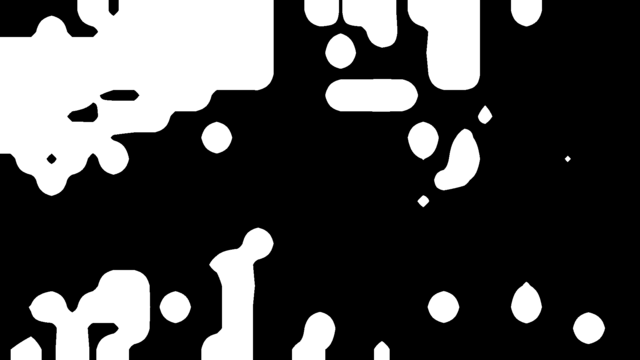}}}
        \fbox{\includegraphics[width=0.100\textwidth]{{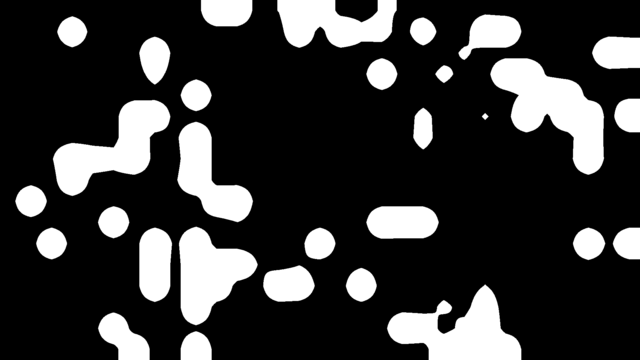}}}
        \fbox{\includegraphics[width=0.100\textwidth]{{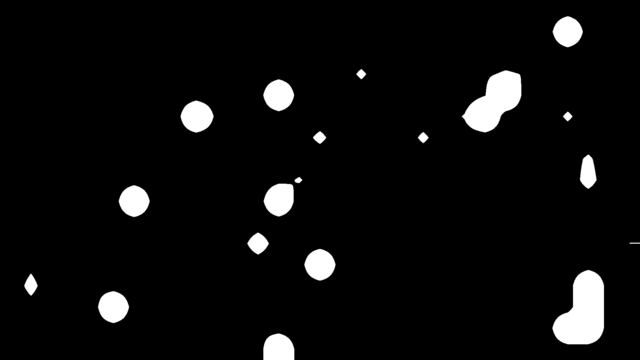}}}
        \fbox{\includegraphics[width=0.100\textwidth]{{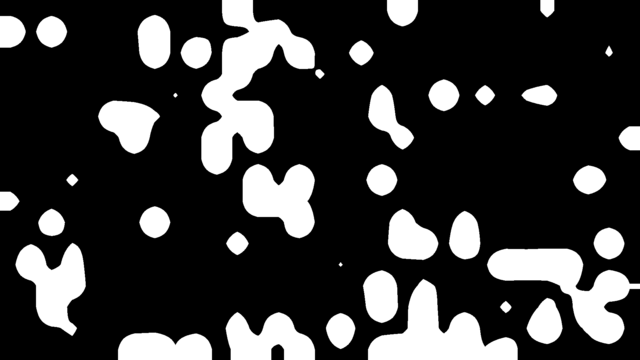}}}
        \smallskip
    \end{minipage}

	\vspace*{-0.1\baselineskip}

    \begin{minipage}[t]{1\textwidth}
        \makebox[0.083\textwidth][r]{\raisebox{15pt}{\smaller Noiseprint\hspace{6pt}}}
        \fbox{\includegraphics[width=0.100\textwidth]{{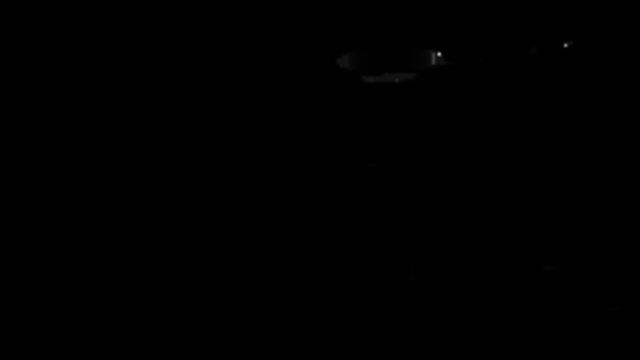}}}
        \fbox{\includegraphics[width=0.100\textwidth]{{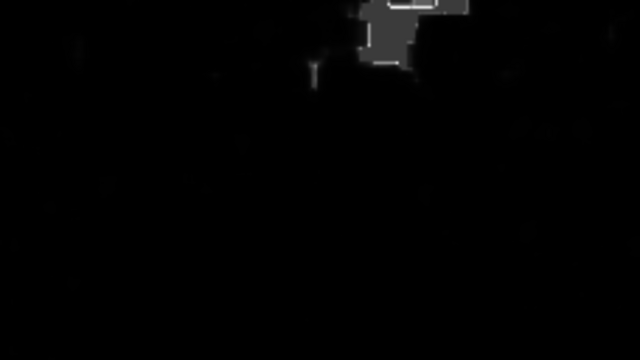}}}
        \fbox{\includegraphics[width=0.100\textwidth]{{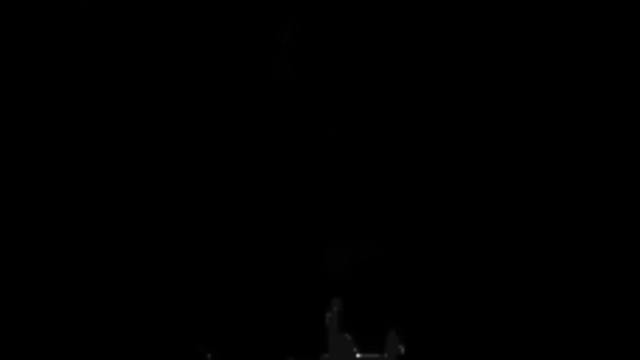}}}
        \fbox{\includegraphics[width=0.100\textwidth]{{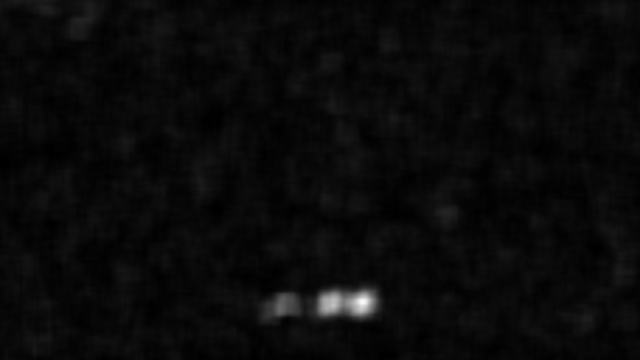}}}
        \fbox{\includegraphics[width=0.100\textwidth]{{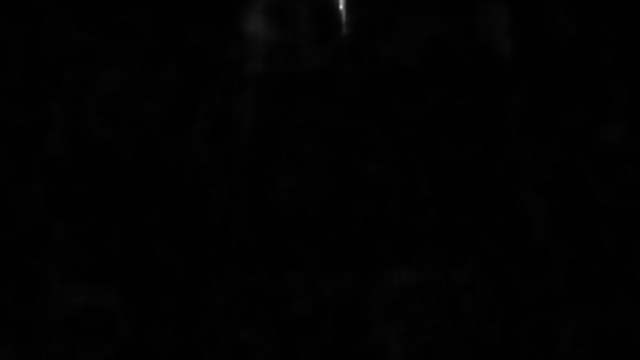}}}
        \fbox{\includegraphics[width=0.100\textwidth]{{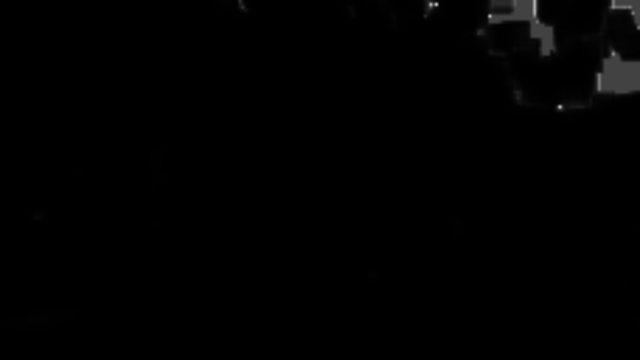}}}
        \fbox{\includegraphics[width=0.100\textwidth]{{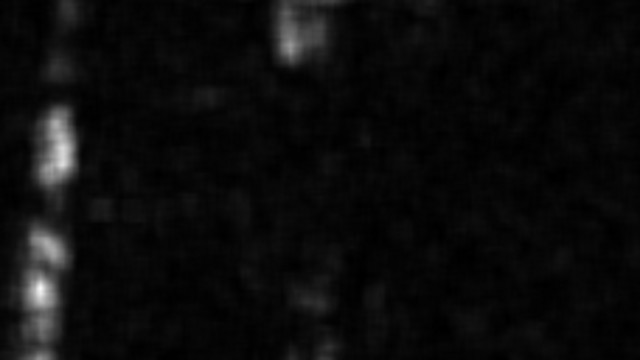}}}
        \fbox{\includegraphics[width=0.100\textwidth]{{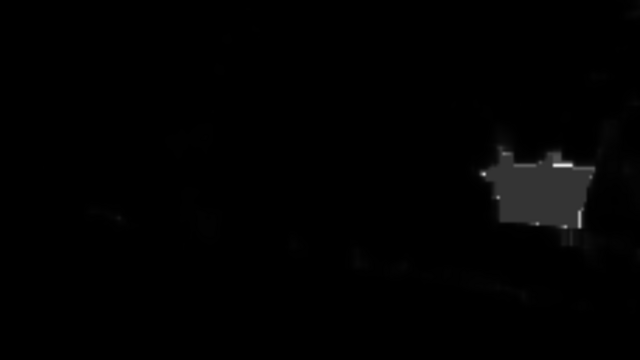}}}
        \smallskip
    \end{minipage}

	\vspace*{-0.1\baselineskip}

    \begin{minipage}[t]{1\textwidth}
        \makebox[0.083\textwidth][r]{\raisebox{15pt}{\smaller ManTra-Net\hspace{6pt}}}
        \fbox{\includegraphics[width=0.100\textwidth]{{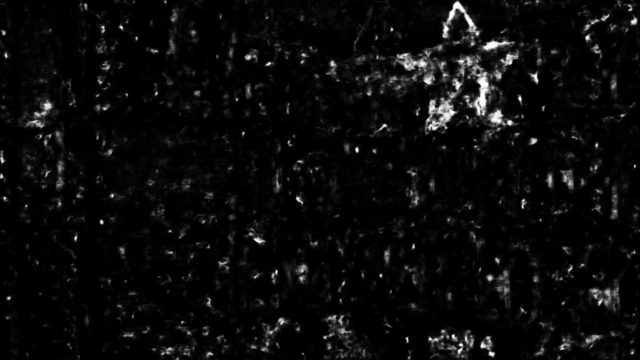}}}
        \fbox{\includegraphics[width=0.100\textwidth]{{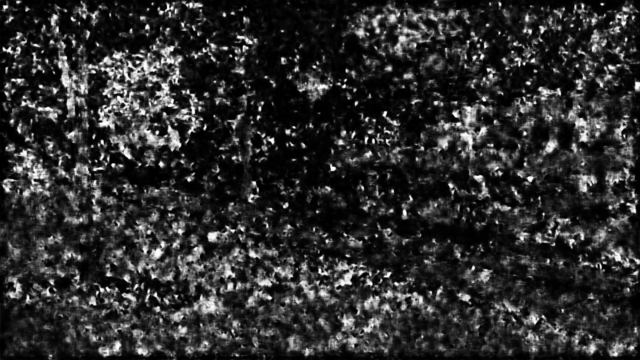}}}
        \fbox{\includegraphics[width=0.100\textwidth]{{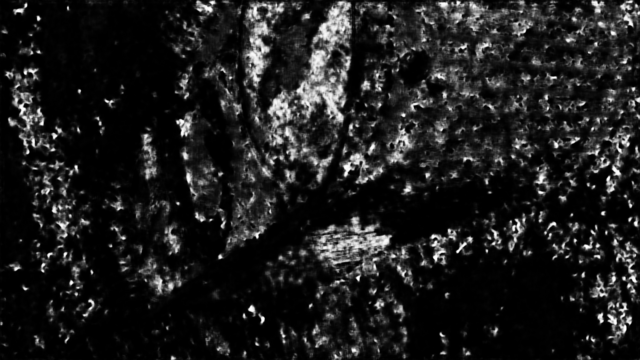}}}
        \fbox{\includegraphics[width=0.100\textwidth]{{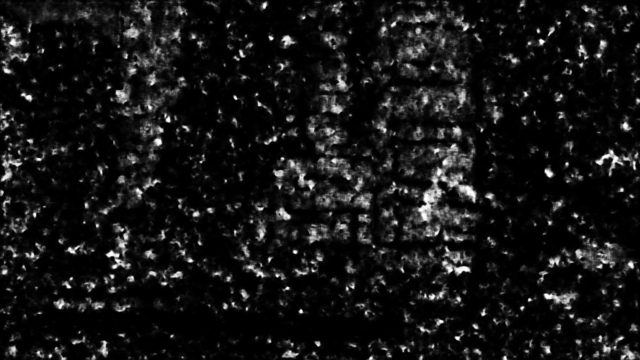}}}
        \fbox{\includegraphics[width=0.100\textwidth]{{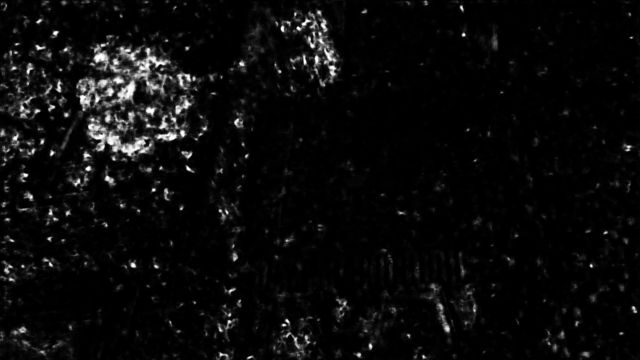}}}
        \fbox{\includegraphics[width=0.100\textwidth]{{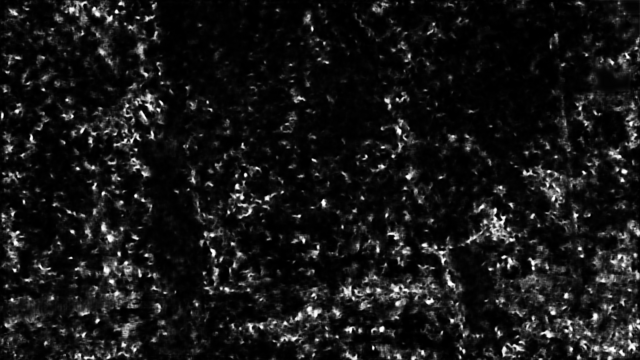}}}
        \fbox{\includegraphics[width=0.100\textwidth]{{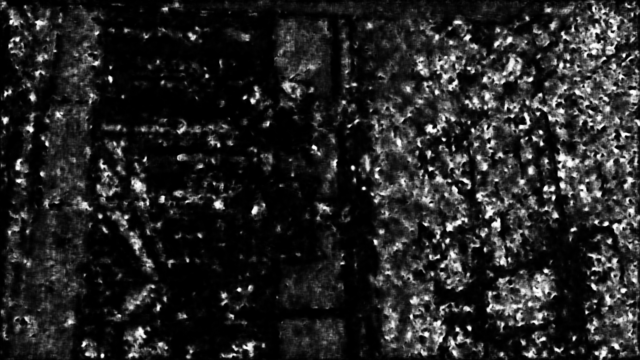}}}
        \fbox{\includegraphics[width=0.100\textwidth]{{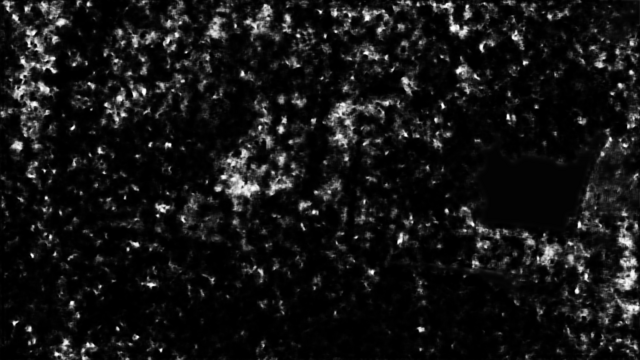}}}
        \smallskip
    \end{minipage}

	\vspace*{-0.1\baselineskip}

    \begin{minipage}[t]{1\textwidth}
        \makebox[0.083\textwidth][r]{\raisebox{15pt}{\smaller MVSS-Net\hspace{6pt}}}
        \fbox{\includegraphics[width=0.100\textwidth]{{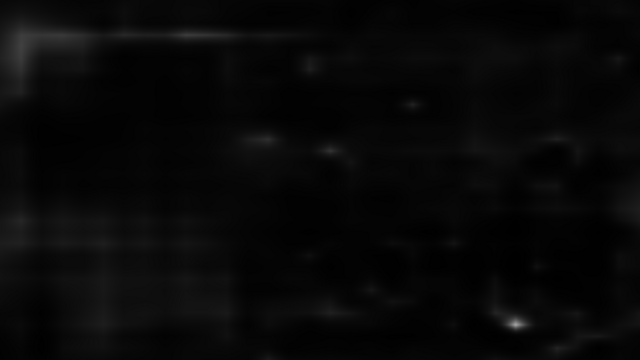}}}
        \fbox{\includegraphics[width=0.100\textwidth]{{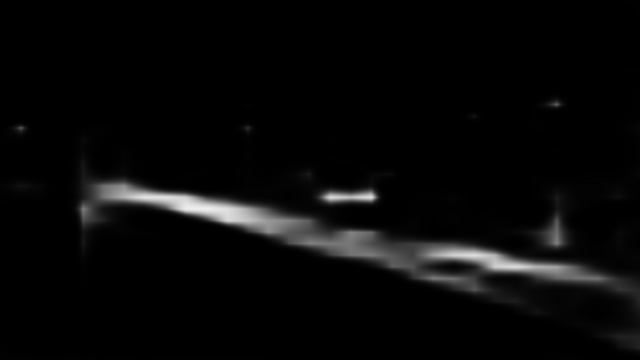}}}
        \fbox{\includegraphics[width=0.100\textwidth]{{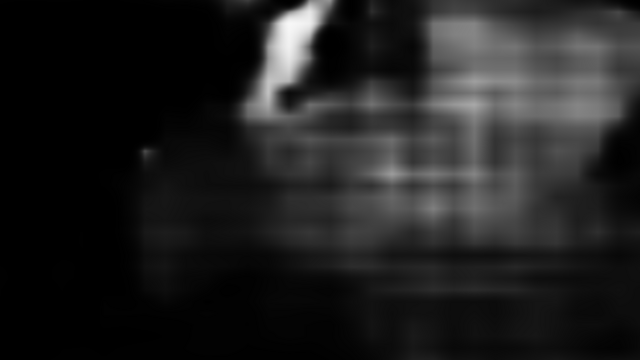}}}
        \fbox{\includegraphics[width=0.100\textwidth]{{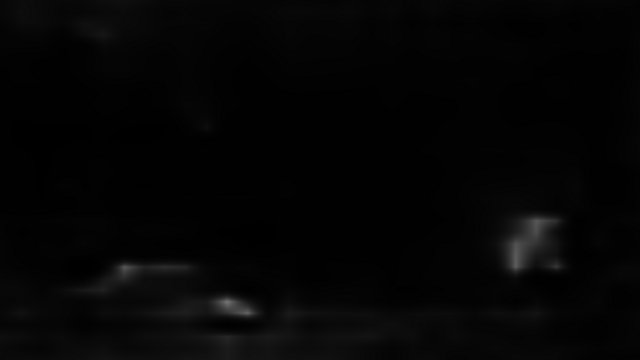}}}
        \fbox{\includegraphics[width=0.100\textwidth]{{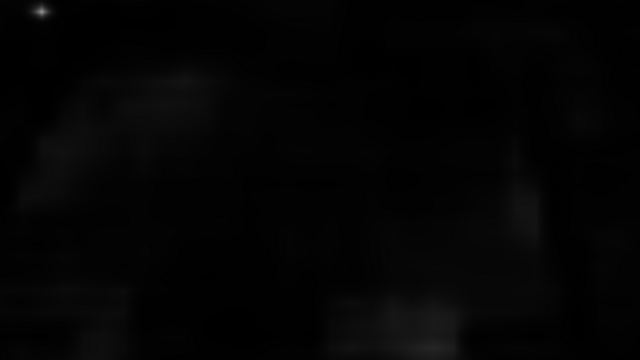}}}
        \fbox{\includegraphics[width=0.100\textwidth]{{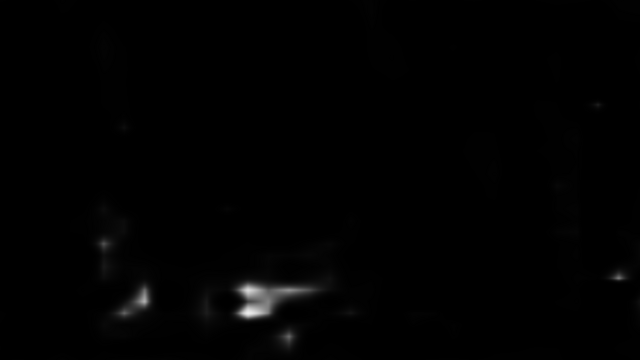}}}
        \fbox{\includegraphics[width=0.100\textwidth]{{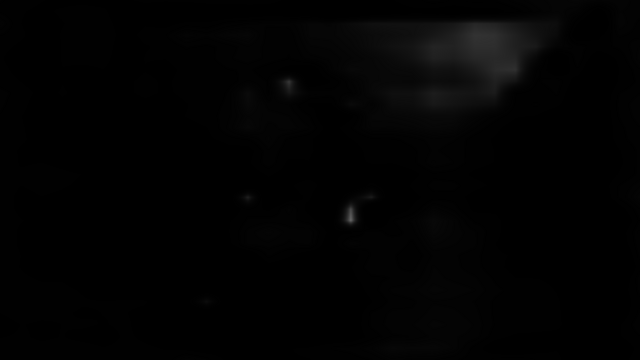}}}
        \fbox{\includegraphics[width=0.100\textwidth]{{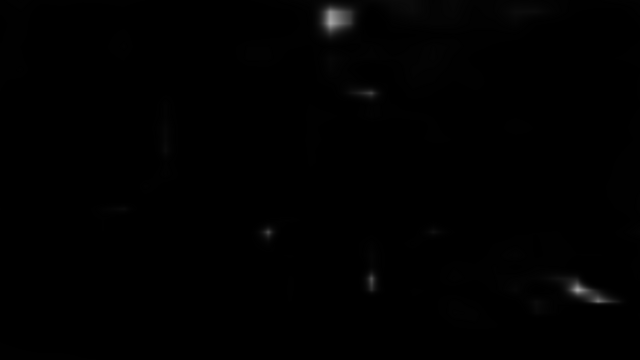}}}
        \smallskip
    \end{minipage}
    
	\vspace*{-0.5\baselineskip}

    \caption{\label{fig:vpim_localization_results} 
    This figure shows the localization results of different networks on the VPIM dataset. Our proposed network's localization results are strong, with some minor false alarms on column 1, 2 and mis-detections on column 4. Differs from VCMS and VPVM, in this dataset, the manipulations' strengths were so low that they are largely perceptually invisible. Hence, in contrast to our network, other competing networks failed to provide any meaningful predictions of the manipulated region.
    }

\end{figure*}


\begin{figure*}[!t]
    \centering
    \setlength{\fboxsep}{0pt}

    \begin{minipage}[t]{1\textwidth}
        \makebox[0.083\textwidth][r]{\raisebox{15pt}{\smaller Frame\hspace{6pt}}}
        \fbox{\includegraphics[width=0.100\textwidth]{{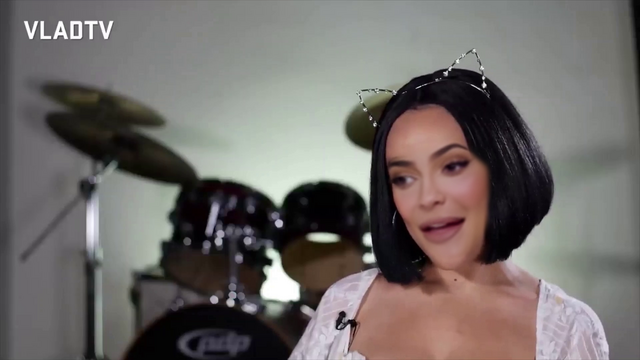}}}
        \fbox{\includegraphics[width=0.100\textwidth]{{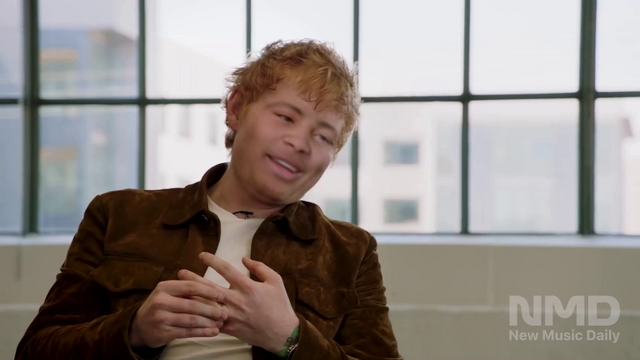}}}
        \fbox{\includegraphics[width=0.100\textwidth]{{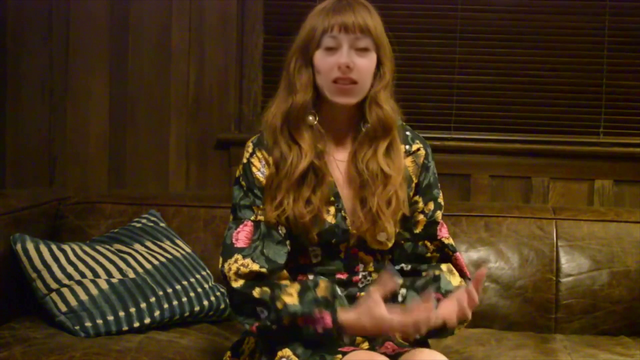}}}
        \fbox{\includegraphics[width=0.100\textwidth]{{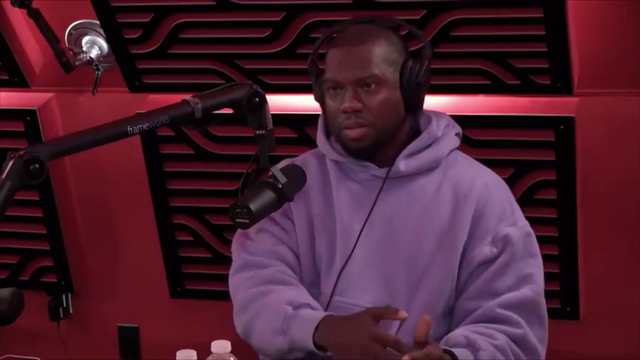}}}
        \fbox{\includegraphics[width=0.100\textwidth]{{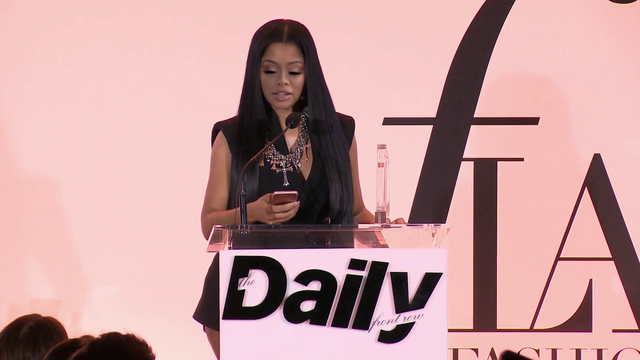}}}
        \fbox{\includegraphics[width=0.100\textwidth]{{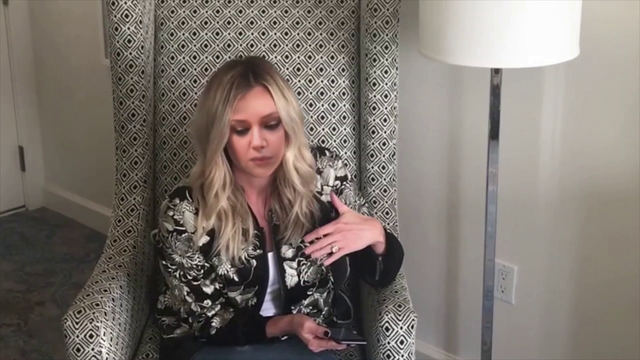}}}
        \fbox{\includegraphics[width=0.100\textwidth]{{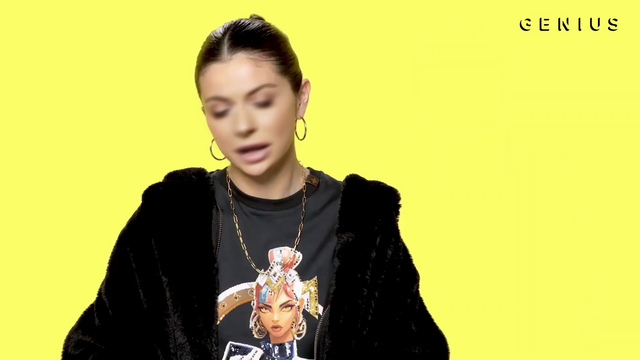}}}
        \fbox{\includegraphics[width=0.100\textwidth]{{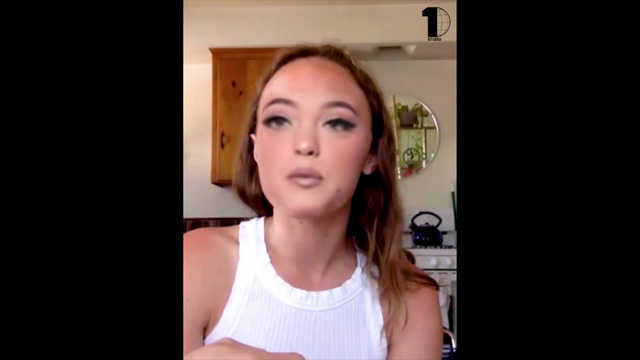}}}
        \smallskip
    \end{minipage}

	\vspace*{-0.1\baselineskip}

    \begin{minipage}[t]{1\textwidth}
        \makebox[0.083\textwidth][r]{\raisebox{15pt}{\smaller Mask\hspace{6pt}}}
        \fbox{\includegraphics[width=0.100\textwidth]{{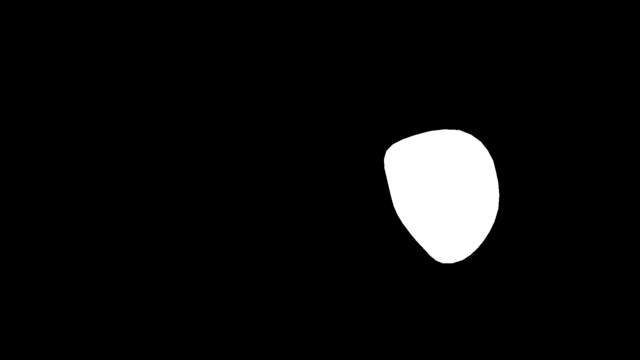}}}
        \fbox{\includegraphics[width=0.100\textwidth]{{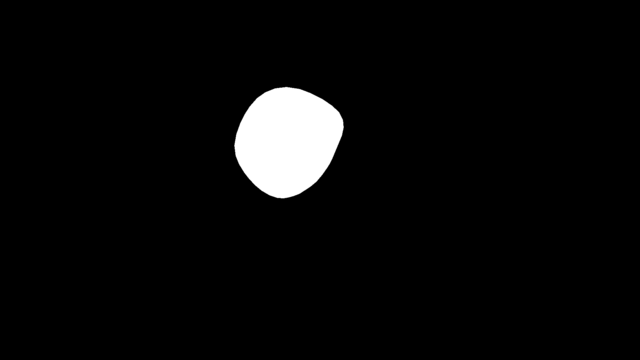}}}
        \fbox{\includegraphics[width=0.100\textwidth]{{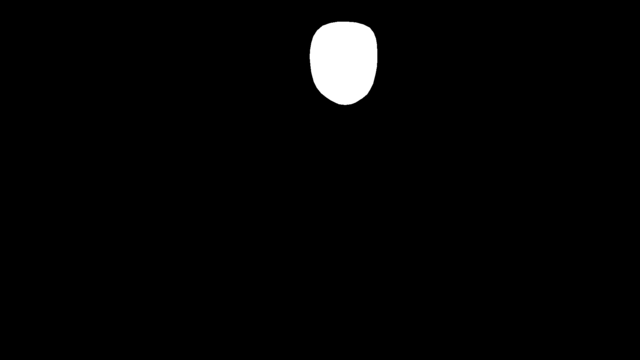}}}
        \fbox{\includegraphics[width=0.100\textwidth]{{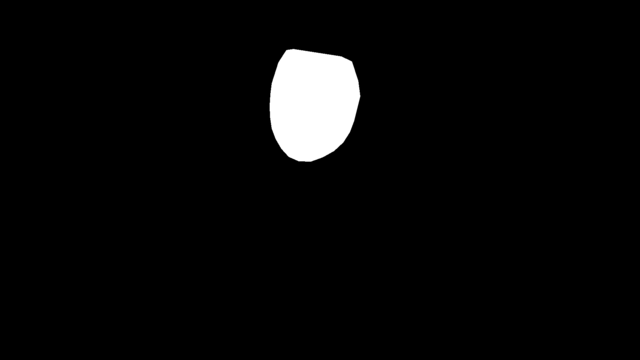}}}
        \fbox{\includegraphics[width=0.100\textwidth]{{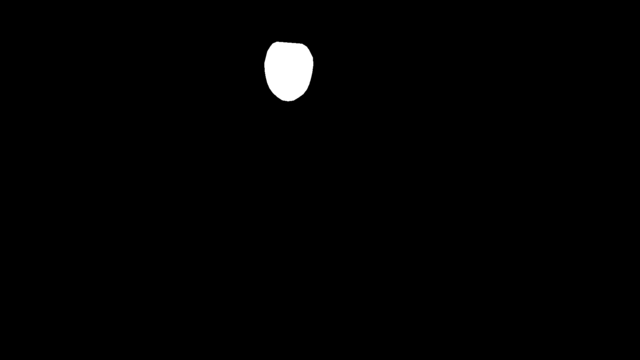}}}
        \fbox{\includegraphics[width=0.100\textwidth]{{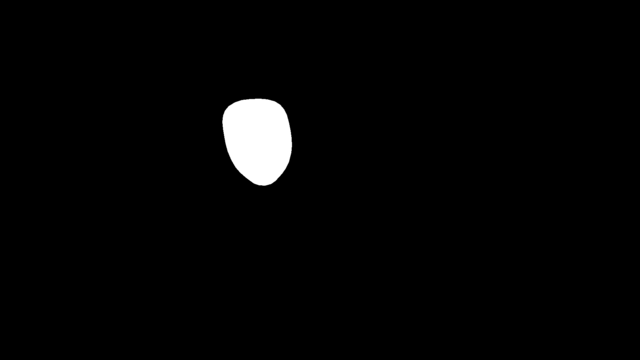}}}
        \fbox{\includegraphics[width=0.100\textwidth]{{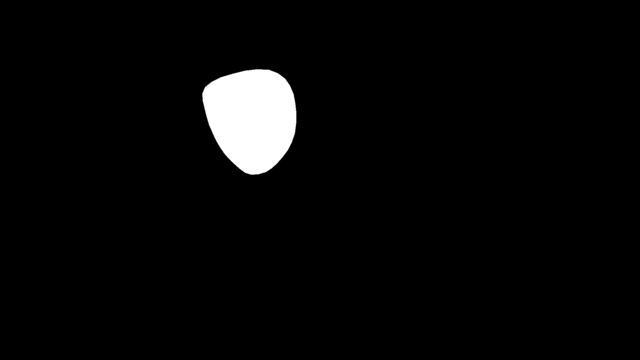}}}
        \fbox{\includegraphics[width=0.100\textwidth]{{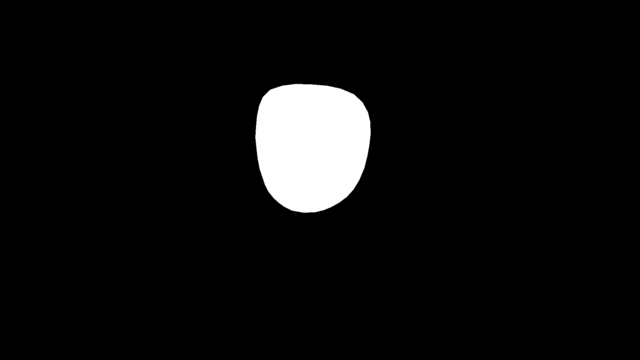}}}
        \smallskip
    \end{minipage}

	\vspace*{-0.1\baselineskip}

    \begin{minipage}[t]{1\textwidth}
        \makebox[0.083\textwidth][r]{\raisebox{15pt}{\smaller Proposed\hspace{6pt}}}
        \fbox{\includegraphics[width=0.100\textwidth]{{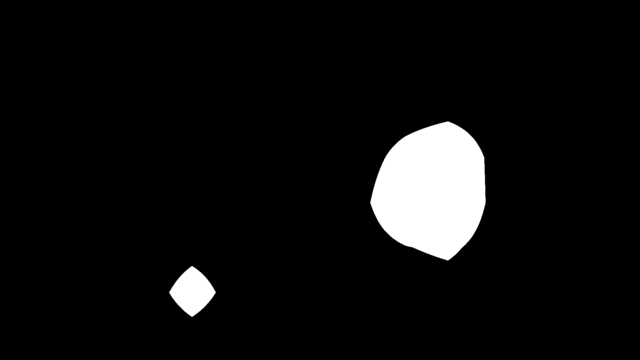}}}
        \fbox{\includegraphics[width=0.100\textwidth]{{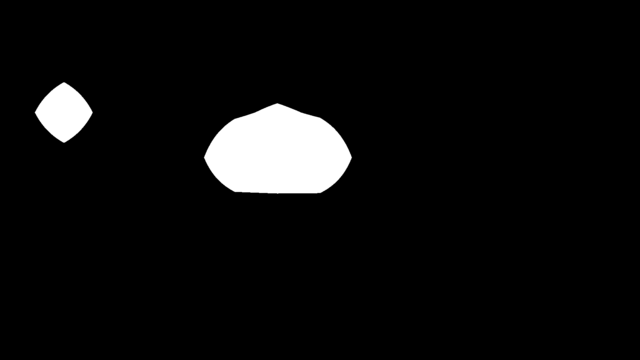}}}
        \fbox{\includegraphics[width=0.100\textwidth]{{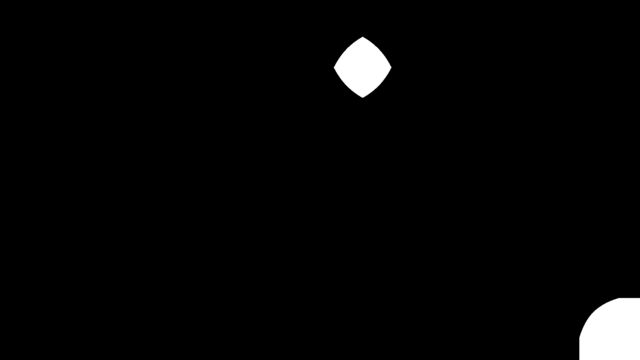}}}
        \fbox{\includegraphics[width=0.100\textwidth]{{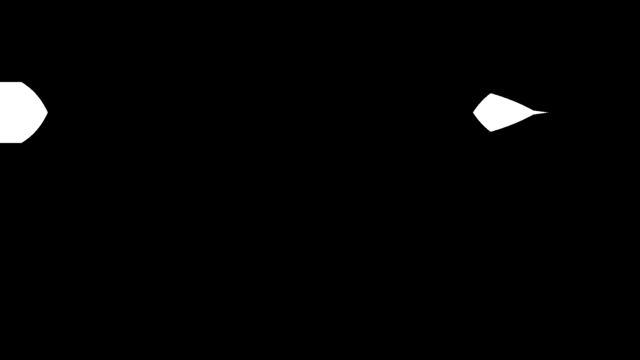}}}
        \fbox{\includegraphics[width=0.100\textwidth]{{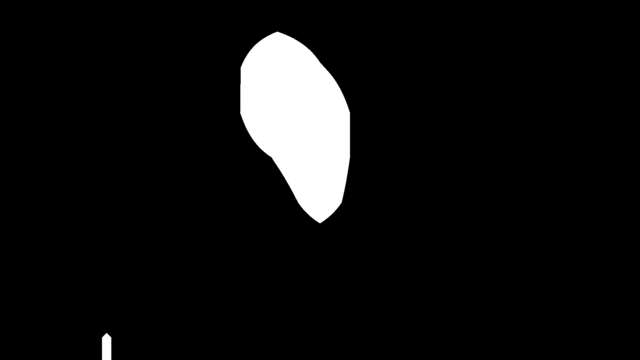}}}
        \fbox{\includegraphics[width=0.100\textwidth]{{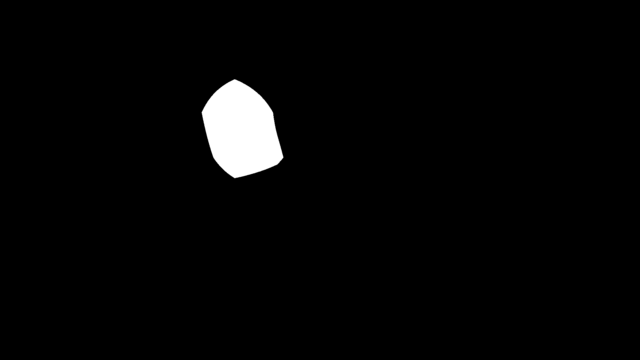}}}
        \fbox{\includegraphics[width=0.100\textwidth]{{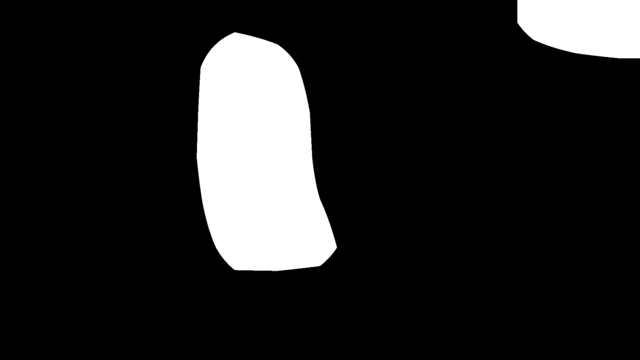}}}
        \fbox{\includegraphics[width=0.100\textwidth]{{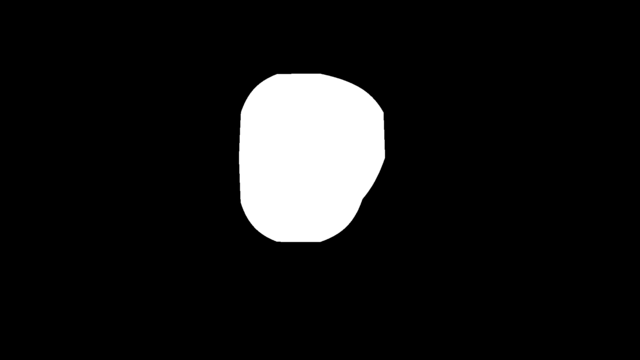}}}
        \smallskip
    \end{minipage}

	\vspace*{-0.1\baselineskip}

    \begin{minipage}[t]{1\textwidth}
        \makebox[0.083\textwidth][r]{\raisebox{15pt}{\smaller FSG\hspace{6pt}}}
        \fbox{\includegraphics[width=0.100\textwidth]{{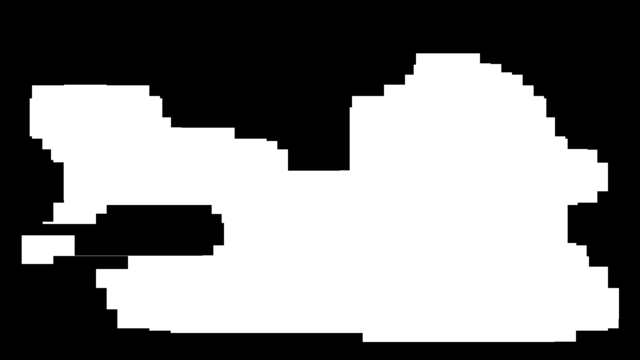}}}
        \fbox{\includegraphics[width=0.100\textwidth]{{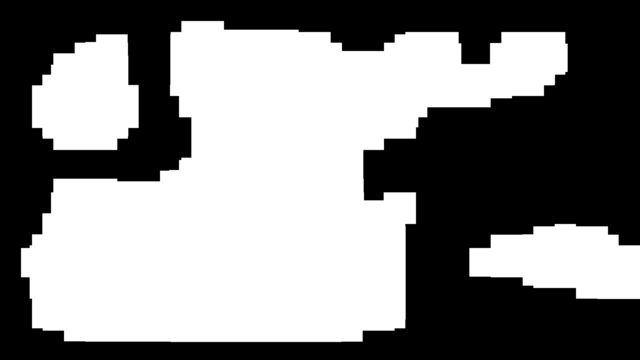}}}
        \fbox{\includegraphics[width=0.100\textwidth]{{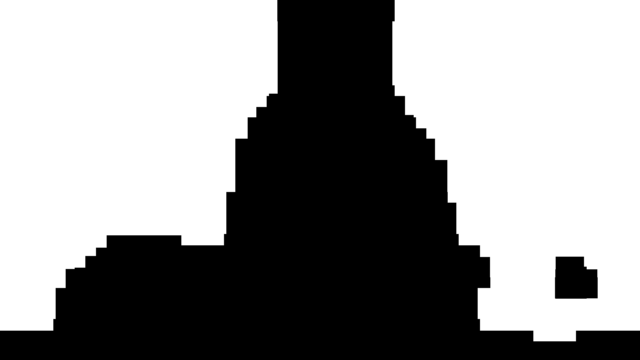}}}
        \fbox{\includegraphics[width=0.100\textwidth]{{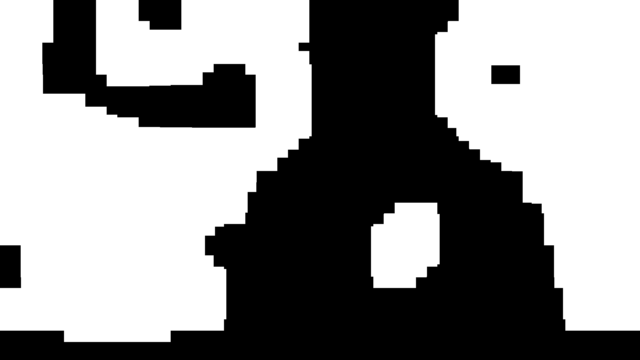}}}
        \fbox{\includegraphics[width=0.100\textwidth]{{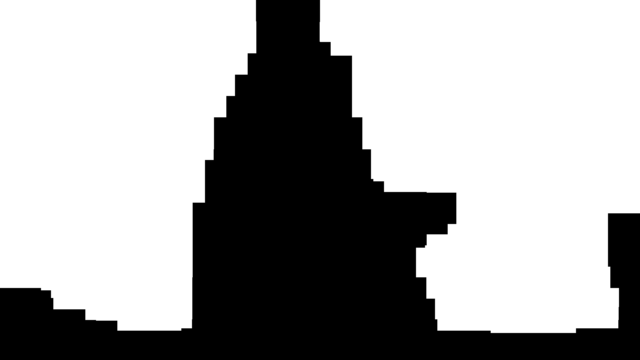}}}
        \fbox{\includegraphics[width=0.100\textwidth]{{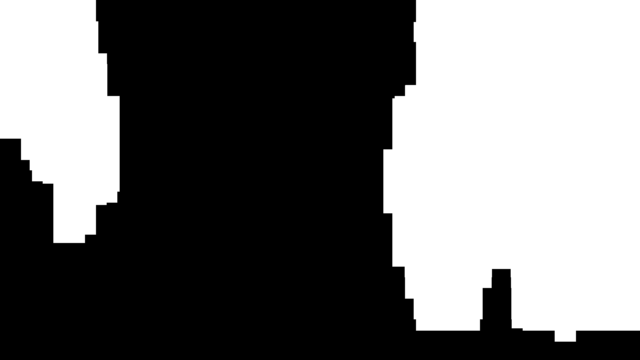}}}
        \fbox{\includegraphics[width=0.100\textwidth]{{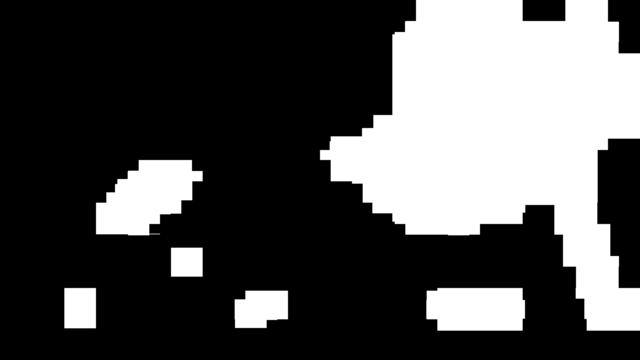}}}
        \fbox{\includegraphics[width=0.100\textwidth]{{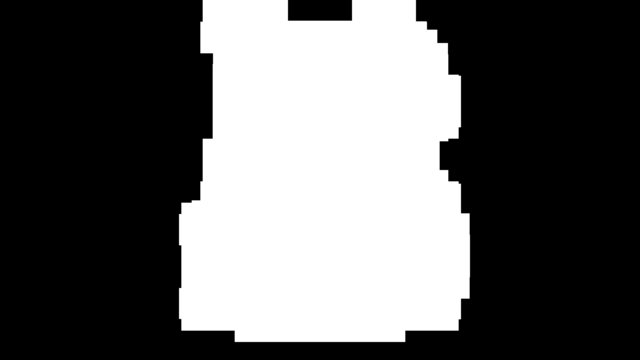}}}
        \smallskip
    \end{minipage}

	\vspace*{-0.1\baselineskip}

    \begin{minipage}[t]{1\textwidth}
        \makebox[0.083\textwidth][r]{\raisebox{15pt}{\smaller EXIFnet\hspace{6pt}}}
        \fbox{\includegraphics[width=0.100\textwidth]{{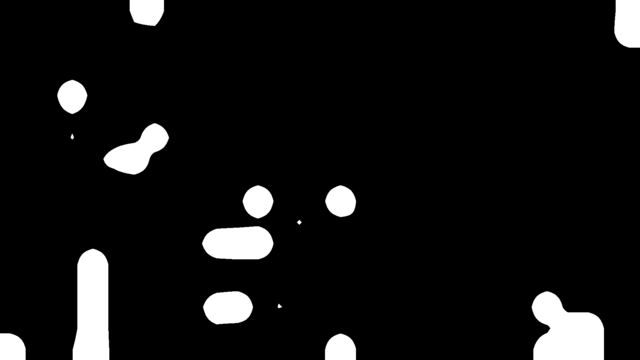}}}
        \fbox{\includegraphics[width=0.100\textwidth]{{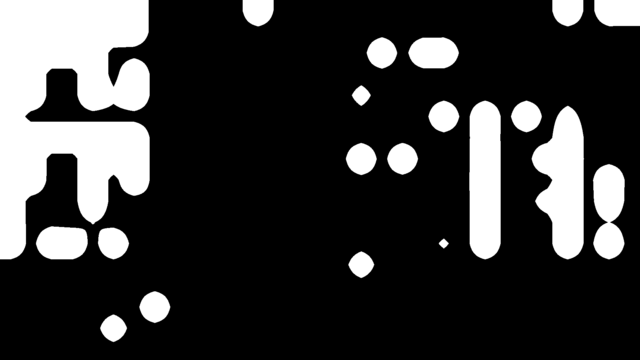}}}
        \fbox{\includegraphics[width=0.100\textwidth]{{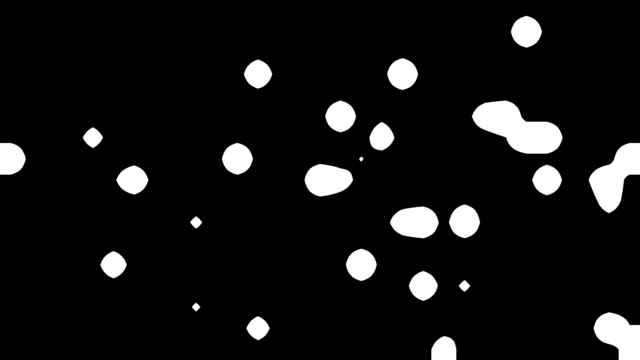}}}
        \fbox{\includegraphics[width=0.100\textwidth]{{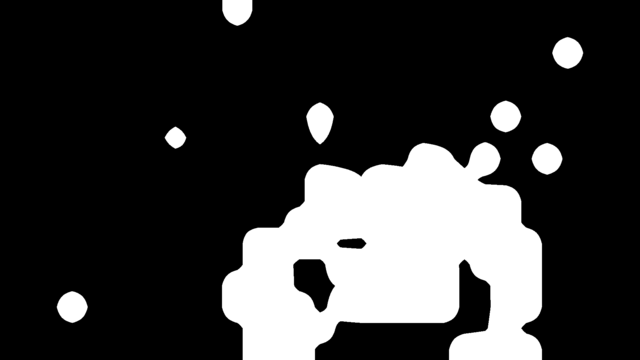}}}
        \fbox{\includegraphics[width=0.100\textwidth]{{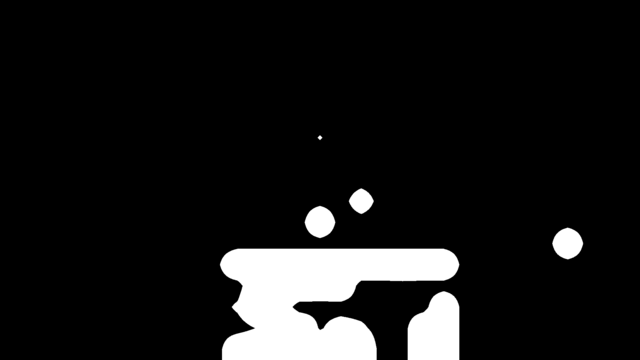}}}
        \fbox{\includegraphics[width=0.100\textwidth]{{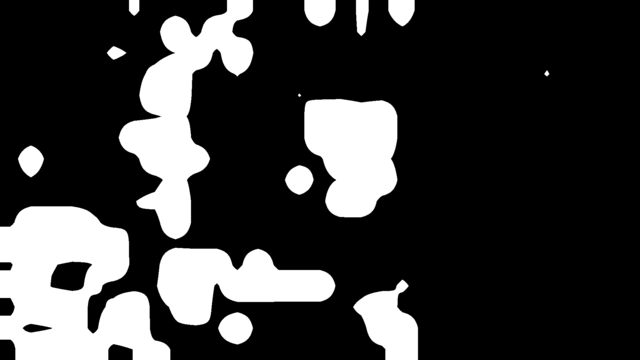}}}
        \fbox{\includegraphics[width=0.100\textwidth]{{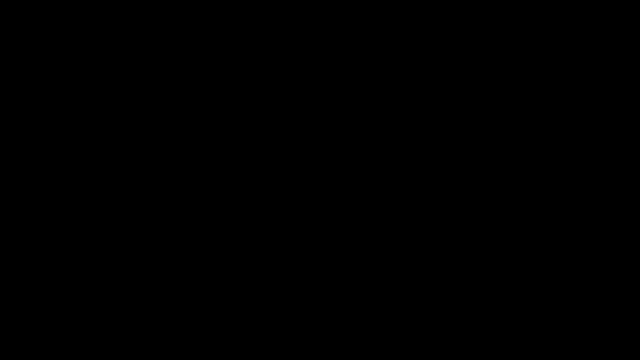}}}
        \fbox{\includegraphics[width=0.100\textwidth]{{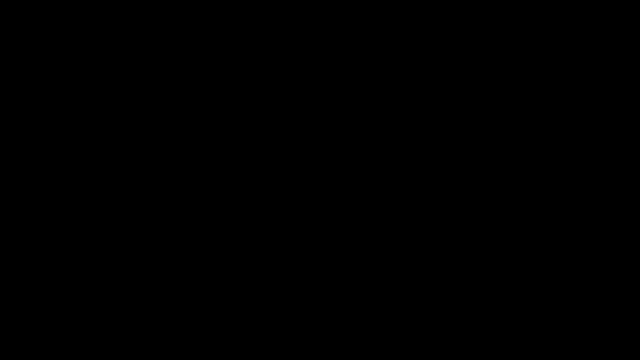}}}
        \smallskip
    \end{minipage}

	\vspace*{-0.1\baselineskip}

    \begin{minipage}[t]{1\textwidth}
        \makebox[0.083\textwidth][r]{\raisebox{15pt}{\smaller Noiseprint\hspace{6pt}}}
        \fbox{\includegraphics[width=0.100\textwidth]{{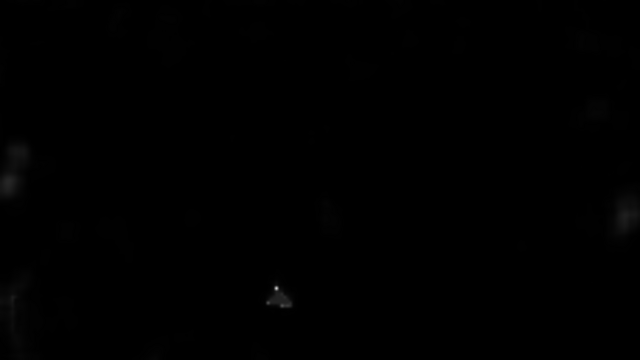}}}
        \fbox{\includegraphics[width=0.100\textwidth]{{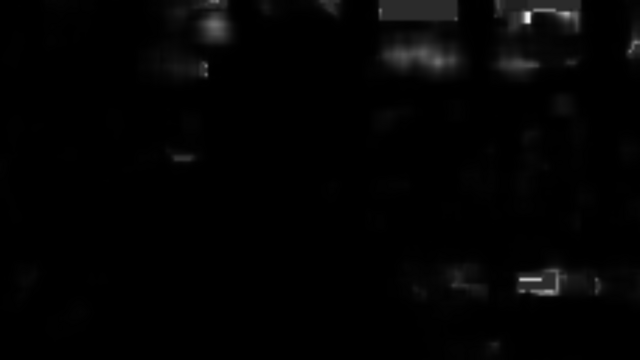}}}
        \fbox{\includegraphics[width=0.100\textwidth]{{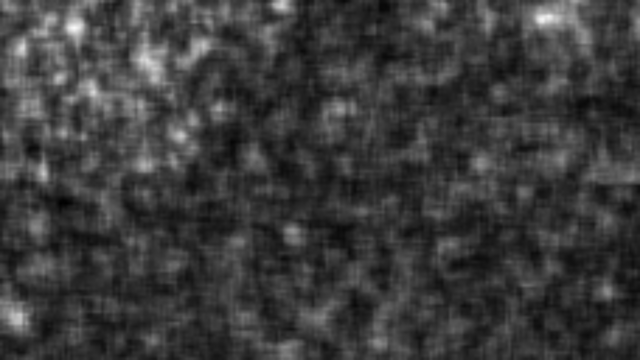}}}
        \fbox{\includegraphics[width=0.100\textwidth]{{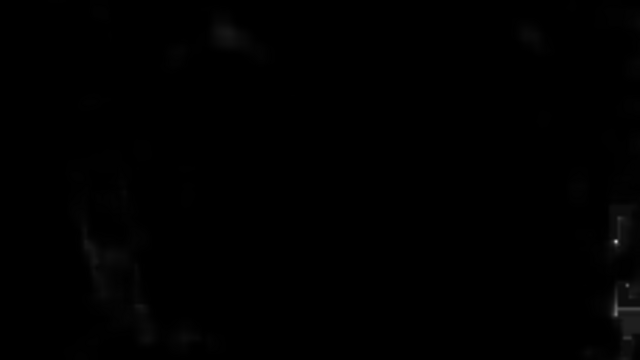}}}
        \fbox{\includegraphics[width=0.100\textwidth]{{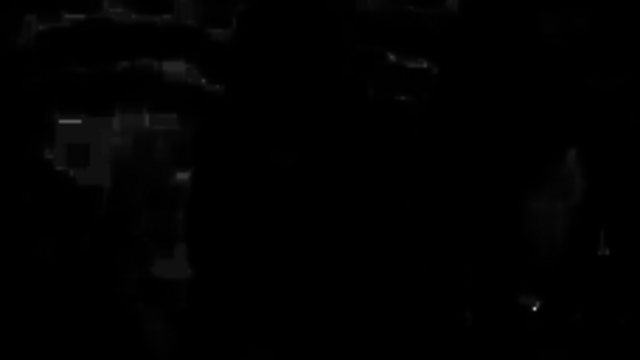}}}
        \fbox{\includegraphics[width=0.100\textwidth]{{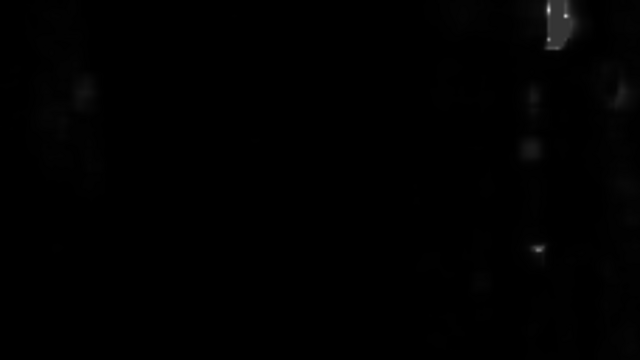}}}
        \fbox{\includegraphics[width=0.100\textwidth]{{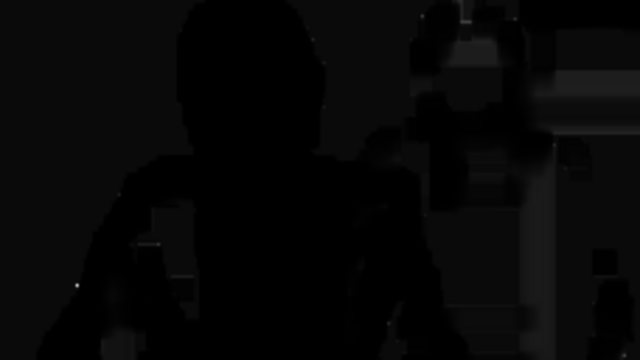}}}
        \fbox{\includegraphics[width=0.100\textwidth]{{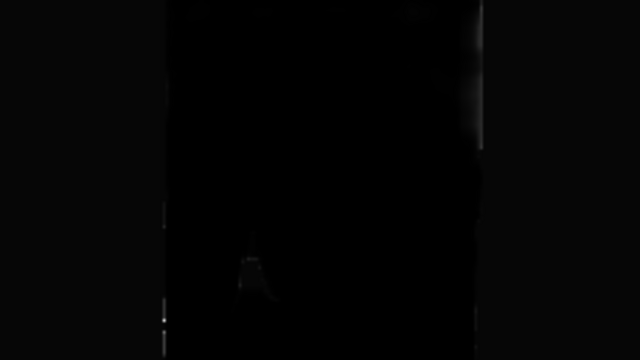}}}
        \smallskip
    \end{minipage}

	\vspace*{-0.1\baselineskip}

    \begin{minipage}[t]{1\textwidth}
        \makebox[0.083\textwidth][r]{\raisebox{15pt}{\smaller ManTra-Net\hspace{6pt}}}
        \fbox{\includegraphics[width=0.100\textwidth]{{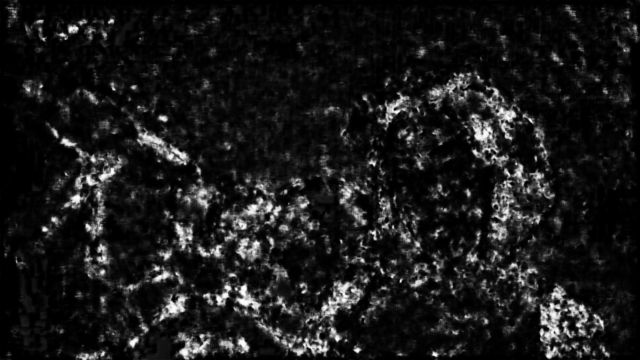}}}
        \fbox{\includegraphics[width=0.100\textwidth]{{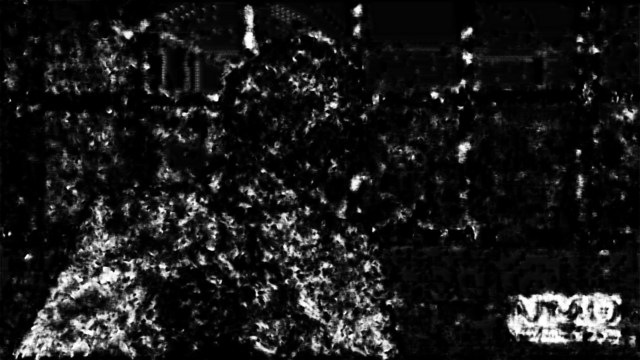}}}
        \fbox{\includegraphics[width=0.100\textwidth]{{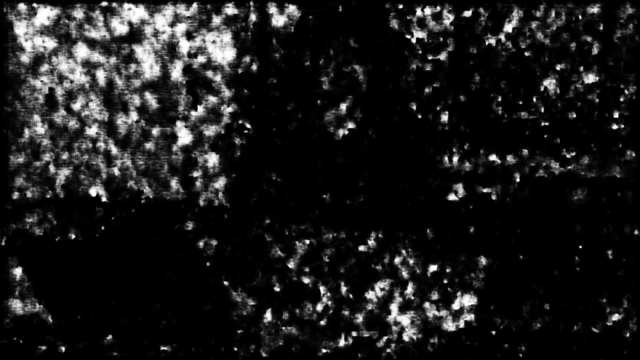}}}
        \fbox{\includegraphics[width=0.100\textwidth]{{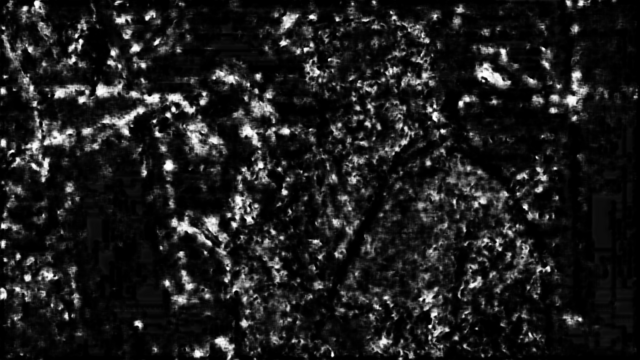}}}
        \fbox{\includegraphics[width=0.100\textwidth]{{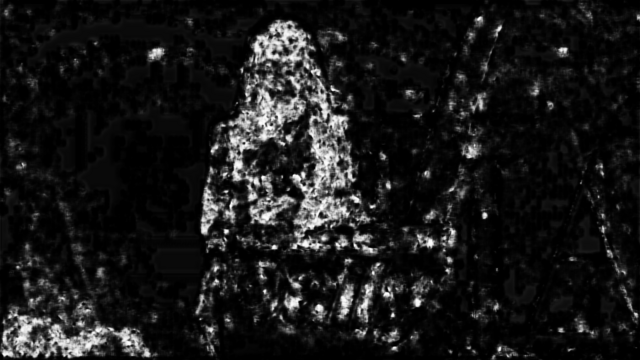}}}
        \fbox{\includegraphics[width=0.100\textwidth]{{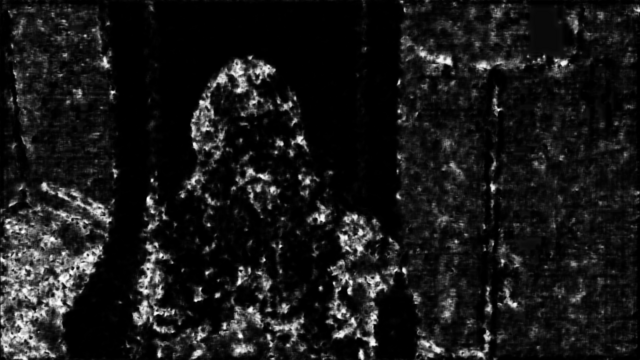}}}
        \fbox{\includegraphics[width=0.100\textwidth]{{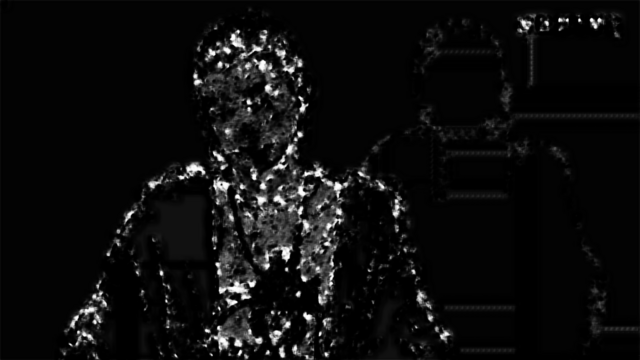}}}
        \fbox{\includegraphics[width=0.100\textwidth]{{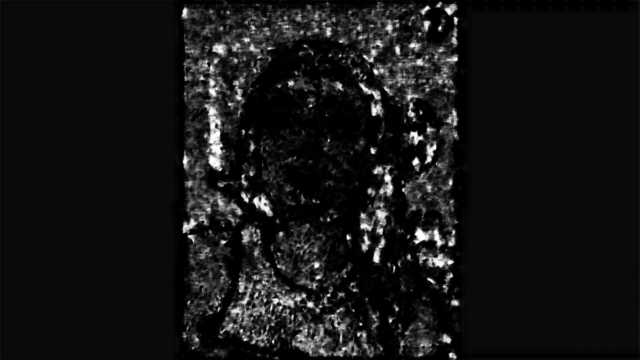}}}
        \smallskip
    \end{minipage}

	\vspace*{-0.1\baselineskip}

    \begin{minipage}[t]{1\textwidth}
        \makebox[0.083\textwidth][r]{\raisebox{15pt}{\smaller MVSS-Net\hspace{6pt}}}
        \fbox{\includegraphics[width=0.100\textwidth]{{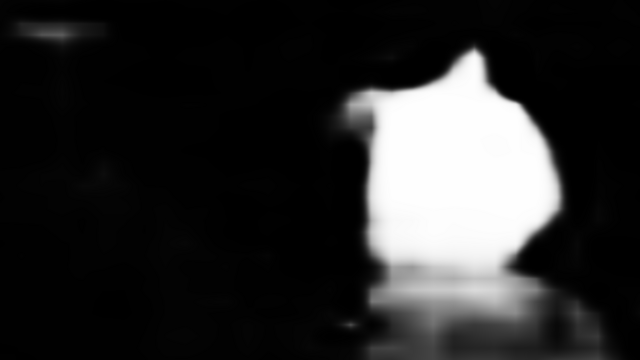}}}
        \fbox{\includegraphics[width=0.100\textwidth]{{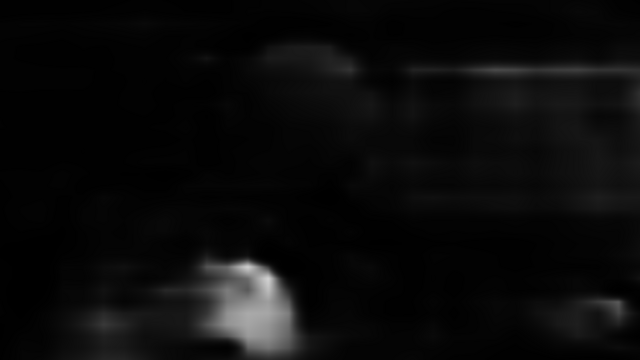}}}
        \fbox{\includegraphics[width=0.100\textwidth]{{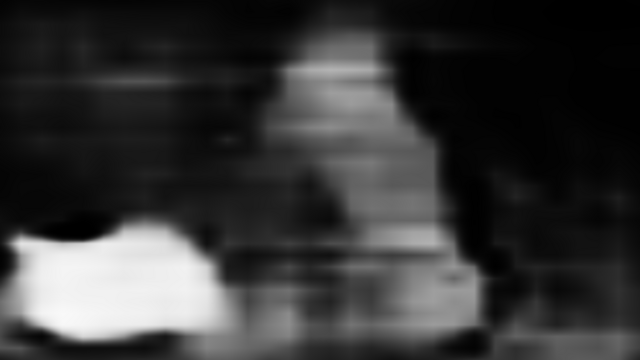}}}
        \fbox{\includegraphics[width=0.100\textwidth]{{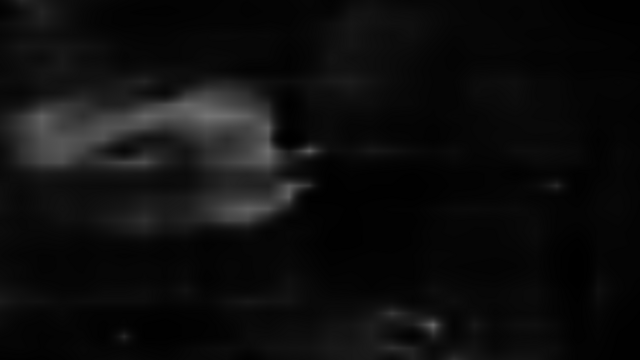}}}
        \fbox{\includegraphics[width=0.100\textwidth]{{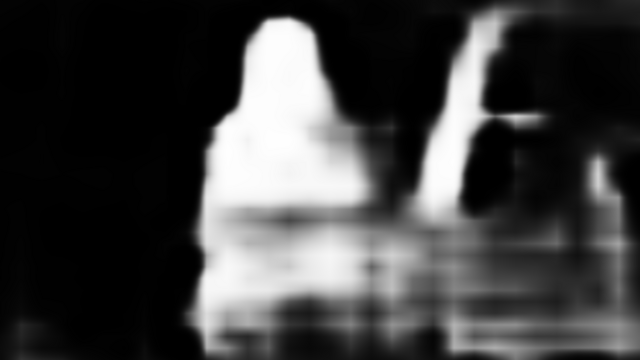}}}
        \fbox{\includegraphics[width=0.100\textwidth]{{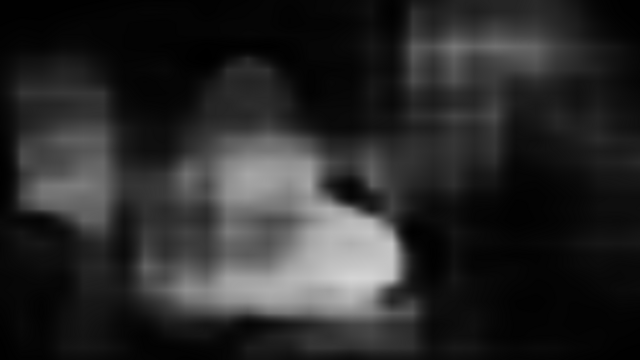}}}
        \fbox{\includegraphics[width=0.100\textwidth]{{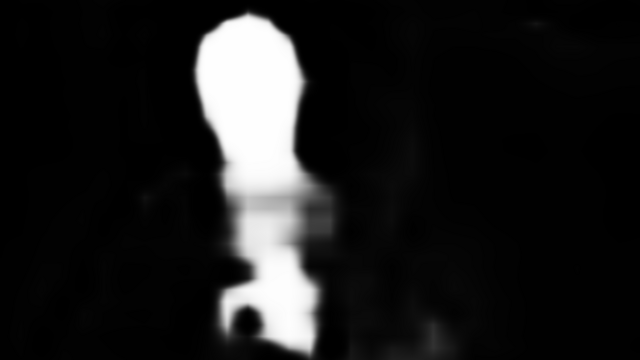}}}
        \fbox{\includegraphics[width=0.100\textwidth]{{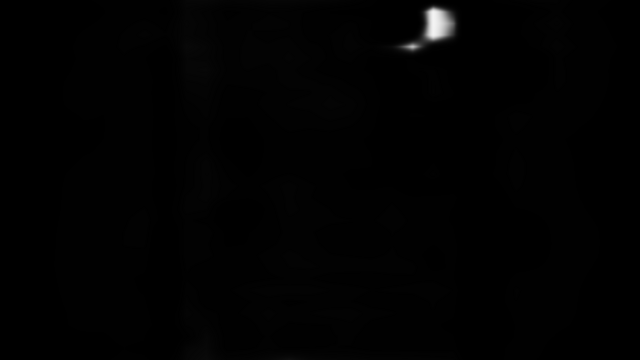}}}
        \smallskip
    \end{minipage}
    
	\vspace*{-0.5\baselineskip}

    \caption{\label{fig:deepfake_localization_results} 
    This figure shows the localization results of different networks on the Deepfake Video dataset. Our proposed network's localization results are reasonable, with minor false alarms on column 1, 2, 3, 7 and mis-detections on column 4. Our network was able to largely identify the deepfaked faces in each of these example. However, for scenes similar to column 4, in which both the manipulated region and its surrounding regions had poor lighting condition, our network's performance seemed to be lower. Notably on column 7, although only the face was manipulated, we were able to detect the watermarked logo on the top right of the frame as well. This prediction is reasonable since the watermark was added in post-processing, hence, it could be considered an additional manipulation. Since other competing networks seemed to be identifying the entire body or the surrounding regions with distinct textures to be manipulated, they did not provide good localization performance in this dataset.  
    }

\end{figure*}


\begin{figure*}[!t]
	\vspace*{-0.7\baselineskip}
	
    \centering
    \setlength{\fboxsep}{0pt}

    \begin{minipage}[t]{1\textwidth}
        \makebox[0.083\textwidth][r]{\raisebox{15pt}{\smaller Frame\hspace{6pt}}}
        \fbox{\includegraphics[width=0.100\textwidth]{{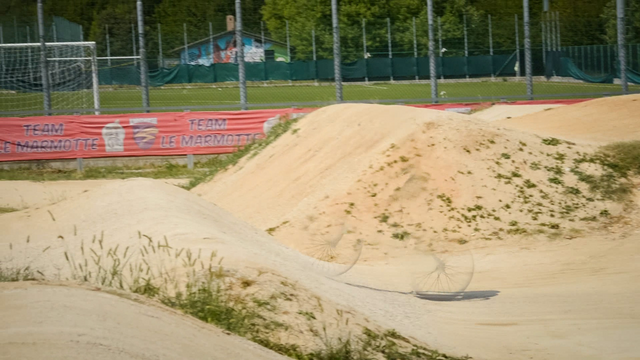}}}
        \fbox{\includegraphics[width=0.100\textwidth]{{}}}
        \fbox{\includegraphics[width=0.100\textwidth]{{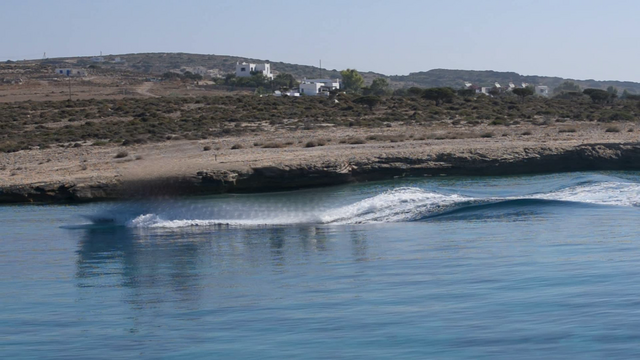}}}
        \fbox{\includegraphics[width=0.100\textwidth]{{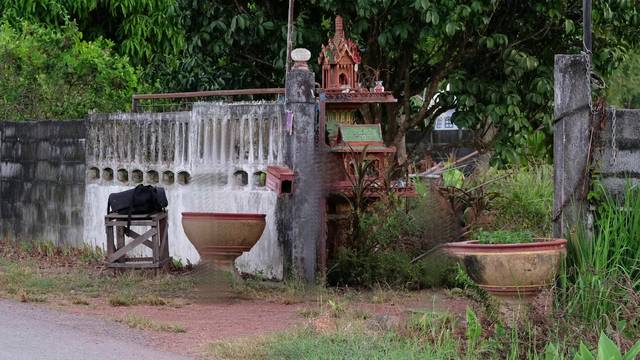}}}
        \fbox{\includegraphics[width=0.100\textwidth]{{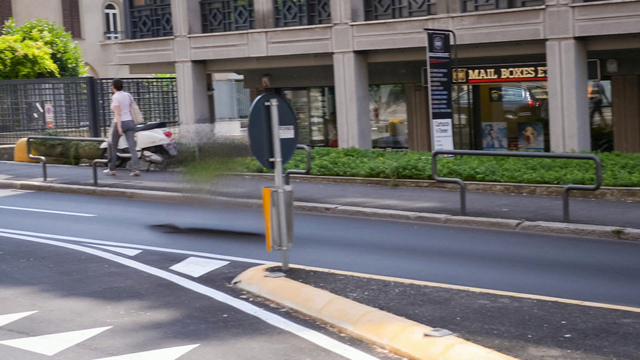}}}
        \fbox{\includegraphics[width=0.100\textwidth]{{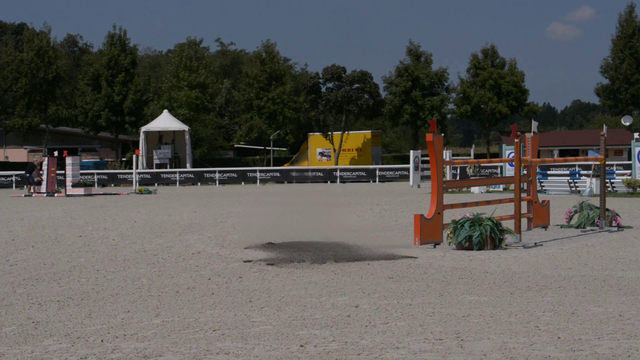}}}
        \fbox{\includegraphics[width=0.100\textwidth]{{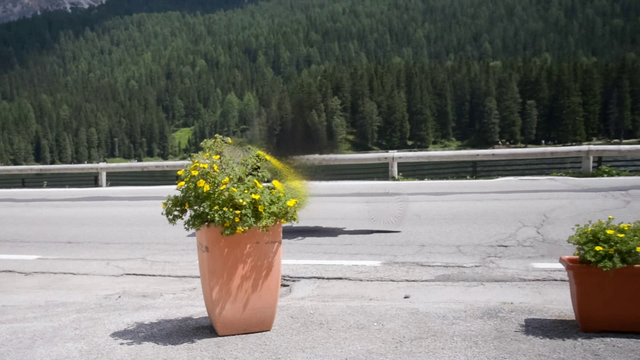}}}
        \fbox{\includegraphics[width=0.100\textwidth]{{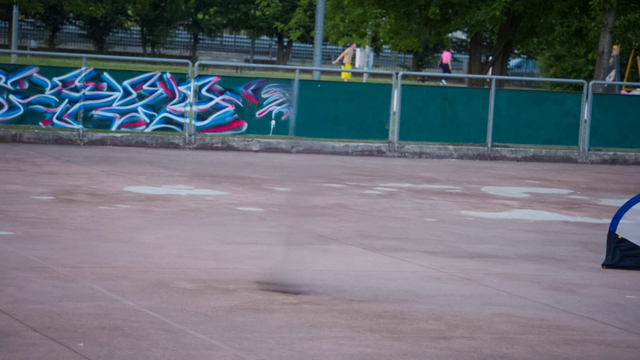}}}
        \smallskip
    \end{minipage}

	\vspace*{-0.1\baselineskip}

    \begin{minipage}[t]{1\textwidth}
        \makebox[0.083\textwidth][r]{\raisebox{15pt}{\smaller Mask\hspace{6pt}}}
        \fbox{\includegraphics[width=0.100\textwidth]{{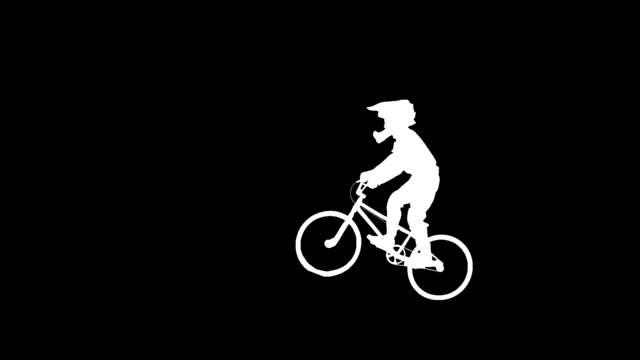}}}
        \fbox{\includegraphics[width=0.100\textwidth]{{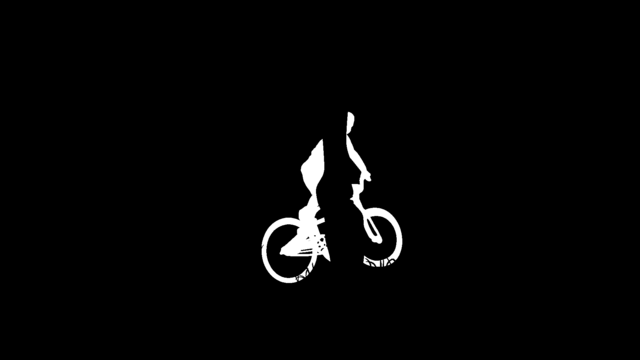}}}
        \fbox{\includegraphics[width=0.100\textwidth]{{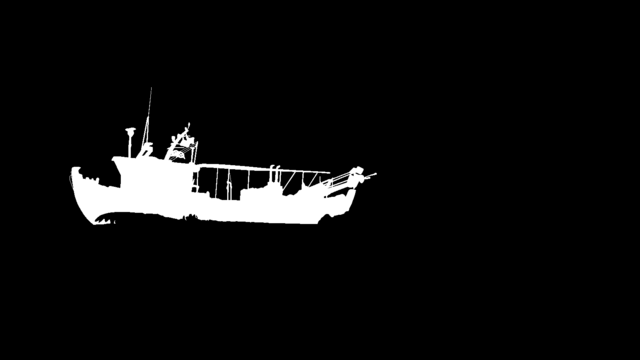}}}
        \fbox{\includegraphics[width=0.100\textwidth]{{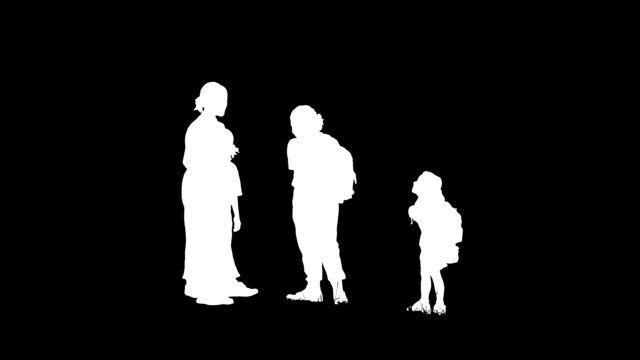}}}
        \fbox{\includegraphics[width=0.100\textwidth]{{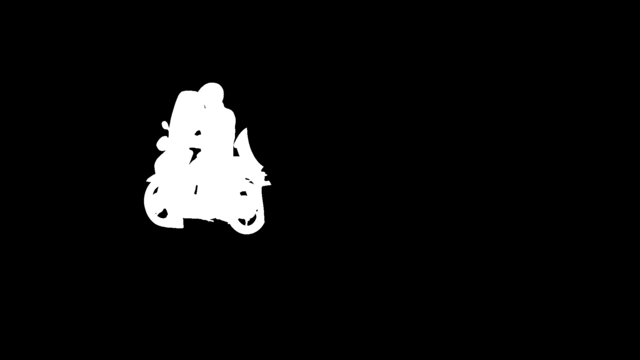}}}
        \fbox{\includegraphics[width=0.100\textwidth]{{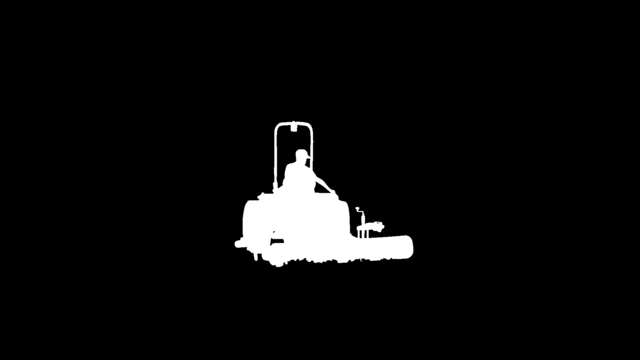}}}
        \fbox{\includegraphics[width=0.100\textwidth]{{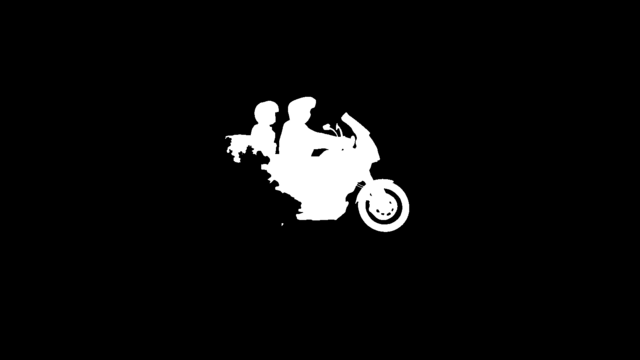}}}
        \fbox{\includegraphics[width=0.100\textwidth]{{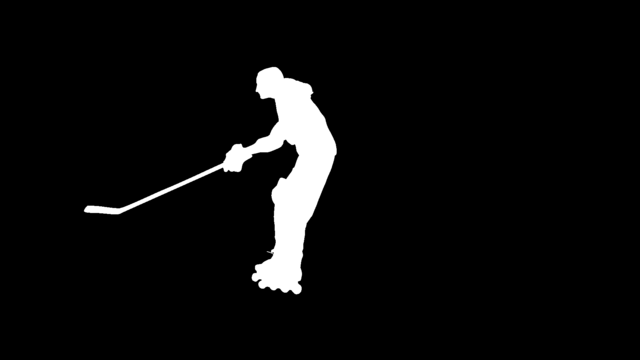}}}
        \smallskip
    \end{minipage}

	\vspace*{-0.1\baselineskip}

    \begin{minipage}[t]{1\textwidth}
        \makebox[0.083\textwidth][r]{\raisebox{15pt}{\smaller Proposed\hspace{6pt}}}
        \fbox{\includegraphics[width=0.100\textwidth]{{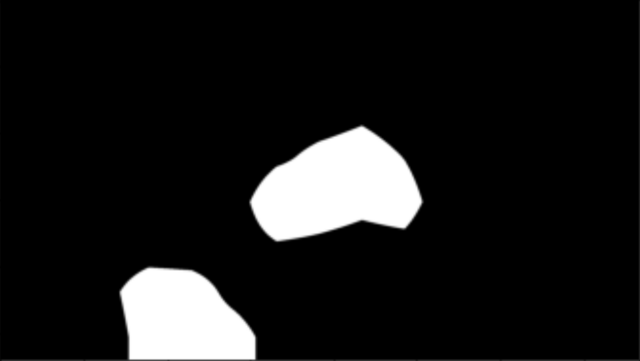}}}
        \fbox{\includegraphics[width=0.100\textwidth]{{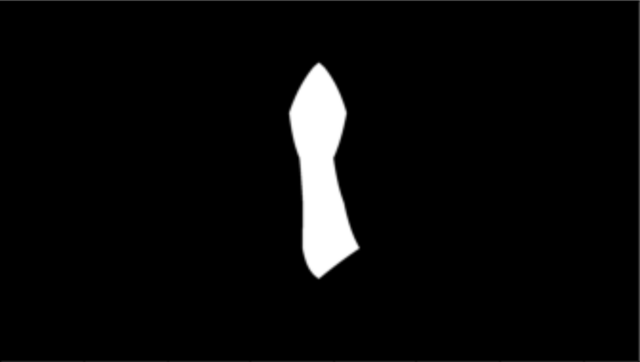}}}
        \fbox{\includegraphics[width=0.100\textwidth]{{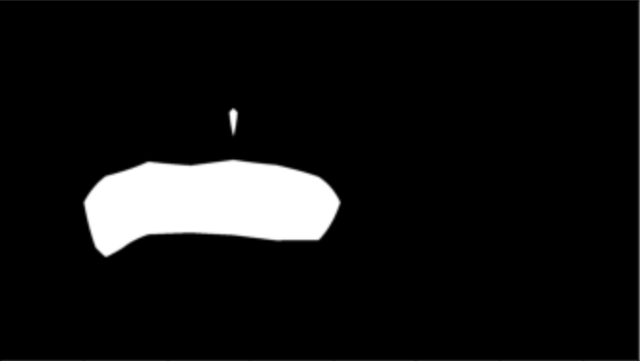}}}
        \fbox{\includegraphics[width=0.100\textwidth]{{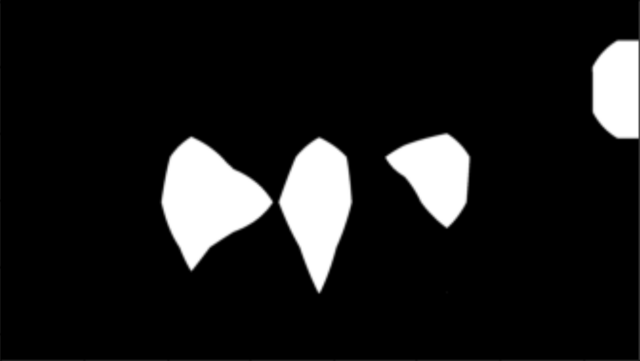}}}
        \fbox{\includegraphics[width=0.100\textwidth]{{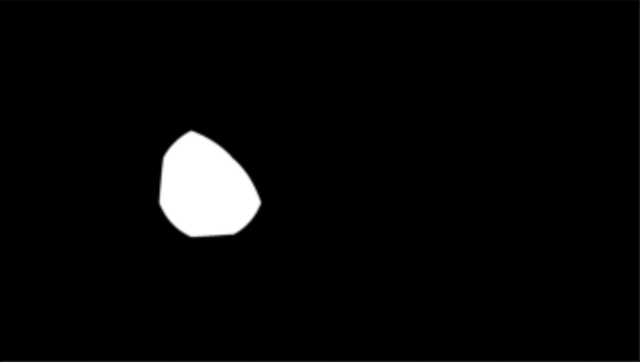}}}
        \fbox{\includegraphics[width=0.100\textwidth]{{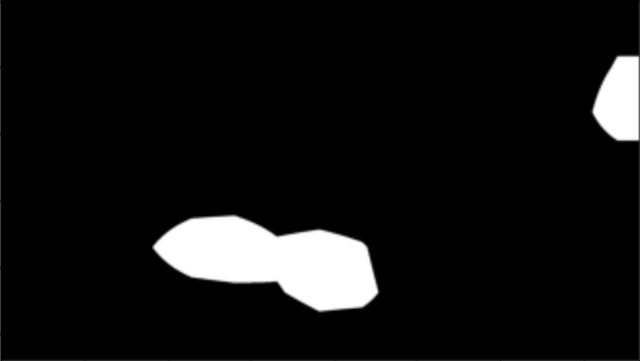}}}
        \fbox{\includegraphics[width=0.100\textwidth]{{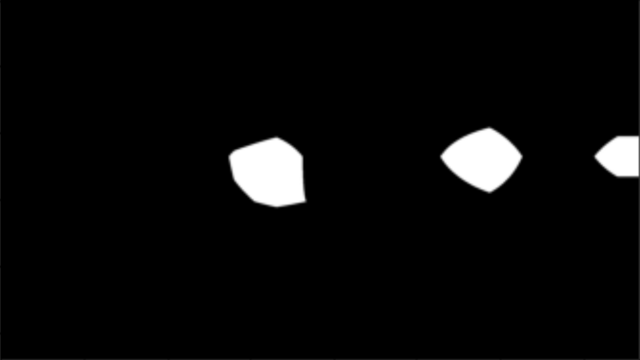}}}
        \fbox{\includegraphics[width=0.100\textwidth]{{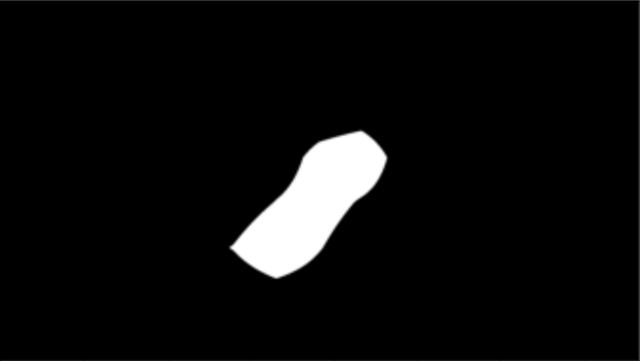}}}
        \smallskip
    \end{minipage}

	\vspace*{-0.1\baselineskip}

    \begin{minipage}[t]{1\textwidth}
        \makebox[0.083\textwidth][r]{\raisebox{15pt}{\smaller FSG\hspace{6pt}}}
        \fbox{\includegraphics[width=0.100\textwidth]{{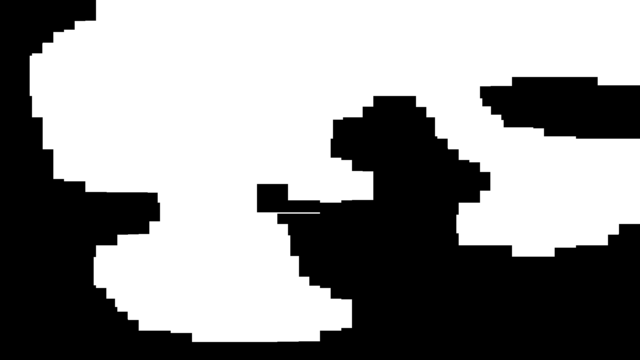}}}
        \fbox{\includegraphics[width=0.100\textwidth]{{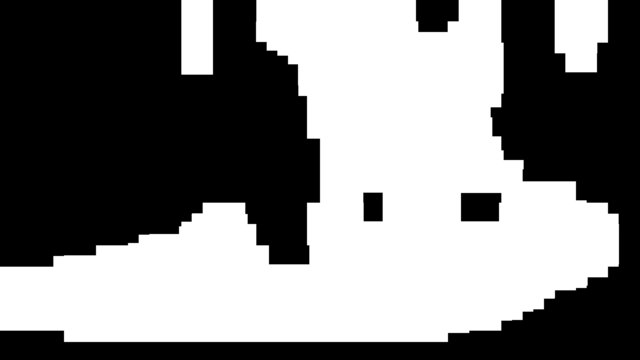}}}
        \fbox{\includegraphics[width=0.100\textwidth]{{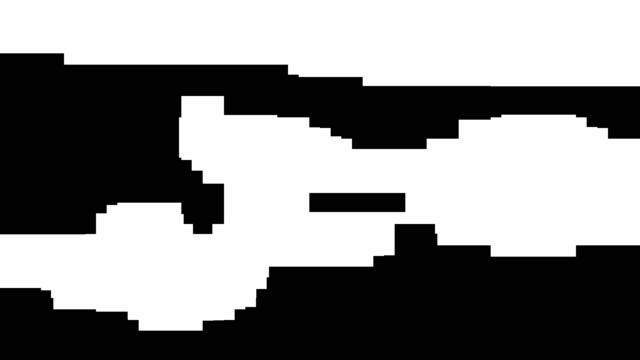}}}
        \fbox{\includegraphics[width=0.100\textwidth]{{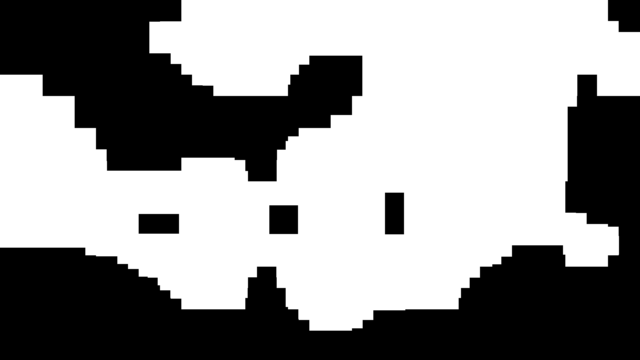}}}
        \fbox{\includegraphics[width=0.100\textwidth]{{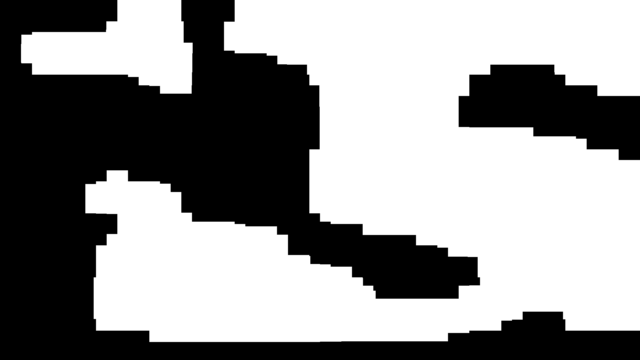}}}
        \fbox{\includegraphics[width=0.100\textwidth]{{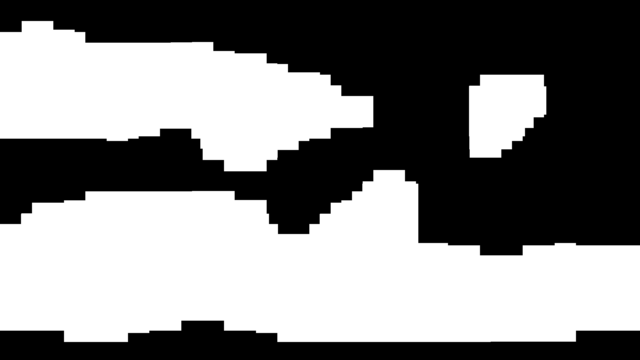}}}
        \fbox{\includegraphics[width=0.100\textwidth]{{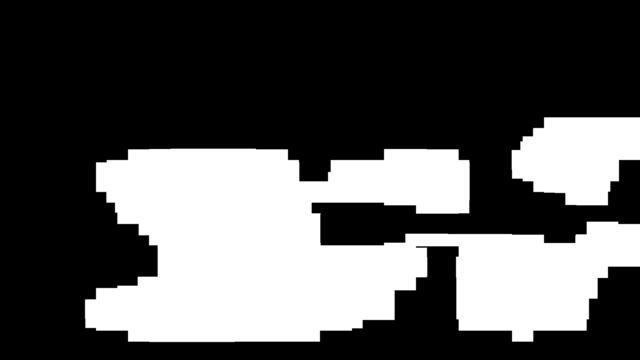}}}
        \fbox{\includegraphics[width=0.100\textwidth]{{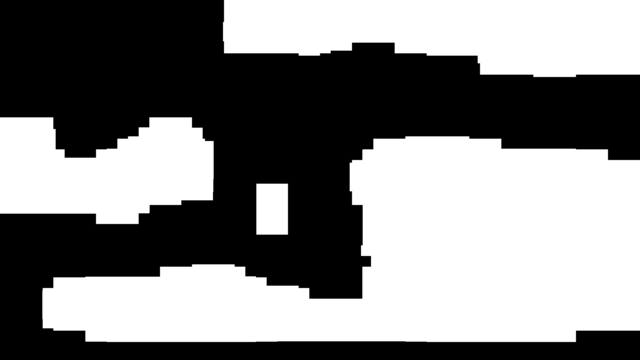}}}
        \smallskip
    \end{minipage}

	\vspace*{-0.1\baselineskip}

    \begin{minipage}[t]{1\textwidth}
        \makebox[0.083\textwidth][r]{\raisebox{15pt}{\smaller EXIFnet\hspace{6pt}}}
        \fbox{\includegraphics[width=0.100\textwidth]{{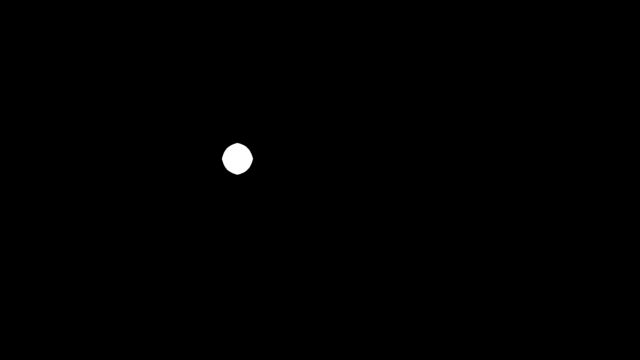}}}
        \fbox{\includegraphics[width=0.100\textwidth]{{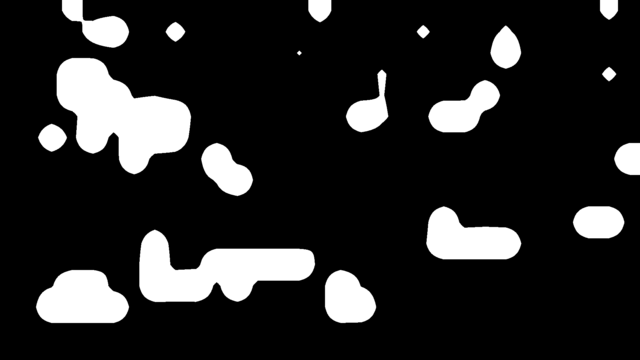}}}
        \fbox{\includegraphics[width=0.100\textwidth]{{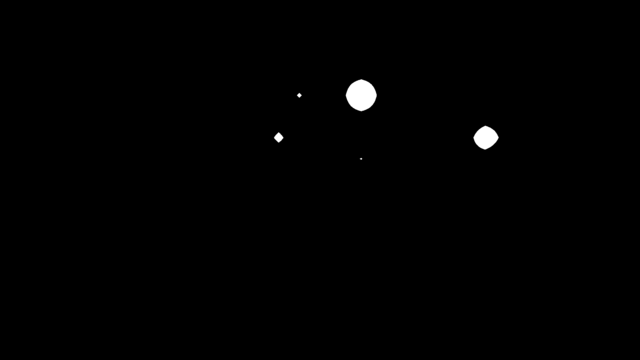}}}
        \fbox{\includegraphics[width=0.100\textwidth]{{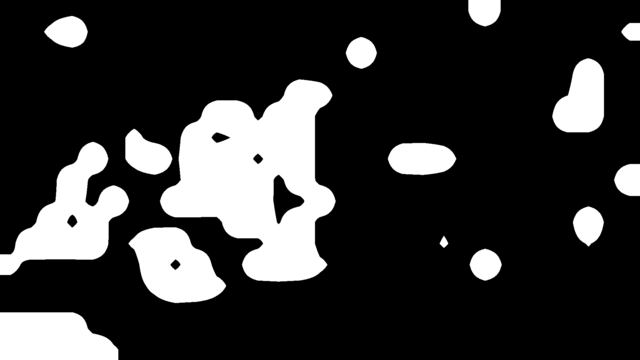}}}
        \fbox{\includegraphics[width=0.100\textwidth]{{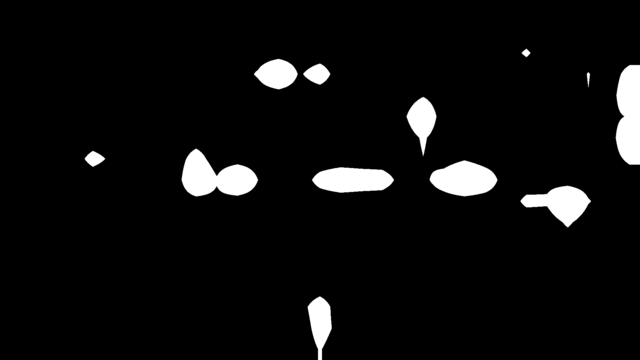}}}
        \fbox{\includegraphics[width=0.100\textwidth]{{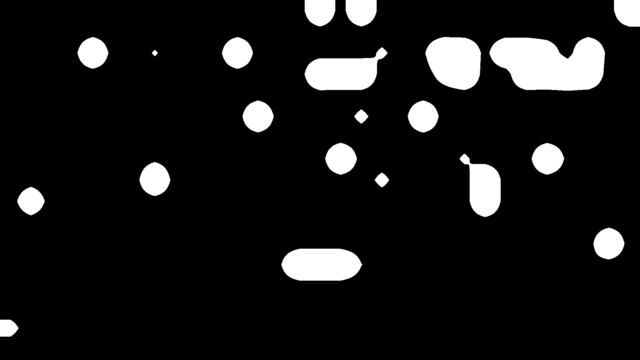}}}
        \fbox{\includegraphics[width=0.100\textwidth]{{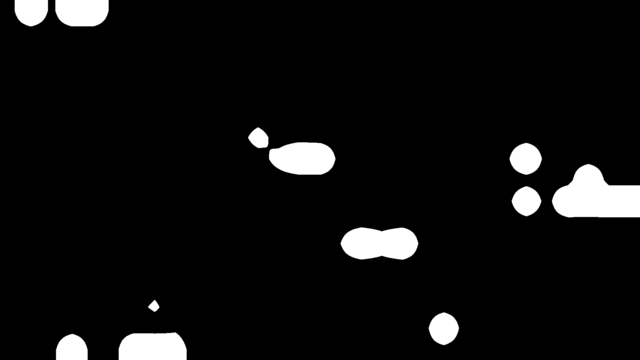}}}
        \fbox{\includegraphics[width=0.100\textwidth]{{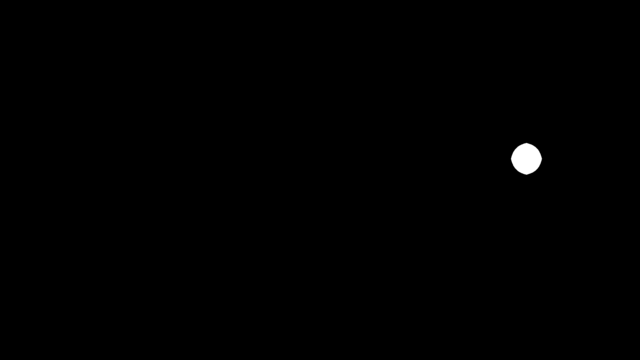}}}
        \smallskip
    \end{minipage}

	\vspace*{-0.1\baselineskip}

    \begin{minipage}[t]{1\textwidth}
        \makebox[0.083\textwidth][r]{\raisebox{15pt}{\smaller Noiseprint\hspace{6pt}}}
        \fbox{\includegraphics[width=0.100\textwidth]{{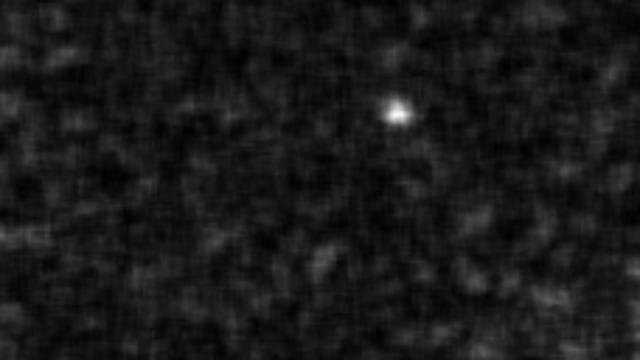}}}
        \fbox{\includegraphics[width=0.100\textwidth]{{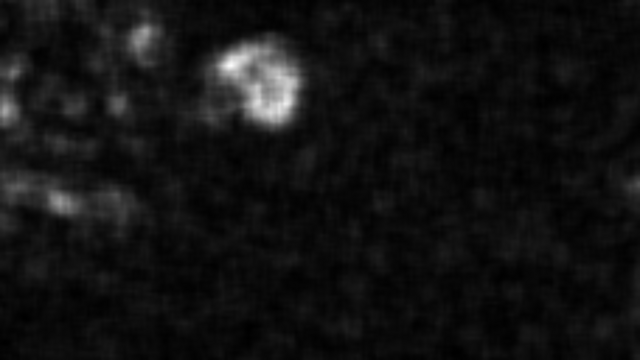}}}
        \fbox{\includegraphics[width=0.100\textwidth]{{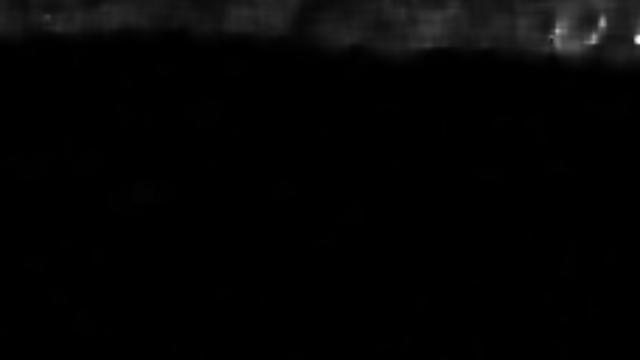}}}
        \fbox{\includegraphics[width=0.100\textwidth]{{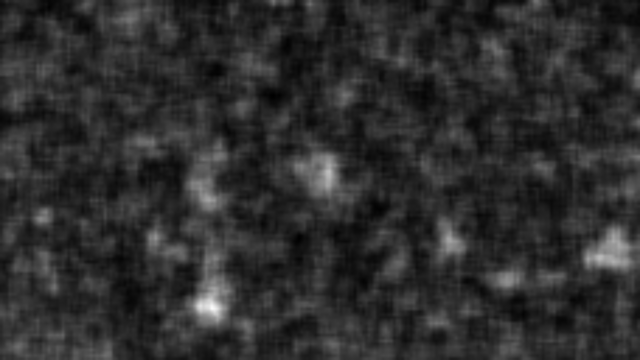}}}
        \fbox{\includegraphics[width=0.100\textwidth]{{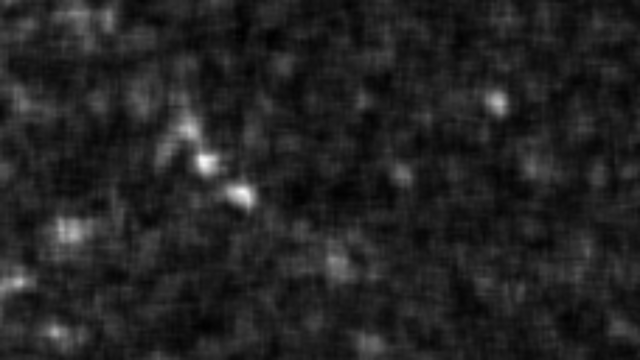}}}
        \fbox{\includegraphics[width=0.100\textwidth]{{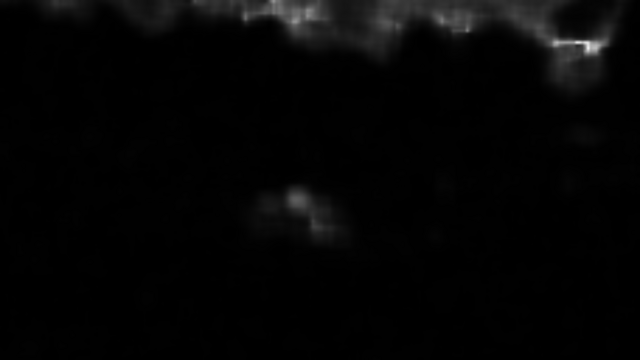}}}
        \fbox{\includegraphics[width=0.100\textwidth]{{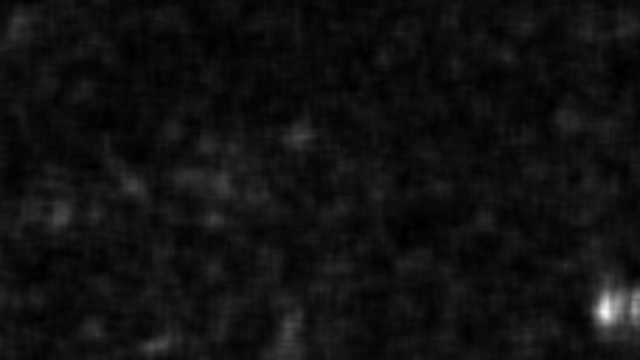}}}
        \fbox{\includegraphics[width=0.100\textwidth]{{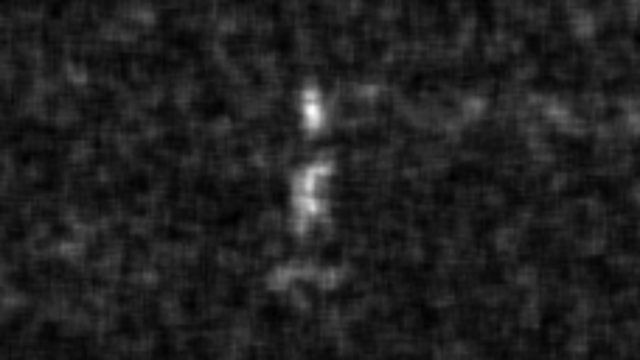}}}
        \smallskip
    \end{minipage}

	\vspace*{-0.1\baselineskip}

    \begin{minipage}[t]{1\textwidth}
        \makebox[0.083\textwidth][r]{\raisebox{15pt}{\smaller ManTra-Net\hspace{6pt}}}
        \fbox{\includegraphics[width=0.100\textwidth]{{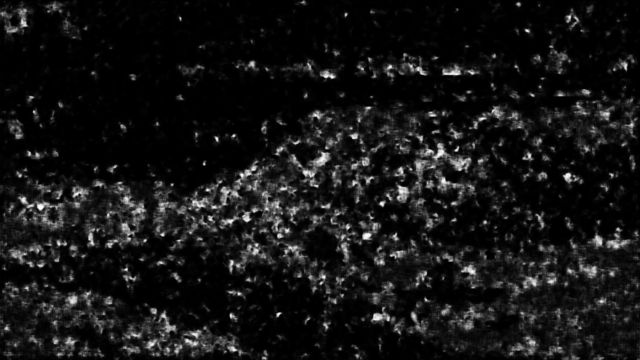}}}
        \fbox{\includegraphics[width=0.100\textwidth]{{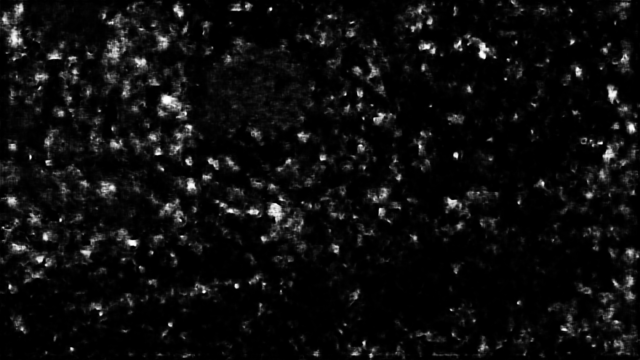}}}
        \fbox{\includegraphics[width=0.100\textwidth]{{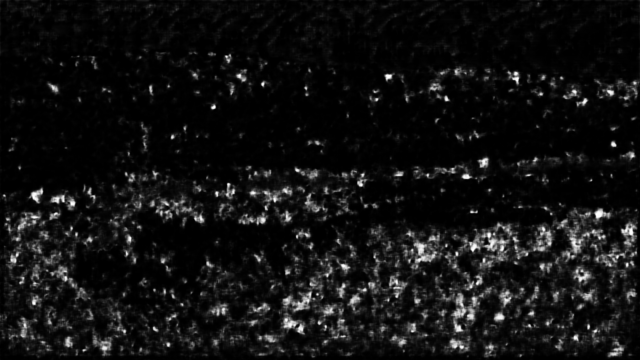}}}
        \fbox{\includegraphics[width=0.100\textwidth]{{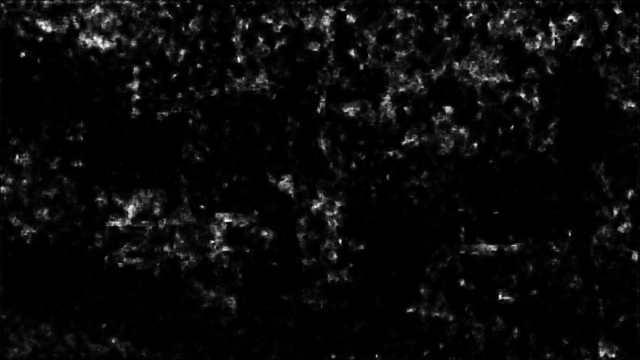}}}
        \fbox{\includegraphics[width=0.100\textwidth]{{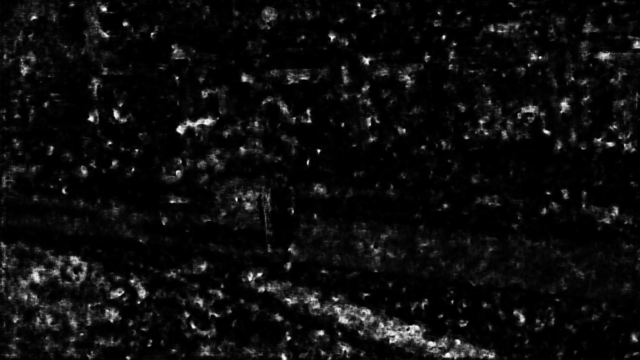}}}
        \fbox{\includegraphics[width=0.100\textwidth]{{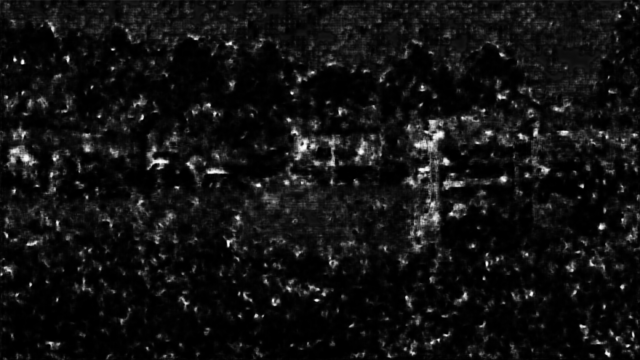}}}
        \fbox{\includegraphics[width=0.100\textwidth]{{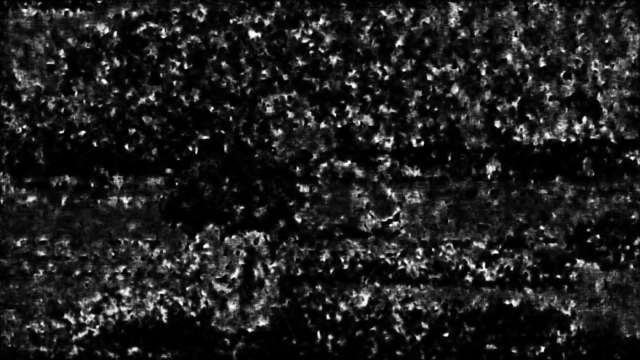}}}
        \fbox{\includegraphics[width=0.100\textwidth]{{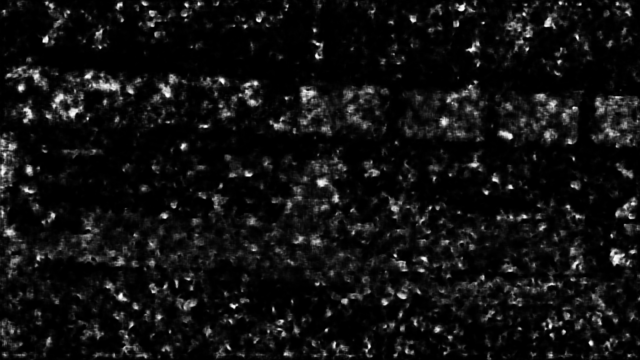}}}
        \smallskip
    \end{minipage}

	\vspace*{-0.1\baselineskip}

    \begin{minipage}[t]{1\textwidth}
        \makebox[0.083\textwidth][r]{\raisebox{15pt}{\smaller MVSS-Net\hspace{6pt}}}
        \fbox{\includegraphics[width=0.100\textwidth]{{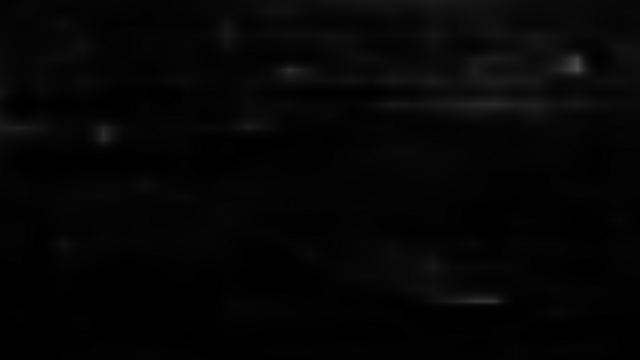}}}
        \fbox{\includegraphics[width=0.100\textwidth]{{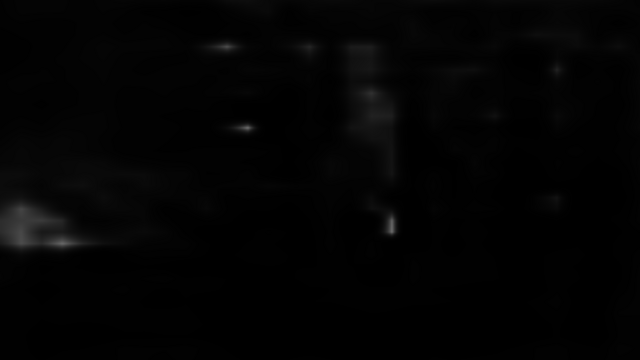}}}
        \fbox{\includegraphics[width=0.100\textwidth]{{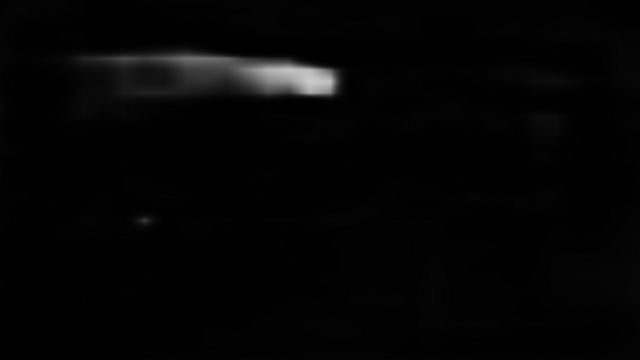}}}
        \fbox{\includegraphics[width=0.100\textwidth]{{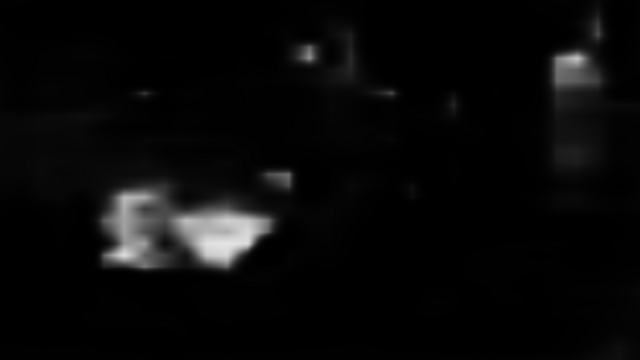}}}
        \fbox{\includegraphics[width=0.100\textwidth]{{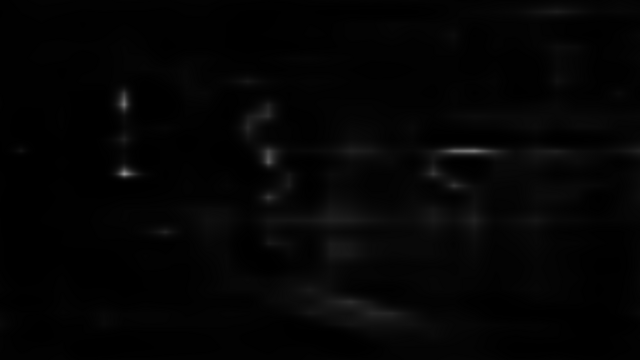}}}
        \fbox{\includegraphics[width=0.100\textwidth]{{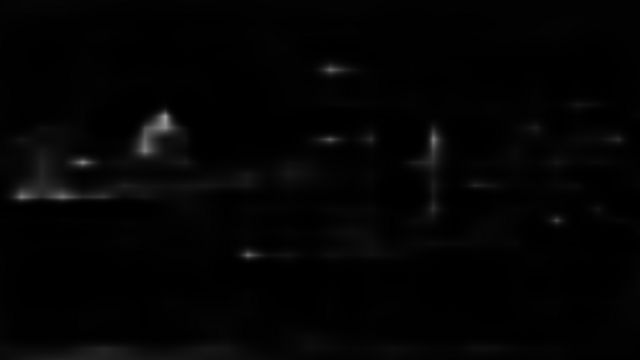}}}
        \fbox{\includegraphics[width=0.100\textwidth]{{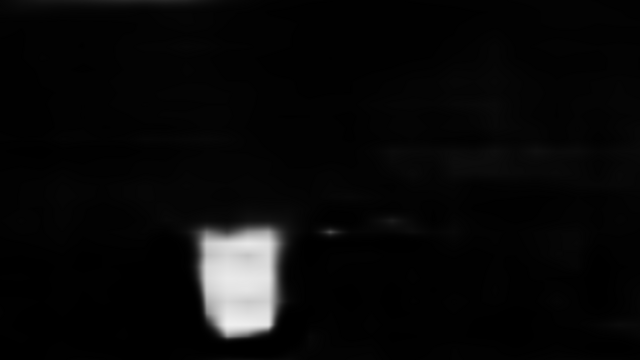}}}
        \fbox{\includegraphics[width=0.100\textwidth]{{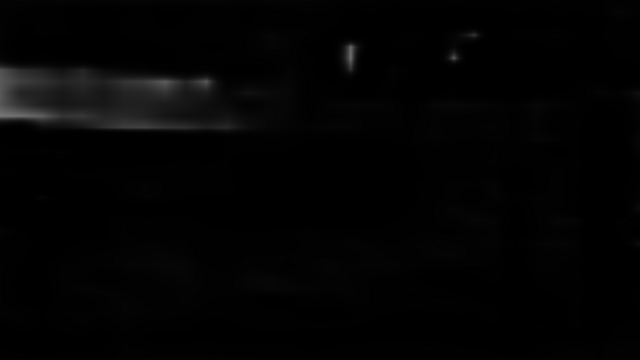}}}
        \smallskip
    \end{minipage}
    
	\vspace*{-0.8\baselineskip}

    \caption{\label{fig:inpaint_localization_results} 
    This figure shows the localization results of different networks on the Inpainted Video dataset. Our proposed network's localization results are reasonable, with false alarms on column 1, 4, 6, 7 and mis-detections on column 7. Our network were able to largely identify the regions where the segmented objects were removed. On column 7, although our network misdetected the removed motorbike on this frame, we were able to occasionally catch it in other frames of the video. Generally, when the video was stable, without a lot of moving objects, we were able to reliably detect the manipulated regions. However, if the camera moved around too quickly, the frame was blurred, which made our network less likely to provide accurate predictions. While our network generalized well over this dataset, other competing networks failed to identify any manipulations in these videos. We suspected that since objects were removed, there existed no edge information for these networks to rely on, and the residual information might be too different for them to behave properly.
    }
    

\end{figure*}


\begin{figure*}[!t]
	\vspace*{0.3\baselineskip}
	
    \centering
    \setlength{\fboxsep}{0pt}

    \begin{minipage}[t]{1\textwidth}
        \makebox[0.083\textwidth][r]{\raisebox{15pt}{\smaller Frame\hspace{6pt}}}
        \fbox{\includegraphics[width=0.100\textwidth]{{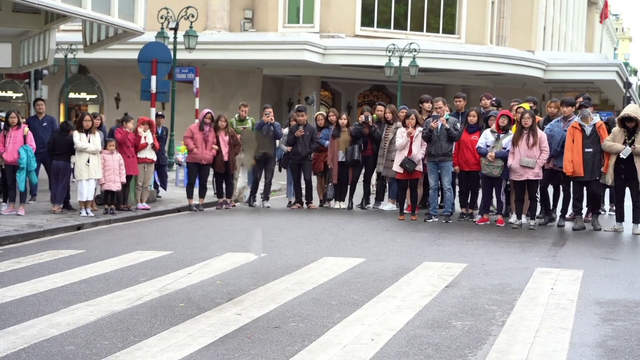}}}
        \fbox{\includegraphics[width=0.100\textwidth]{{figs/loc_comparisons/video_sham_adobe/video_transformer_manip_0087_0140.png}}}
        \fbox{\includegraphics[width=0.100\textwidth]{{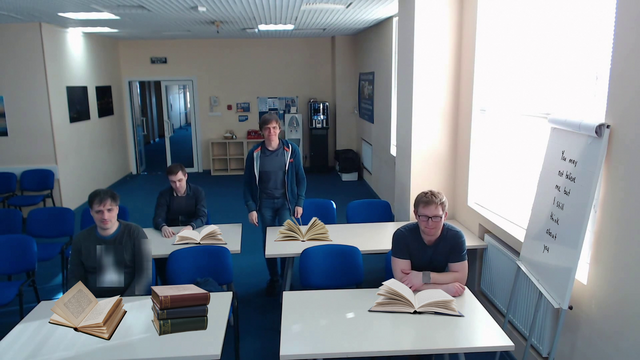}}}
        \fbox{\includegraphics[width=0.100\textwidth]{{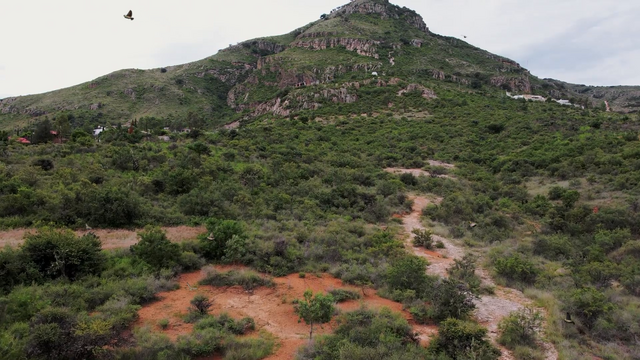}}}
        \fbox{\includegraphics[width=0.100\textwidth]{{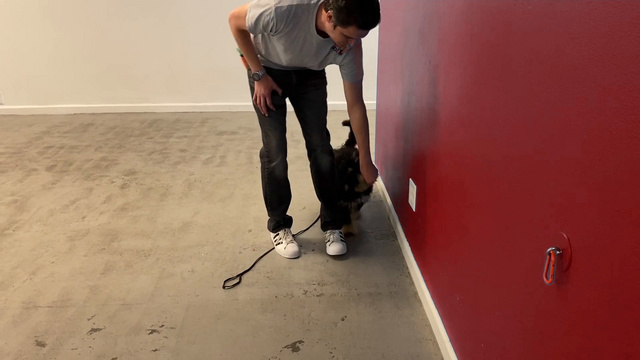}}}
        \fbox{\includegraphics[width=0.100\textwidth]{{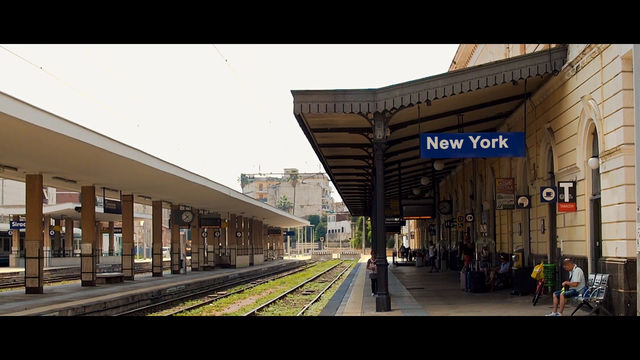}}}
        \fbox{\includegraphics[width=0.100\textwidth]{{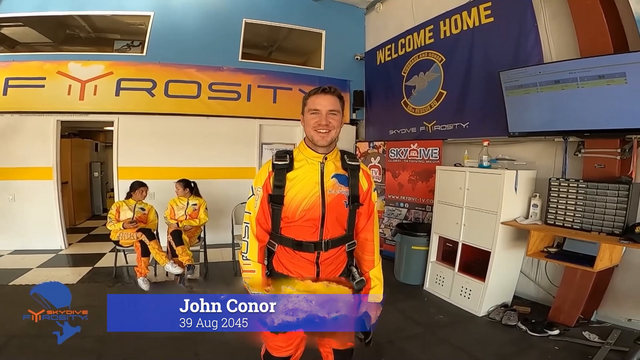}}}
        \fbox{\includegraphics[width=0.100\textwidth]{{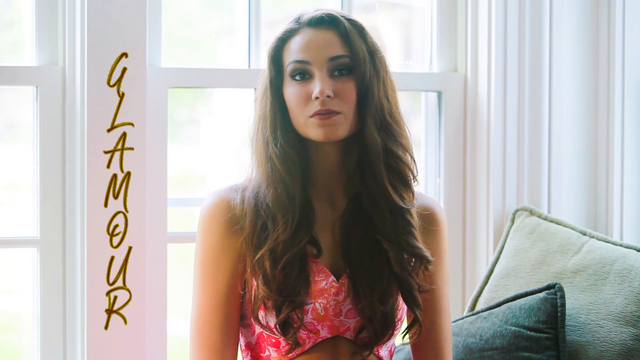}}}
        \smallskip
    \end{minipage}

	\vspace*{-0.1\baselineskip}

    \begin{minipage}[t]{1\textwidth}
        \makebox[0.083\textwidth][r]{\raisebox{15pt}{\smaller Mask\hspace{6pt}}}
        \fbox{\includegraphics[width=0.100\textwidth]{{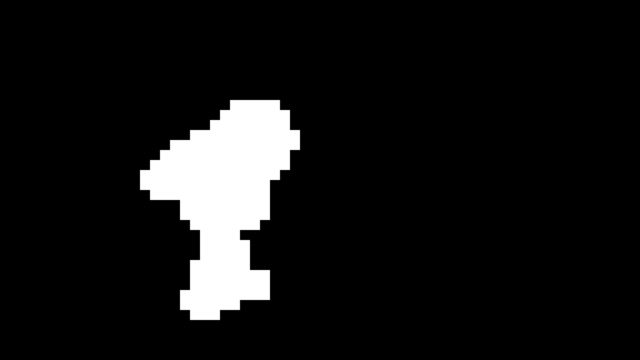}}}
        \fbox{\includegraphics[width=0.100\textwidth]{{figs/loc_comparisons/video_sham_adobe/video_transformer_manip_0087_0140_gt_mask.png}}}
        \fbox{\includegraphics[width=0.100\textwidth]{{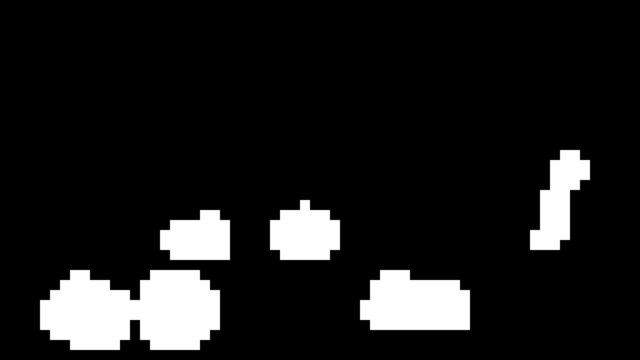}}}
        \fbox{\includegraphics[width=0.100\textwidth]{{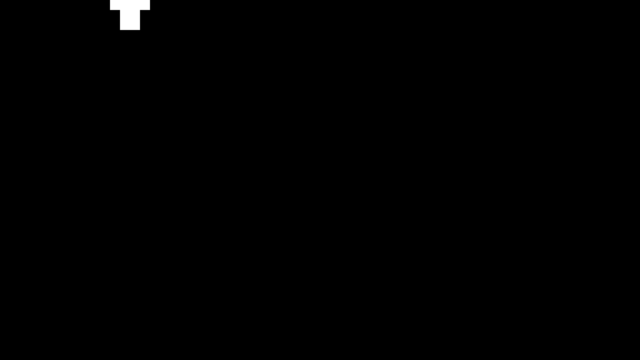}}}
        \fbox{\includegraphics[width=0.100\textwidth]{{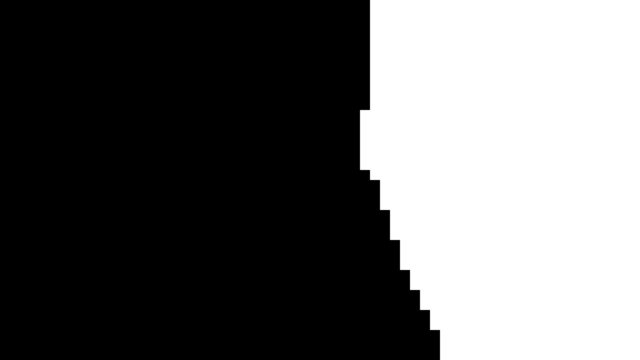}}}
        \fbox{\includegraphics[width=0.100\textwidth]{{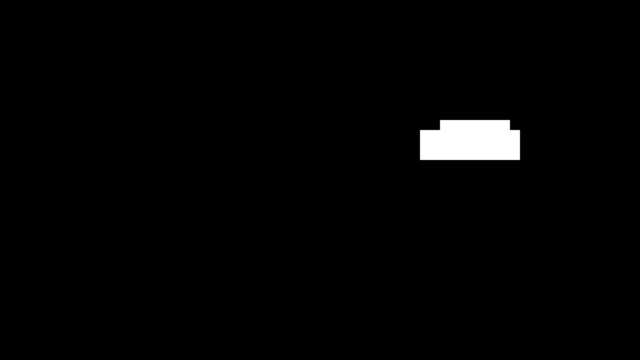}}}
        \fbox{\includegraphics[width=0.100\textwidth]{{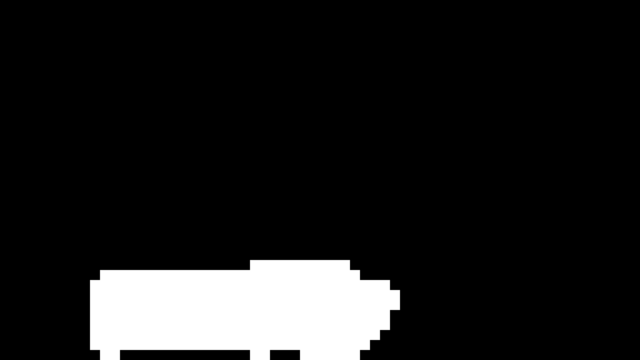}}}
        \fbox{\includegraphics[width=0.100\textwidth]{{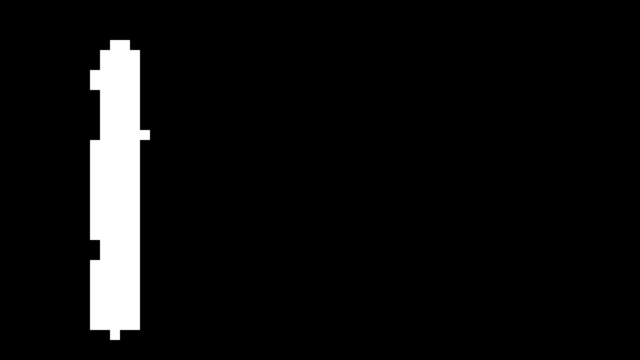}}}
        \smallskip
    \end{minipage}

	\vspace*{-0.1\baselineskip}

    \begin{minipage}[t]{1\textwidth}
        \makebox[0.083\textwidth][r]{\raisebox{15pt}{\smaller Proposed\hspace{6pt}}}
        \fbox{\includegraphics[width=0.100\textwidth]{{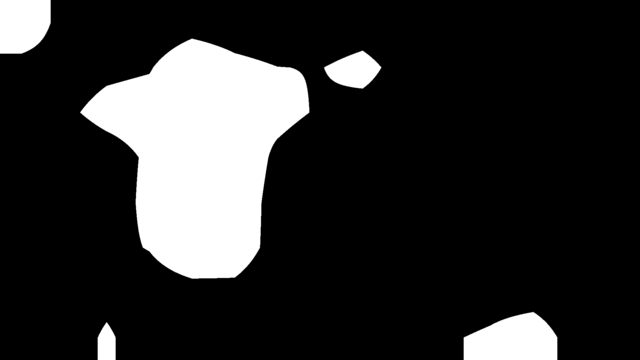}}}
        \fbox{\includegraphics[width=0.100\textwidth]{{figs/loc_comparisons/video_sham_adobe/video_transformer_manip_0087_0140_pred_mask.png}}}
        \fbox{\includegraphics[width=0.100\textwidth]{{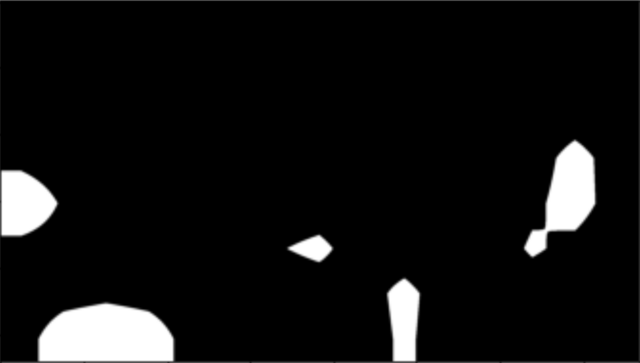}}}
        \fbox{\includegraphics[width=0.100\textwidth]{{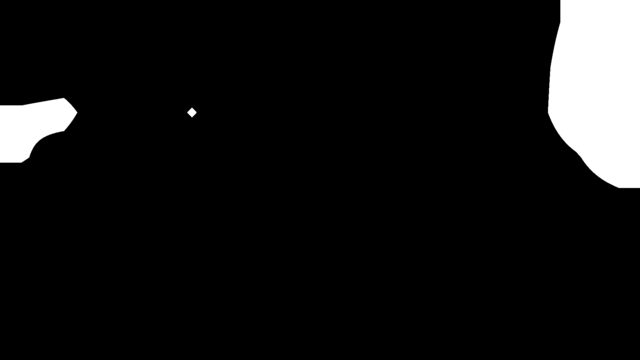}}}
        \fbox{\includegraphics[width=0.100\textwidth]{{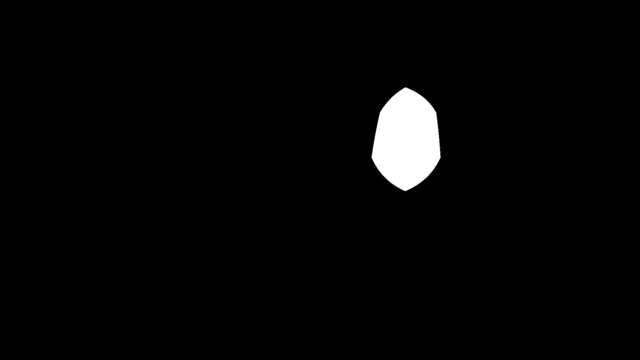}}}
        \fbox{\includegraphics[width=0.100\textwidth]{{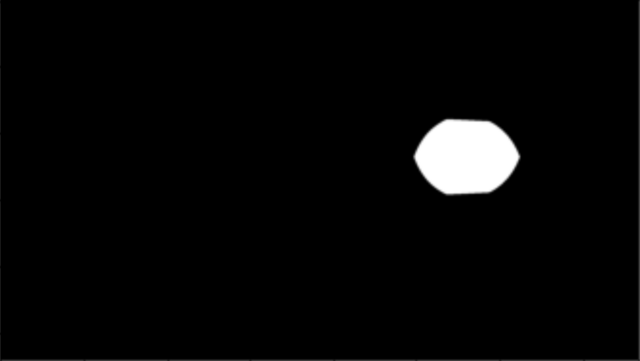}}}
        \fbox{\includegraphics[width=0.100\textwidth]{{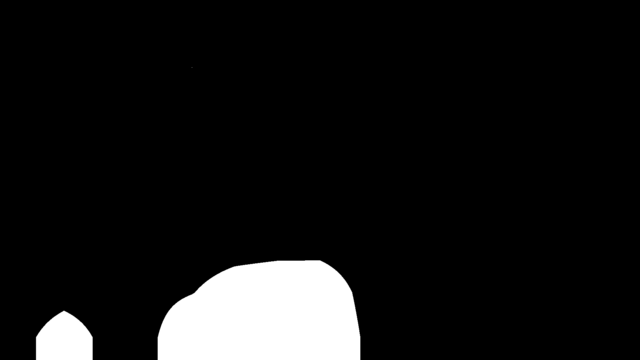}}}
        \fbox{\includegraphics[width=0.100\textwidth]{{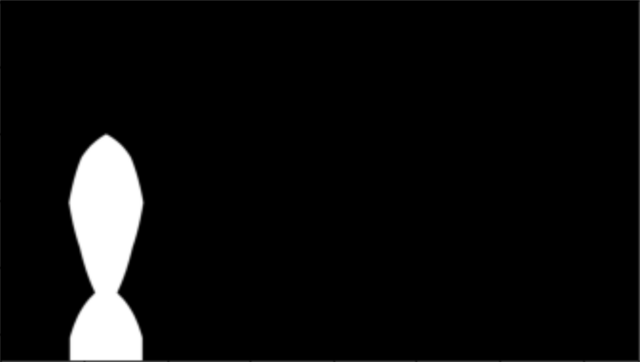}}}
        \smallskip
    \end{minipage}

	\vspace*{-0.1\baselineskip}

    \begin{minipage}[t]{1\textwidth}
        \makebox[0.083\textwidth][r]{\raisebox{15pt}{\smaller FSG\hspace{6pt}}}
        \fbox{\includegraphics[width=0.100\textwidth]{{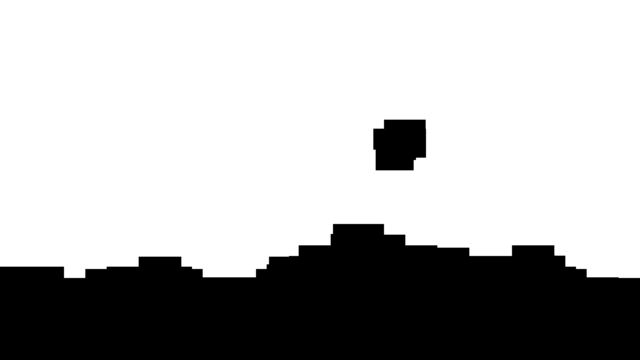}}}
        \fbox{\includegraphics[width=0.100\textwidth]{{figs/loc_comparisons/video_sham_adobe/fsg_manip_0087_0140_pred_mask.png}}}
        \fbox{\includegraphics[width=0.100\textwidth]{{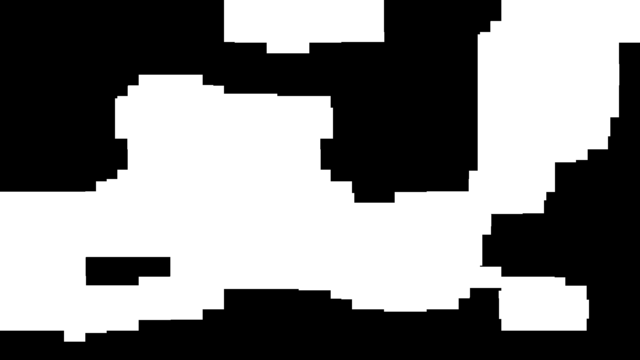}}}
        \fbox{\includegraphics[width=0.100\textwidth]{{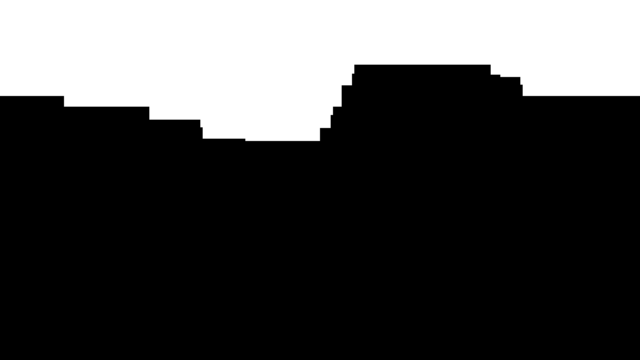}}}
        \fbox{\includegraphics[width=0.100\textwidth]{{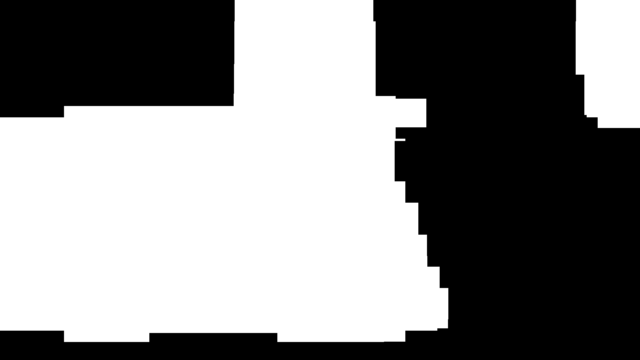}}}
        \fbox{\includegraphics[width=0.100\textwidth]{{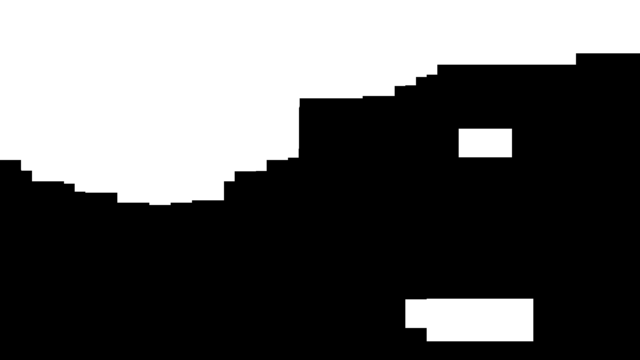}}}
        \fbox{\includegraphics[width=0.100\textwidth]{{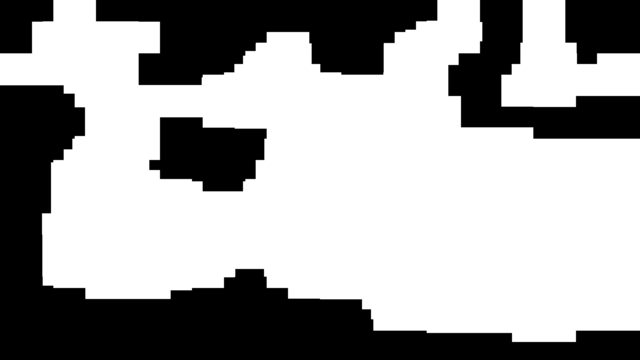}}}
        \fbox{\includegraphics[width=0.100\textwidth]{{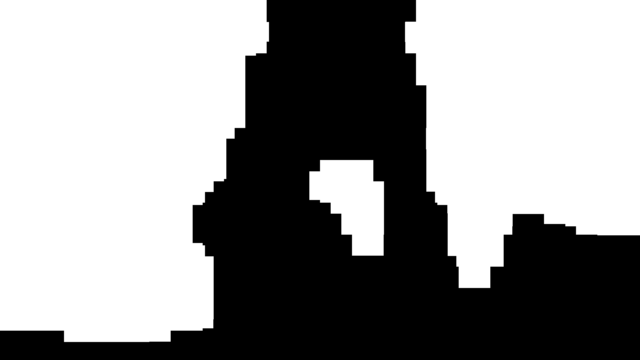}}}
        \smallskip
    \end{minipage}

	\vspace*{-0.1\baselineskip}

    \begin{minipage}[t]{1\textwidth}
        \makebox[0.083\textwidth][r]{\raisebox{15pt}{\smaller EXIFnet\hspace{6pt}}}
        \fbox{\includegraphics[width=0.100\textwidth]{{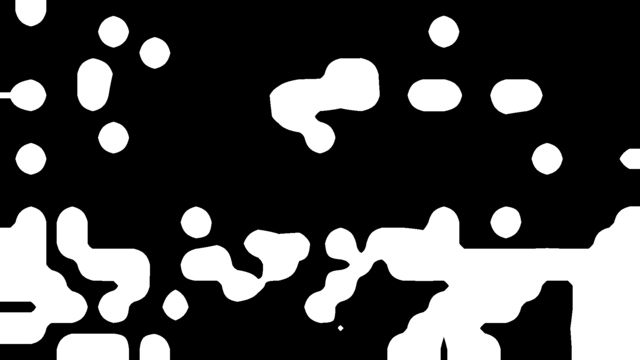}}}
        \fbox{\includegraphics[width=0.100\textwidth]{{figs/loc_comparisons/video_sham_adobe/exif_manip_0087_0140_pred_mask.png}}}
        \fbox{\includegraphics[width=0.100\textwidth]{{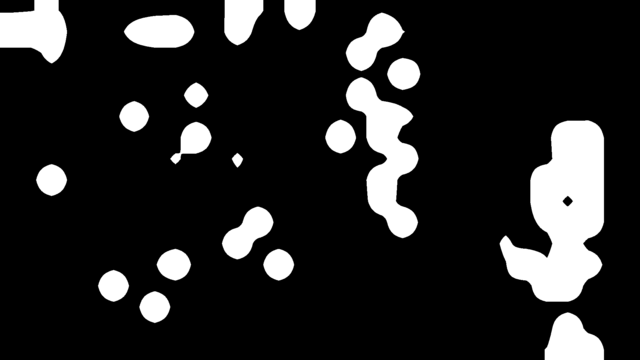}}}
        \fbox{\includegraphics[width=0.100\textwidth]{{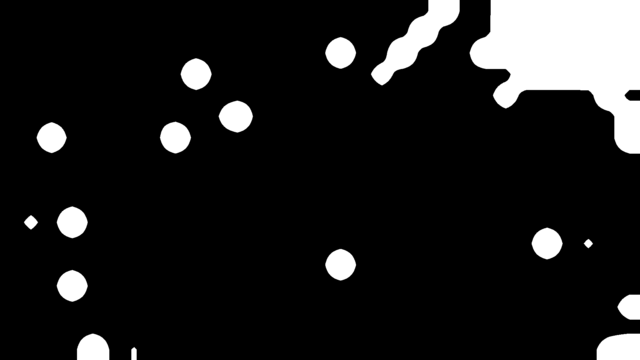}}}
        \fbox{\includegraphics[width=0.100\textwidth]{{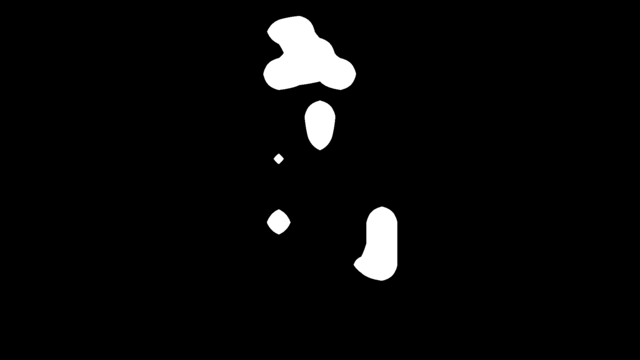}}}
        \fbox{\includegraphics[width=0.100\textwidth]{{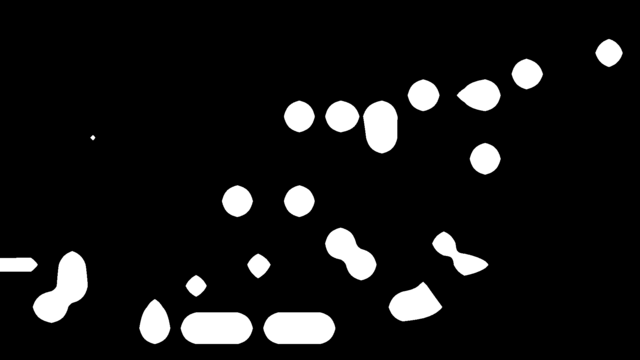}}}
        \fbox{\includegraphics[width=0.100\textwidth]{{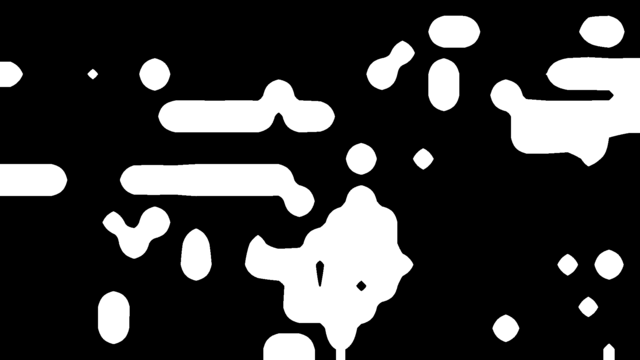}}}
        \fbox{\includegraphics[width=0.100\textwidth]{{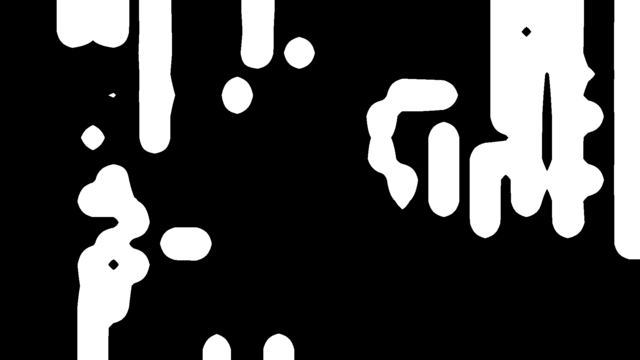}}}
        \smallskip
    \end{minipage}

	\vspace*{-0.1\baselineskip}

    \begin{minipage}[t]{1\textwidth}
        \makebox[0.083\textwidth][r]{\raisebox{15pt}{\smaller Noiseprint\hspace{6pt}}}
        \fbox{\includegraphics[width=0.100\textwidth]{{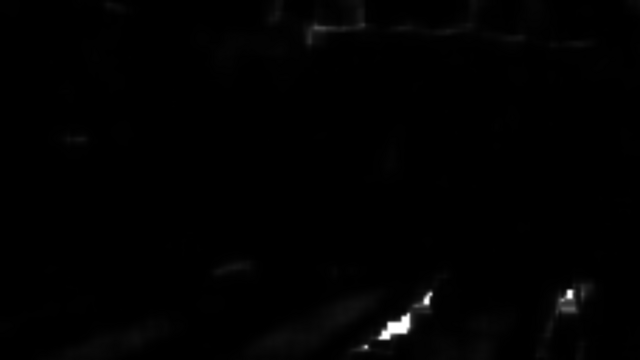}}}
        \fbox{\includegraphics[width=0.100\textwidth]{{figs/loc_comparisons/video_sham_adobe/noiseprint_manip_0087_0140_pred_mask.png}}}
        \fbox{\includegraphics[width=0.100\textwidth]{{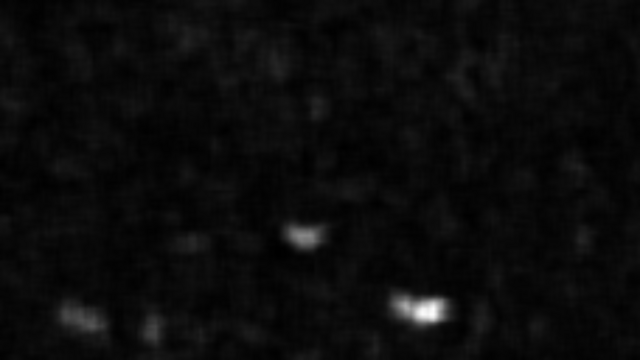}}}
        \fbox{\includegraphics[width=0.100\textwidth]{{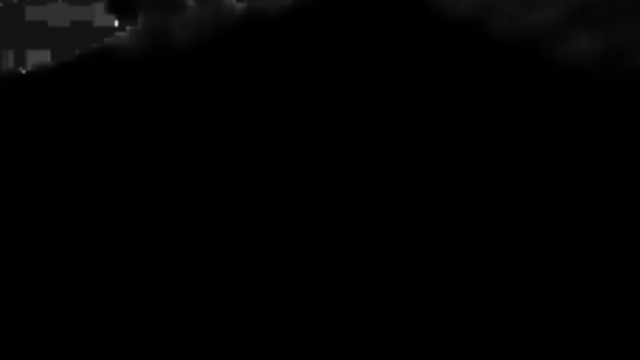}}}
        \fbox{\includegraphics[width=0.100\textwidth]{{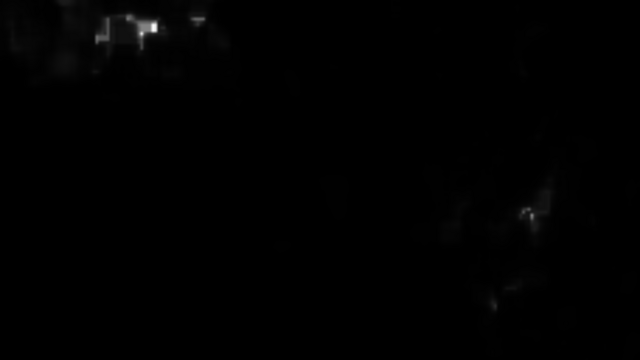}}}
        \fbox{\includegraphics[width=0.100\textwidth]{{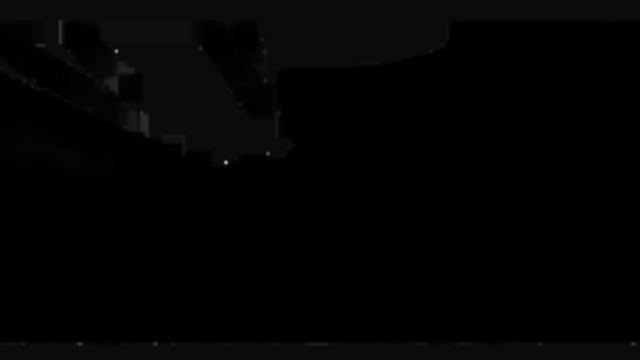}}}
        \fbox{\includegraphics[width=0.100\textwidth]{{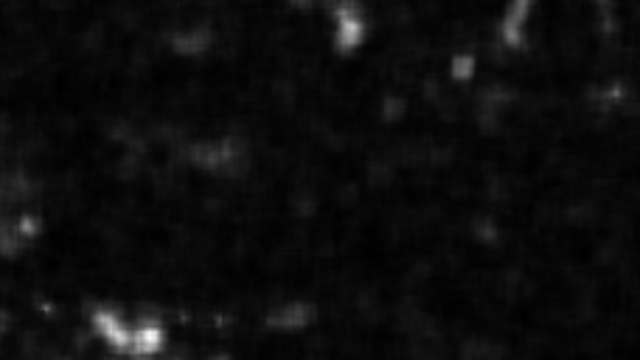}}}
        \fbox{\includegraphics[width=0.100\textwidth]{{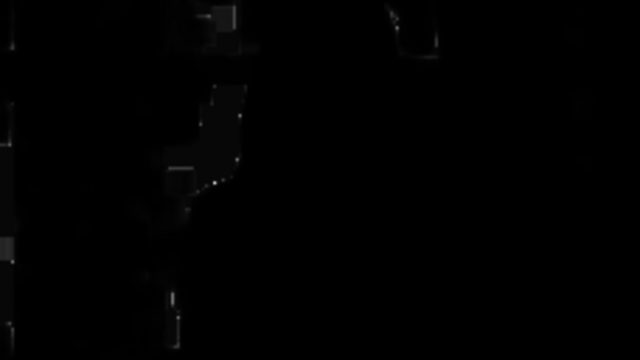}}}
        \smallskip
    \end{minipage}

	\vspace*{-0.1\baselineskip}

    \begin{minipage}[t]{1\textwidth}
        \makebox[0.083\textwidth][r]{\raisebox{15pt}{\smaller ManTra-Net\hspace{6pt}}}
        \fbox{\includegraphics[width=0.100\textwidth]{{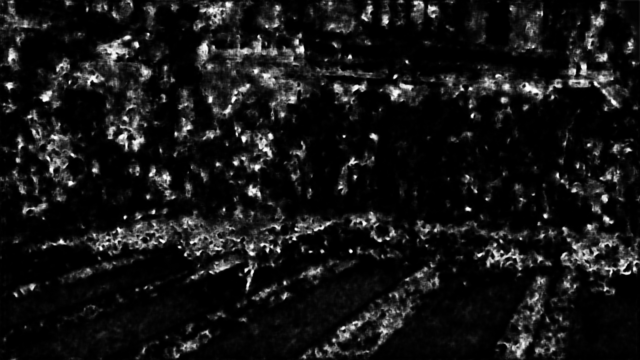}}}
        \fbox{\includegraphics[width=0.100\textwidth]{{figs/loc_comparisons/video_sham_adobe/mantranet_manip_0087_0140_pred_mask.png}}}
        \fbox{\includegraphics[width=0.100\textwidth]{{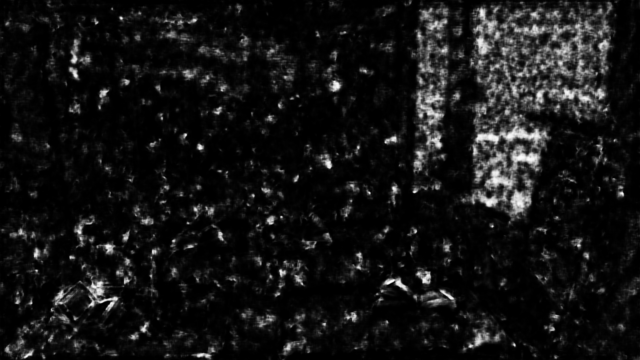}}}
        \fbox{\includegraphics[width=0.100\textwidth]{{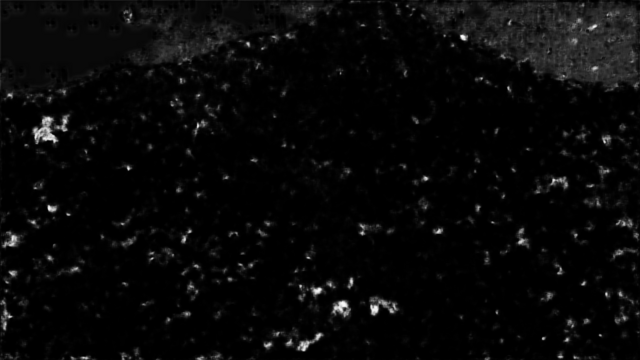}}}
        \fbox{\includegraphics[width=0.100\textwidth]{{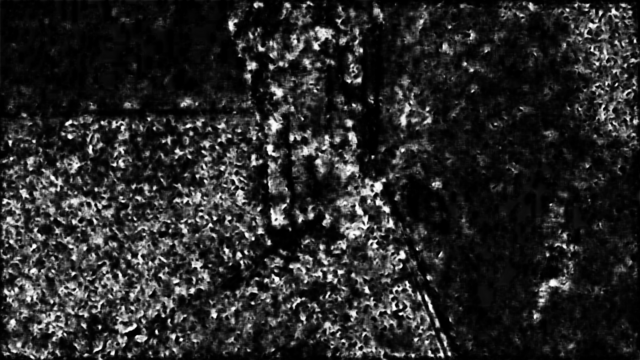}}}
        \fbox{\includegraphics[width=0.100\textwidth]{{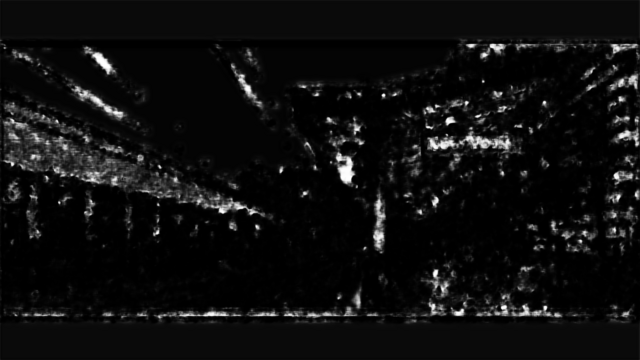}}}
        \fbox{\includegraphics[width=0.100\textwidth]{{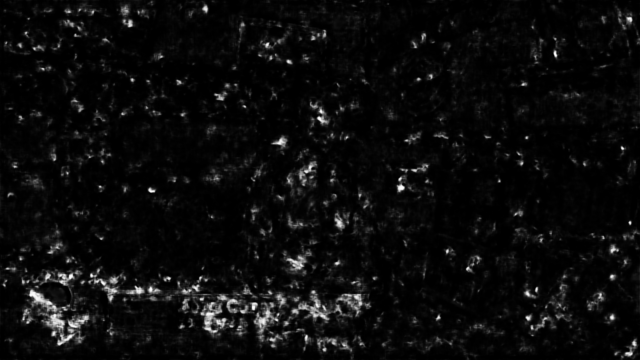}}}
        \fbox{\includegraphics[width=0.100\textwidth]{{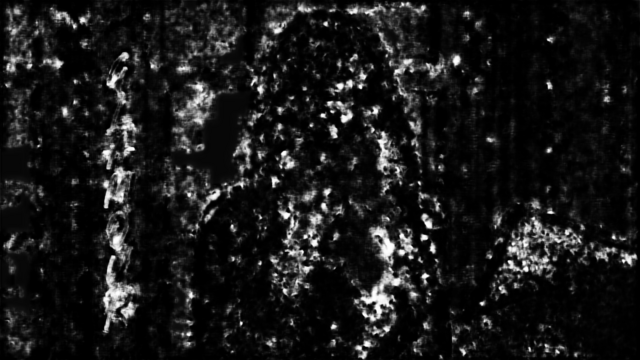}}}
        \smallskip
    \end{minipage}

	\vspace*{-0.1\baselineskip}

    \begin{minipage}[t]{1\textwidth}
        \makebox[0.083\textwidth][r]{\raisebox{15pt}{\smaller MVSS-Net\hspace{6pt}}}
        \fbox{\includegraphics[width=0.100\textwidth]{{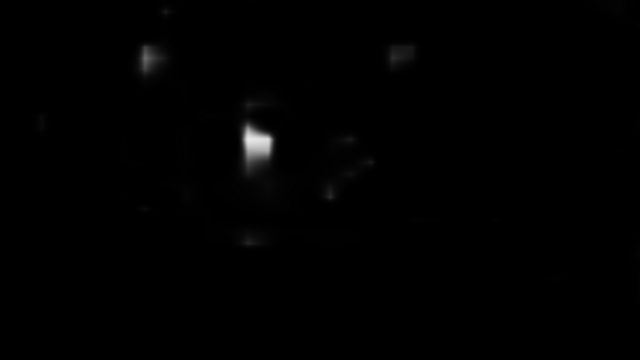}}}
        \fbox{\includegraphics[width=0.100\textwidth]{{figs/loc_comparisons/video_sham_adobe/mvss_manip_0087_0140_pred_mask.png}}}
        \fbox{\includegraphics[width=0.100\textwidth]{{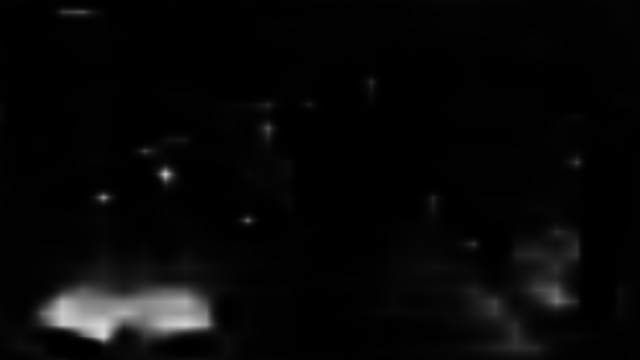}}}
        \fbox{\includegraphics[width=0.100\textwidth]{{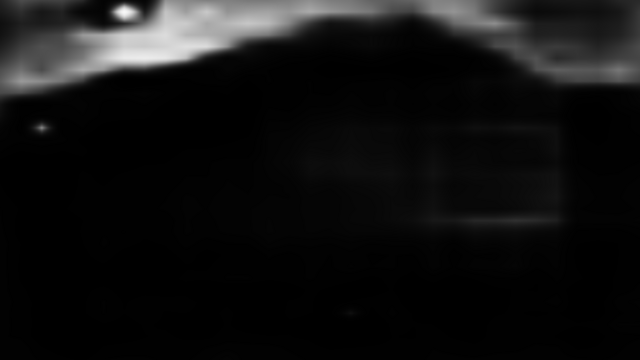}}}
        \fbox{\includegraphics[width=0.100\textwidth]{{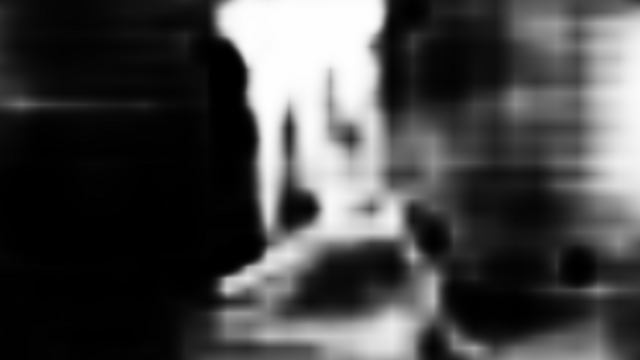}}}
        \fbox{\includegraphics[width=0.100\textwidth]{{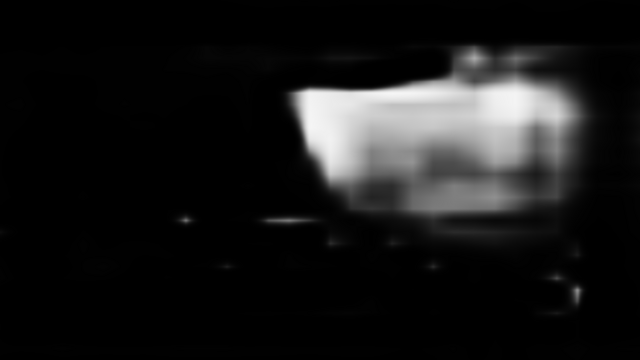}}}
        \fbox{\includegraphics[width=0.100\textwidth]{{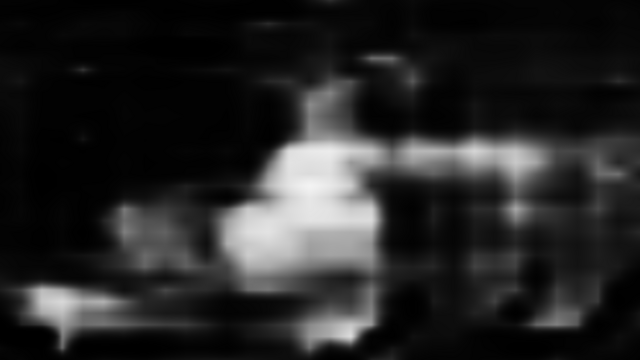}}}
        \fbox{\includegraphics[width=0.100\textwidth]{{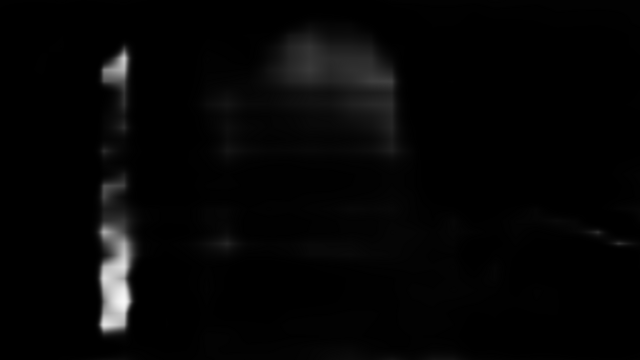}}}
        \smallskip
    \end{minipage}
    
	\vspace*{-0.8\baselineskip}

    \caption{\label{fig:videosham_localization_results} 
    This figure shows the localization results of different networks on the VideoSham dataset. Our proposed network's localization results are reasonable, with minor false alarms on column 1, 2, 3 and mis-detections on column 3, 4, 5. Our network were able to largely identify the manipulated regions in each of these example. However, on scenes like column 4 where the manipulated region was too small, it was not possible for our network and other competing networks to detect. Our network also missed two added books in the back on columns 3, likely because their sizes were small. Additionaly, on column 5, where the color of the wall on the right was changed from blue to red, our network failed to identify the entire manipulated region. Other than these errors, our network produced reasonable localization results, while other competing methods largely failed to identify any manipulations.
    }

	\vspace*{-0.5\baselineskip}

\end{figure*}


 \fi

\end{document}